\documentclass{article}

\usepackage[nonatbib, final]{neurips_2023}
\usepackage[numbers, sort, compress]{natbib}
\usepackage[dvipsnames, table]{xcolor}
\usepackage[utf8]{inputenc} 
\usepackage[T1]{fontenc}    
\usepackage{hyperref}       
\usepackage{url}            
\usepackage{booktabs}       
\usepackage{amsfonts}       
\usepackage{nicefrac}       
\usepackage{listings}
\usepackage{microtype}      
\usepackage{mathtools}
\usepackage{refcount}

\usepackage{algorithm}
\usepackage{algorithmic}
\usepackage{amsmath,amssymb, amsthm}
\usepackage{physics}
\usepackage[subtle, mathdisplays=tight, charwidths=normal, leading=normal]{savetrees}
\usepackage{caption}
\usepackage[bottom]{footmisc}
\usepackage{subcaption}
\usepackage{wrapfig}

\usepackage{amsmath,amsfonts,bm, amsthm}

\newtheorem{theorem}{Theorem}

\def\eqref#1{equation~\ref{#1}}

\def\1{\bm{1}}

\def\rvg{{\mathbf{g}}}

\def\rvx{{\mathbf{x}}}
\def\rvy{{\mathbf{y}}}

\def\rmD{{\mathbf{D}}}

\def\vg{{\bm{g}}}

\def\vo{{\bm{o}}}

\def\vx{{\bm{x}}}

\def\mD{{\bm{D}}}

\def\mX{{\bm{X}}}

\DeclareMathAlphabet{\mathsfit}{\encodingdefault}{\sfdefault}{m}{sl}
\SetMathAlphabet{\mathsfit}{bold}{\encodingdefault}{\sfdefault}{bx}{n}

\def\sD{{\mathbb{D}}}

\def\sG{{\mathbb{G}}}
\def\sH{{\mathbb{H}}}

\def\sK{{\mathbb{K}}}

\def\sO{{\mathbb{O}}}

\def\sX{{\mathbb{X}}}
\def\sY{{\mathbb{Y}}}

\newcommand{\E}{\mathbb{E}}

\newcommand{\R}{\mathbb{R}}

\DeclareMathOperator*{\argmax}{arg\,max}
\DeclareMathOperator*{\argmin}{arg\,min}

\newcommand{\todo}[1]{\textcolor{red}{\textbf{TODO: #1}}}
\newcommand{\custompar}[1]{\noindent{\bf #1.}\:}

\newcommand{\loss}{f}

\usepackage{float}
\usepackage{enumitem}
\usepackage{xurl} 
\usepackage{placeins} 
\usepackage{amsmath}
\usepackage{amssymb}
\usepackage{mathtools}
\usepackage{amsthm}

\theoremstyle{plain}

\theoremstyle{definition}
\newtheorem{definition}[theorem]{Definition}
\theoremstyle{remark}

\newcommand{\pred}{\hat{y}}
\newcommand{\instance}{\vx}
\newcommand{\group}{\vg}
\newcommand{\olabel}{o}
\newcommand{\olabels}{\sO}
\newcommand{\instances}{\sX}
\newcommand{\labels}{\sY}
\newcommand{\hatlabels}{\hat{\labels}}
\newcommand{\datasets}{\sD}
\newcommand{\dataset}{\mD}
\newcommand{\hatdataset}{\hat{\dataset}}
\newcommand{\hatdatasets}{\hat{\datasets}}
\newcommand{\datasetk}{\dataset_k}
\newcommand{\hatdatasetk}{\hatdataset_k}
\newcommand{\hclass}{\sH}
\newcommand{\tproc}{\mathcal{A}}
\newcommand{\model}{h}
\newcommand{\modelk}{\model_{\datasetk}}
\newcommand{\hatmodel}{\hat{h}}

\newcommand{\possiblemodels}{\mu}
\newcommand{\regressor}{r}
\newcommand{\regressork}{\regressor_{\datasetk}}
\newcommand{\boot}{B}

\newcommand{\err}{\texttt{Err}}
\newcommand{\fp}{\texttt{FP}}

\newcommand{\fn}{\texttt{FN}}
\newcommand{\tp}{\texttt{TP}}
\newcommand{\tn}{\texttt{TN}}

\newcommand{\fpr}{\texttt{FPR}}
\newcommand{\hatpr}{\hat{\texttt{PR}}}
\newcommand{\fnr}{\texttt{FNR}}
\newcommand{\haterr}{\hat{\texttt{Err}}}
\newcommand{\hatfpr}{\hat{\texttt{FPR}}}
\newcommand{\hatfnr}{\hat{\texttt{FNR}}}
\newcommand{\costfp}{C_{\text{01}}}
\newcommand{\costfn}{C_{\text{10}}}

\newcommand{\variance}{\texttt{var}\big(\tproc, \datasets, (\instance, \group)\big)}
\newcommand{\hatvariance}{\hat{\texttt{var}}\big(\tproc, \hatdatasets, (\instance, \group)\big)}
\newcommand{\hatvar}{\hat{\texttt{var}}}
\newcommand{\consistency}{\texttt{SC}\big(\tproc, \datasets, (\instance, \group)\big)}
\newcommand{\hatconsistency}{\hat{\texttt{SC}}\big(\tproc, \{\hatdataset_b\}_{b=1}^\boot, (\instance, \group)\big)}
\newcommand{\hatsc}{\hat{\texttt{SC}}}
\newcommand{\hatar}{\hat{\texttt{AR}}}

\newif\ifredacted
\newif\ifnotredacted
\notredactedtrue
\ifredacted
\newcommand{\code}{[REDACTED]}
\else 
\newcommand{\code}{\texttt{https://github.com/pasta41/variance}}
\fi

\ifredacted
\newcommand{\hmda}{[REDACTED]}
\else 
\newcommand{\hmda}{\texttt{https://github.com/pasta41/hmda}}
\fi

\title{Arbitrariness and Social Prediction: \\ The Confounding Role of Variance in Fair Classification}

\author{%
  A. Feder Cooper,\thanks{Corresponding author. \url{https://afedercooper.info}; \url{https://genlaw.org}}\\
  The GenLaw Center\\
  Cornell University\\
  \And
  Katherine Lee,\\
  The GenLaw Center\\
  Cornell University\\
  \And
  Madiha Zahrah Choksi,\\
  Cornell University\\
  \And
  Solon Barocas,\\
  Microsoft Research\\
  Cornell University\\
  \And
  Christopher De Sa,\\
  Cornell University
  \And James Grimmelmann,\\
  Cornell University\\
  The GenLaw Center
  \And Jon Kleinberg,\\
  Cornell University
  \And Siddhartha Sen,\\
  Microsoft Research
  \And Baobao Zhang\\
  Syracuse University
}

\begin{document}

\maketitle

\begin{abstract} 
Variance in predictions across different trained models is a significant, under-explored source of error in 
fair binary classification. 
In practice, the variance on some data examples is so large that decisions can be effectively \emph{arbitrary}. 
To investigate this problem, we take an experimental approach and make four overarching contributions.  
We: 1) Define a metric called \emph{self-consistency}, derived from variance, which we use as a proxy for measuring and reducing arbitrariness; 
2) Develop an ensembling algorithm that abstains from classification when a prediction would be arbitrary; 
3) Conduct the largest to-date empirical study of the role of variance (\emph{vis-a-vis} self-consistency and arbitrariness) in fair binary classification; and, 
4) Release a toolkit that makes the US Home Mortgage Disclosure Act (\texttt{HMDA}) datasets easily usable for future research. 
Altogether, our experiments reveal shocking insights about the reliability of conclusions on benchmark datasets. 
\textbf{Most fair binary classification benchmarks are \emph{close-to-fair} when taking into account the amount of arbitrariness present in predictions 
--- 
\emph{before} we even try to apply \emph{any} fairness interventions.} 
This finding calls into question the practical utility of common algorithmic fairness methods, and in turn suggests that we should reconsider how we choose to measure fairness in binary classification.\looseness=-1
\end{abstract}
\section{Introduction}\label{sec:intro}
A goal of algorithmic fairness is to develop techniques that measure and mitigate discrimination in automated decision-making. 
In fair binary classification, this often involves training a model to satisfy a chosen \emph{fairness metric}, which typically defines fairness as parity between model error rates for different demographic groups in the dataset~\citep{barocas2019textbook}. 
However, even if a model's classifications satisfy a particular fairness metric, it is not necessarily the case that the model is equally confident in each classification.\looseness=-1

\begin{figure}[t!]
    \centering
    \includegraphics[width=.45\textwidth]{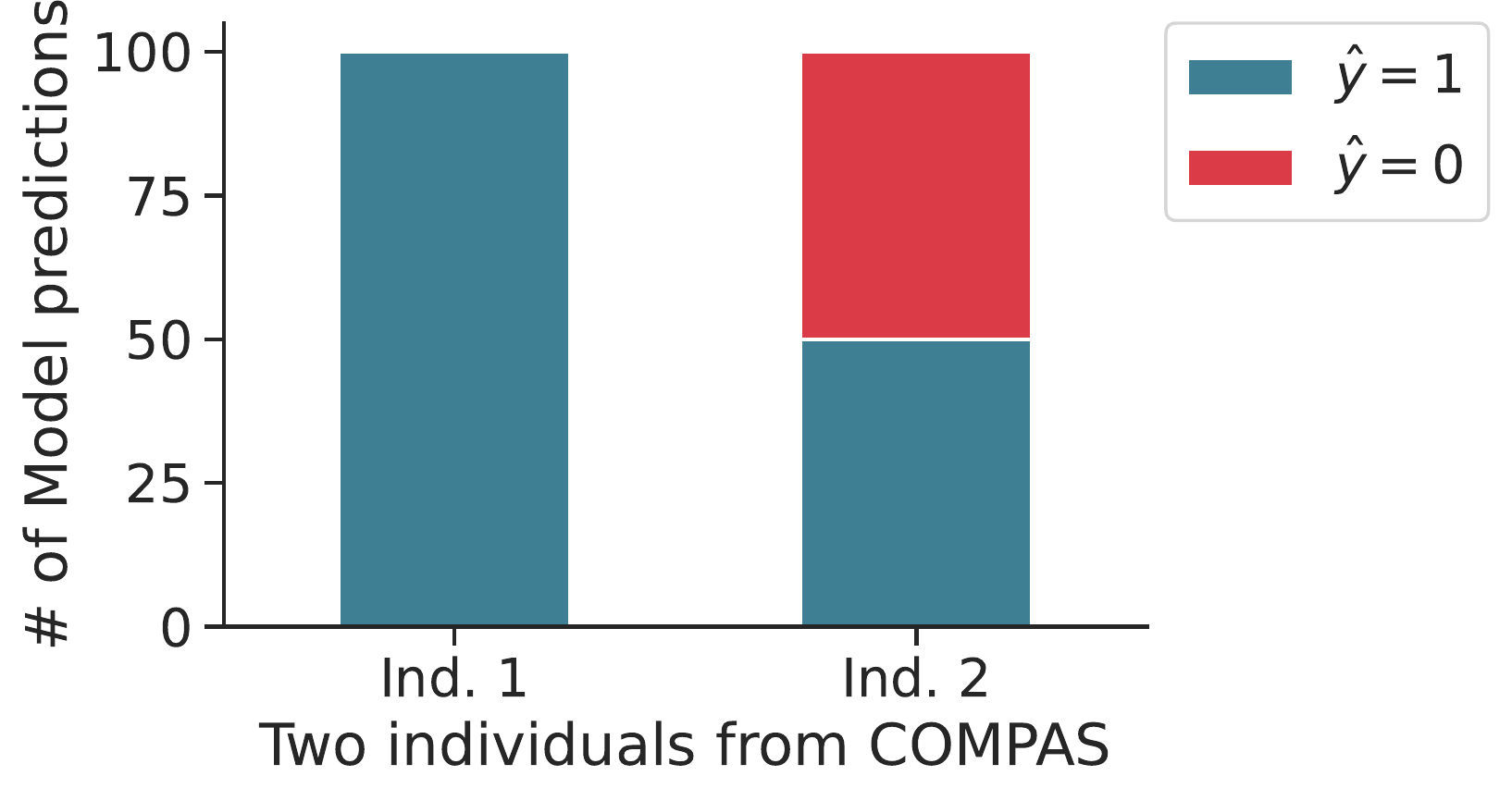}
    \caption{100 bootstrapped logistic regression models 
    show models can be very consistent in predictions $\hat{y}$ for some individuals (Ind. 1) and arbitrary for others (Ind. 2).}
    \label{fig:vote}
  \vspace{-.5cm}
\end{figure}

To provide an intuition for what we mean by confidence, consider the following experiment: 
We fit 100 logistic regression models using the same learning process, which draws different subsamples of the training set from the \texttt{COMPAS} prison recidivism dataset~\citep{larson2016propublica, friedler2019datasets}, and we compare the resulting classifications for two individuals in the test set. 
Figure~\ref{fig:vote} shows a difference in the consistency of predictions for both individuals:  
the 100 models  agree completely to classify Individual 1 as ``will recidivate'' and disagree completely on whether to classify Individual 2 as ``will'' or ``will not recidivate.'' 
If we were to pick one model at random to use in practice, there would be no effect on how Individual 1 is classified;
%
yet, for Individual 2, the prediction is effectively random. 
We can interpret this disagreement to mean that the learning process that produced these predictions is not sufficiently confident to justify assigning Individual 2 \emph{either decision outcome}. 
In practice, instances like Individual 2 exhibit so little confidence that their classification is effectively \emph{arbitrary}~\citep{cooper2022accountability,cooper2022lawless,creel2022leviathan}.
Further, this arbitrariness can also bring about discrimination if classification decisions are \emph{systematically more arbitrary} for individuals in certain demographic groups.\looseness=-1

A key aspect of this example is that we use only one model to make predictions. 
This is the typical setup in fair binary classification: 
Popular metrics are commonly applied to evaluate the fairness of a \emph{single model}~\citep{hardt2016eo, pleiss2017calibration, kleinberg2017impossibility}. 
However, as is clear from the example learning process in Figure~\ref{fig:vote}, using only a single model can mask the arbitrariness of predictions. 
Instead, to reveal arbitrariness, we must examine \emph{distributions over possible models for a given learning process}. 
With this shift in frame, we ask:\looseness=-1 

\vspace{-.15cm}
\begin{center}
\emph{What is the empirical role of arbitrariness in  fair binary classification tasks?}
\end{center}
\vspace{-.2cm}

\noindent To study this question, we make four contributions: 
\begin{enumerate}[topsep=2pt, itemsep=0pt, leftmargin=.5cm]
    \item \textbf{Quantify arbitrariness.} We formalize a metric called \emph{self-consistency}, derived from statistical variance, which we use as a quantitative proxy for arbitrariness of model outputs. Self-consistency is a simple yet powerful tool for empirical analyses of fair classification (Section~\ref{sec:significance}). 
    
    \item \textbf{Ensemble to improve self-consistency.} We extend~\citeauthor{breiman1996bagging}'s classic bagging to allow for abstaining from classifying instances for which self-consistency is low. 
    This improves overall self-consistency (i.e., reduces variance), and improves accuracy (Section~\ref{sec:algorithms}).\looseness=-1

    \item \textbf{Perform a comprehensive experimental study of variance in fair binary classification.\looseness=-1} 
    We conduct the largest-to-date such study, through the lens of self-consistency and its relationship to arbitrariness. 
    \textbf{Surprisingly, we find that most benchmarks 
    are \textit{close-to-fair} when taking into account the amount of arbitrariness present in predictions --- \emph{before} we even try to apply \emph{any} fairness interventions} (Section~\ref{sec:empirical}). 
    This shocking finding has huge implications for the field: it casts doubt on the reliability of prior work that claims there is baseline unfairness in these benchmarks, in order to demonstrate that methods to improve fairness work in practice. 
    We instead find that such methods are often empirically unnecessary to improve fairness 
    (Section~\ref{sec:related}).\looseness=-1

    \item  \textbf{Release a large-scale fairness dataset package.} 
    We observe that variance, particularly in small datasets, can undermine the reliability of conclusions about fairness.  
    We therefore  open-source a package that makes the large-scale US Home Mortgage Disclosure Act datasets (\texttt{HMDA}) easily usable for future research. 
\end{enumerate}
\section{Preliminaries on Fair Binary Classification}\label{sec:prelim}

To analyze arbitrariness in the context of fair binary classification, we first need to establish our background 
definitions. This material is likely familiar to most readers. Nevertheless, we highlight particular details that are important for understanding the experimental methods that enable our contributions. We present the fair-binary-classification problem formulation and associated empirical approximations, with an emphasis on the \emph{distribution over possible models} that could be produced from training on different subsets of data drawn from the same data distribution.\looseness=-1

\subsection{Problem formulation}\label{sec:prelim:form}

Consider a distribution $q(\cdot)$ from which we can sample \textit{examples} $(\instance, \group, \olabel)$. 
The $\instance \in \instances \subseteq \R^m$ are feature \textit{instances} and $\group \in \sG$ is a group of \textit{protected attributes} that we do not use for learning (e.g., race, gender).\footnote{We examine the common setting in which $|\group|=1$, and abuse notation by treating $\group$ like a scalar with $\sG = \{0, 1\}$.} 
The $\olabel \in \olabels$ are the associated \textit{observed labels}, and $\olabels \subseteq \labels$, where $\labels =  \{0, 1\}$ is the label space. 
From $q(\cdot)$ we can sample training datasets $\{(\instance, \group, \olabel)\}_{i=1}^n$, with $\datasets$ representing the set of all  $n$-sized datasets. 
To reason about the possible models of a hypothesis class $\hclass$ that could be learned from the different subsampled datasets $\datasetk \in \datasets$, we define a \textit{learning process}:\looseness=-1

\begin{definition}
    \label{def:learningprocess}
    A \textbf{learning process} is a randomized function that runs instances of a \textbf{training procedure} $\tproc$ on each $\datasetk \in \datasets$ and a model specification, in order to produce \textbf{classifiers} $\modelk \in \hclass$. 
    A particular run $\tproc(\datasetk) \rightarrow \modelk$, where $\modelk: \instances \rightarrow \labels$, which is  deterministic mapping from the instance space $\instances$ to the label space $\labels$. 
    All such runs over $\datasets$ produce a distribution over possible trained models, $\possiblemodels$. 
\end{definition}

Reasoning about $\possiblemodels$, rather than individual models $\modelk$, enables us to contextualize arbitrariness in the data, which, in turn, is captured by learned models (Section~\ref{sec:significance}).\footnote{Model multiplicity has similar aims, but ultimately relocates  the arbitrariness we describe to model selection (Section~\ref{sec:related}; Appendix \ref{app:sec:mm}).} 
Each particular model $\modelk\sim\possiblemodels$ deterministically produces classifications $\pred = \modelk(\instance)$. 
The classification rule is $\modelk(\instance) = \1[\regressork(\instance) \geq \tau]$, for some threshold $\tau$, where regressor $\regressork: \instances \rightarrow [0, 1]$ computes the probability of positive classification. 
Executing $\tproc(\datasetk)$ produces $\modelk \sim \possiblemodels$ by minimizing the \textit{loss} of predictions $\pred$ with respect to their associated observed labels $\olabel$ in $\datasetk$. 
This loss is computed by a chosen \textit{loss function} $\loss:\labels \times \labels \mapsto \R$.  
We compute predictions for a \textit{test set} of fresh examples and calculate their loss. 
The loss is an estimate of the \textit{error} of $\modelk$, which is dependent on the specific dataset $\datasetk$ used for training. 
To generalize to the error of all possible models produced by a specific learning process (Definition~\ref{def:learningprocess}), we consider the \textit{expected error}, $\err(\tproc, \datasets, (\instance,\;\group,\; \olabel)) =\E_{\rmD}[\loss(\olabel,\; \pred)|\rvx = \instance]$.\looseness=-1 

In fair classification, it is common to use  \textit{0-1 loss} $\triangleq \1[\pred \ne \olabel]$ or \textit{cost-sensitive loss}, which assigns asymmetric costs $\costfp$ for false positives \fp{} and $\costfn$ for false negatives \fn{}~\citep{elkan2001cost}. 
These costs are related to the classifier threshold $\tau = \frac{\costfp}{\costfp + \costfn}$, with $\costfp, \costfn \in \R^+$ (Appendix~\ref{app:sec:prelim:costs}). 
Common fairness metrics, such as Equality of Opportunity~\citep{hardt2016eo}, further analyze error by computing disparities across group-specific error rates $\fpr_{\group}{}$ and $\fnr_{\group}$. 
For example, $\fpr_\group \triangleq p_\possiblemodels[\regressor_{\rmD}(\rvx) \geq \tau| \olabel = 0, \rvg = \group] = p_\possiblemodels[\pred=1 | \olabel = 0, \rvg = \group]$. 
Model-specific $\fpr_\group$ and $\fnr_\group$ are further-conditioned on the dataset used in training, i.e., $\rmD = \datasetk$.

\subsection{Empirical approximation of the formulation}\label{sec:prelim:emp}

We typically only have access to one dataset, not the data distribution $q(\cdot)$. 
In fair binary classification experiments, it is common to estimate expected error by performing \textit{cross validation} (CV) on this dataset to produce a small handful of models~\citep[e.g.]{chen2018tradeoff, jiang2020wass, corbettdavies2017cost}. 
\textbf{CV can be unreliable when there is high variance}; 
it can produce error estimates that are themselves high variance, and does not reliably estimate expected error with respect to possible models $\possiblemodels$ (Section~\ref{sec:empirical}). 
For more details, see~\citet{efron1997boot, efron1993bootsrap} and \citet{wager2020cv}. 

To get around these reliability 
issues, one can \textit{bootstrap}.\footnote{We could use MCMC~\citep{zhang2020amagold}, but optimization is the standard tool that allows use of standard models  in fairness.\looseness=-1} 
Bootstrapping splits the available data into train and test sets, and simulates drawing different training datasets from a distribution by resampling the train set $\hatdataset$, generating replicates $\hatdataset_1, \hatdataset_2, \ldots, \hatdataset_\boot \coloneqq \hatdatasets$. 
We use these replicates $\hatdatasets$ 
to approximate the learning process on $\datasets$ (Def.~\ref{def:learningprocess}). 
We treat the resulting $\hatmodel_{\hatdataset_1}, \hatmodel_{\hatdataset_2}, \ldots, \hatmodel_{\hatdataset_\boot}$ as our empirical estimate for the distribution $\hat{\mu}$, and evaluate their predictions for the \emph{same reserved test set}. 
This enables us to produce comparisons of classifications across test instances like in Fig.~\ref{fig:vote} (Appendix~\ref{app:sec:prelim:boot}).\looseness=-1 
\section{Variance, Self-Consistency, and  Arbitrariness\looseness=-1}\label{sec:significance}

From these preliminaries, we can now pin down arbitrariness more precisely. 
We develop a quantitative proxy for measuring arbitrariness, called \emph{self-consistency} (Section~\ref{sec:var:sc}), which is derived from a definition of statistical \emph{variance} between different model predictions (Section~\ref{sec:var:intuition}). 
We then illustrate how self-consistency is a simple-yet-powerful tool for revealing the role of arbitrariness in fair classification (Section~\ref{sec:var:arbitrary}). 
Next, we will introduce an algorithm to improve self-consistency (Section~\ref{sec:algorithms}) and compute self-consistency on popular fair binary classification benchmarks (Section~\ref{sec:empirical}).

\subsection{Arbitrariness resembles statistical variance}\label{sec:var:intuition}

In Section~\ref{sec:prelim}, we discussed how common fairness metrics analyze error by computing  
false positive rate (\fpr{}) and false negative rate (\fnr{}). Another common way to formalize error is as a 
decomposition of different statistical sources: \emph{noise}-, \emph{bias}-, and \emph{variance}-induced error~\citep{abumostafa2012learning, geman1992bvd}. 
To understand our metric for self-consistency (Section~\ref{sec:var:sc}), we first describe how the arbitrariness in Figure~\ref{fig:vote} (almost, but not quite) resembles variance.\looseness=-1

Informally, variance-induced error quantifies fluctuations in individual example predictions for different models $\modelk \sim \possiblemodels$. Variance is the error in the learning process that comes from training on different datasets $\datasetk \in \datasets$. In theory, we measure variance by imagining training all possible $\modelk \sim \possiblemodels$, testing them all on the same test instance $(\instance, \group)$, and then quantifying how much the resulting classifications for $(\instance, \group)$ deviate \emph{from each other}. More formally,

\begin{definition}
\label{def:variance}
\looseness=-1
For all pairs of possible models $\model_{\dataset_i}, \model_{\dataset_j}\sim\possiblemodels \;(i\neq j)$, the \textbf{variance} for a 
test 
$(\instance, \group)$ is\looseness=-1
{
\begin{align*}
\variance &\triangleq \E_{\model_{\dataset_i} \sim \possiblemodels, \model_{\dataset_j} \sim \possiblemodels}\Big[\loss\Big(\model_{\dataset_i}(\instance), \model_{\dataset_j}(\instance)\Big)\Big].
\end{align*}
}%
\end{definition}

We can approximate variance directly by using the bootstrap method (Section~\ref{sec:prelim:emp}, Appendix~
\ref{app:sec:noisebias}). For 0-1 and cost-sensitive loss with costs $\costfp, \costfn \in \R^+$ (Section~\ref{sec:prelim:form}), we can generate $\boot$ replicates to train $\boot$ concrete models that serve as our approximation for the distribution  $\hat{\possiblemodels}$. For $\boot= \boot_0 + \boot_1 > 1$, where $\boot_0$ and $\boot_1$ 
denote the number of $0$- and $1$-class predictions for $(\instance, \group)$,\looseness=-1 

{\small
\begin{align}
\label{eq:hatvar}
\hatvariance \coloneqq \frac{1}{\boot(\boot-1)} \sum_{i \neq j} \loss\Big(\hatmodel_{\hatdataset_i}(\instance), \hatmodel_{\hatdataset_j}(\instance)\Big) 
= \frac{(\costfp + \costfn)\boot_0\boot_1}{\boot(\boot-1)}.
\end{align}
}%

We derive (\ref{eq:hatvar}) in Appendix~
\ref{app:sec:ourvariance} and show that, for increasingly large $\boot$, $\hatvar$ is defined on $[0, \frac{\costfp + \costfn}{4} + \epsilon]$.

\subsection{Defining self-consistency from variance}\label{sec:var:sc} 

It is clear from above that, in general, variance (\ref{eq:hatvar}) is unbounded. We can always increase the maximum possible $\hatvar$ by increasing the magnitudes of our chosen $\costfp$ and $\costfn$.\footnote{Because $\tau = \frac{\costfp}{\costfp + \costfn}$, for a given $\tau$ we can scale costs arbitrarily and have the same decision rule  (Section~\ref{sec:prelim:form}). Relative, not absolute, costs affect the number of classifications $\boot_0$ and $\boot_1$.} 
However, as we can see from our intuition for arbitrariness in Figure~\ref{fig:vote}, the most important takeaway is the amount of (dis)agreement, reflected in the counts $\boot_0$ and $\boot_1$. Here, there is no notion of the cost of misclassifications. So, variance (\ref{eq:hatvar}) does not exactly measure what we want to capture. Instead, we want to focus unambiguously on the (dis)agreement part of variance, which we call \emph{self-consistency of the learning process}:\looseness=-1

\begin{definition}
\label{def:sc}
\looseness=-1
For all pairs of possible models $\model_{\dataset_i}, \model_{\dataset_j}\sim\possiblemodels \;(i\neq j)$, the \textbf{self-consistency of the learning process} for a 
test 
$(\instance, \group)$ is\looseness=-1
{
\begin{align}
\label{eq:sc}
\consistency \triangleq \E_{\model_{\dataset_i} \sim \possiblemodels, \model_{\dataset_j} \sim \possiblemodels}\Big[\model_{\dataset_i}(\instance) = \model_{\dataset_j}(\instance)\Big] 
 = p_{\model_{\dataset_i} \sim \possiblemodels, \model_{\dataset_j} \sim \possiblemodels}\big(\model_{\dataset_i}(\instance) = \model_{\dataset_j}(\instance)\big).
\end{align}
}%
\end{definition}

In words, (\ref{eq:sc}) models the probability that two models produced by the same learning process on different $n$-sized training datasets agree on their predictions for the same test instance.\footnote{(\ref{eq:sc}) follows from 
it being equally likely to draw any two $\dataset_i, 
\dataset_j \in \datasets$ in a learning process (Appendix~
\ref{app:sec:consistency}).} Like variance, we can derive an empirical approximation of \texttt{SC}. Using the bootstrap method with $\boot = \boot_0 + \boot_1 > 1$,\looseness=-1 

{\small
\begin{align}
\label{eq:hatsc}
\hat{\texttt{SC}}\big(\mathcal{A}, \hatdatasets, (\instance, \group)\big) \coloneqq \frac{1}{\boot(\boot-1)} \sum_{i \neq j} \1\Big[\hatmodel_{\hatdataset_i}(\instance) = \hatmodel_{\hatdataset_j}(\instance)\Big]  
= 1 - \frac{2\boot_0\boot_1}{\boot(\boot-1)}.
\end{align}
}%

For increasingly large $\boot$, $\hat{\texttt{SC}}$ is defined on $[0.5 - \epsilon, 1]$ (Appendix~\ref{app:sec:consistency}). Throughout, we use the shorthand \emph{self-consistency}, but it is important to note that Definition~\ref{def:sc} is a property of the distribution over possible models $\mu$ produced by the learning process, not of individual models. We summarize other important takeaways below:

\setlength{\tabcolsep}{6pt}
\begin{figure*}[t!]
\begin{minipage}{.495\linewidth}
\centering
\hspace{-.4cm}
        \includegraphics[width=.85\linewidth]{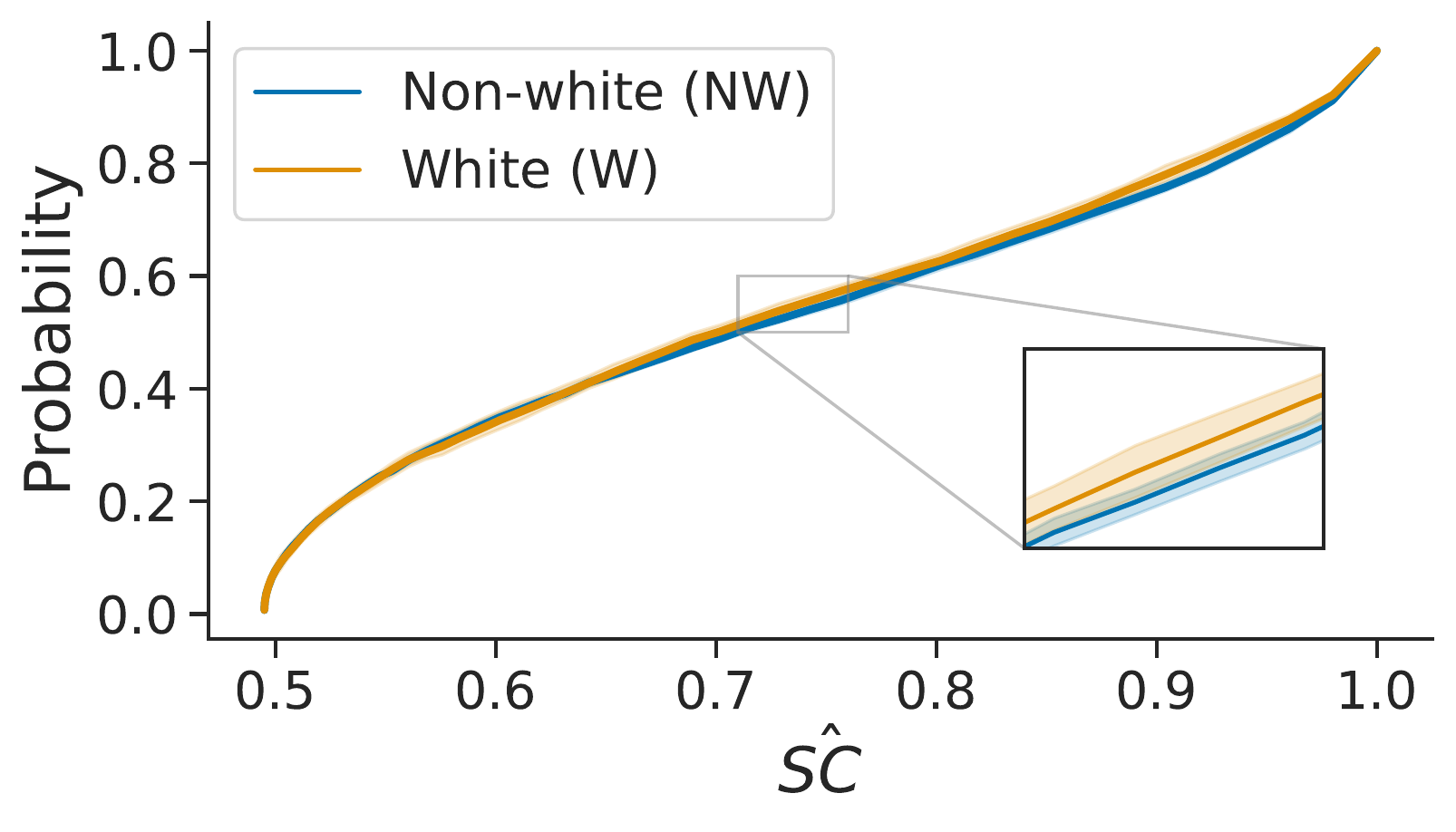}\\
        \vspace{-.1cm}%
        \footnotesize
        \begin{tabular}{lccc}
        \toprule
             & \textbf{$\Delta\haterr$} & \textbf{$\Delta\hatfpr$} & \textbf{$\Delta\hatfnr$} 
             \\ \cmidrule{2-4}
            & $1.0\pm1.4\%$ & $2.0\pm1.4\%$ & $0.9\pm1.4\%$ 
            \\
            \midrule
            & \textbf{$\haterr$} & \textbf{$\hatfpr$} & \textbf{$\hatfnr$} 
            \\ \midrule
            \textbf{Total} & $36.6\pm0.5\%$ & $17.3\pm0.8\%$ & $19.3\pm0.7\%$ 
            \\ \midrule
             $\text{NW}$ & $36.9\pm0.5\%$ & $18.0\pm0.7\%$ & $19.0\pm0.8\%$ 
             \\ \midrule 
            $\text{W}$ & $35.9\pm1.3\%$ & $16.0\pm1.2\%$ & $19.9\pm1.1\%$
            \\ 
              \bottomrule
        \end{tabular}
        \subcaption{\texttt{COMPAS} split by $\texttt{race}$; random forests (RFs)}
        \label{subfig:compas-cdf-rfc}
\end{minipage}%
\hspace{.25cm}
\begin{minipage}{.495\linewidth}
\centering
\hspace{-.2cm}
        \includegraphics[width=.85\linewidth]{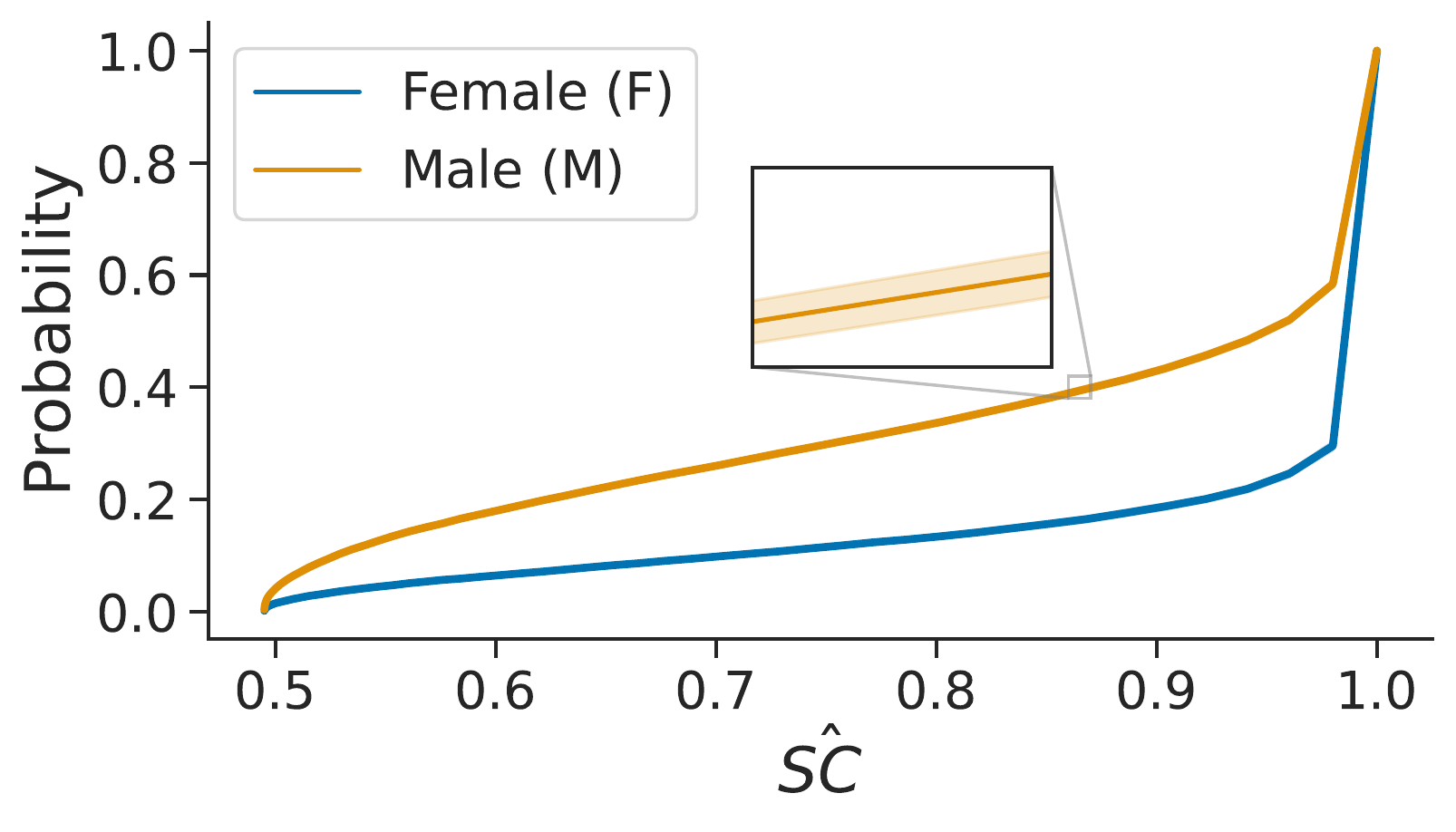}\\\vspace{-.1cm}
        \footnotesize
        \begin{tabular}{lccc}
        \toprule
             & \textbf{$\Delta\haterr$} & \textbf{$\Delta\hatfpr$} & \textbf{$\Delta\hatfnr$} 
             \\ \cmidrule{2-4}
            & $12.2\pm0.4\%$ & $6.0\pm0.3\%$ & $6.3\pm0.3\%$ 
            \\ 
            \midrule
             & \textbf{$\haterr$} & \textbf{$\hatfpr$} & \textbf{$\hatfnr$} 
             \\ \midrule
             \textbf{Total} & $17.3\pm0.3\%$ & $7.7\pm0.3\%$ & $9.6\pm0.1\%$ 
             \\ \midrule
             $\text{F}$\;\;\;\; & $9.0\pm0.3\%$ & $3.7\pm0.1\%$ & $5.3\pm0.3\%$ 
             \\ \midrule
             $\text{M}$ & $21.2\pm0.3\%$ & $9.7\pm0.3\%$ & $11.6\pm0.1\%$ 
             \\ 
             \bottomrule
        \end{tabular}
        \subcaption{\texttt{Old Adult} split by $\texttt{sex}$; random forests (RFs)}
        \label{subfig:adult-cdf-rfc}
\end{minipage}%
\vspace{.2cm}
\caption{$\hatsc$ CDFs for \texttt{COMPAS} (\ref{subfig:compas-cdf-rfc}) and  \texttt{Old Adult} (\ref{subfig:adult-cdf-rfc}). We train random forests  ($B=101$ replicates), and repeat with 10 train/test splits to produce (very tight) confidence intervals. 
$\hatsc$ is effectively identical across subgroups $\group$ in \texttt{COMPAS}; \texttt{Old Adult} exhibits \textbf{systematic} differences in \textbf{arbitrariness} across $\group$. Tables show mean $\pm$ STD of the relative disparities, 
e.g., $\Delta\haterr=|\haterr_0 - \haterr_1|$ (top); and, the absolute $\haterr, \hatfpr, \hatfnr, $ and $\hatsc$, also broken down by $\group$ (bottom) (Appendix~
\ref{app:sec:empirical}).\looseness=-1}
\label{fig:adult-compas-cdf-rfc}
\vspace{-.6cm}
\end{figure*}

\custompar{Terminology} In naming our metric, we intentionally evoke related notions of ``consistency'' in logic and the law~(\citet{fuller1965law, stalnaker2006logic}; Appendix~
\ref{app:sec:consistency}).\looseness=-1 

\custompar{Interpretation}  Definition~\ref{def:sc} is defined on $[0.5, 1]$, which coheres with the intuition in Figure~\ref{fig:vote}:  $0.5$ and $1$ respectively reflect minimal (Individual 2) and maximal (Individual 1) possible $\texttt{SC}$. $\texttt{SC}$, unlike \fpr{} and \fnr{} (Section~\ref{sec:prelim:form}), does \emph{not} depend on the observed label $\olabel$. It captures the learning process's confidence in a classification $\pred$, but says nothing directly about $\pred$'s accuracy. By construction, \textbf{low self-consistency indicates high variance, and vice versa}. We derive empirical $\hatsc$ (\ref{eq:hatsc}) from $\hatvar$ (\ref{eq:hatvar}) by leveraging observations about the definition of $\hatvar$ for 0-1 loss (Appendix~
\ref{app:sec:consistency}). While there are no costs $\costfp$, $\costfn$ in computing (\ref{eq:hatsc}), they still affect empirical 
measurements of $\hatsc$. Because $\costfp$ and $\costfn$ affect $\tau$ (Section~\ref{sec:prelim:form}), they control the concrete number of $\boot_0$ and $\boot_1$, and thus the $\hatsc$ we measure in experiments.\looseness=-1  

\custompar{Empirical focus} Since self-consistency depends on the particular data subsets used in training, conclusions about its relevance 
vary according to task. \textbf{This is why we take a practical approach for our main results  --- of running a large-scale experimental study on many different datasets to extract 
general observations about $\hatsc$'s practical effects} (Section~\ref{sec:empirical}). In our experiments, we typically use $\boot=101$, which yields a $\hatsc$ range of $[\approx 0.495, 1]$ in practice.\footnote{\citet{efron1993bootsrap} recommend $\boot \in \{50 \ldots 200\}$.\looseness=-1}\looseness=-1 

\custompar{Relationship to other fairness concepts} Self-consistency is qualitatively different from traditional fairness metrics. 
Unlike \fpr{} and \fnr{}, \texttt{SC} does not depend on observed label $\olabel$. This has two important implications. First,  while calibration 
also measures a notion of confidence, it is 
different: calibration reflects confidence with respect to \emph{a model} predicting $\olabel$, but says nothing about the relative confidence in predictions $\hat{y}$ produced by the \emph{possible models} $\possiblemodels$ that result from the learning process~\citep{pleiss2017calibration}. 
Second, a common assumption in algorithmic fairness is that there is \emph{label bias} --- that unfairness is due in part to discrimination reflected in recorded, observed decisions $\olabel$~\citep{friedler2016impossibility, cooper2021emergent}. 
As a result, it is arguably a nice side effect that self-consistency does not depend on $\olabel$. 
However, it is also possible to be perfectly self-consistent and inaccurate (e.g., $\forall k, \hat{y}_k \neq \olabel$; Section~\ref{sec:related}).

\subsection{Illustrating self-consistency in practice}\label{sec:var:arbitrary}
$\hatsc$ enables us to evaluate arbitrariness in classification experiments. It is straightforward to compute $\hatsc$ (\ref{eq:hatsc}) with respect to multiple test instances $(\instance, \group)$ --- for all instances in a test set or for all instances conditioned on membership in $\group$. Therefore, beyond visualizing $\hatsc$ for individuals (Figure~\ref{fig:vote}), we can also do so across sets of individuals. 
We plot the cumulative distribution (CDF) of $\hatsc$ for the groups $\group$ in the test set (i.e., the $x$-axis shows the range of $\hatsc$ for $\boot=101$, $[\approx0.495, 1]$). In Figure~\ref{fig:adult-compas-cdf-rfc}, we provide illustrative examples from two of the most common fair classification benchmarks~\citep{fabris2022datasets}, \texttt{COMPAS} and \texttt{Old Adult} using random forests (RFs). 
We split the available data into train and test sets, and bootstrap the train set $B=101$ times to train models $\hat{\model_1}, \hat{\model_2}, \ldots, \hat{\model_{101}}$ (Section~\ref{sec:prelim:emp}). We repeat this process on 10 train/test splits, and the resulting confidence intervals (shown in the inset) indicate that our $\hatsc$ estimates are stable. We group observations regarding these examples into two 
categories:\looseness=-1

\custompar{Individual arbitrariness} Both CDFs show that $\hatsc$ varies drastically across test instances. For random forests on the \texttt{COMPAS} dataset, about one-half of instances are under $.7$ self-consistent. \textbf{Nearly one-quarter of test instances are effectively $.5$ self-consistent; they resemble Individual 2 in Figure~\ref{fig:vote}, meaning that their predictions are essentially arbitrary.} These differences in $\hatsc$ across the test set persist even though the 101 models exhibit relatively small average disparities $\Delta\haterr$, $\Delta\hatfpr$, and $\Delta\hatfnr$ (Figure~\ref{subfig:compas-cdf-rfc}, bottom; Section~\ref{sec:empirical-repro}). This supports our motivating claim: it is possible to come close to satisfying fairness metrics, while the learning process exhibits very different levels of confidence for the underlying classifications that inform those metrics 
(Section~\ref{sec:intro}).\looseness=-1

\custompar{Systematic arbitrariness} 
We can also highlight $\hatsc$ according to groups. 
The $\hatsc$ plot for \texttt{Old Adult} shows that it is possible for the degree of arbitrariness to be \emph{systematically worse} for a particular demographic $\group$ (Figure~\ref{subfig:adult-cdf-rfc}). While the lack of $\hatsc$ is not as extreme as it is for \texttt{COMPAS} (Figure~\ref{subfig:compas-cdf-rfc}) --- the majority of test instances exhibit over $.9\;\;\; \hatsc$ --- there is more arbitrariness in the \texttt{Male} subgroup. 
We can quantify such \emph{systematic arbitrariness} using a measure of distance between probability distributions. We use 
the Wasserstein-1 distance ($\mathcal{W}_1$), which has a closed form for CDFs~\citep{ramdas2015wass}.  The $\mathcal{W}_1$ distance has an intuitive interpretation for measuring systematic arbitrariness: it computes the total disparity in \texttt{SC} by examining all possible \texttt{SC} levels $\kappa$ at once (Appendix~
\ref{app:sec:consistency}). 
For two groups $\group=0$ and $\group=1$ with respective 
\texttt{SC} CDFs $F_0$ and $F_1$, $\mathcal{W}_{1} \triangleq \int_{\R} |F_0(\kappa) - F_1(\kappa)| \; d\kappa$. 
For \texttt{Old Adult}, empirical $\hat{\mathcal{W}_1}=0.127$; for  \texttt{COMPAS}, which 
does not show systematic arbitrariness, $\hat{\mathcal{W}_1}=0.007$.
\section{Accounting for Self-Consistency}\label{sec:algorithms}

By definition, low $\hatsc$ signals that there is high $\hatvar$ (Section~\ref{sec:var:sc}). It is therefore a natural idea to use variance reduction techniques to improve $\hatsc$ (and thus reduce arbitrariness).

\begin{figure}[h!]
    \vspace{-.2cm}
  \begin{minipage}{0.49\linewidth}
      As a starting point for improving $\hatsc$, we perform variance reduction with~\citet{breiman1996bagging}'s \emph{bootstrap aggregation}, or \emph{bagging}, ensembling algorithm. 
      Bagging involves bootstrapping to produce a set of $\boot$ models (Section~\ref{sec:prelim:emp}), and then, for each test instance, producing an aggregated prediction $\pred_A$, which takes the \textbf{majority vote} of the $\pred_1, \ldots, \pred_\boot$ classifications.
      This procedure is practically effective for classifiers with high variance~\citep{breiman1996bagging, breiman1998ac}. 
      \textbf{However, by taking the majority vote, bagging embeds the idea that having slightly-better-than-random classifiers is sufficient for improving ensembled predictions, $\pred_A$}. 
      Unfortunately, there exist instances like Individual 2 (Figure~\ref{fig:vote}), where the classifiers in the ensemble are evenly split between classes. This means that bagging alone cannot overcome arbitrariness (Appendix~\ref{app:algo:sc}).\vspace{.2cm}

      To remedy this, we add the option to abstain from prediction if $\hatsc$ is low (Algorithm~\ref{algo:bagging-confidently}). 
      A minor adjustment to (\ref{eq:hatsc}) accounts for abstentions, and a simple proof follows that Algorithm~\ref{algo:bagging-confidently} improves $\hatsc$ (Appendix~\ref{app:sec:algorithm}). 
      We bootstrap as usual, but pro-
  \end{minipage}
  \hfill
  \vspace{-.2cm}
  \begin{minipage}{0.48\linewidth}
  \vspace{-.65cm}
    \setlength{\textfloatsep}{3pt}
\begin{algorithm}[H]
\caption{$\hatsc$ Ensembling with Abstention}\label{algo:bagging-confidently}
{\footnotesize
\textbf{Input}: training data $(\mX, \vo)$, $\tproc$, $\boot$, $\kappa \in [0.5, 1]$, $\instance_\text{test}$\\
\textbf{Output}: $\pred$ with $\hatsc \geq \kappa$ or \texttt{Abstain}}%
\\\vspace{-.2cm}
    \begin{algorithmic}[1]
    {\footnotesize
    \STATE $\pred_A \coloneqq \mathsf{list}()$ $ \hspace{1em} \rhd$ To store ensemble predictions
    \FOR{$1 \ldots \boot$} 
                \STATE $\dataset_\boot \leftarrow \mathsf{Bootstrap}\big((\mX, \vo)\big)$
                \STATE $\rhd$ $\hat{\model}_{\dataset_\boot}$ can itself be a bagged model, with $\tproc$
                \STATE \hspace{.2cm} bagging on $\dataset_\boot$ as the dataset to bootstrap \\\vspace{.1cm}
                \STATE $\hat{\model}_{\dataset_\boot} \leftarrow \tproc(\dataset_\boot)$
                \STATE $\pred_A.\mathsf{append}\big(\hat{\model}_{\dataset_\boot}(\instance_\text{test})\big) \hspace{1em} \rhd \pred_A = [\pred_{1}, \ldots, \pred_{\boot}]$
    \ENDFOR
    \STATE \textbf{return} $\mathsf{Aggregate}(\pred_A, \kappa)$
    \vspace{.2cm}
    \STATE $\rhd$ Returns $\kappa$-majority prediction or abstains
    \STATE \textbf{function $\mathsf{Aggregate}\big(\pred_{1}, \ldots, \pred_{\boot}, \kappa\big)$}
        \STATE \hspace{.5em} $\rhd$ Compute $\hatsc$ (\ref{eq:hatsc})
        \STATE \hspace{.2cm} \textbf{if } $\mathsf{SelfConsistency}(\pred_{1}, \ldots, \pred_{\boot}) \geq \kappa$
            \STATE \hspace{.4cm} \textbf{return} $\argmax_{y'\in\sY} \Big[\sum_{i=1}^\boot \1[y' = \pred_{i}]\Big]$
        \STATE \hspace{.2cm} \textbf{end if}
        \STATE \hspace{.2cm} \textbf{return} \texttt{Abstain}
    \STATE \textbf{end function}
    }
\end{algorithmic}
\end{algorithm}
\setlength{\textfloatsep}{3pt}
  \end{minipage}
  \vspace{-.2cm}
\end{figure}

duce a prediction $\pred \in [0, 1]$ for $\instance$ only if $\instance$ surpasses a user-specified minimum level $\kappa$ of $\hatsc$; 
otherwise, if an instance fails to achieve $\hatsc$ of at least $\kappa$, we \texttt{Abstain} from predicting. 
For evaluation, we divide the test set into two subsets: we group together the instances we \texttt{Abstain} on in an \emph{abstention set} and those we predict on in a \emph{prediction set}. 
This method improves self-consistency through two complementary mechanisms: 1) variance reduction (due to bagging, see Appendix \ref{app:sec:algorithm}) and 2) abstaining from instances that exhibit low $\hatsc$ (thereby raising the overall amount of $\hatsc$ for the prediction set, see Appendix \ref{app:sec:algorithm}).\looseness=-1

Further, since variance is a component of error (Section~\ref{sec:significance}), variance reduction also 
tends to improve accuracy~\citep{breiman1996bagging}. 
This leads to an important observation: the abstention set, by definition, exhibits high variance; we can therefore expect it to exhibit higher error than the prediction set (Section~\ref{sec:empirical}, Appendix~
\ref{app:sec:empirical}). So, while at first glance it may seem odd that our solution for arbitrariness is to \emph{not predict}, it is worth noting that \textbf{we often would have predicted incorrectly on a large portion of the abstention set anyway} (Appendix~
\ref{app:sec:algorithm}). In practice, we test two versions of our method:\looseness=-1

\custompar{Simple ensembling} We run Algorithm~\ref{algo:bagging-confidently} to build ensembles of typical hypothesis classes in algorithmic fairness. For example, running with $\boot=101$ decision trees and $\kappa=0.75$ produces a bagged classifier that contains $101$ underlying decision trees, for which the bagged classifier abstains from predicting on test instances that exhibit less than $0.75$ $\hatsc$. If overall $\hatsc$ is low, then simple ensembling will lead to a large number of abstentions. For example, almost half of all test instances in \texttt{COMPAS} using random forests would fail to surpass the threshold $\kappa=0.75$ (Figure~\ref{subfig:compas-cdf-rfc}). The potential for large abstention sets informs our second approach.\looseness=-1

\custompar{Super ensembling} We run Algorithm~\ref{algo:bagging-confidently} on \emph{bagged} models $\hat{\model}$. When there is 
low $\hatsc$ (i.e., high $\hatvar$) it can be 
beneficial to do an initial pass of variance reduction. We produce bagged classifiers using traditional bagging, but without abstaining (at Algorithm~\ref{algo:bagging-confidently}, lines 4-5); \emph{then} we $\mathsf{Aggregate}$ using those bagged classifiers as the underlying models $\hat{\model}$. The first round of bagging raises the overall $\hatsc$ before the second round, which is when we decide whether to \texttt{Abstain} or not. We therefore expect this approach to abstain less; however, it may potentially incur higher error, if, by happenstance, simple-majority-vote bagging chooses $\hat{y}\neq o$ for instances with very low $\hatsc$ (Appendix~
\ref{app:sec:algorithm}).\footnote{We could recursively super ensemble, but do not in this work.\looseness=-1}
We also experiment with an $\mathsf{Aggregate}$ rule that averages the output probabilities of the underlying regressors $\regressork$, and then applies threshold $\tau$ to produce ensembled predictions. We do not observe major differences in results.

\section{Experiments}\label{sec:empirical}
We release an extensible package  of different $\mathsf{Aggregate}$ methods, with which we trained and compared several million different models (all told, taking on the order of $10$ hours of compute).
We include results covering common 
datasets and models: \texttt{COMPAS}, \texttt{Old Adult}, \texttt{German} and \texttt{Taiwan Credit}, and 3 large-scale \texttt{New Adult - CA} tasks on logistic regression (LR), decision trees (DTs), random forests (RFs), 
MLPs, and SVMs (Appendix~
\ref{app:sec:empirical}). 
\textbf{Our results are shocking: 
by using Algorithm~\ref{algo:bagging-confidently}, we happened to observe close-to-fairness in nearly every task. Mitigating arbitrariness leads to fairness, \emph{\underline{without}} applying common fairness-improving interventions} (Section~\ref{sec:empirical-repro}, Appendix~
\ref{app:sec:empirical}). 

\setlength{\tabcolsep}{6pt}
\begin{figure*}[t!]
\vspace{-.5cm}
\begin{minipage}{.495\linewidth}
\centering
\hspace{-.4cm}
        \includegraphics[width=.85\linewidth]{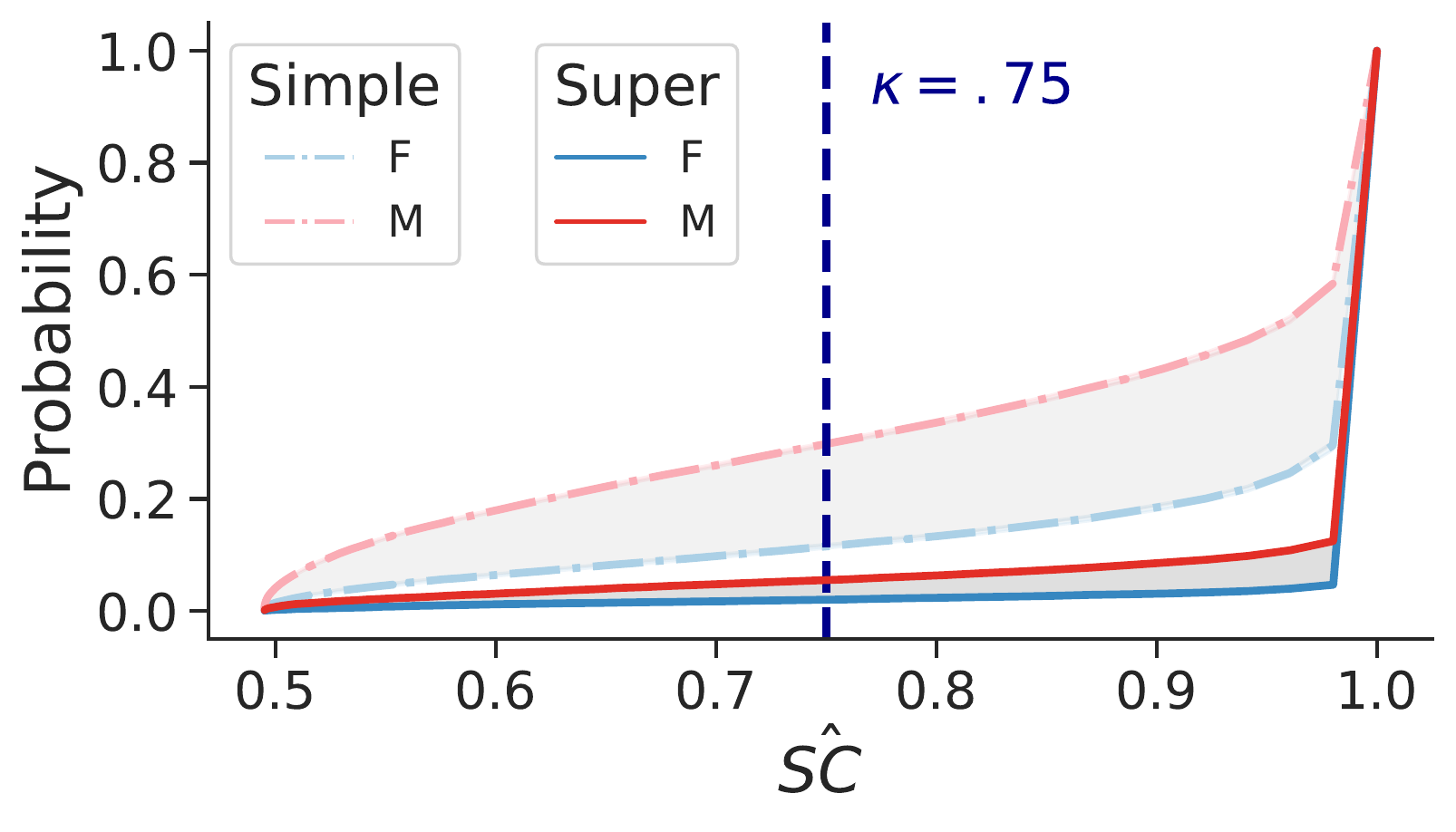}\\\vspace{-.1cm}
        \begin{tabular}{lccc}
        \toprule
            & \textbf{Baseline} & \textbf{Simple} 
            & \textbf{Super}
            \\ \midrule
            \textbf{$\Delta\hatfnr$} & $\;\;6.3\pm.3\%$ & $4.1\pm.3\%$ 
            & $\;\;5.8\pm.4\%$ 
            \\ \midrule
             \textbf{$\hatfnr_\text{F}$} & $\;\;5.3\pm.3\%$ & $3.5\pm.1\%$ 
             & $\;\;4.9\pm.2\%$ 
             \\ \midrule 
            \textbf{$\hatfnr_\text{M}$} & $11.6\pm.1\%$ & $7.6\pm.3\%$  
            & $10.7\pm.3\%$ 
            \\ 
              \bottomrule
        \end{tabular}
        \subcaption{\texttt{Old Adult} split by $\texttt{sex}$}
        \label{subfig:adult-ens}
\end{minipage}%
\hspace{.25cm}
\begin{minipage}{.495\linewidth}
\centering
\hspace{-.2cm}
        \includegraphics[width=.85\linewidth]{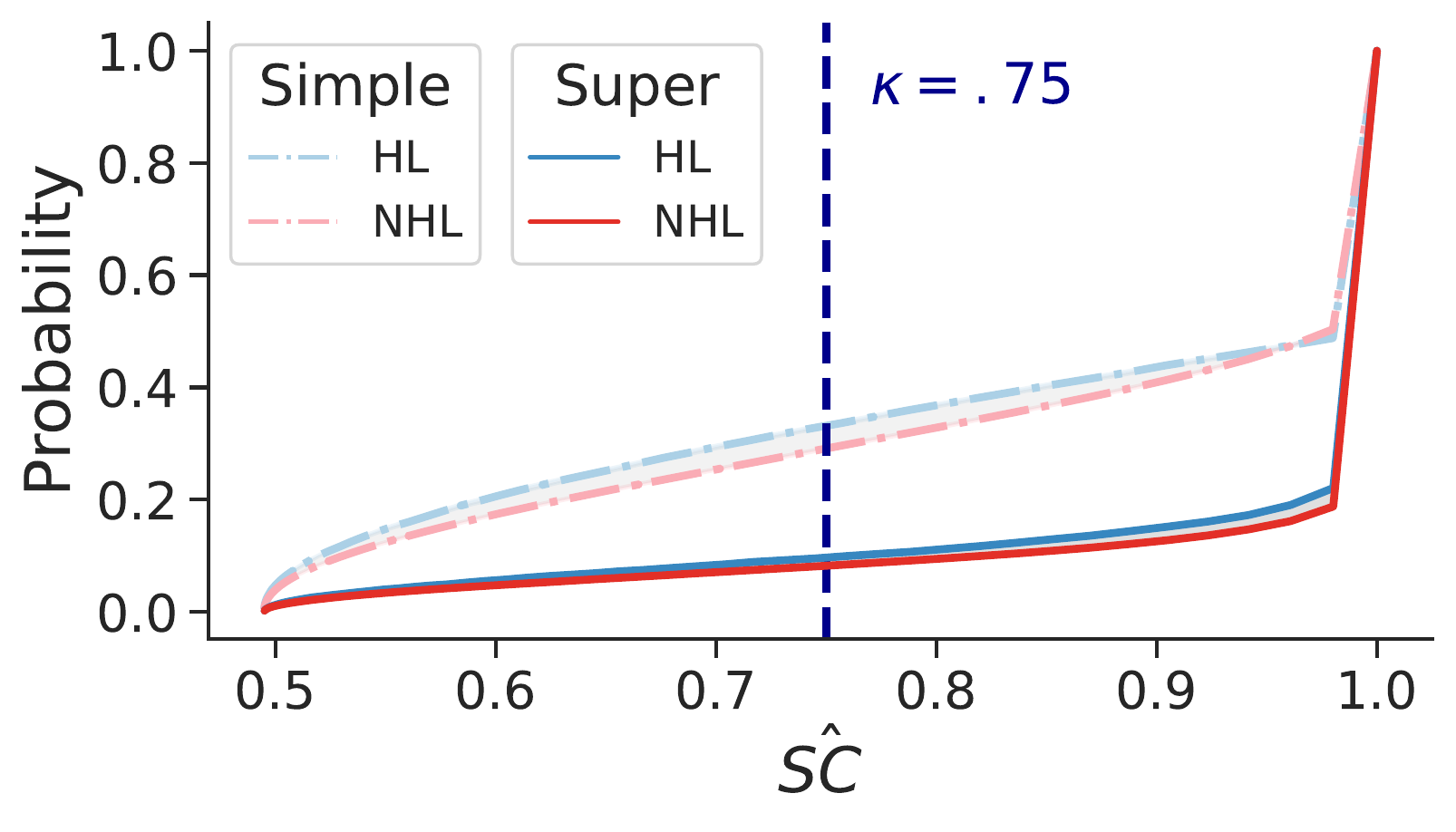}\\\vspace{-.1cm}
        \begin{tabular}{lccc}
        \toprule
            & \textbf{Baseline} & \textbf{Simple} 
            & \textbf{Super}
            \\ \midrule
            \textbf{$\Delta\hatfnr$} & $\;\;0.7\pm.2\%$ & $1.1\pm.3\%$ 
            & $2.2\pm.3\%$ 
            \\ \midrule
             \textbf{$\hatfnr_\text{HL}$} & $10.1\pm.2\%$ & $3.3\pm.3\%$ 
             & $8.0\pm.3\%$ 
             \\ \midrule 
            \textbf{$\hatfnr_\text{NHL}$} & $\;\;9.4\pm.1\%$ & $2.2\pm.1\%$  
            & $5.8\pm.1\%$ 
            \\ 
              \bottomrule
        \end{tabular}
        \subcaption{\texttt{HMDA-NY-2017} split by $\texttt{ethnicity}$}
        \label{subfig:hmda-ens}
\end{minipage}%
\caption{Algorithm~\ref{algo:bagging-confidently}: \textbf{simple} and \textbf{super ensembling} random forests (RFs) for \texttt{Old Adult} (\ref{subfig:adult-ens}) and 
\texttt{HMDA-NY-2017} (\ref{subfig:hmda-ens}). Tables 
show \textbf{$\hatfnr$} 
(mean $\pm$ STD) for individual models (\textbf{Baseline}) and each ensembling method's prediction set; 
$\boot=101$, 10 train/test splits (Appendix~
\ref{app:sec:empirical}). To highlight systematic arbitrariness (Section~\ref{sec:var:arbitrary}), we shade in gray the area between group-specific $\hatsc$ CDFs for each 
method. An initial pass of variance reduction in \textbf{super} significantly decreases the systematic arbitrariness in \texttt{Old Adult}.\looseness=-1} 
\label{fig:adult-hmda-ens}
\vspace{-.2cm}
\end{figure*}

\custompar{Releasing an \texttt{HMDA} toolkit} A possible explanation is that most fairness benchmarks are small ($<25,000$ examples) and therefore exhibit high variance. 
We therefore clean a larger, more diverse, and newer dataset for investigating fair binary classification  --- the Home Mortgage  Disclosure Act (\texttt{HMDA}) 2007-2017 
datasets~\cite{ffiec2022housingdata} --- and release them with a standalone, easy-to-use software package.\footnote{It is repeatedly argued that the field needs such datasets~\citep[e.g.]{ding2021adult}. \texttt{HMDA} meets this need, but is less commonly used. It requires engineering effort to manipulate --- a barrier we remove.\looseness=-1} 
In this paper, we examine the \texttt{NY} and \texttt{TX 2017} subsets of \texttt{HMDA}, which have $244,107$ and $576,978$ examples, respectively, and \textbf{we still find close-to-fairness} (Section~\ref{sec:empirical-algo}, Appendix~
\ref{app:sec:empirical}).\looseness=-1 

\custompar{Presentation} To visualize Algorithm~\ref{algo:bagging-confidently}, we plot the CDFs of the $\hatsc$ of the underlying models used in each ensembling method. We simultaneously plot the results of \textbf{simple ensembling} (dotted curves) and \textbf{super ensembling} (solid curves). Instances to the left of the vertical line (the minimum $\hatsc$ threshold $\kappa$) form the abstention set. 
We also provide corresponding mean $\pm$ STD fairness and accuracy metrics for individual models (our expected, but not-necessarily-practically-attainable \textbf{baseline}) and for both \textbf{simple} and \textbf{super} ensembling. 
For ensembling methods, we report these metrics on the prediction set, along with the \textbf{abstention rate ($\hatar$)}.\looseness=-1 

We necessarily defer most of our results to Appendix~\ref{app:sec:empirical}. 
In the main text, we exemplify two overarching themes: 
the effectiveness of both ensembling variants (Section~\ref{sec:empirical-algo}), and 
how our results reveal shocking insights about reliability in fair binary classification research (Section~\ref{sec:empirical-repro}). For all experiments, we illustrate Algorithm~\ref{algo:bagging-confidently} with $\kappa=0.75$, but note that $\kappa$ is task-dependent in practice. 

\subsection{Validating Algorithm~\ref{algo:bagging-confidently}}\label{sec:empirical-algo}

\begin{figure}[t!]
\centering
    \begin{subfigure}{.49\linewidth}
    \centering
        \includegraphics[width=.9\linewidth]{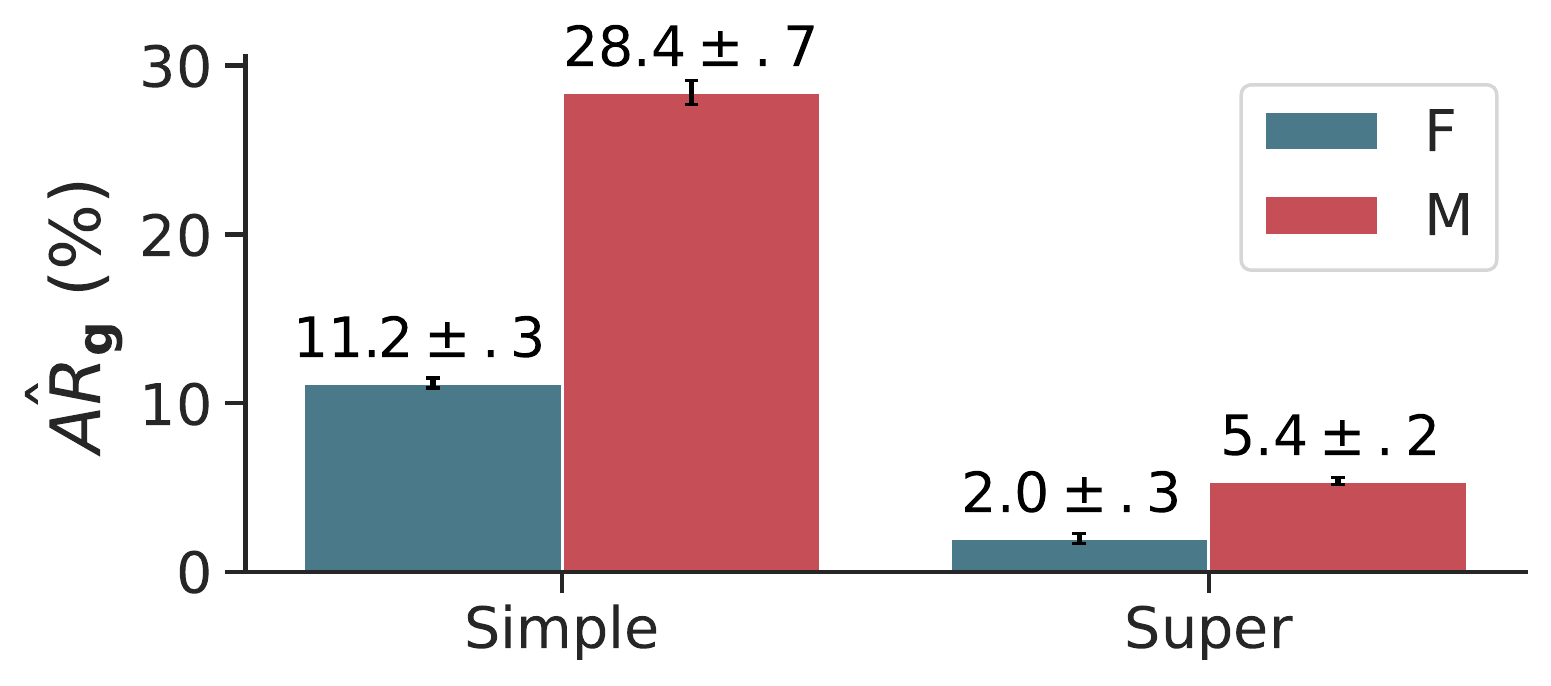}
        \caption{\texttt{Old Adult}, $\group=\texttt{sex}$}
        \label{fig:abstention-adult}
    \end{subfigure}
    \begin{subfigure}{.49\linewidth}
    \centering
        \includegraphics[width=.9\linewidth]{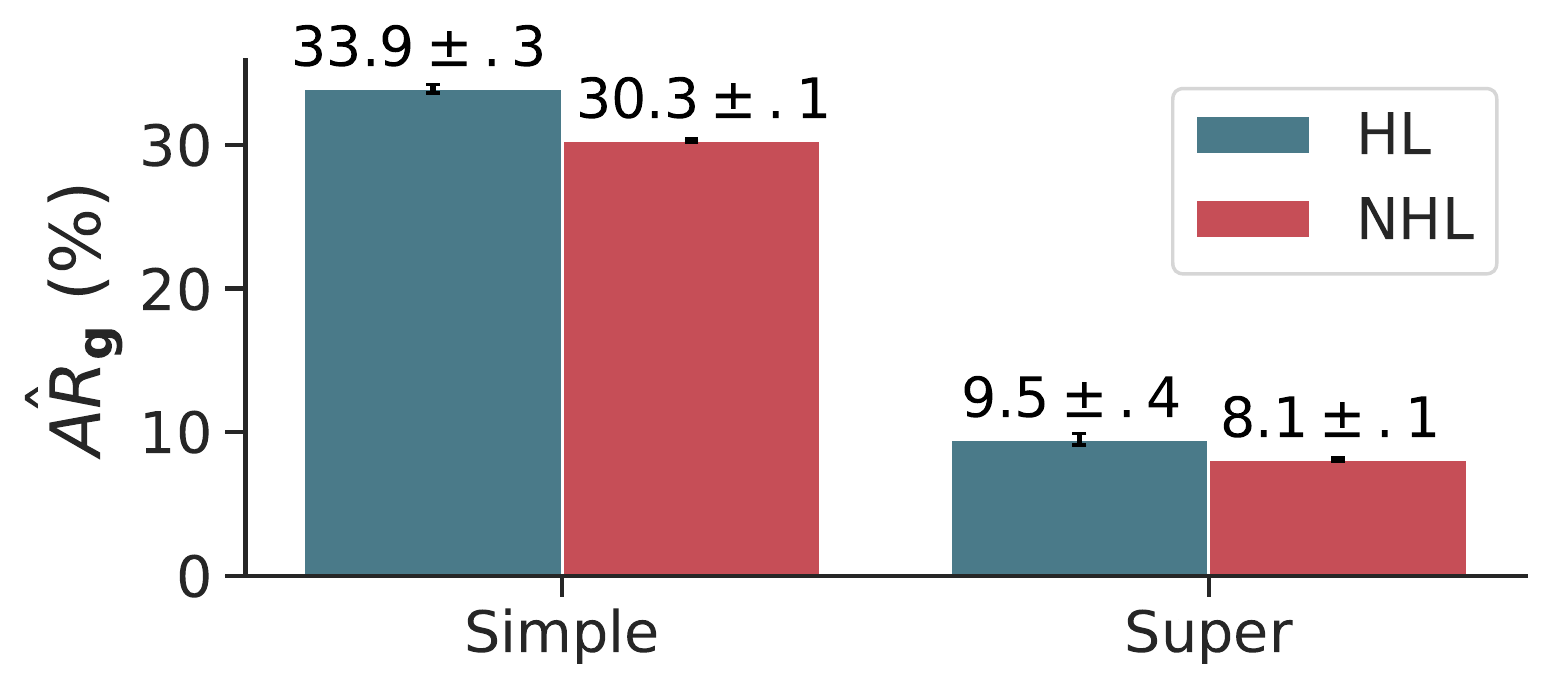}
        \caption{\texttt{HMDA-NY-2017}, $\group=\texttt{ethnicity}$}
    \end{subfigure}
\caption{Group-specific abstention rates \textbf{$\hatar_\group$} for each algorithm. 
\textbf{Super ensembling} abstains less overall, and more equally than \textbf{simple ensembling}. \texttt{HMDA-NY-2017}, which exhibits less systematic arbitrariness than \texttt{Old Adult}(Figure~\ref{fig:adult-hmda-ens}), exhibits roughly equal abstention rates across subgroups.} 
\label{fig:abstention}
\end{figure}

We highlight results for two  illustrative examples: \texttt{Old Adult} 
and \texttt{HMDA\--NY\--2017} for \texttt{ethnicity} (Hispanic or Latino (HL), Non\--Hispanic or Latino (NHL)). We plot $\hatsc$ CDFs and show $\hatfnr$ metrics 
using random forests (RFs). For \texttt{Old Adult}, the expected  disparity of the RF baseline is $\Delta\hatfnr=6.3\%$. The dashed set of curves plots the underlying $\hatsc$ for these RFs (Figure~\ref{subfig:adult-ens}). When we apply \textbf{simple} to these RFs, 
overall $\haterr$ decreases (Appendix~
\ref{app:sec:empirical}), 
shown in part by the decrease in $\hatfnr_\text{F}$ and  $\hatfnr_\text{M}$. Fairness also improves:  $\Delta\hatfnr$ decreases to $4.1\%$. However, the corresponding $\hatar$ is quite high, especially for the \texttt{Male} subgroup ($\group=\text{M}$, Figure~\ref{fig:abstention-adult}).

As expected, \textbf{super} improves overall $\hatsc$ through a first pass of variance reduction (Section~\ref{sec:algorithms}).  The $\hatsc$ CDF curves are brought down, indicating a lower proportion of the test set exhibits low $\hatsc$. Abstention rate $\hatar$ is lower and more equal (Figure~\ref{fig:abstention-adult}); however, error, while still lower than the baseline RFs, has gone up for all metrics. There is also a decrease in systematic arbitrariness (Section~\ref{sec:var:arbitrary}): the dark gray area for \textbf{super} ($\hat{\mathcal{W}_1}=.014$) is smaller than the light gray area for \textbf{simple} ($\hat{\mathcal{W}_1}=.063$) (\ref{app:sec:consistency}, \ref{app:sec:experiments-algo}).\looseness=-1

For \texttt{HMDA} (Figure~\ref{subfig:hmda-ens}), \textbf{simple} similarly improves $\hatfnr$, but has a less beneficial effect on fairness ($\Delta\hatfnr$). \textbf{However, note that since the baseline is the empirical expected error over thousands of RF models, the specific $\Delta\hatfnr$ is not necessarily attainable by any individual model}. In this respect, \textbf{simple} has the benefit of actually obtaining a specific (ensemble) model that yields this disparity reliably in practice: $\Delta\hatfnr=1.1\%$ is the mean over $10$ simple ensembles. Notably, this is extremely low, even without applying traditional fairness techniques. Similar to \texttt{Old Adult}, \textbf{simple} exhibits high $\hatar$, which decreases with \textbf{super} at the cost of higher error. $\hatfnr$ still improves for both $\group$ in comparison to the baseline, but the benefits are unequally applied: $\hatfnr_\text{W}$ has a larger benefit, so $\Delta\hatfnr$ increases slightly.

\custompar{Abstention set error} As an example, the average 
$\haterr$ in the \texttt{Old Adult} \textbf{simple} abstention set is close to $40\%$ --- compared to $17\%$ for the RF baseline, and $8\%$ for \textbf{simple} and $14\%$ for \textbf{super} prediction sets (Appendix~\ref{app:sec:adult-algo}). As expected, beyond reducing arbitrariness, we abstain from predicting for many instances for which we also would have been more inaccurate (Section~\ref{sec:algorithms}).\looseness=-1 

\custompar{A trade-off} 
Our results support that there is indeed a trade-off between abstention rate and error 
(Section~\ref{sec:algorithms}). This is because Algorithm~\ref{algo:bagging-confidently} identifies low-$\hatsc$ instances for which ML prediction does a poor job, and abstains from predicting on them. Nevertheless, it may be infeasible for some applications to tolerate a high $\hatar$. Thus the choice of $\kappa$ and ensembling method should be considered a 
context-dependent decision.\looseness=-1 

\custompar{Unequal abstention rates} When there is a high degree of systematic arbitrariness, $\hatar$ can vary a lot by $\group$ (Figure~\ref{fig:abstention}). With respect to improving $\hatsc$,  error, and fairness, this may be a reasonable outcome: it is arguably better to abstain unevenly --- deferring a final classification to non-ML decision processes --- than to predict more inaccurately and arbitrarily for one group. More importantly, we rarely observe systematic arbitrariness; 
unequal $\hatar$ is uncommon in practice (Section~\ref{sec:related}).\looseness=-1

\subsection{A problem of empiriclal algorithmic fairness}\label{sec:empirical-repro}

\begin{figure}[t]
    \vspace{-.2cm}
    \centering
    \hspace{-1.75cm}
  \begin{minipage}{0.55\linewidth}
    \centering
    \includegraphics[width=.85\linewidth]{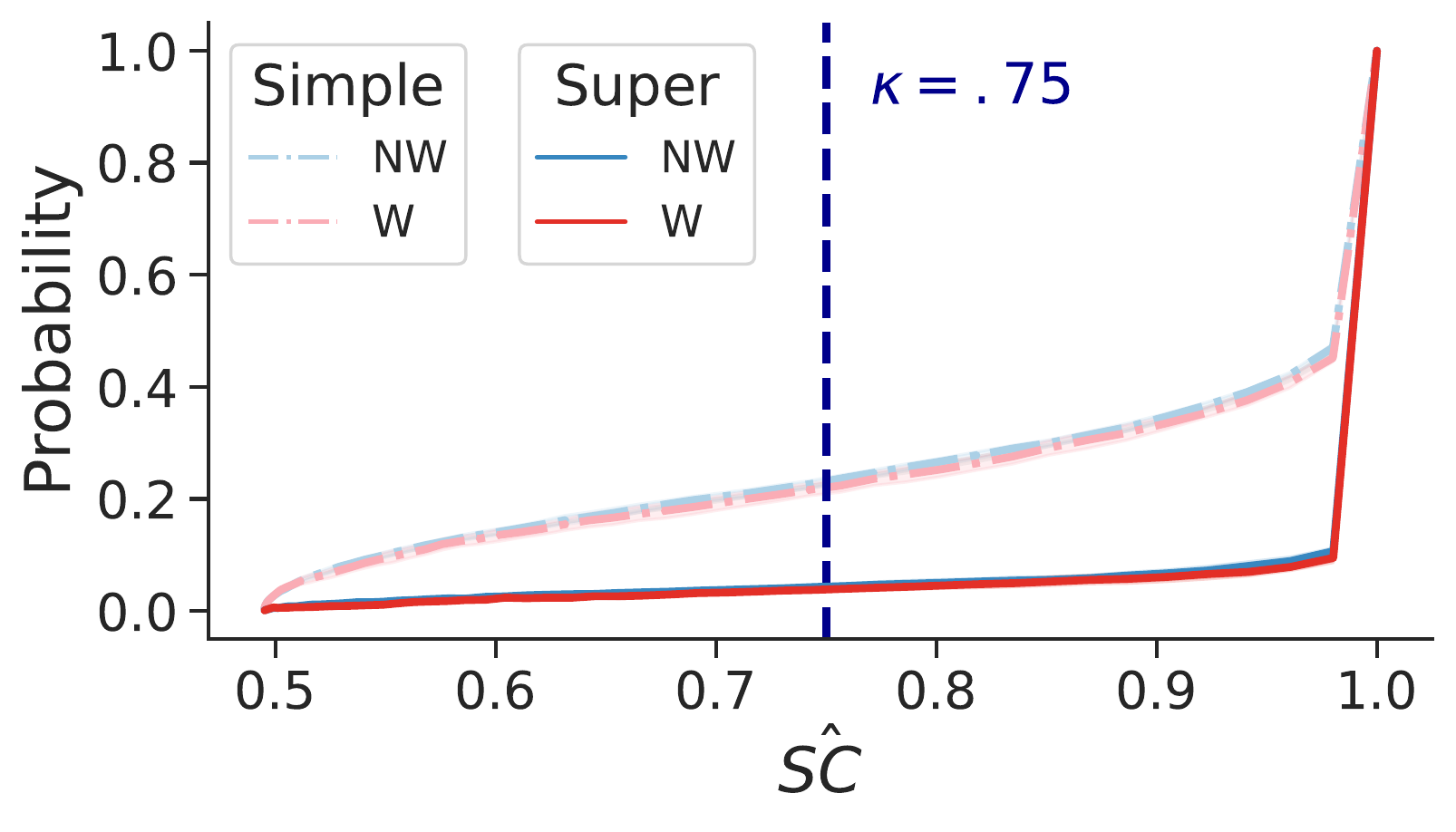}
  \end{minipage}
  \begin{minipage}{0.42\linewidth}
    \footnotesize
    \setlength{\tabcolsep}{6pt}
    \begin{tabular}{lccc}
        \toprule
            & \textbf{Baseline} & \textbf{Simple} 
            & \textbf{Super} 
            \\ \midrule
            \textbf{$\Delta\hatfpr$} & $\;\;2.1\pm1.8\%$ & $\;\;3.0\pm1.4\%$ 
            & $\;\;\;1.8\pm1.0\%$ 
            \\ \midrule
             \textbf{$\hatfpr_\text{NW}$} & $14.7\pm1.3\%$ & $11.4\pm1.0\%$ 
             & $12.9\pm.8\%$ 
             \\ \midrule 
            \textbf{$\hatfpr_\text{W}$} & $12.6\pm1.3\%$ & $\;\;8.4\pm1.0\%$ 
            & $11.1\pm.6\%$ 
            \\ 
              \bottomrule
    \end{tabular} 
  \end{minipage}
  \caption{Algorithm~\ref{algo:bagging-confidently}: \textbf{simple} and \textbf{super ensembling} 
  logistic regression on \texttt{COMPAS}. $\boot=101$, 10 train/test splits. Table shows mean \textbf{$\hatfpr$} $\pm$ STD for individual models (\textbf{Baseline}) and each ensembling method's prediction set. The $\hatsc$ CDFs are effectively identical across $\group$. 
  \looseness=-1}
    \label{fig:compas-ens}
\end{figure}

We also highlight results for \texttt{COMPAS}, 1 of the 3 most common fairness datasets~\citep{fabris2022datasets}. 
Algorithm~\ref{algo:bagging-confidently} is similarly very effective at reducing arbitrariness (Figure~\ref{fig:compas-ens}), and is able to obtain state-of-the-art accuracy~\cite{lin2020compas} with $\Delta\hatfpr$ between $1.8-3\%$. Analogous results for \texttt{German Credit} indicate statistical equivalence in fairness metrics (Appendices~
\ref{app:sec:german-algo} and~\ref{app:sec:algo-discussion}).\looseness=-1

These low-single-digit 
disparities do not cohere with much of the literature on fair binary classification, which often reports much larger fairness violations~\citep[notably]{larson2016propublica}. However, most work on fair classification examines individual models, selected via cross-validation with a handful of random seeds (Section~\ref{sec:prelim}). Our results suggest that selecting between a few individual models in fair binary classification experiments is unreliable. 
\textbf{When we instead estimate expected error by ensembling, we have difficulty reproducing unfairness in practice.} 
Variance in the underlying models in $\hat{\possiblemodels}$ seems to be the culprit. The individual models we train on these tasks exhibit radically different group-specific error rates (Appendix~\ref{app:sec:algo-discussion}). 
Our strategy of shifting focus to the overall behavior of the distribution $\hat{\possiblemodels}$ provides a solution: 
we not only mitigate arbitrariness, we \textbf{also improve accuracy \emph{and} usually average away most underlying, individual-model unfairness} 
(Appendix~
\ref{app:sec:fair}).\looseness=-1  

\section{Discussion and Related Work}\label{sec:related}

In this paper, we advocate for a shift in thinking about \emph{individual} models to the \emph{distribution over possible models} in fair binary classification. 
This shift surfaces 
arbitrariness in underlying model decisions. We suggest a metric of \emph{self-consistency} as a proxy for arbitrariness (Section~\ref{sec:significance}) and an intuitive, elegantly simple extension of the classic bagging algorithm to mitigate it (Section~\ref{sec:algorithms}). 
Our approach is tremendously effective with respect to improving $\hatsc$, accuracy, and fairness metrics in practice (Section~\ref{sec:empirical}, Appendix~
\ref{app:sec:empirical}).

Our findings contradict accepted truths in algorithmic fairness. For example, much work posits that there is an inherent analytical trade-off between fairness and accuracy~\citep{corbettdavies2017cost, menon2018cost}.
Instead, our experiments complement prior work that disputes the practical relevance of this formulation~\citep{rodolfa2021tradeoff}. We show it is in fact 
typically possible to achieve accuracy (via variance reduction) 
and close-to-fairness --- and to do so \textbf{without} using fairness-focused interventions.\looseness=-1

Other research also calls attention to the need for metrics 
beyond fairness and accuracy. Model multiplicity reasons about sets of models that have similar accuracy~\citep{breiman2001multiplicity}, but differ in underlying properties due to variance in decision rules~\citep{black2022multiplicity, marx2020mm, watson2023multiplicity}. 
This work emphasizes developing criteria for 
selecting an \emph{individual} model from that set.   
Instead, our work uses the  \emph{distribution over possible models} (with no normative claims about model accuracy 
or other selection criteria) to reason about arbitrariness (Appendix~\ref{app:sec:mm}). 
Some related work considers the role of uncertainty and variance in fairness~\citep{ali2021uncertainty, chen2018tradeoff, khan2023fairness, bhatt2021uncertainty}.   
Notably, \citet{black2022selective} concurrently investigates abstention-based ensembling, 
employing a strategy that (based on their choice of variance definition) 
ultimately does not 
address the arbitrariness we describe and mitigate 
(Appendix~
\ref{app:sec:othervariance}).  
After our work, \citet{ko2023fairensemble} 
built on prior work that studies fairness and variance in deep learning tasks~\citep{forde2021disparate, qian2021variance}, and 
find that fairness emerges in deep ensembles (Appendix~\ref{app:sec:concurrent}).\looseness=-1 

Most importantly, we take a comprehensive experimental approach missing from prior work.
It is this approach that uncovers our 
alarming results: 
\textbf{almost all tasks and settings demonstrate close-to or complete statistical equality in fairness metrics, after accounting for arbitrariness} (Appendix~
\ref{app:sec:experiments-algo}). \texttt{Old Adult} (Figure~\ref{subfig:adult-ens}) is one of two exceptions. 
These results hold  
for larger, newer datasets like  
\texttt{HMDA}, which we clean and release. 
Altogether, our findings indicate that \textbf{variance is undermining the reliability of conclusions in fair binary classification experiments}. It is worth revisiting all prior  experiments that depend on cross validation or few models.\looseness=-1 

\vspace{-.3cm}
\subsection*{What does this mean for fairness research?}
\vspace{-.2cm}

While the field has put forth numerous theoretical results about (un)fairness regarding single models --- impossibility of satisfying multiple metrics~\citep{kleinberg2017impossibility}, post-processing individual models to achieve a particular metric~\citep{hardt2016eo} --- these results seem to miss the point. By examining individual models, arbitrariness remains latent; when we account for arbitrariness in practice, most measurements of unfairness vanish. 

We are not suggesting that there are no reasons to be concerned with the fairness of machine-learning models. We are not challenging the idea that actual, reliable violations of standard fairness metrics should be of concern. 
Instead, we are suggesting that common formalisms and methods for measuring fairness can lead to false conclusions about the degree to which such violations are happening in practice (Appendix~\ref{app:sec:future}). Worse, they can conceal a tremendous amount of arbitrariness, which should itself 
be an important concern 
when examining the social impact of automated decision-making.

\section*{Ethical Statement}

This work raises important ethical concerns regarding the practice of fair-binary-classification research. We organize these concerns into several themes below.

\subsection*{Arbitrariness and legitimacy} 
On common research benchmarks, we show that many classification decisions are effectively arbitrary. 
Intuitively, this is unfair, but is a type of unfairness that largely has gone unnoticed in the algorithmic-fairness community. 
Such arbitrariness raises serious concerns about the legitimacy of automated decision-making. 
Fully examining these implications is the subject of current work that our team is completing. 
Complementing prior work on ML and arbitrariness~\citep{creel2022leviathan, cooper2022lawless}, we are working on a law-review piece that clarifies the due process implications of arbitrariness in ML-decision outcomes. 
For additional notes on future work in this area, see Appendix~\ref{app:sec:future}.

\subsection*{Misspecification, mismeasurement, and fairness} 

Much prior work has emphasized theoretical contributions and problem formulations for how to study fairness in ML. 
A common pattern is to study unequal model error rates between demographic subgroups in the available data. 
Typically, experimental validation of these ideas has relied on using just a handful of models. 
Our work shows that this is not empirically sound: 
it can lead to drawing unreliable conclusions about the degree of unfairness (defined in terms of error rates). 
Most observable unfairness seems due to inadequately modeling or measuring the role of variance in learned models on common benchmark tasks.\looseness=-1 

Other than indicating serious concerns about the rigor of experiments in fairness research, our findings suggest  ethical issues about the role of mismeasurement in identifying and allocating resources to specific research problems~\citep{jacobs2021measurement}. 
A lot of resources and research effort have been allocated to the study of these problem formulations. 
In turn, they have had profound social influence and impact, both in research and in the real world, with respect to how we reason broadly about fairness in automated decision-making. 

In response to the heavy investment in these ideas, many have noted that there are normative and ethical reasons why such formulations are inadequate for the task of aligning with more just or equitable outcomes in practice. 
Our work shows that normative and ethical considerations extend even further. 
\textbf{Even if we take these formulations at face value in theory, they are very difficult to replicate in practice on common fairness benchmarks when we account for variance in predictions across trained models.} 
When we perform due diligence with our experiments, we have trouble validating the hypothesis that popular ML-theoretical formulations of fairness are capturing a meaningful practical phenomenon. 

This should be an incredibly alarming finding to anyone in the community that is concerned about the practice, not just the theory, of fairness research. 
Using bootstrapping, we observe serious problems with respect to the reliability of how fairness and accuracy are measured in work that relies on cross-validation of just a few models. 
We also find little empirical evidence of a trade-off between fairness and accuracy (another common formulation in the community), which complements prior work that has made similar observations~\citep{rodolfa2021tradeoff}.\looseness=-1

\subsection*{Project aims, reduction of scope} 

We emphasize that this was an unintended outcome of our original research objectives. 
We aimed to study arbitrariness as a latent issue in problem formulations that have to do with fair classification. 
This included broader methodological aims: studying many sources of non-determinism that could impact arbitrariness in practice (e.g., minibatching, example ordering). 
Instead, our initial results of close-to-fair expected performance for individual models made us refocus our work on issues of mismeasurement and fairness. 

We did not expect to find that dealing with arbitrariness would make almost all unfairness (again, as measured by common definitions) vanish in practice. 
Regardless of our intention, these results indicate the limits of theory in a domain that, by centering social values like fairness, cannot be separated from practice. 
We believe that problem formulations are only as good as they are useful.
Based on our work, it is unclear how useful our existing formulations are if they do not bear out in experiments.\looseness=-1

\subsection*{Reproducibility and project aims} 
In the course of this study, we also tried to reproduce the experiments of many well-cited theory-focused works. 
We almost always could not do so: code was almost always unavailable. 
We therefore pivoted from making reproducibility an explicit aspect of the present paper, which we will pursue in future work that focuses solely on reproducibility and fairness. 
Nevertheless, our work raises concerns about the validity of some of this past work. 
At the very least, past work that makes claims about baseline unfairness in fairness benchmarks (in order to demonstrate that proposed methods improve upon these baselines) should be subject to experimental scrutiny. 

Further along these lines, \textbf{in our opinion, this project should not have been possible or necessary in 2022}. 
We believe that the novel findings we present here should have surfaced long ago, and likely would have surfaced if experimental contributions had been more evenly balanced with theoretical work. 

\subsection*{The limits of prediction} 

Lastly, it has not escaped our notice that  our results signal severe limits to prediction in social settings. 
It is true that our method performs reasonably well with respect to both fairness and accuracy metrics; however, arbitrariness is such a rampant problem, it is arguably unreasonable to assign these metrics much value in practice.

\section*{Acknowledgments}
This work was done as part of an internship at Microsoft Research. 
A. Feder Cooper is supported by Prof. Christopher De Sa's NSF CAREER grant, Prof. Baobao Zhang, and Prof. James Grimmelmann. 
A. Feder Cooper is affiliated with The Berkman Klein Center for Internet \& Society at Harvard University.
The authors would like to thank The Internet Society Project at Yale Law School, Artificial Intelligence Policy and Practice at Cornell University, Jack Balkin, Emily Black, danah boyd, Sarah Dean, Fernando Delgado, Hoda Heidari, Ken Holstein, Jessica Hullman, Abigail Z. Jacobs, Meg Leta Jones, Michael Littman, Kweku Kwegyir-Aggrey, Rosanne Liu, Emanuel Moss, Kathy Strandburg, Hanna Wallach, and Simone Zhang for their feedback.

\bibliography{references}
\bibliographystyle{plainnat}

\newpage
\appendix
\onecolumn
\newcommand{\appprelim}{Extended Preliminaries}
\newcommand{\appprelimnotation}{Notes on notation and on our choice of terminology}
\newcommand{\appsetuplimitations}{Constraints on our setup}
\newcommand{\appprelimcosts}{Costs and the classification decision threshold}
\newcommand{\appprelimbootstrap}{The bootstrap method}

\newcommand{\appvariance}{Additional Details on Variance and Self-Consistency}
\newcommand{\appprelimenoisebias}{Other statistical sources of error}
\newcommand{\appprelimvariance}{Our variance definition}
\newcommand{\appconsistencyderiving}{Deriving self-consistency from variance}
\newcommand{\appconsistencycost}{Cost-independence of self-consistency}
\newcommand{\appconsistencydetails}{Additional details on our choice of self-consistency metric}

\newcommand{\appothervariance}{Related Work and Alternative Notions of Variance}
\newcommand{\appmain}{Defining variance in relation to a ``main prediction''}

\newcommand{\appprelimcomparison}{Why we choose to avoid computing the main prediction}
\newcommand{\appmaincost}{The main prediction and cost-sensitive loss}

\newcommand{\appmodelm}{Putting our work in conversation with research on model multiplicity}
\newcommand{\appconcurrent}{Concurrent work}

\newcommand{\appalgorithm}{Additional Details on Our Algorithmic Framework}
\newcommand{\appsecconfidence}{Self-consistent ensembling with abstention}

\newcommand{\appexperiments}{Additional Experimental Results and Details for Reproducibility}

\newcommand{\appexperimentsdata}{Hypothesis classes, datasets, and code}
\newcommand{\apphmda}{The standalone \texttt{HMDA} tookit}
\newcommand{\appcluster}{Cluster environment details}
\newcommand{\appillustrativedetails}{Details on motivating examples in the main paper}
\newcommand{\appadditionalillustrative}{Additional illustrative results}
\newcommand{\appcompasadult}{\texttt{COMPAS} and \texttt{Old Adult}}
\newcommand{\appgerman}{\texttt{South German Credit}}
\newcommand{\apptaiwan}{\texttt{Taiwan Credit}}
\newcommand{\appnewadult}{\texttt{New Adult - CA}}
\newcommand{\apphmdaillustrative}{\texttt{HMDA}}
\newcommand{\appillustrativediscussion}{Extended discussion of illustrative examples of self-consistency}
\newcommand{\appexperimentsalgo}{Validating our algorithm in practice}
\newcommand{\appcompas}{\texttt{COMPAS}}
\newcommand{\appoldadult}{\texttt{Old Adult}}
\newcommand{\appwasserstein}{Measuring the distance between empirical self-consistency curves}
\newcommand{\appalgodiscussion}{Discussion of extended results for Algorithm~\ref{algo:bagging-confidently}}
\newcommand{\appfair}{Reliability and fairness metrics in \texttt{COMPAS} and \texttt{South German Credit}}

\newcommand{\appfuture}{Brief notes on future research}

\newpage
\onecolumn
\appendix

\section*{Appendix Overview}

The Appendix goes into significantly more detail than the main paper. It is organized as follows:\looseness=-1

\vspace{.5cm}

\noindent\textbf{Appendix~\ref{app:sec:prelim}: \appprelim} 
\begin{itemize}[topsep=3.5pt, itemsep=0pt, leftmargin=.5cm]
	\item \ref{app:sec:prelim:note}: \appprelimnotation
	\item \ref{app:sec:limitations}: \appsetuplimitations
	\item \ref{app:sec:prelim:costs}: \appprelimcosts
	\item \ref{app:sec:prelim:boot}: \appprelimbootstrap
\end{itemize}

\noindent\textbf{Appendix~\ref{app:sec:variance}: \appvariance}
\begin{itemize}[topsep=3.5pt, itemsep=0pt, leftmargin=.5cm]
	\item \ref{app:sec:noisebias}: \appprelimenoisebias
	\item \ref{app:sec:ourvariance}: \appprelimvariance
	\item \ref{app:sec:consistency}: \appconsistencyderiving
	\begin{itemize}[topsep=3.5pt, itemsep=0pt, leftmargin=.5cm]
		\item \ref{app:sec:consistency:details}: 
		\appconsistencydetails
	\end{itemize}
\end{itemize}

\noindent\textbf{Appendix~\ref{app:sec:othervariance}: \appothervariance}
\begin{itemize}[topsep=3.5pt, itemsep=0pt, leftmargin=.5cm]
	\item \ref{app:sec:other:main}: \appmain
	\item \ref{app:sec:comparison}: \appprelimcomparison
	\begin{itemize}[topsep=3.5pt, itemsep=0pt, leftmargin=.5cm]
		\item \ref{app:sec:maincost}: 
		\appmaincost
	\end{itemize}
	\item \ref{app:sec:mm}: \appmodelm
	\item \ref{app:sec:concurrent}: \appconcurrent
\end{itemize}

\noindent\textbf{Appendix~\ref{app:sec:algorithm}: \appalgorithm}
\begin{itemize}[topsep=3.5pt, itemsep=0pt, leftmargin=.5cm]
	\item \ref{app:algo:sc}: \appsecconfidence
\end{itemize}

\noindent\textbf{Appendix~\ref{app:sec:empirical}: \appexperiments}
\begin{itemize}[topsep=3.5pt, itemsep=0pt, leftmargin=.5cm]
	\item \ref{app:sec:experimentsdata}: \appexperimentsdata
	\begin{itemize}[topsep=3.5pt, itemsep=0pt, leftmargin=.5cm]
		\item \ref{app:sec:hmda}: \apphmda
	\end{itemize}
	\item \ref{app:sec:cluster}: \appcluster
	\item \ref{app:sec:illustrativedetails}: \appillustrativedetails
	\item \ref{app:sec:experiments-algo}: \appexperimentsalgo
	\begin{itemize}[topsep=3.5pt, itemsep=0pt, leftmargin=.5cm]
		\item \ref{app:sec:compas-algo}:
		\appcompas
		\item \ref{app:sec:adult-algo}: 
		\appoldadult
		\item \ref{app:sec:german-algo}: 
		\appgerman
		\item \ref{app:sec:taiwan-algo}: 
		\apptaiwan
		\item \ref{app:sec:newadult-algo}: 
		\appnewadult
		\item \ref{app:sec:hmda-algo}: 
		\apphmdaillustrative
		\item \ref{app:sec:algo-discussion}: 
		\appalgodiscussion
	\end{itemize}
	\item \ref{app:sec:fair}: \appfair
\end{itemize}
\noindent\textbf{Appendix~\ref{app:sec:future}: \appfuture}

\newpage
\section{\appprelim}\label{app:sec:prelim}

\subsection{\appprelimnotation}\label{app:sec:prelim:note}

\custompar{Terminology} Traditionally, what we term ``observed labels'' $\olabel$ are often referred to instead as the ``ground truth'' or ``correct'' labels~\citep[e.g.]{abumostafa2012learning, hastie2009statistical, kong1995decomp}. We avoid this terminology because, as the work on label bias has explained, these labels are often unreliable or contested~\citep{friedler2016impossibility, cooper2021emergent}. 

\custompar{Sets, random variables, and instances} We use bold non-italics letters to denote random variables (e.g., $\rvx$, $\rmD$), capital block letters to denote sets (e.g., $\instances$, $\labels$), lower case italics letters to denote scalars (e.g., $\olabel$), bold italics lower case letters to denote vectors (e.g., $\instance$), and bold italics upper case to denote matrices (e.g., $\datasetk$). For a complete example, $\instance$ is an arbitrary instance's feature vector, $\instances$ is the set representing the space of instances  $\instance$ ($\instance \in \instances$), and $\rvx$ is the random variable that can take on specific values of $\instance \in \instances$. We use this notation consistently, and thus do not always define all symbols explicitly.


\subsection{\appsetuplimitations}\label{app:sec:limitations}

Our setup, per our definition of the learning process (Definition~\ref{def:learningprocess}) is deliberately limited to studying the effects of variance due to changes in the underlying training dataset, with such datasets drawn from the same distribution.
For this reason, Definition~\ref{def:learningprocess} does not include the data collection process or hyperparameter optimization (HPO), which can further introduce non-determinism to machine learning, and are thus assumed to have been  already be completed.

Relatedly, variance-induced error can of course have other sources due to such non-determinism. 
For example, stochastic optimization methods, such as SGD and Adam, can cause fluctuations in test error; as, too, can choices in HPO configurations~\citep{cooper2021hpo}. While each of these decision points is worthy of investigation with respect to their impact on fair classification outcomes, we aim to fix as many sources of randomness as possible in order to highlight the particular kind of arbitrariness that we describe in Sections~\ref{sec:intro} and~\ref{sec:significance}. As such, we use the Limited-memory BFGS solver and fix our hyperparameters based on the results of an initial search (Section~\ref{sec:empirical}), for which we selected a search space through consulting related work such as \citet{chen2018tradeoff}.

\subsection{\appprelimcosts}\label{app:sec:prelim:costs}

For reference, we provide a bit more of the basic background regarding the relationship between the classification decision threshold $\tau$ and costs of false positives \fp{} ($\costfp$) and false negatives \fn{} ($\costfn$). We visualize the loss as follows:
\begin{table}[h!]
	\begin{center}
		\caption{Confusion matrix for cost-sensitive loss $\loss$, adapted from \citet{elkan2001cost}.}
		\label{app:tab:confusion}
		\begin{tabular}{lll}
			\toprule %
			{} & \textbf{$\pred = 0$} & \textbf{$\pred = 1$}\\
			\midrule
			\textbf{$\olabel = 0$} & \tn: $0$ & \fp: $\costfp$ \\
			\midrule %
			\textbf{$\olabel = 1$} & \fn: $\costfn$ & \tp: $0$ \\
			\bottomrule %
		\end{tabular}
	\end{center}
\end{table}

\noindent 0-1 loss treats the cost of different types of errors equally $\costfp = \costfn = 1)$; false positives and false negatives are quantified as equivalently bad -- they are \emph{symmetric}; the case for which $\costfp \neq \costfn$ is \emph{asymmetric} or \emph{cost-sensitive}.



Altering the asymmetric of costs shifts the classification decision threshold $\tau$ applied to the underlying regressor $\regressork$. We can see this by examining the behavior of $\regressork$ that we learn. $\regressork$ estimates the probability of a each label given $\instance$ (since we do not learn using $\group$), i.e., that we develop a good approximation of the distribution $p(\rvy | \rvx)$. Ideally, $\regressork$ will be similar to the Bayes optimal classifier (for which the classification rule produces classifications $y^*$ that yield the smallest weighted sum of the loss, where the weights are the probabilities of a particular label $\rvy=i$ for a given $(\instance, \group)$, i.e., sums over\looseness=-1

\begin{align}
	\label{app:eq:optsum}
	p(\rvy=i |\rvx=\instance)\; \loss(i, y').
\end{align}

\noindent For binary classification, 
the terms of (\ref{app:eq:optsum}) in the sum for a particular $y'$ 
yield two cases: 

\begin{itemize}
	\item \textbf{$i = y'$}: By definition, $\loss(i, y') = 0$; therefore, (\ref{app:eq:optsum}) $=0$.\looseness=-1
	
	\item \textbf{$i \ne y'$}: By definition, $\loss(i, y') = \costfp$ or $\ell(i, y') = \costfn$. 
	So, (\ref{app:eq:optsum}) will weight the cost by the probability $p(\rvy=i | \rvx=\instance)$.
\end{itemize}

We can therefore break down the Bayes optimal classifier into the following decision rule, which we hope to approximate through learning. For an arbitrary $(\instance, \group)$ and $\labels = \{0,1\}$, 

\begin{align*}
	&\min \Big(\overbrace{\overbrace{p(\rvy = 0| \rvx = \instance)}^{\text{Probability of \fp}} \times \costfp + \overbrace{p(\rvy = 1| \rvx = \instance)}^{\text{Probability of \tp}}\times 0}^{\text{Weighted cost of predicting positive (1) class }}, \;\; \overbrace{\overbrace{p(\rvy = 0| \rvx = \instance)}^{\text{Probability of \tn}} \times 0 + \overbrace{p(\rvy = 1| \rvx = \instance)}^{\text{Probability of \fn}}\times \costfn}^{\text{Weighted cost of predicting negative (0) class}}\Big) \\ 
	= &\min \Big(\overbrace{p(\rvy = 0| \rvx = \instance)}^{\text{Probability of \fp}} \times \costfn, \;\; \overbrace{p(\rvy = 1| \rvx = \instance)}^{\text{Probability of \fn}} \times \costfn\Big)
\end{align*}

That is, to predict label $1$, the cost of mis-predicting $1$ (i.e., the cost of a false positive \fp) must be be smaller than the cost of mis-predicting $0$ (i.e, the cost of a false negative \fn). In binary classification $p(\rvy| \rvx=\instance)= p(\rvy=1| \rvx=\instance) + p(\rvy=0 | \rvx=\instance)=1.$ So, we can assign $p(\rvy=1|\rvx=\instance) = \tau$ and $p(\rvy=0|\rvx=\instance) = 1-\tau$, and rewrite the above as

\begin{align}
	\label{app:eq:decision-rule}
	\min \Big( (1-\tau)\,\costfp, \;\; \tau \costfn\Big).
\end{align}

\noindent The decision boundary is the case for which both of the arguments to $\min$ in (\ref{app:eq:decision-rule}) are equivalent (i.e., the costs of predicting a false positive and a false negative are equal), i.e., 

\begin{align*}
	(1-\tau)\,\costfp &=  \tau \costfn \Rightarrow \tau = \frac{\costfp}{\costfp + \costfn}, \text{ so,}\\
	\modelk(\instance) = \1[\regressork(\instance) \geq \tau] &= 
	\begin{cases}
		1,& \text{if } p(\rvy = 1 | \rvx = \instance) \ge \tau = \frac{\costfp}{\costfp + \costfn}\\
		0,              & \text{otherwise}.
	\end{cases}
\end{align*}


\noindent For 0-1 loss, in which $\costfp = \costfn = 1$, $\tau$ evaluates to $\frac{1}{2}$. 
If we want to model asymmetric costs, then we need to change this decision threshold to account for which type of error is more costly. For example, let us say that false negatives are more costly than false positives, with $\costfp = 1$ and $\costfn = 3$. This leads to a threshold of $\frac{1}{4}$, which  biases $\modelk$ toward choosing the (generally cheaper to predict/more conservative) positive class. 

\subsection{\appprelimbootstrap}\label{app:sec:prelim:boot}

In the bootstrap method, we treat each dataset $\hatdatasetk \in \hat{\datasets}$ as equally likely.  For each set aside test example $(\instance, \group, \olabel)$, we can approximate $\err(\tproc, \datasets, (\instance, \group, \olabel))$ empirically by computing 
\begin{align}
	\label{eq:errorsumapprox}
	\haterr\big(\tproc, \hatdatasets, (\instance, \group, \olabel)\big) &= \frac{1}{\boot} \sum_{i=1}^{\boot} \ell\big(\olabel, \hatmodel_{\hatdataset_i}(\instance)\big)
\end{align}
\vspace{-.15cm}

\noindent for a concrete number of replicates $\boot$. We estimate overall error $\haterr(\tproc, \hatdatasets)$ for the test set by additionally summing over each example instance $(\instance, \group, \olabel)$, which we can further delineate into $\hatfpr$ and $\hatfnr$, or into group-specific $\haterr_\group$, $\hatfpr_\group$, and $\hatfnr_\group$ by computing separate averages according to $\group$.\looseness=-1 

The bootstrap method exhibits less variance than cross-validation, but can be biased --- in particular, pessimistic --- with respect to estimating expected error. To reduce this bias, one can follow our setup in Definition~\ref{def:learningprocess}, which splits into train and test sets before resampling. For more information comparing the two methods, see \citet{efron1997boot, efron1993bootsrap}. Further, recent work shows that, in relation to studying individual models, CV is in fact asymptotically uninformative regarding expected error~\citep{wager2020cv}.
\section{\appvariance}\label{app:sec:variance}

In this appendix, we provide more details on other types of statistical error (Appendix~\ref{app:sec:noisebias}), on variance (Appendix~\ref{app:sec:ourvariance}) and self-consistency (Appendix~\ref{app:sec:consistency}). Following this longer presentation of our metrics, we then provide some additional information on other definitions of variance that have been used in work on fair classification, and contextualize issues with these definitions that encouraged us to deviate from them in order to derive our definition of self-consistency (Appendix~\ref{app:sec:othervariance}).

\subsection{\appprelimenoisebias}\label{app:sec:noisebias}

\custompar{Noise} 
Noise is traditionally understood as \textit{irreducible error}; it is due to inherent randomness in the data, 
which 
cannot be captured perfectly accurately by a deterministic decision rule $\modelk$. 
Notably, noise is an aspect of the data collection pipeline, not the learning process (Definition~\ref{def:learningprocess}). 
It is \textit{irreducible} in the sense that it does not depend on our choice of training procedure $\tproc$ or how we draw datasets for training from $\datasets$, either in theory or in practice. Heteroskedastic noise across demographic groups is often hypothesized to be a source of unfairness in machine learning~\citep{cooper2021emergent, chen2018tradeoff}. 
	Importantly, albeit somewhat confusingly, 
	this is commonly referred to as label bias, where ``bias'' connotes discrimination, as opposed to the statistical bias that we mention here.
	
	Unlike noise, bias and variance are traditionally understood as sources of epistemic uncertainty. 
	These sources of error are \textit{reducible} because they are contingent on the modeling choices we make in the learning process; if we knew how to model the task at hand more effectively, in principle, we could reduce bias and variance error. 
	
	\custompar{Bias} Within the amount of reducible error, bias reflects the error associated with the chosen hypothesis class $\sH$, and is therefore governed by decisions concerning the training procedure $\tproc$ in the learning process (Definition~\ref{def:learningprocess}). This type of error is persistent because it takes effect at the level of possible models in $\sH$; in expectation, all models $\modelk \in \sH$ have the same amount of bias-induced error. 
	
	Whereas variance depends on stochasticity in the underlying training data, noise and bias error are traditionally formulated in relation to the Bayes optimal classifier --- the best possible classifier that machine learning could produce for a given task~\citep{geman1992bvd, abumostafa2012learning, domingos2000icml}. Since the Bayes optimal classifier is typically not available in practice, we often cannot estimate noise or bias directly in experiments.
	
	Of the three types of statistical error, it is only variance that seems to reflect the intuition in Figure~\ref{fig:vote} concerning the behavior of different possible models $\modelk$. This is because noise is a property of the data distribution; for a learning process (Definition~\ref{def:learningprocess}), in expectation we can treat noise error as constant. Bias can similarly be treated as constant for the learning process: It is a property of the chosen hypothesis class $\hclass$, and thus is in expectation the same for each $\modelk \in \hclass$. In Figure~\ref{fig:vote}, we are keeping the data distribution constant and $\hclass$ constant; we are only changing the underlying subset of training data to produce different models $\modelk$.
	
	\subsection{\appprelimvariance}\label{app:sec:ourvariance}
	
	We first provide a simple proof that explains the simplified version for our empirical approximation for variance in (\ref{eq:hatvar}).
	
	\vspace{-.25cm}
	\begin{proof}
		For the models $\{\model_{\dataset_b}\}_{b=1}^{\boot}$ that we produce, we denote $\hatlabels$ to be the multiset of their predictions on $(\instance, \group)$. $|\hatlabels| = \boot = \boot_0 + \boot_1$, where $\boot_0$ and $\boot_1$ represent the counts of $0$ and $1$-predictions, respectively. We also set the cost of false positives to be $\loss(0, 1) = \costfp$ and the cost of false negatives to be $\loss(1, 0) = \costfn$. 
		
		Looking at the sum in $\hatvar$ (i.e., $\sum_{i \neq j}$), each of the $\boot_0$ $0$-predictions will get compared to the other $\boot_0 - 1$ $0$-predictions and to the $\boot_1$ $1$-predictions.  By the definition of $\loss$, each of the $\boot_0 - 1$  computations of $\loss(0, 0)$ evaluates to $0$ and each of the $\boot_1$ computations of $\loss(0, 1)$ evaluates to $\costfp$. Therefore, the $\boot_0$ $0$-predictions contribute 
		\begin{align*}
			\textstyle
			\boot_0 \times \big[\big(0 \times (\boot_0 - 1)\big) + \costfp 
			\times \boot_1 \big] = \costfp\boot_0\boot_1
		\end{align*}
		
		\noindent to the sum in $\hatvar$, and,  by similar reasoning, 
		$\boot_1 \times \big[\big(0 \times (\boot_1 - 1)\big) + \costfn \times \boot_0 \big] = \costfn\boot_0\boot_1.$ It follows that the total sum in $\hatvar$ is
		\begin{align*}
			&\sum_{i \neq j} \loss\Big(\hatmodel_{\hatdataset_i}(\instance), \hatmodel_{\hatdataset_j}(\instance)\Big) = (\costfp + \costfn)\boot_0\boot_1. \text{ Therefore}\\
			&\overbrace{\frac{1}{\boot(\boot-1)} \sum_{i \neq j} \loss\Big(\hatmodel_{\hatdataset_i}(\instance), \hatmodel_{\hatdataset_j}(\instance)\Big)}^{\hatvariance} = \overbrace{\frac{(\costfp + \costfn)\boot_0\boot_1}{\boot(\boot-1)}}^{\text{(\ref{eq:hatvar})}}
		\end{align*}
	\end{proof}
	
	\custompar{The effect of $\tau$ on variance} As discussed in Appendix~\ref{app:sec:prelim:costs}, $\costfp$ and $\costfn$ can be related to changing $\tau$ applied to $\regressork$ to produce classifier $\modelk$. We analyze the range of minimal and maximal empirical variance by examining the behavior of $B\to\infty$, i.e.,
	\begin{align}
		\label{app:eq:lim-variance}
		\lim_{\boot\to\infty} \frac{(\costfp + \costfn)\boot_0\boot_1}{\boot(\boot-1)}.
	\end{align}
	
	
	\custompar{Minimal variance} Either $\boot_0$ or $\boot_1$ (exclusively, since $\boot_0 + \boot_1 > 1$) will be $=0$, with the other being $=\boot$, making (\ref{app:eq:lim-variance}) equivalent to
	\begin{align*}
		\lim_{\boot\to\infty} \frac{(\costfp + \costfn)\times 0}{\boot(\boot-1)} = 0, \text{regardless of the value of $\costfp + \costfn$.}
	\end{align*}
	
	\custompar{Maximal variance} $\boot_0$ will represent half of $\boot$, with $\boot_1$ representing the other half. More particularly, $\boot_0 = \frac{\boot}{2}$ and $\boot_1 = \frac{\boot}{2}$; or, without loss of generality, $\boot_0=\frac{\boot-1}{2}$ and $\boot_1=\frac{\boot+1}{2}$. This means that
	\begin{align*}
		\frac{(\costfp + \costfn)\boot_0\boot_1}{\boot(\boot-1)} &= \frac{(\costfp + \costfn)(\frac{\boot}{2})^2} {\boot(\boot-1)} & \text{\small{\bigg(Or, $= \frac{(\costfp + \costfn)(\frac{\boot-1}{2})(\frac{\boot+1}{2})} {\boot(\boot-1)}$\bigg)}}\\
		&= \frac{(\costfp + \costfn)(\frac{\boot^2}{4})}{\boot^2 - \boot} & \text{\small{\bigg(Or, $= \frac{(\costfp + \costfn)(\frac{(\boot^2 - 1}{4})}{\boot(\boot-1)}$; it will not matter in the limit\bigg)}}\\
		&= \frac{(\costfp + \costfn)\boot^2} {4\boot^2 - 4\boot}.
	\end{align*}
	
	\noindent And, therefore, 
	\begin{align}
		\label{app:eq:maxvariance}
		\lim_{\boot\to\infty}\frac{(\costfp + \costfn)\boot^2} {4\boot^2 - 4\boot} &= \frac{\costfp + \costfn}{4}.
	\end{align}
	It follows analytically that variance will be in the range  $[0, \frac{\costfp + \costfn}{4})$. However, empirically, 
	for concrete $\boot$, $\hatvariance \rightarrow [0, \frac{\costfp + \costfn}{4} + \epsilon]$, for smaller positive $\epsilon$ as the number of models $\boot$ increases. The maximal variance will better approximate $\frac{\costfp + \costfn}{4}$ as $\boot$ gets larger, but will be $>\frac{\costfp + \costfn}{4}$. For example, for 0-1 loss $\frac{\costfp + \costfn}{4} = \frac{2}{4} = 0.5$. For $\boot=100$, the maximal 
	$\hatvariance = \frac{2 \times 50 \times 50}{100 \times 99} = \frac{50}{99} \approx .505$.

	\subsection{\appconsistencyderiving}\label{app:sec:consistency}
	
	In this appendix, we describe the relationship between variance (Definition~\ref{def:variance}) and self-consistency (Definition~\ref{def:sc}) in more detail, and show that $\hatsc\big(\tproc, \{\dataset_b\}_{b=1}^\boot, (\instance, \group)\big) \rightarrow [0.5 - \epsilon, 1$] for small positive $\epsilon$ as the number of models $\boot$ increases. 

	\begin{proof}

		Note that, by the definition of 0-1 loss, $\costfp = \costfn = 1$, so 
		
		\begin{align}
			\label{app:eq:01variance}
			\hatvariance_{\text{0-1}} &= \frac{1}{\boot(\boot-1)}\sum_{i\neq j}\1[\model_{\dataset_i}(\instance) \ne \model_{\dataset_j}(\instance)]  = \frac{2\boot_0\boot_1}{\boot(\boot-1)}.
		\end{align}
		
		By the definition of the indicator function $\1$,
		
		\begin{align*}
			1 &= \frac{1}{\boot(\boot-1)}\sum_{i\neq j}\Big[\overbrace{\1[\model_{\dataset_i}(\instance) \ne \model_{\dataset_j}(\instance)]}^{\text{From } \hatvariance_{\text{0-1}}} + \:\:\: \overbrace{\1[\model_{\dataset_i}(\instance) = \model_{\dataset_j}(\instance)]}^{\text{From } \hatconsistency}\Big]\\
			&= \overbrace{\frac{2\boot_0\boot_1}{\boot(\boot-1)}}^{(\ref{app:eq:01variance})} + \frac{1}{\boot(\boot-1)}\sum_{i\neq j}\1[\model_{\dataset_i}(\instance) = \model_{\dataset_j}(\instance)]. 
		\end{align*}
    Therefore, rearranging,
		\begin{align*}
			\hat{\texttt{SC}}\big(\mathcal{A}, \hatdatasets, (\instance, \group)\big) = \frac{1}{\boot(\boot-1)}\sum_{i\neq j}\1[\model_{\dataset_i}(\instance) = \model_{\dataset_j}(\instance)] =  1 - \frac{2\boot_0\boot_1}{\boot(\boot-1)}.
		\end{align*}
		
	\end{proof}
	
	We note that $\hatsc$ (\ref{eq:sc}) is independent of specific costs $\costfp$ and $\costfn$. Nevertheless, the choice of decision threshold $\tau$ will of course impact the values of $\boot_0$ and $\boot_1$ in practice. In turn, this will impact the degree of self-consistency that a learning process exhibits empirically. In short, the measured degree of self-consistency in practice will depend on the choice of $\loss$. Further, following an analysis similar to what we can show that $\hatsc$ will be a value in $[0.5 + \epsilon, 1]$, for small positive $\epsilon$. This reality is reflected in the results that we report for our experiments, for which $B=101$ yields minimal $\hatsc \approx 0.495$. 

	\paragraph{\appconsistencycost} Intuitively, \emph{self}-consistency of a learning process is a relative metric; it is a quantity that is measured relative to the learning process. We therefore conceive of it as a metric that is normalized with respect to the learning process (Definition~\ref{def:learningprocess}). Such a process can be maximally $100\%$ self-consistent, but it does not make sense for it to be more than that (reflected by the maximum value of $1$).
	
	In contrast, as discussed in Appendix~\ref{app:sec:variance}, variance can measure much greater than 1, depending on the magnitude of the sum of the costs $\costfp$ and $\costfn$, in particular, for $\costfp + \costfn > 4$ (\ref{app:eq:maxvariance}). However, it is not necessarily meaningful to compare the magnitude of variance across classifiers. Recall that the effect of changing costs $\costfp$ and $\costfn$ corresponds to a change in the binary classification decision threshold, with $\tau = \frac{\costfp}{\costfp + \costfn}$. It is the \emph{relative} costs that change the decision threshold; not the costs themselves. For example, the classifier with costs $\costfp = 1$ and $\costfn = 3$ is equivalent to the classifier with costs $\costfp = \frac{1}{2}$ and $\costfn = \frac{3}{2}$ (for both, $\tau = \frac{1}{4}$), but the former would measure a larger magnitude for variance.
	
	It is this observation that grounds our cost-independent definition of self-consistency in Section~\ref{sec:significance} and Appendix~\ref{app:sec:consistency}. Given the fact that the magnitude of variance measurements can complicate our comparisons of classifiers, as discussed above, we focus on the part of variance that encodes information about arbitrariness in a learning process: its measure of (dis)agreement between classification decisions that result from changing the training dataset. We could alternatively conceive of self-consistency as the additive inverse of normalized variance, but this is more complicated because it would require a computation that depends on the specific costs, $\hatvariance_{\text{normalized}} = \frac{\hatvariance}{\hatvariance_\text{max}}$.\looseness=-1

    \subsubsection{\appconsistencydetails}\label{app:sec:consistency:details}
	
	\custompar{Terminology}  In logic, the idea of consistent belief has to do with ensuring that we do not draw conclusions that contradcit each other. This is much like the case that we are modeling with self-consistency --- the idea that underlying changes in the dataset can lead to predictions that are directly in contradition~\citep{smullyan1986belief, hintikka1962doxastic, stalnaker2006logic}. Ideas of consistency in legal rules have a similar flavor; legal rules should not contradict each other; legal judgments should not contradict each other (this is at least an aspiration for the law, based on common ideas in legal theory~\citep{fuller1965law, tamanaha2004law}. For both of these reasons, the term ``consistent'' has a natural mapping to our usage of it in this paper. This is especially true in the legal theory case, given that inconsistency in the law is often considered arbitrary and a source of discrimination. 
	
	We nevertheless realize that the word ``consistent'' is overloaded with many meanings in statistics and different subfields computer science like distributed computing \citep[e.g.,]{zhang2020amagold, abadi2012tradeoff}. Nevertheless, due to the clear  relationship between our purposes concerning arbitrariness and discrimination, and definitions in logic and the law, we believe that it is the most appropriate term for our work. 
	
	\custompar{Quantifying systematic arbitrariness} We depict \emph{systematic arbitrariness} using the Wasserstein-1 distance~\cite{ramdas2015wass}. This is the natural distance for us to consider because it has a closed form when being applied to CDFs. For our purposes, it should be interpreted as computing the total disparity in self-consistency by examining all possible self-consistency levels $\kappa$ at once.  
	
	Formally,\footnote{We consider the Wasserstein distance for one-dimensional distributions. More generally, the $p$-th Wasserstein distance for such distributions, $\mathcal{W}_p$, requires the inverse CDFs to be well-defined (i.e.,  the CDFs need to be strictly monotonic). This is fine to assume for our purposes. We have to relax the formal definition of the Wasserstein distance, anyway, when we estimate it in practice with a discrete number of samples.} for two groups $\group=0$ and $\group=1$ with respective $\texttt{SC}$ CDFs $F_0$ and $F_1$, 
	\begin{align*}
		\mathcal{W}_{1} &= \int_{\R} |F_0(\kappa) - F_1(\kappa)| \; d\kappa.
	\end{align*}
	
	For self-consistency, which we have defined on $[0.5, 1]$, this is just 
	
	\begin{align*}
		\mathcal{W}_{1} &= \int_{0.5}^1 |F_0(\kappa) - F_1(\kappa)| \; d\kappa.
	\end{align*}
	
	Empirically, we can approximate this with 
	
	\begin{align*}
		\hat{\mathcal{W}_1} \coloneqq \frac{1}{|\hat{\sK}|}\sum_{\hat{\sK}} | \hat{F}_0(\hat\kappa) - \hat{F}_1(\hat\kappa)|, \hspace{.5em} \text{where } \hat{\sK} = \biggl\{1 - \frac{2\boot_0\boot_1}{\boot(\boot-1)} \bigg| \boot_0 \in \{0 \ldots \boot\} \land \boot_1 \in \{0 \ldots \boot\} \land \boot_0 + \boot_1 = \boot \biggr\}.
	\end{align*}
	
	We typically set $\boot=101$, and thus 
	
	\begin{align*}
		\hat{\sK} = [&0.49505, 0.49545, 0.49624, 0.49743, 0.49901, 0.50099, 0.50337, 0.50614, 0.50931, 0.51287,\\
        &0.51683, 0.52119, 0.52594, 0.53109, 0.53663, 0.54257, 0.54891, 0.55564, 0.56277, 0.57030,\\
        &0.57822, 0.58653, 0.59525, 0.60436, 
		0.61386, 0.62376, 0.63406, 0.64475, 0.65584, 0.66733,\\
        &0.67921, 0.69149, 0.70416, 0.71723, 0.73069, 0.74455,
		0.75881, 0.77347, 0.78851, 0.80396,\\
        &0.81980, 0.83604, 0.85267, 0.86970, 0.88713, 0.90495, 0.92317, 0.94178, 0.96079, 0.9802, 1.0],
	\end{align*}
	
	\noindent which we use to produce our CDF plots.
	
	When measuring systematic arbitrariness with abstention, we set the probability mass for $<\kappa$ to $0$ it. This makes sense because we are effectively re-defining the $\hatsc$ CDFs to not include instances that exhibit below a minimal amount of $\hatsc$. This also makes comparing systematic arbitrariness across CDFs for different interventions more interpretable. It allows us to keep the number of experimental samples for the empirical CDF measures constant when computing averages, so abstaining would then always have the effect of decreasing systematic arbitrariness. If we did not do this, because the Wasserstein-1 distance is an average, changing the set $\hat{\sK}$, of course, would change the amount of Wasserstein-1 distance --- possibly leading to a relative \emph{increase} (if there are greater discrepancies between $\group$-condition CDF curves at $\geq \kappa$). 
\section{\appothervariance}\label{app:sec:othervariance}

As noted in Section~\ref{sec:related}, prior work that discusses variance and fair classification often relies on the definition of variance from \citet{domingos2000icml}. We deviate from prior work and provide our own definition for two reasons: 
1) variance in \citet{domingos2000icml, domingos2000report} does not cleanly extend to cost-sensitive loss, and 2) the reference point for measuring variance in \citet{domingos2000icml,domingos2000report} --- the 
\emph{main prediction} --- can be unstable/ brittle in practice. We start by explaining the \citet{domingos2000icml, domingos2000report} definitions, and then use these definitions to support our rationale. 

\subsection{\appmain}\label{app:sec:other:main} 

To begin, we restate the definitions from \citet{domingos2000icml, domingos2000report} concerning the expected model (called the \emph{main predictor}).  We change the notation from \citeauthor{domingos2000icml} to align with our own, as we believe these changes provide greater clarity concerning meaning, significance, and consequent takeaways. Nevertheless, these definitions for quantifying error are equivalent to those in~\citet{domingos2000report}, and they fundamentally depend on human decisions for setting up the learning process.

\citet{domingos2000icml, domingos2000report} define predictive variance in relation to this single point of reference. 
This reference point captures the general, expected behavior of models that could be produced by the chosen learning process. 
We can think of each prediction of this point of reference as the ``central tendency'' of the predictions made by all possible models in $\possiblemodels$ for $(\instance, \group)$.  
Formally,\looseness=-1

\begin{definition}
	\label{def:main}
	The \textbf{main prediction} $\pred$ is the prediction value $y' \in \labels$ that generates the minimum average loss with respect to all of the predictions $\pred \in \hat{\labels}$ generated by the different possible models in $\possiblemodels$. It is defined as the expectation over training sets $\datasets$ for a loss function $\loss$, given an example instance $(\instance, \group)$. That is,
	\begin{align}
		\label{eq:main}
		\overline{y} = \argmin_{y'}\E_{\rmD}[\loss(\pred, y') | \rvx=\instance, \rvg=\group]. 
	\end{align}
	The \textit{main predictor} $\overline{\model}: \instances \rightarrow \labels$ produces the main prediction $\overline{y}$ for each $(\instance, \group)$.
\end{definition}


What 
(\ref{eq:main}) evaluates to in practice of course depends on the loss function $\loss$. For squared loss, the main prediction is defined as the mean prediction of all the $\modelk$~\cite{domingos2000icml, kong1995decomp}. 
Following~\citet{kong1995decomp}, for 0-1 loss \citet{domingos2000icml} defines the main prediction as the mode/majority vote --- the most frequent prediction for an example instance $(\instance, \group)$.  We provide a more formal discussion of why this is the case when we discuss problems with the main prediction for cost-sensitive loss (Appendix~\ref{app:sec:comparison}). \citet{domingos2000icml, domingos2000report} then define variance in relation to specific models $\modelk$ and the main predictor $\overline{\model}$:  

\begin{definition}
	\label{def:app-variance-domingos}
	The \textit{variance}-induced error for fresh example instance $(\instance,\group)$ is
	\begin{align*}
		\variance = \E_\rmD[\loss(\overline{y}, \pred)|\rvx=\instance, \rvg=\group],
	\end{align*}
	where $\overline{y} = \overline{\model}(\instance)$ is the main prediction and the $\pred$ are the predictions for the different $\modelk \sim \possiblemodels$.
\end{definition}

\noindent That is, for a specific $(\instance,\group)$, it is possible to compare the individual predictions $\pred = \modelk(\instance)$ to the main prediction $\overline{y} = \overline{\model}(\instance)$. Using the main prediction as a reference point, one can compute the extent of disagreement of individual predictions with the main prediction as a source of error. It is this definition (Definition~\ref{def:app-variance-domingos}) that prior work on fair classification tends to reference when discussing variance~\citep{chen2018tradeoff, black2022selective}. However, as we discuss in more detail below (Appendix~\ref{app:sec:comparison}), many of the theoretical results in \citet{chen2018tradeoff} follow directly from the definitions in \citet{domingos2000icml}, and the experiments do not actually use those results in practice. \citet{black2022selective}, in contrast, presents results that rely heavily on the main prediction in \citet{domingos2000icml}. 

\subsection{\appprelimcomparison}\label{app:sec:comparison}

We now compare our definition of variance (Definition~\ref{def:variance}) to the one in \citet{domingos2000icml, domingos2000report} (Definition~\ref{def:app-variance-domingos}). This comparison makes clear in detail why we deviate from prior work that relies on \citet{domingos2000icml, domingos2000report}.

\custompar{No decomposition result} Following from above, it is worth noting that by not relying on the main prediction, we lose the applicability of the decomposition result that \citet{domingos2000icml, domingos2000report} develop. However, we believe that this is fine for our purposes, as we are interested in the impact of empirical variance specifically on fair classification outcomes. We do not need to reason about bias or noise in our results to understand the arbitrariness with which we are concerned (Section~\ref{sec:var:intuition}). It is also worth noting that prior work on fair classification that leverages \citet{domingos2000icml} also does not leverage the decomposition, either. \citet{chen2018tradeoff} extends the decomposition to subgroups in the context of algorithmic fairness,\footnote{This just involves splitting the conditioning on an example instance of features $\instance$ into conditioning on an example instance whose features are split into $(\instance, \group)$.} and then informally translates the takeaways of the \citet{domingos2000icml} result to a notion of a ``level of discrimination.''  Moreoever, unlike our work, these prior studies do not actually measure variance directly in its experiments.

\custompar{No need to compute a ``central tendency''} In \citet{domingos2000icml, domingos2000report}, variance is defined in terms of both the loss function $\loss$ and the main prediction $\overline{y}$. This assumes that the main prediction is well-defined for the loss function, and that it is well-behaved. While there is a simple interpretation of the main prediction for squared loss (the mean) and for 0-1 loss (the mode/majority vote), it is significantly messier for cost-sensitive loss, which is a more general formulation that includes 0-1 loss. \citet{domingos2000icml, domingos2000report} does not discuss this explicitly, so we derive the main prediction for cost-sensitive loss ourselves below. In summary:
\begin{itemize}[topsep=3.5pt, itemsep=0pt, leftmargin=.5cm]
	\item The behavior of the main prediction for cost-sensitive loss reveals that the decomposition result provided in the extended technical report (Theorem 4, \citet{domingos2000report}) is in fact very carefully constructed. We believe that this construction is so specific that it is not practically useful (it is, in our opinion, hardly ``unified'' in a more general sense, as it is so carefully adapted to specific loss functions and their behavioral special cases).
	\item By decoupling from the need to compute a main prediction as a reference point, our variance definition is ultimately much simpler and more general, with respect to how it accommodates different loss functions.\footnote{This reveals a subtle ambiguity in the definition of the loss  $\loss$ in \citet{domingos2000icml, domingos2000report}. Neither paper explicitly defines the signature of $\loss$. For the main prediction (Definition~\ref{def:main}) and variance (Definition~\ref{def:app-variance-domingos}), there is a lack of clarity in what constitutes a valid domain for $\loss$. Computing the main prediction $\overline{y}$ suggests $\loss: \labels \times \labels \rightarrow \R_{\ge0}$, where $\overline{y} \in \labels$, but, since $\hat{\labels} \subseteq \labels$, it is possible that $\overline{y} \not\in \labels$. However, the definition of variance suggests that $\loss: \labels \times \hat{\labels} \rightarrow \R_{\ge0}$. Since  $\hat{\labels} \subseteq \labels$, it is not guaranteed that $\hat{\labels} = \labels$. This may be fine in practice, especially for squared loss and 0-1 loss (the losses with which \citet{domingos2000icml} explicitly contends), but it does arguably present a problem formally with respect to generalizing.}
\end{itemize}

\custompar{Brittleness of the main prediction} For high variance instances, the main prediction can flip-flop from $\pred=1$ to $\pred=0$ and back. While the strategy in \citet{black2022selective} is to abstain on the prediction in these cases, we believe that a better alternative is to understand that the main prediction is not very meaningful more generally for high-variance examples. That is, for these examples, the ability (and reliability) of breaking close ties to determine the main (simple majority) prediction is not the right approach. Instead, we should ideally be able to embed more confidence into our process than a simple-majority-vote determination.\footnote{This is also another aspect of the simplicity of not needing to define and compute a ``central tendency'' prediction. We do not need to encode a notion of a tie-breaking vote to determine a ``central tendency.'' The main prediction can be unclear in cases for which there is no ``main outcome'' (e.g., Individual 2 in Figure~\ref{fig:vote}), as the vote is split exactly down the middle. By avoiding the need to vote on a main reference point, we also avoid having to ever choose that reference point arbitrarily.} Put different, in cases for which we can reliably estimate the main prediction, but the vote margin is slim, we believe that the main prediction is still uncertain, based on our understanding of variance, intuited in Figure~\ref{fig:vote}. \textbf{The main prediction can be reliable, but it can still, in this view, be arbitrary} (Section~\ref{sec:related}). With a simple-majority voting scheme, there can be huge differences between predictions that are mostly in agreement, and those that are just over the majority reference point. Freeing ourselves of this reference point via our self-consistency metric, we can define thresholds of self-consistency as our criterion for abstention (where simple-majority voting is one instantiation of that criterion).\looseness=-1\footnote{This problem is worse for cost-sensitive loss, where the main prediction is not always the majority vote (see below).} 


\subsubsection{\appmaincost}\label{app:sec:maincost}

We show here that, for cost-sensitive loss, the main prediction depends on the majority class being predicted, the asymmetry of the costs, and occasional tie-breaking, such that the main prediction can either be the majority vote or the minority vote. 
\citet{domingos2000report} provides an error decomposition in Theorem 4, but does not explain the effects on the main prediction. We do so below, and also call attention to 0-1 loss as a special case of cost-sensitive loss, for which the costs are symmetric (and equal to 1). We first summarize the takeaways of the analysis below:

\begin{itemize}[itemsep=0pt, leftmargin=.5cm]
\item \textbf{Symmetric loss}: The main prediction is the \textbf{majority vote}.
\item \textbf{Asymmetric loss}: Compute 1) the relative cost difference (i.e., $\frac{\costfp-\costfn}{\costfn}$), 2) the majority class (and, as a result, the minority class) for the $\pred \in \hat{\labels}$, and 3) the relative difference in the number of votes in the majority and minority classes (i.e., what we call the \emph{vote margin}; below, $\frac{(i + 2j + 1) - i}{i}$)
\begin{itemize}[itemsep=0pt, leftmargin=.5cm]
	\item If the \textbf{majority class} in $\hat{\labels}$ has the \textbf{lower cost} of misclassification, then the main prediction is the \textbf{majority vote}.\looseness=-1
	\item If the \textbf{majority class} in $\hat{\labels}$ has the \textbf{higher cost} of misclassification, then the main prediction \textbf{depends on the asymmetry of the costs and the vote margin}, i.e.,
	\begin{itemize}[itemsep=0pt, leftmargin=.5cm]
		\item If $\frac{\costfp-\costfn}{\costfn} = \frac{(i + 2j + 1) - i}{i}$, we can choose the main prediction to be \textbf{either class} (but must make this choice consistently).
		\item If $\frac{\costfp-\costfn}{\costfn} > \frac{(i + 2j + 1) - i}{i}$, the \textbf{minority vote} is the main prediction.
		\item If $\frac{\costfp-\costfn}{\costfn} < \frac{(i + 2j + 1) - i}{i}$, the \textbf{majority vote} is the main prediction.
	\end{itemize}     
\end{itemize}
\end{itemize}

\begin{proof}
Let us consider cost-sensitive loss for binary classification, for which $\loss(0,0) = \loss(1, 1) = 0$ and we have potentially-asymmetric loss for misclassifications, i.e. $\loss(1,0) = \costfn$ and $\loss(0, 1) = \costfp$, with $\costfp, \costfn \in \R^+$. 0-1 loss is a special case for this type of loss, for which $\costfp = \costfn = 1$.     

Let us say that the total number of models trained is $k$, which we evaluate on an example instance $\instance$. Let us set $|\hat{\labels}| = k = 2i + 2j + 1$, with $i \geq 0$ and $j \geq 0$. We can think of $i$ as the common number of votes that each class has, and $2j + 1$ as the margin of votes between the two classes. Given this setup, this means that $k \geq 1$, i.e., we always have the predictions of at least 1 model to consider, and $k$ is always odd. This means that there is always a strict majority classification.

Without loss of generality, on $\instance$, of these $k$ model predictions $\pred \in \hat{\labels}$ , there are $i$ class-$0$ predictions and $i + 2j + 1$ class-$1$ predictions (i.e., we do our analysis with class $1$ as the majority prediction). To compute the main prediction $\overline{y}$, each $\pred \in \hat{\labels}$ will get compared to the values of possible predictions $y' \in \labels=\{0, 1\}$. That is, there are two cases to consider:

\begin{itemize}[topsep=5pt, leftmargin=0.5cm]
	\item \textbf{Case $y' = 0$}:  $y'=0$ will get compared $i$ times to the $i$ $\pred = 0$s in $\hat{\labels}$, for which $\loss(0,0)=0$; $y'=0$ will similarly get compared $i + 2j + 1$ times to the $1$s in in $\hat{\labels}$, for which (by Definition~\ref{def:main}) the comparison is $\loss(1,0)=\costfn$. By definition of expectation, the expected loss is\looseness=-1
	\begin{align}
		\label{eq:case-y'-0}
		\frac{i \times 0 + (i + 2j+ 1) \times \costfn}{2i + 2j + 1} = \frac{\costfn(i + 2j + 1)}{2i + 2j + 1}. 
	\end{align}
	
	\item \textbf{Case $y' = 1$}: Similarly, the label $1$ will also get compared $i$ times to the $0$s in $\hat{\labels}$, for which the comparison is $\loss(0,1) = \costfp$; $y'=1$ will also be compared $i + 2j + 1$ times to the $1$s in $\hat{\labels}$, for which $\loss(1,1)=0$. The expected loss is 
	\begin{align}
		\label{eq:case-y'-1}
		\frac{i \times \costfp + (i + 2j + 1) \times  0}{2i + 2j + 1} = \frac{\costfp i}{2i + 2j +1}.
	\end{align}
\end{itemize}

We need to compare these two cases for different possible values of $\costfn$ and $\costfp$ to understand which expected loss is minimal, which will determine the main prediction $\overline{y}$ that satisfies Equation (\ref{eq:main}). The three different possible relationships for values of $\costfn$ and $\costfp$ are $\costfn=\costfp$ (symmetric loss), and $\costfn > \costfp$ and $\costfn < \costfp$ (asymmetric loss). Since the results of the two cases above share the same denominator, we just need to compare their numerators, $\costfn(i+2j+1)$ (\ref{eq:case-y'-0})  and $\costfp i$ (\ref{eq:case-y'-1}).\looseness=-1 

\custompar{Symmetric Loss (0-1 Loss)} When $\costfn = \costfp = 1$, the numerators in (\ref{eq:case-y'-0}) and (\ref{eq:case-y'-1}) yield expected losses $i + 2j + 1$ and $i$, respectively. We can rewrite the numerator for (\ref{eq:case-y'-1}) as

\vspace{-.5cm}
\begin{align*}
	i + \overbrace{2j + 1}^{\geq 1, \text{ given $j \geq 0$}} &\geq i + 1,
\end{align*}

\noindent which makes the comparison of numerators $i < i + 1$, i.e., we are in the case (\ref{eq:case-y'-1}) $<$ (\ref{eq:case-y'-0}). This means that the case of $y' = 1$ (\ref{eq:case-y'-1}) is the minimal one; the expected loss for class $1$, the most frequent class, is the minimum, and thus the most frequent/ majority vote class is the main prediction. An analogous result holds if we instead set the most frequent class to be $0$. More generally, this holds for all symmetric losses, for which $\costfn = \costfp$.

\noindent$\blacktriangleright$ For \textbf{symmetric losses}, the main prediction $\overline{y}$ is \textbf{majority vote} of the predictions in $\hat{\labels}$.

\custompar{Asymmetric Loss} For asymmetric/ cost-sensitive loss, we need to examine two sub-cases:  $\costfn > \costfp$ and $\costfn < \costfp$. 

\begin{itemize}[topsep=5pt, leftmargin=0.5cm]
	\item \textbf{Case $\costfn > \costfp$}: $\costfp i < \costfn(i + \overbrace{2j + 1}^{\ge 1})$, given that $j \geq 0$. Therefore, since $\costfp i$ is minimal and associated with class $1$ (the most frequent class in our setup), the majority vote is the main prediction. We can achieve an analogous result if we instead set $0$ as the majority class.
	
	\noindent \vspace{5pt}$\blacktriangleright$ For \textbf{asymmetric losses}, the main prediction $\overline{y}$ is the \textbf{majority vote} of the predictions in $\hat{\labels}$, \textbf{if the majority class has a cheaper cost associated with misclassification} (i.e., if the majority class is $1$ and $\costfn < \costfp$, or if the majority class is $0$ and $\costfp < \costfn$).
	
	\item \textbf{Case $\costfn < \costfp$}:  If $\costfn < \costfn$, it depends on how asymmetric the costs are and how large the vote margin (i.e., $2j + 1$) between class votes is. There are 3 sub-cases:
	\begin{itemize}[topsep=5pt, leftmargin=0.5cm]
		\item \textbf{Case $\costfp i = \costfn(i + 2j + 1)$, i.e. cost equality}:  We can look at the relative asymmetric cost difference of the minority class cost (above $\costfp$, without loss of generality) and the majority class cost (above $\costfn$, without loss of generality), (above $\frac{\costfp - \costfn}{\costfn}$, without loss of generality). If that relative cost difference is equal to the relative difference of the votes between the majority and minority classes (i.e., $\frac{(i + 2j + 1) - i}{i}$), then the costs of predicting either $1$ or $0$ are equal. That is, we can rearrange terms as a ratio of costs to votes:
		\begin{align}
			\costfp i &= \costfn(i + \overbrace{2j + 1}^{\geq 1}) & \text{(The terms in this equality are $>0$)} \nonumber\\
			\frac{\costfp}{\costfn} &= \frac{i + 2j + 1}{i} & \text{(Given the above, $\costfp i > 0$ so $i > 0$)} \nonumber\\
			&= 1 + \frac{2j + 1}{i}\nonumber\\
			\frac{\costfp}{\costfn} - 1 &= \frac{2j + 1}{i}\nonumber\\
			\frac{\costfp - \costfn}{\costfn} &= \frac{2j + 1}{i} = \frac{(i + 2j + 1) - i}{i} \geq \frac{1}{i}
			\label{eq:app:comp}
		\end{align}
		\vspace{5pt}$\blacktriangleright$ For asymmetric loss \textbf{when the majority-class-associated cost is less than the minority-class associated cost and if the expected losses are equal}, then the \textbf{main prediction $\overline{y}$ is either $1$ or $0$}, (and we must make this choice consistently).\looseness=-1
		\item \textbf{Case $\costfp i > \costfn(i + 2j + 1)$}: We can look at the relative asymmetric cost difference of the minority class cost (above $\costfp$, without loss of generality) and the majority class cost (above $\costfn$, without loss of generality), (above $\frac{\costfp - \costfn}{\costfn}$, without loss of generality). If that relative cost difference is greater than the relative difference of the votes between the majority and minority classes (i.e., $\frac{(i + 2j + 1) - i}{i}$), then the \textit{minority vote} yields the minimum cost and is the main prediction $\overline{y}$ (above $\overline{y} = 0$, without loss of generality; an analogous result holds if we had set the majority vote to be $0$ and the minority vote to be $1$).  Following (\ref{eq:app:comp}) above, this is the same as \looseness=-1
		\begin{align*}
			\frac{\costfp - \costfn}{\costfn} &> \frac{(i + 2j + 1) - i}{i}
		\end{align*}
		\vspace{5pt}$\blacktriangleright$ For asymmetric loss \textbf{when the majority-class-associated cost is less than the minority-class associated cost}, it is possible for the \textbf{minority class} to have a greater associated loss. In this case, the \textbf{\textit{minority vote} is the main prediction $\overline{y}$}.\looseness=-1    
		
		\item \textbf{Case $\costfp i < \costfn(i + 2j + 1)$}: We can look at the relative asymmetric cost difference of the minority class cost (above $\costfp$, without loss of generality) and the majority class cost (above $\costfn$, without loss of generality), (above $\frac{\costfp - \costfn}{\costfn}$, without loss of generality).  If that relative cost difference s less than the relative difference of the votes between the majority and minority classes (i.e., $\frac{(i + 2j + 1) - i}{i}$), then the majority vote yields to minimum cost and is the main prediction $\overline{y}$ (above $\overline{y} = 1$, without loss of generality; an analogous result holds if we had set the majority vote to be $0$ and the minority vote to be $1$). Following (\ref{eq:app:comp}) above, this is the same as\looseness=-1
		\begin{align*}
			\frac{\costfp - \costfn}{\costfn} &< \frac{(i + 2j + 1) - i}{i}
		\end{align*}
		\vspace{5pt}$\blacktriangleright$ For asymmetric loss \textbf{when the majority-class-associated cost is less than the minority-class associated cost}, it is possible for the \textbf{majority class} to have a greater associated loss. In this case, the \textbf{\textit{majority vote} is the main prediction $\overline{y}$}.\looseness=-1  
	\end{itemize}
\end{itemize}
\end{proof}

\subsection{\appmodelm}\label{app:sec:mm}

A line of related work to ours concerns \emph{model multiplicity} and fairness~\citep{watson2023multiplicity, marx2020mm, black2022multiplicity}. This work builds off of an observation made by \citet{breiman2001multiplicity} regarding how there are multiple possible models of the same problem that exhibit similar degrees of accuracy. This set of multiple possible models of similar accuracy is referred to as the Rashomon set~\citep{breiman2001multiplicity}.

Work on model multiplicity has recently become fashionable in algorithmic fairness. In an effort to develop more nuanced model selection metrics beyond looking at just fairness and accuracy for different demographic groups, work at the intersection of model multiplicity and fairness tends to examine other properties of models in the Rashomon set in order to surface additional metrics for determining which model to use in practice. 

At first glance, this work may seem similar to what we investigate here, but we observe four key differences:\footnote{We defer discussion of \citet{black2022selective} to~\ref{app:sec:concurrent}.}
\begin{enumerate}
    \item  Model multiplicity places conditions on accuracy and fairness in order to determine the Rashomon set. We place no such conditions on the models that a learning process (Definition~\ref{def:learningprocess}) produces; we simulate the distribution over possible models $\possiblemodels$ without making any claims about the associated properties of those models.
    \item Model multiplicity makes observations about the Rashomon set with the aim of still ultimately putting forward criteria for helping to select \emph{a single model}. While the metrics used to inform these criteria include variance, most often work on model multiplicity still aims to choose one model to use in practice.
    \item Much of the work on model multiplicity emphasizes theoretical contributions, whereas our emphasis is on more experimental contributions. In conjunction with the first point, of ultimately trying to arrive at a single model, this work is also trying to make claims with respect to the Bayes-optimal model. Given our empirical focus --- of what we can actually produce in practice --- claims about optimality are not our concern. 
    \item   We focus specifically on variance reduction as a way to mitigate arbitrariness. We rely on other work, coincidentally contributions also made by \citeauthor{breiman1996bagging}, to study arbitrariness~\citep{breiman1996bagging}, and emphasize the importance of using ensemble models to produce predictions or abstention from prediction. We do not study the development of model selection criteria to pick a single model to use in practice; we use self-consistency to give a sense of predictive confidence about when to predict or not. We always select an ensemble model --- regardless of whether that model is produced by simple or super ensembling (Section~\ref{sec:algorithms}) --- and then use a user-specified level of self-consistency $\kappa$ to determine when that model actually produces predictions.
\end{enumerate}

These differences ultimately lead to very different methods for making observations about fairness. Importantly, we can study the arbitrariness of the underlying laerning process with a bit more nuance. For example, it could be the case that a particular task is just impossible to get right for some large subset of the test data (and this would be reflected in the Rashomon set of models), but for some portion of it there is a high amount of self-consistency for which we may still want to produce predictions.

Further, based on our experimental approach, we highlight completely different normative problems than those highlighted in work on model multiplicity (notably, see \citet{black2022multiplicity}). So, in short, while model multiplicity deals with related themes as our work --- issues of model selection, problem formulation, variance, etc. --- the goals of that work are ultimately different, but potentially complementary, from those in our paper. 

For example, a potentially interesting direction for future work would be to measure how metrics from work on model multiplicity behave in practice in light of the ensembling methods we present here. We could run experiments using Algorithm~\ref{algo:bagging-confidently} and investigate model multiplicity metrics for the underlying ensembled models. However, we ultimately do not see a huge advantage to doing this. Our empirical results indicate that variance is generally high, and has led to reliability issues regarding conclusions about fairness and accuracy. In fairness settings and available benchmarks, we find that the most important point is that variance has muddled conclusions. Under these circumstances, ensembling with abstention based on self-consistency seems a reasonable solution, in contrast to finding a single best model in the Rashomon set that attains other desired criteria. 

\subsection{\appconcurrent}\label{app:sec:concurrent}

There are several related papers that either preceded or came after this work's public posting. Some of this work is clearly concurrent, given the time frame. Other works that came after ours are not necessarily concurrent, but are either independent and unaware of our paper, or build on our work. 

\custompar{Setting the stage in 2021} The present work was scoped in 2021, in direct response to the initial study by \citet{forde2021disparate} and critical review by \citet{cooper2021emergent}. \citet{forde2021disparate} was one of the first (if not the first) paper to note that variance is overlooked in problem formulations that consider fairness. However, it was limited in scope and also dealt with deep learning settings, which have multiple sources of non-determinism that can be difficult to tease apart with respect to their effects on variance. 

\citet{cooper2021emergent} notes important, overlooked normative assumptions in the fairness-accuracy trade-off problem formulation, and suggests that this formulations is tautological. Our work is a natural direction for future research, in this respect -- to see how, in practice, the fairness-accuracy trade-off behaves after we account for variance. Indeed, we find that there is often no such trade-off, but for different reasons than those suggested by \citet{cooper2021emergent}. We expected there to be residual label bias that contributes to noise-induced error, but ultimately did not really observe this in practice. In these respects, our work both strengthens and complements these prior works. We support their claims, and go significantly beyond the work they did in order to provide such support. Further, our results suggest additional conclusions about experimental reliability in algorithmic fairness.\looseness=-1

\custompar{Variance and abstention-based ensembling} \citet{black2022selective} is concurrent work that slightly preceded our public posting. This work is similarly is interested in variance reduction, ensembling, and abstention in fairness settings, but fundamentally studies these topics in a different manner. We address four differences:

\begin{enumerate}
    \item \citet{black2022selective} does not take the wide-ranging experimental approach that we take. While we both study variance and fairness, our work also considers \emph{the practice of fair classification research} as an object of study. It is for these reasons that we do so many experiments on benchmark datasets, and clean and release another dataset for others to use.
    \item They rely on the definition of variance from \citet{domingos2000icml} in their work, likely building on the choice made by \citet{chen2018tradeoff} to use this defintion. Much of this Appendix is devoted to discussing \citet{domingos2000icml, domingos2000report} and his definition of variance. The overarching takeaway from our discussion is that 1) there are technical problems with this definition (which have been noted by others that investigated the bias-variance-noise trade-off for 0-1 loss in the early 2000s), 2) the definition does not naturally extend to cost-sensitive loss, 3) the main prediction can be unstable in practice and thus should not be the criterion for investigating arbitrariness (indeed, relying on the main prediction just pushes arbitrariness into that definition). While \citet{black2022selective} observes that variance is an important consideration for fairness, they ultimately focus on reliable estimation of the main prediction as the criterion for abstention in their ensembling method. While this kind of reliability is important, it does not deal with the general problem of arbitrary predictions (i.e., it is possible to have a reliable main prediction that is still effectively arbitrary). As a result, the nature of when and how to abstain is very different from ours. We instead base our criterion on a notion of confidence in the prediction, and we allow for flexibility around when to abstain when predictions are too arbitrary.
    \item As a result of the above two differences, the claims and conclusions in both of our works are different. While there are similar terms used in both works (e.g., variance, abstention), which may make the works seem overlapping with a cursory read, our definitions, methods, claims, and conclusions are non-overlapping.  For example, as stated in 1., while \citet{black2022selective}'s use of successful ensembles is intended to address individual-level arbitrariness, by relying on traditional bagging (simple-majority vote ensembling) and the definition of variance from \citet{domingos2000icml} that encodes a main prediction, arbitrariness gets pushed into the aggregation rule. If they can estimate the mode prediction reliably, they do not abstain; the mode, however, may still be effectively arbitrary. Our measure of arbitrariness is more direct and more configurable. We can avoid such degenerate situations, as in the example we give for making reliable but arbitrary predictions in \citet{black2022selective}.
    \item We also describe a method for recursively ensembling in order to achieve different trade-offs between abstention and prediction. This type of strategy is absent from \citet{black2022selective}.
\end{enumerate}

\custompar{Deep learning} \citet{qian2021variance} is work that came after \citet{forde2021disparate}. They, too, do a wide-ranging empirical study of variance and fairness, but focus on deep learning settings. As a result, they are not examining the fair classification experimental setup that is most common in the field. They therefore make different claims about reliability, which have a similar flavor as those that we make here. However, because of our setup, we are able to probe these claims much deeper (due in part to model/ problem size and being able to limit non-determinism solely to sampling the training data). We mention this work because of its close relationship to \citet{forde2021disparate}, which in part inspired this study.

\citet{ko2023fairensemble} is another deep learning fairness paper. It was posted publicly months after our study, and examines non-overlapping settings and tasks. While the results are similar --- we find fairness after ensembling --- it is again fundamentally different (along the lines of \citet{qian2021variance} and \citet{forde2021disparate}) because it does not study common non-deep-learning setups. They also do not study arbitrariness, which is one of the main purposes of our paper. 

\custompar{Variance in fair classification} \citet{khan2023fairness} is concurrent work that studies the same problem that we study, but also takes a different approach. For one, they bake in a notion of 0-1 loss into their definitions. In this respect, our definition of self-consistency generalizes the definitions in their paper. While they run more types of models than we do (we initially ran more, but ultimately stopped because the results were largely similar with more common model types), they do not cover as many datasets as we do. They also do not study arbitrariness or abstention-based ensembling to deal with it, and they do not release a dataset. Further, based on the fact that they study fewer empirical tasks than we do, and that they do not examine abstention-based ensembling, they do not surface or make claims about the experimental reliability issues that we observe. They do not make claims about the fundamental problem that we observe: \textbf{That variance is the culprit for much observed algorithmic unfairness in classification; in practice, we do not seem to learn very confident decisions for large portions of the datasets we examine, and this is a key problem that has been masked by current common experimental practices in the field}. We make notes about this in our Ethics Statement. 

\custompar{Other work} Any other work on variance and fairness \textbf{comes after} the present study. We have made a significant attempt to keep our related work section up-to-date in response to this new work. We have used a detailed and robust mixed of Google alerts and scraping arXiv to find new related work. We used this same procedure to make sure we  found (ideally) all related work on fairness and variance when we conducted this project. There are some studies, which directly build on ours, which we choose not to cite.\looseness=-1

\section{\appalgorithm}\label{app:sec:algorithm}

A natural question is to see if we can improve self-consistency, with the hope that doing so would reduce arbitrariness in the learning process, improve accuracy, and, for the cases in which there is different self-consistency across subgroups, also perhaps improve fairness. To do so, we consider ways of reducing variance, as, based on our definitions (Definition~\ref{def:variance} and~\ref{def:sc}), doing so should improve self-consistency. 

We  consider the classic  \textit{b}ootstrap \textit{agg}regation --- or, \textit{bagging} --- algorithm~\cite{breiman1996bagging} as a starting point. 
It has been well-known since \citet{breiman1996bagging} that \textit{bagging}  can improve the performance of unstable predictors. That is, for models produced by a learning process that is sensitive to the underlying training data, it is (theoretically-grounded) good practice to train an ensemble of models using bootstrapping (Appendix~\ref{app:sec:prelim:boot};~\citet{efron1979bootstrap, efron1993bootsrap}). When classifying an example instance, we then leverage the whole ensemble by aggregating the predictions produced by its members. This aggregation process identifies the most common prediction in the ensemble, and returns that label as the classification. Put differently, we have combined the information of a lot of unstable classifiers, and averaged over their behavior in order to generate more stable classifications. 

Given the the relationship between variance (Definition~\ref{def:variance}) and self-consistency (Definition~\ref{def:sc}), reducing variance will improve self-consistency. However, rather than relying on a simple-majority-vote to decide the aggregated prediction, we also  will instill a notion of confidence in our predictions by requiring a minimum level of self-consistency, which is described in Algorithm~\ref{algo:bagging-confidently}.

\subsection{Self-consistent ensembling with abstention}\label{app:algo:sc}

We present a framework that alters the semantics of classification outputs to $0$, $1$, and \texttt{Abstain}, and employ ensembling to determine the $\hatsc$-level that guides the output process. We modify bagging from using a simple-majority-vote because this type of aggregation rule still allows for arbitrariness. If, for example, we happen to train $\boot=101$ classifiers, it is possible that 50 of them yield one classification and the other 51 yield the other classification for a particular example. Bagging would select the classification that goes along with the 51 underlying models; however, if we happened to train $\boot=103$ models, it is perhaps the case that the majority vote would flip. In short, the bagging aggregation rule bakes in the idea that simple-majority voting is a sufficient strategy for making decisions. And while this may generally be true for variance reduction in high-variance classifiers, it does not address the problem of arbitrariness that we study. It just encodes arbitrariness in the aggregation rule --- it picks classifications, in some cases, that are no better than a coin flip.

Instead, Algorithm~\ref{algo:bagging-confidently} is more flexible. It suggests many possible ways to produce bagged classifiers that do not have to rely on simple-majority voting, by allowing for abstentions. For example, we can change the aggregation rule in regular bagging to use a self-consistency level $\kappa$ rather than majority vote. Instead of relying on votes, we can bag the underlying prediction probabilities and then apply $\kappa$ a filter. We could take the top-$n$ most consistent predictions and let a super-ensemble of underlying bagged classifiers decide whether to abstain or predict.

In the experiments in the paper, we provide two examples: Changing the underlying bagging vote aggregation rule (simple ensembling), and applying a round of regular bagging to do variance reduction and then bagging the bagged outputs (super ensembling) to apply a self-consistency threshold. Our ensemble model will not produce predictions for examples for which the lack of self-consistency is too high. We describe our procedure more formally in Algorithm~\ref{algo:bagging-confidently}.

\custompar{Simple proof that abstention improves self-consistency (by construction)} We briefly show the simple proof that any method that meets the semantics of Algorithm~\ref{algo:bagging-confidently} will be more self-consistent than its counterpart that cannot \texttt{Abstain}. 

We define abstentions to be in agreement with both $0$ and $1$ predictions. This makes sense intuitively: Algorithm~\ref{algo:bagging-confidently} abstains to avoid making predictions that lack self-consistency, so abstaining should not increase disagreement between predictions. 

It follows that we can continue to use Definition~\ref{def:sc} and associated empirical approximations $\hatsc$ (\ref{eq:hatsc}), 
but with one small adjustment. Instead of the total number of predictions $\boot = \boot_0 + \boot_1$, with $\boot_0$ and $\boot_1$ corresponding to $0$ and $1$ predictions, respectively, we now allow for $\boot \geq \boot_0 + \boot_1$, in order to account for possibly some non-zero number of abstentions. 

In more detail, let us denote $\hat{\sY}$ to be the multiset of predictions for models $\model_{\dataset_1}, \model_{\dataset_2}, \ldots, \model_{\dataset_\boot}$ on $(\instance, \group)$, with $|\hat{\sY}| = \boot = \boot_0 + \boot_1 + \boot_{\texttt{Abstain}}$. This is where we depart from our typical definition of self-consistency, for which $\boot= \boot_0 + \boot_1$ (Section~\ref{sec:significance}, Appendix~\ref{app:sec:consistency}). We continue to let $\boot_0$ and $\boot_1$ represent the counts of $0$ and $1$ predictions, respectively, and now include $\boot_{\texttt{Abstain}}$ to denote the (possibly nonzero) number of abstentions. This leads to the following adjustment of (\ref{eq:hatsc}): \looseness=-1

\begin{align}
	\label{eq:sc-abstain}
	\hatconsistency = 1 - \frac{2(\boot_0\boot_1 + \boot_0\boot_{\texttt{Abstain}} + \boot_1\boot_{\texttt{Abstain}})}{\boot(\boot-1)}.
\end{align}

\noindent Equation (\ref{eq:sc-abstain}) follows from a similar analysis of comparing $0$s, $1$s, and abstentions for Definition~\ref{def:sc}, which lead us to derive (\ref{eq:hatsc}) in Appendix~\ref{app:sec:consistency}. However, since the costs of $0$-to-\texttt{Abstain} comparisons and $1$-to-\texttt{Abstain} comparisons are both 0, the $\boot_0\boot_{\texttt{Abstain}}$ and $\boot_1\boot_{\texttt{Abstain}}$ terms in (\ref{eq:sc-abstain}) reduce to 0. As a result, we yield our original definition for self-consistency (\ref{eq:hatsc}), with the possibility that $\boot=\boot_0 + \boot_1 + \boot_{\texttt{Abstain}} > \boot_0 + \boot_1$, if there is a nonzero number of abstentions $\boot_{\texttt{Abstain}}$. 

Since $\boot>1$ and $\boot_0, \boot_1, \boot_{\texttt{Abstain}} \ge 0$, it is always the case that option to \texttt{Abstain} is at least as self-consistent as not having the option to do so. This follows from the fact that $\boot_0 + \boot_1 + \boot_{\texttt{Abstain}} = \boot \geq \boot_0 + \boot_1$, which would make the denominator in (\ref{eq:sc-abstain}) greater than or equal to the corresponding method that cannot \texttt{Abstain}; when subtracted from 1, this would produce a $\hatsc$ that is no smaller than the value for the corresponding method without that cannot \texttt{Abstain}.

Now, it follows that, given the choice between \texttt{Abstain} and predicting a label that is in disagreement with an existing prediction label, choosing to \texttt{Abstain} will always lead to higher self-consistency. This is because the cost to \texttt{Abstain} is less than disagreeing, so it will always be the minimal choice that maximizes $\hatsc$. 



\custompar{Error and the abstention set} It is very straightforward to see that the \emph{abstention set} will generally exhibit higher than the \emph{prediction set}. When we ensemble and measure $\hatsc$, the exmaples that exhibit low $\hatsc$ contain higher variance-induced error. Let us call the size of the abstention set $U$ (which incurs error $u$), the size of the prediction set $V$ (which incurs error $v$), and the size of the test set $T$ (which incurs error $t$). We can relate the total number of misclassified examples as $T * t = U * u + V * v,$ with $T = U + V$. If we assume the bias and noise are equally distributed across the test and abstention sets (this is a reasonable assumption, on average, in our setup), then splitting off the high variance instances from the low variance (high $\hatsc$ instances) requires that $u > v$. The error on the abstention set necessarily has to be larger than the error on the prediction set, in order to retain the above relationship. 
\section{\appexperiments}\label{app:sec:empirical}

The code for the examples in Sections~\ref{sec:intro}, \ref{sec:significance} and \ref{sec:empirical} can be found in \code. This repository also contains necessary and sufficient information concerning reproducibility. At the time of writing, we use \texttt{Conda} to produce environments with associated package-versioning information, so that our results can be exactly replicated and independently verified. We also use the \texttt{Scikit-Learn}~\citep{pedregosa2011} toolkit for modeling and optimization. More details on our choice of models and hyperparameter optimization can be found in our code repository, cited above. In brief, we consulted prior related work (e.g., \citet{chen2018tradeoff}) and performed our own validation for reasonable hyperparameters per model type. We keep these settings fixed to reduce impact on our results, in order to observe in isolation how different training data subsets impact our results. 

During these early runs, we collected information on train accuracy, not just test accuracy; while models ultimately have similar test accuracy in most cases for the same task, they can vary significantly in terms of train accuracy (e.g., for logistic regression, \texttt{COMPAS} is in the low .70s; for random forests, it is in the mid .90s). We do not include these results for the sake of space.\looseness=-1 

This section is organized as follows. We first present information on our datasets, models and code, including our \texttt{HDMA} toolkit (Appendix~\ref{app:sec:experimentsdata}). We then provide details on our setup for running experiments on our cluster (Appendix~\ref{app:sec:cluster}). Appendix~\ref{app:sec:illustrativedetails} contains  more detailed information concerning the experiments performed to produce Figures~\ref{fig:vote} and~\ref{fig:adult-compas-cdf-rfc} in the main paper. 
In Appendix~\ref{app:sec:experiments-algo}, we provide more details on the results presented in Section~\ref{sec:empirical}, as well as additional experiments. Lastly, in  ppendix~\ref{app:sec:fair}, we discuss implications of these results for common fairness Abenchmarks like \texttt{South German Credit}. We conclude that in many cases, without adequate attention to error estimation, it is likely that training and post-processing a single model for fairness on these models likely is a brittle approach to achieve generalizable fairness (and accuracy) performance. Based on our experiments, it seems like high variance can be a significant confounding factor when using a small set of models to draw conclusions about performance --- whether fairness or accuracy. There is an urgent need for future work concerning reproducibility. More specifically, our results indicate that it would be useful to revisit key algorithmic strategies in fair classification to see how they perform in context with more reliable expected error estimation and variance reduction.

\custompar{Note on CDF figures} We show our results in terms of the $\hatsc$ of the underlying bagged models because doing so conveys how Algorithm~\ref{algo:bagging-confidently} makes decisions to predict or abstain.\footnote{The $\hatsc$ CDF of Algorithm~\ref{algo:bagging-confidently}, computed via a \emph{third} round of bootstrapping, has nearly all mass at $\hatsc=1$; it is difficult to visualize.} For both types of ensembling, Algorithm~\ref{algo:bagging-confidently} predicts for all examples captured by the area to the right of the $\kappa$ reference line, and abstains for all examples on the left.\looseness=-1

It is also worth noting (though hopefully obvious) that our CDF plots of $\hatsc$ are not continuous, yet we choose to plot them as interpolated curves. This are discrete because we train a concrete number of models (individual models or bags) --- typically 101 of them --- that we treat as our approximation for $\boot$ when computing $\hatsc$. This means that there are a finite number of $\kappa$-values for $\hatsc$, for which we plot a corresponding concrete number of heights $y$ corresponding to the cumulative proportion of the test set. In this respect, it would perhaps be more precise to plot our curves using a step function, exemplified below (see Appendix~\ref{app:sec:consistency} for the values in $\hat{\sK}$):

\begin{figure}[ht!]
    \centering
    \includegraphics[width=.4\textwidth]{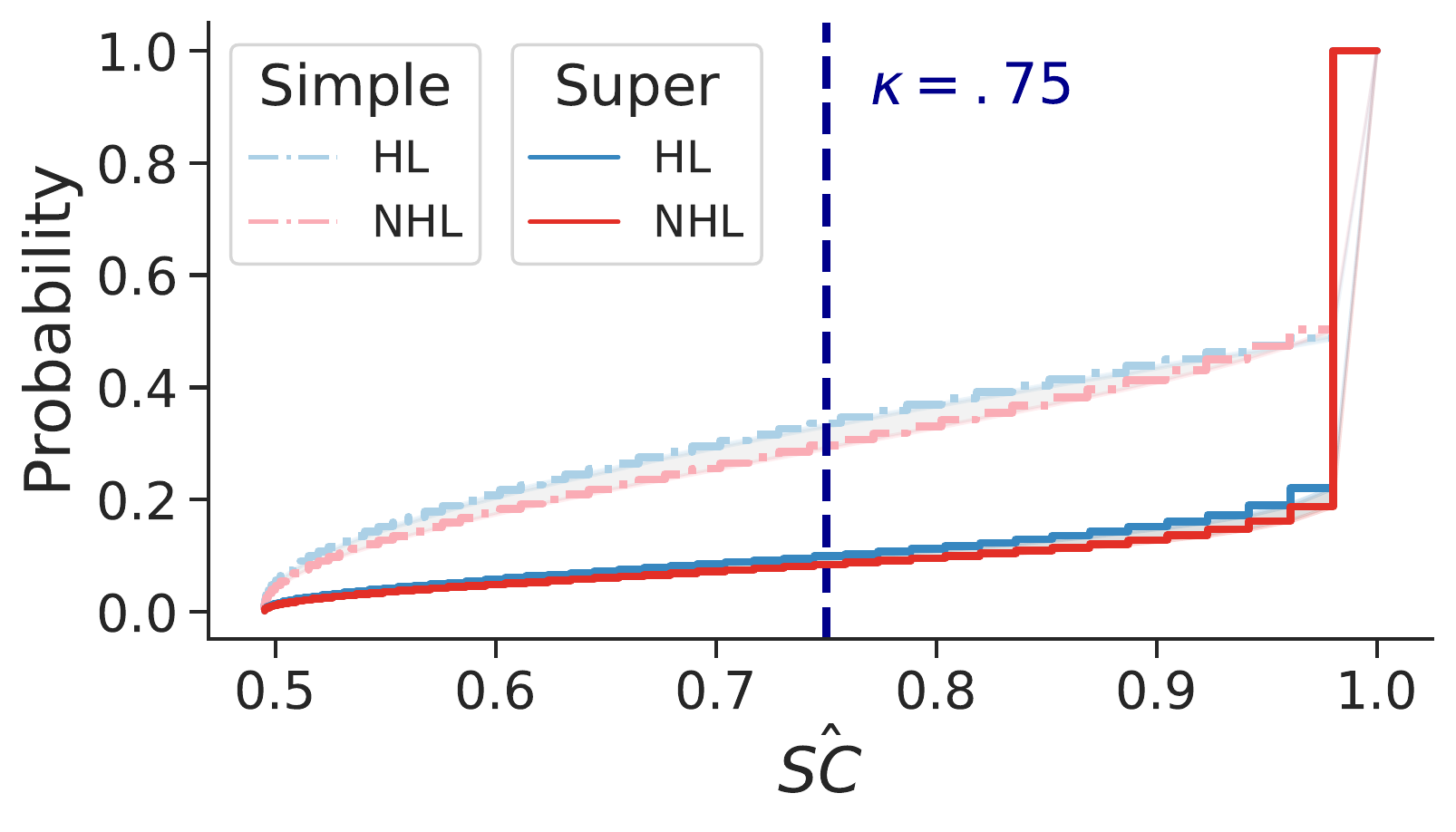}
    \caption{Plotting $\hatsc$ with an emphasis on discrete levels $\kappa$.}
\end{figure}

We opted not to do this for two reasons. First, plotting steps for some of our figures, in our opinion, can make the figures more difficult to understand. Second, in experiments for which we increase the number of models used to estimate $\hatsc$ (e.g., Appendix~\ref{app:sec:fair}), we found that the curves for 101 models were a reasonable approximation of the overall CDF. We therefore concluded that plotting the figures without steps was worth the clarity of presentation, with a sacrifice in correctness for the overall takeaways that we intend with these figures.

\custompar{A remark on cost} It can be considerably more computationally intensive to train an ensemble of models to compute $\hatsc$ than to train a handful of models and perform cross-validation, as is the standard practice in fair classification. 
However, as our empirical analysis demonstrates, 
this cost comes with a huge benefit: It enables us to improve self-consistency and to root out the arbitrariness of producing predictions that are effectively close-to-random, which is especially important in high-stakes fairness settings~\citep{cooper2021eaamo}. Moreover, for common fair classification datasets, the increased cost on modern hardware is relatively small; 
(super-) ensembling with confidence takes under an hour to execute (Appendix~\ref{app:sec:experiments-algo}).

\subsection{\appexperimentsdata}\label{app:sec:experimentsdata}

\custompar{Models} According to a comprehensive recent survey study~\cite{fabris2022datasets}, as well as related work like \citet{chen2018tradeoff}, we conclude that some of the most common models used in fair classification are logistic regression, decision tree classifiers, random forest classifiers, SVMs, and MLPs. We opted to include comprehensive results for the first three, since they capture different complexities, and therefore encode different degrees of statistical bias, that we expected to have an impact on the underlying sources of error. We provide some results for SVMs and MLPs, which we include in this Appendix. Since we choose not to use stochastic optimizers to reduce the sources of randomness, for our results, training MLPs is slower than it could be. We consistently use a decision threshold of 0.5 (i.e., 0-1 loss) for our experiments, though our results can easily be extended to other thresholds, as discussed in Section~\ref{sec:significance}. Depending on the dataset, we reserve between 20\% and 30\% of the available data for the test set. This is consistent with standard fair classification training settings, which we validated during our initial experiments to explore the space (for which we also did preliminary hyperparameter optimization, before fixing the hyperparameters for our presented results).\footnote{Please refer to \code{} for more details.}

\custompar{Datasets}  Also according to \citet{fabris2022datasets}, the most common tasks in fair classification are \texttt{Old Adult}~\cite{kohavi1996oldadult}, \texttt{COMPAS}~\cite{larson2016propublica}, and \texttt{South German Credit}~\cite{gromping2019german}.\footnote{Technically, \citet{gromping2019german} is an updated and corrected version of the dataset from 2019.} These three datasets arguably serve as a \emph{de facto} benchmark in the community, so we felt the need to include them in the present work. 
In recognition of the fact that these three datasets, however standard, have problems, we also run experiments on $3$ tasks in the \texttt{New Adult} dataset, introduced by \citet{ding2021adult} to replace \texttt{Old Adult}. We subset to the \texttt{CA} (California) subset of the dataset, and run on \texttt{Income}, \texttt{Employment}, and \texttt{Public Coverage}, and consider \texttt{sex} and \texttt{race} as protected attributes, which we binarize into \{Male, Female\} and \{White, Non-white\}. These are all large-scale tasks, at least in the domain of algorithmic fairness --- on the order of hundreds of thousands of example instances. However, the $3$ tasks do share example instances and some features. In summary, concerning common tasks in fair classification:

\begin{itemize}[topsep=2pt, itemsep=2pt]
\item \texttt{COMPAS}~\cite{larson2016propublica}. We run on the commonly-used version of this dataset from \citet{friedler2019datasets}, which has 6167 example instances with 404 features. The target is to predict recidivism within 2 years ($1$ corresponding to Yes, and $0$ to No). The protected attribute is \texttt{race}, binarized into ``Non-white'' ($0$) and ``White'' ($1$) subgroups.

\item \texttt{Old Adult}~\cite{kohavi1996oldadult}. We run on the commonly-used version of this dataset from \citet{friedler2019datasets}, which has 30,162 examples with 97 features. This version of the dataset removes instances with missing values from the original dataset, and changes the encoding of some of the features (\citet{kohavi1996oldadult} has 48842 example instnaces with 88 features). The target is to predict $<\$50,000$  income ($0$) $>=\$50,000$ income ($1$). The protected attribute is \texttt{sex}, binarized into ``Female'' ($0$) and ``Male'' ($1$) subgroups.

\item \texttt{South German Credit}~\cite{gromping2019german}. We download the dataset from UCI\footnote{See \texttt{https://archive.ics.uci.edu/ml/datasets/South+German+Credit+\%28UPDATE\%29}} and process the data ourselves. We use the provided \texttt{codetable.txt} to ``translate'' the features from German to English. We say ``translate'' because the authors took some liberties, e.g., the column converted to ``credit\_history'' is labeled ``moral'' in the German, which is not a translation. There are four categories in the protected attribute ``personal\_status\_sex'' column, one of which ($2$) is used for both ``Male (single)'' and ``Female (non-single).'' We therefore remove rows with this value, and binarize the remaining three categories into ``Female'' ($0$) and ``Male'' ($1$). What results is a dataset with 690 example instances (of the original 1000) with 19 features. The target is ``good'' credit ($1$) and ``bad'' credit ($0$).

\item \texttt{Taiwan Credit}~\cite{yeh2009taiwan}. This task is to predict default on credit card payments ($1$) or not ($0$). There are 30,000 example instances and 24 features. The protected attribute is binary \texttt{sex}. We download this dataset from UCI.\footnote{See  \texttt{https://archive.ics.uci.edu/ml/datasets/default+of+credit+card+clients}}.

\item \texttt{New Adult}~\cite{ding2021adult}. This dataset contains millions of example instances from US Census data, which can be used for several different targets/tasks. We select three of them (listed below). These tasks share some features, and therefore are not completely independent. Further, given the size of the whole dataset, we subset to  \texttt{CA} (California), the most populous state in the US. There are two protected attribute columns that we use: \texttt{sex}, which is binarized ``Female'' ($0$) and ``Male'' ($1$) subgroups, and \texttt{race}, which we binarize into ``Non-white'' ($0$) and ``White'' ($1$). In future work, we would like to explore extending our results beyond binary subgroups.
\begin{itemize}
    \item \texttt{Income}. This task is designed to be analogous to \texttt{Old Adult}~\cite{kohavi1996oldadult}. As a result, the target is to predict  $<\$50,000$  income ($0$) $>=\$50,000$ income ($1$). In the \texttt{CA} subset, there are 195,665 example instances with 8 features.
    \item \texttt{Employment}. This task is to predict whether an individual is employed ($1$) or not ($0$). In the \texttt{CA} subset, there are 378,817 example instances with 14 features. 
    \item \texttt{Public Coverage}. This task is to predict whether an individual is on public health insurance ($1$) or not ($0$). In the \texttt{CA} subset, there are 138,554
    example instances with 17 features.

\end{itemize}
\end{itemize}

\subsubsection{\apphmda}\label{app:sec:hmda}
In addition to the above standard tasks, we include experiments that use the \texttt{NY} and \texttt{TX} 2017 subsets of the the Home Mortgage Data Disclosure Act (\texttt{HMDA}) 2007-2017 dataset~\cite{ffiec2022housingdata}. These two datasets have 244,107 and 576,978 examples, respectively, with 18 features. The \texttt{HMDA} datasets together contain over 140 million examples of US home mortgage loans from 2007-2017 (newer data exists, but in a different format). We developed a toolkit, described below, to make this dataset easy to use for classification experiments. Similar to \texttt{New Adult}, we enable subsetting by US state. For the experiments in this paper, we run on the \texttt{NY} (New York) and \texttt{TX} (Texas) 2017 subset, in order to add some geographic diversity to complement our \texttt{New Adult} experiments. We additionally chose \texttt{NY} and \texttt{TX} because they are two of the most populous states in the US, alongside \texttt{CA}.\footnote{Per the 2020 Census, the top-4-most-populous states are \texttt{CA}, \texttt{TX}, \texttt{FL}, and \texttt{NY}~\cite{mackun2021census}.}

The target variable, \texttt{action\_taken}, concerning loan origination has 8 values, 2 of which we cannot meaningful conclude approval or denial decisions. They are: Action Taken: $1$ -- Loan originated, $2$ -- Application approved but not accepted, $3$ -- Application denied by financial institution, $4$ -- Application withdrawn by applicant, $5$ -- File closed for incompleteness, $6$ -- Loan purchased by the institution, $7$ -- Preapproval request denied by financial institution, and $8$ -- Preapproval request approved but not accepted (optional reporting). We filter out $4$ and $6$, and binarize into \texttt{grant}=$\{1, 2, 8\}=1$ and \texttt{reject}=$\{3, 5, 7\}=0$. There are three protected attributes that we consider: \texttt{sex}, \texttt{race}, and \texttt{ethnicity}:
\begin{itemize}[topsep=2pt]
    \item \texttt{sex} has 5 possible values, 2 of which correspond to categories/non-missing values: Male -- $1$ and Female -- $2$. We binarize \texttt{sex} into $\text{F}=0$ and $\text{M}=1$. 
    \item \texttt{race} has 8 possible values, 5 of which correspond to categories/ non-missing information: $1$ -- American Indian or Alaska Native, $2$ -- Asian, $3$ -- Black or African American, $4$ -- Native Hawaiian or Other Pacific Islander, and $5$ -- White. There are 5 fields for applicant race, which model an applicant belonging to more than one racial group. For our experiments, we only look at the first field. When we binarize \texttt{race}, $\text{NW}=0$ and $\text{W}=1$. 
    \item \texttt{ethnicity} has 5 possible values, 2 of which correspond to categories/ non-missing information: $1$ -- Hispanic or Latino and $2$ -- Not Hispanic or Latino. We binarize \text{ethnicity} to be $\text{HL}=0$ and $\text{NHL}=1$. 
\end{itemize}

After subsetting to only include examples that have values that do not correspond to missing information, \texttt{HMDA} has 18 features. The \texttt{NY} dataset has 244,107 examples; the \texttt{TX} dataset has 576,978 examples, making it the largest dataset in our experiments. As with our experiments using \texttt{New Adult}, we would like to extend our results beyond binary subgroups and binary classification in future work.

\custompar{Releasing a standalone toolkit} These datasets are less-commonly used in current algorithmic fairness literature~\cite{fabris2022datasets}. We believe this is likely due to the fact that the over-100-million data examples are only available in bulk files, which are on the order of 10s of gigabytes and therefore not easily downloadable or explorable on most personal computers. Following the example of \citet{ding2021adult}, one of our contributions is to pre-process all of these datasets --- all locations and years --- and release them with a software toolkit. The software engineering effort to produce this toolkit was substantial. Our hope is that wider access to this dataset will further reduce the community's dependency on small (and dated) datasets. Please refer to \hmda{} for the latest information on this standalone software package. Our release aligns with the terms of service for this dataset.\looseness=-1

\subsection{\appcluster}\label{app:sec:cluster}

While most of the experiments run in this paper can be easily reproduced on a modern laptop, for efficiency, we ran all of our experiments (except the one to produce Figure~\ref{fig:vote}) in a cluster environment. This enabled us to easily execute train/test splits $n$ in parallel on different CPUs, serialize our results, and then reconstitute and combine them to produce plots locally. Our cluster environment runs Ubuntu 20.04 and uses Slurm v20.11.8 to manage jobs. We ran all experiments using \texttt{Anaconda3}, which is why we used \texttt{Conda} to reproduce environments for easy replicability. 

The experiments using \texttt{New Adult} and \texttt{HMDA} rely on datasets that are (in some cases) orders of magnitude larger than the traditional algorithmic fairness tasks. This is one of the reasons why we recommend running on a cluster, and therefore do not include Jupyter notebooks in our repository for these tasks. We also limit our modeling choices to logistic regression, decision tree classifiers, and random forest classifiers for these results due to the expense of training on the order of thousands of models for each experiment. 

\subsection{\appillustrativedetails}\label{app:sec:illustrativedetails}

This appendix provides extended results for the experiments associated in Sections~\ref{sec:intro} and \ref{sec:significance}, which give an intuition for individual- and subgroup-level consistency. The experimental results in the main paper are for logistic regression. We expand the set of models we examine, and  associated discussion of how to interpret comparisons between these results. 

\custompar{Reproducing Figure~\ref{fig:vote}} The experiment to produce this figure in Section~\ref{sec:intro} (also shown in Appendix~\ref{app:sec:consistency}) trains $\boot=10$ logistic regression models on the \texttt{COMPAS} dataset (Appendix~\ref{app:sec:experimentsdata}) using 0-1 loss. We use the bootstrap method to produce each model, which we evaluate on the same test set. We then search for a maximally consistent and minimally consistent individual in the test set, i.e., an individual with $10$ predictions that agree and an individual with $5$ predictions in each class, which we plot in the bar graph. Please refer to the README in \code{} regarding which \texttt{Jupyter} notebook to run to produce the underlying results and figure. The experiments to reproduce this figure can be easily replicated on a laptop.\looseness=-1 

\custompar{Reproducing Figure~\ref{fig:adult-compas-cdf-rfc}} These figures were produced by executing $S=10$ runs of $\boot=101$ bootstrap training replicates to train random forest classifiers for \texttt{Old Adult} and \texttt{COMPAS}. We reproduce these figures below, so that they can be examined and treated in relation to our additional results for decision tree classifiers and logistic regression. For each $s$ run, we take train/test split, bootstrap the train split $\boot=101$ times, and evaluate the resulting model classification decisions on the test set. $\hatsc$ can be estimated from the results across those $101$ models. We Run this process $S=10$ times to produce confidence intervals, shown in the figures below. The intervals are not always clearly visible; there is not a lot of variance at the level of comparing whole runs to each other. Please refer to the README in \code{} regarding which \texttt{Jupyter} notebook to run to produce the underlying results and figure. There are also scripted version of these experiments, which enable them to be run in parallel in a cluster environment.

\custompar{Self-consistency of incorrectly-classified instances} Last, we include figures that underscore how self-consistency is independent from correctness that is measured in terms of observed label alignment. That is, it is possible for an instance $(\instance, \group)$ to be self-consistent and classified incorrectly, with respect to its observed label $\olabel$. We show this using stacked bar plots. For the above experiments, we find the test examples that have the majority of their classifications incorrect ($\pred\neq \olabel$, for $\boot=101$, we find the instances with $\ge 51$ incorrect classifications) and the majority of their classification correct (similarly), and we examine how self-consistent they are. We bucket self-consistency into different levels, and then plot the relative proportion of majority-incorrectly and majority-correctly classified examples according to subgroup. Subgroups in \texttt{COMPAS} exhibit a similar trend, while subgroups in \texttt{Adult Old} exhibit differences, with the heights of the bars corresponding to the trends we plot in our CDF plots. As we note briefly in Section~\ref{sec:significance}, it may be interesting to examine patterns in examples about which learning processes are confident (i.e., highly self-consistent) but wrong in terms of label alignment. If such issues correlate with subgroup, it may be worth testing the counterfactual that such labels are indicative of label bias. We leave such thoughts to future work. 

\begin{figure*}[!h]
    \begin{center}
    \begin{minipage}{\textwidth}
        \centering
        \includegraphics[width=0.7\linewidth]{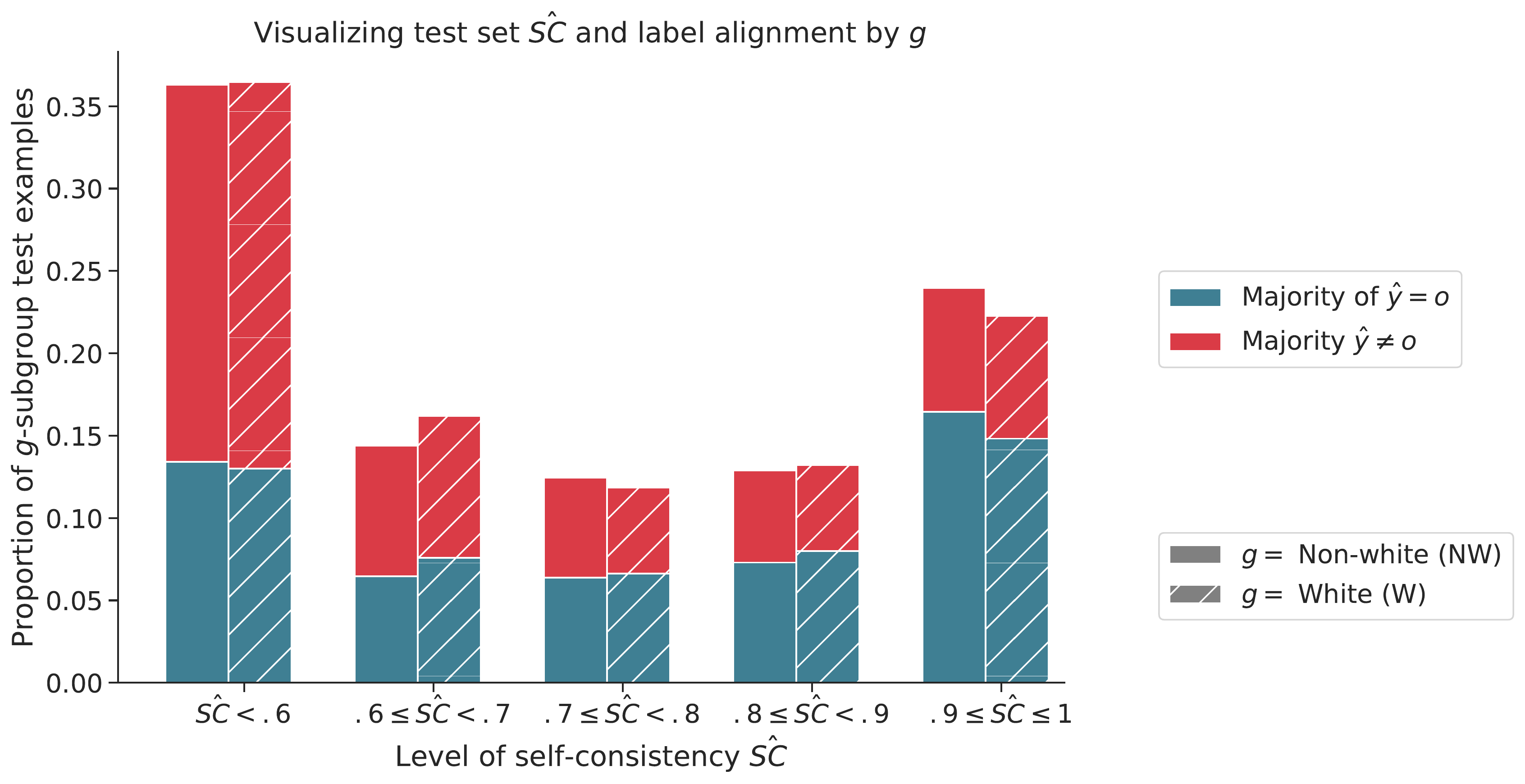}
        \subcaption{\texttt{COMPAS}}
        \label{app:subfig:compas-sc-acc}
        \includegraphics[width=0.7\linewidth]{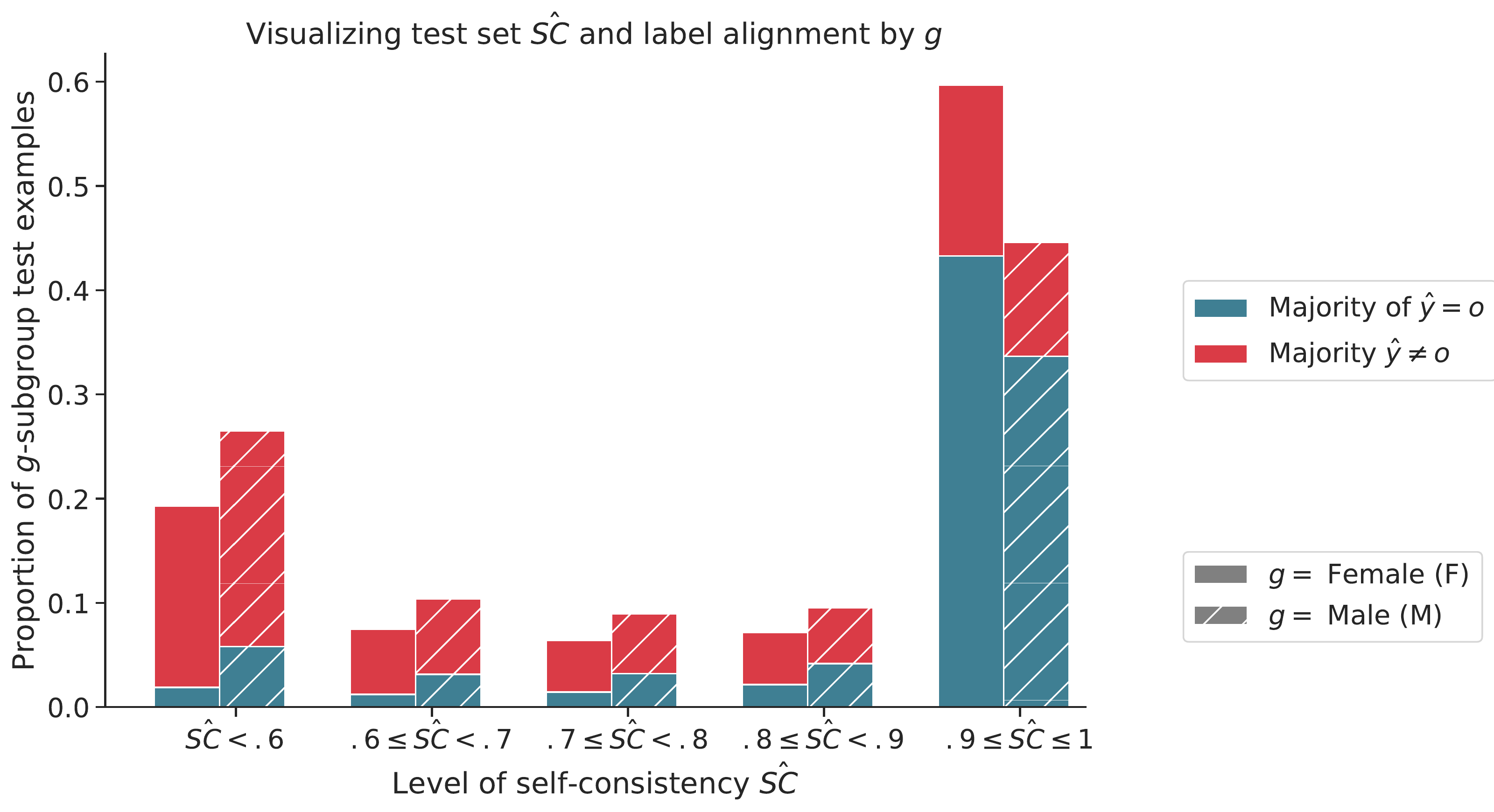}
        \subcaption{\texttt{Adult Old}}
        \label{app:subfig:adult-old-sc-acc}
    \end{minipage}%
    \end{center}
    \vspace{-.35cm}
    \caption{$\hatsc$ broken down by $\group$ and label alignment with the observed label $\olabel$. For each train/test split, and for each $\hatsc$ range ($x$-axis), we find the examples that are incorrectly classified the majority of time ($\geq 5$ splits, we find that $\pred \ne \olabel$), and the examples that are correctly classified the majority of the time ($> 5$, we find that $\pred = \olabel$). We compute the average the proportion over (over splits) in each $\hatsc$ range ($y$-axis). We plot these proportions with respect to subgroup $\group$ (where the sums of the heights of bars for by each $\group$ is equal to $1$).\looseness=-1}
\end{figure*}
\FloatBarrier 
\pagebreak
\newpage
\newpage
\subsection{\appexperimentsalgo}\label{app:sec:experiments-algo}


\subsubsection{\appcompas}\label{app:sec:compas-algo}

$\hatsc$ CDFs for \texttt{COMPAS} ($\group=\texttt{race}$) and associated error metrics on the prediction set. \textbf{Baseline} metrics computed with $\boot=101$ models. For \textbf{simple}, $\boot=101$ models; for \textbf{super}, $\boot=101$ ensemble models, each composed of $51$ underlying models. We repeat for $10$ test/train splits. We also report abstention rate $\hatar$. 

\setlength{\tabcolsep}{6pt}
\begin{figure*}[h!]
\begin{minipage}{.495\linewidth}
\centering
\hspace{-.4cm}
        \includegraphics[width=\linewidth]{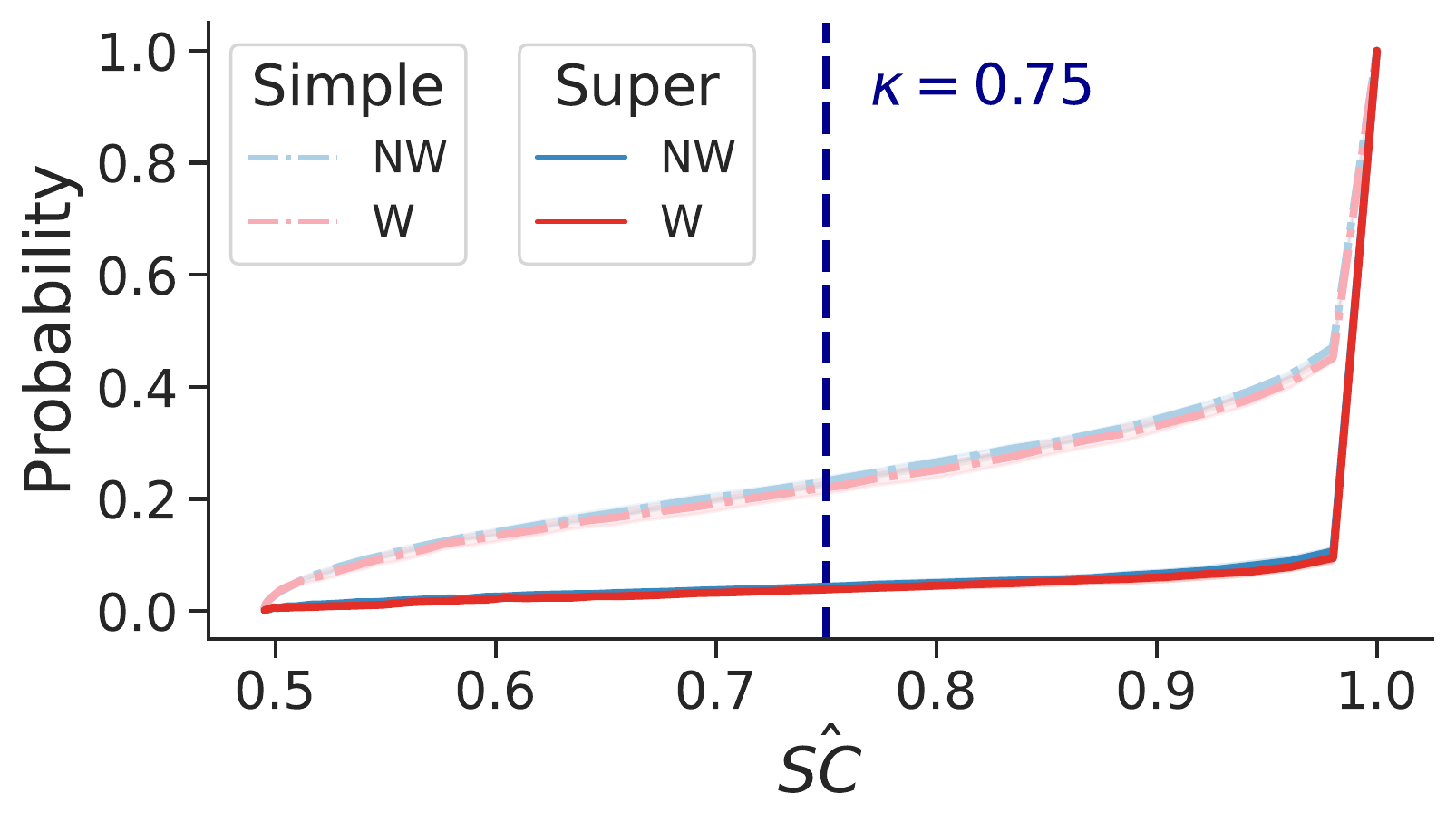}\\
        \vspace{-.1cm}%
	\begin{tabular}{lcc}
		\toprule
		\multicolumn{3}{c}{\textbf{Abstention set metrics}} \\ \cmidrule(lr){1-3}
		 & \textbf{Simple}     & \textbf{Super}      \\ \midrule
		 \textbf{$\Delta\hatar$} & \textcolor{blue}{$1.1\pm0.9\%$} & \textcolor{blue}{$0.5\pm0.0\%$}\\ \midrule
		 \textbf{$\hatar_\text{NW}$} & $23.2\pm1.3\%$ & $4.3\pm0.5\%$\\ \midrule
		 \textbf{$\hatar_\text{W}$} & $22.1\pm2.2\%$ & $3.8\pm0.5\%$\\ \bottomrule
	\end{tabular}
\end{minipage}%
\hspace{.25cm}
\begin{minipage}{.495\linewidth}
\centering
\hspace{-.2cm}
	\begin{tabular}{lccc}
		\toprule
		\multicolumn{4}{c}{\textbf{Logistic regression prediction set metrics}} \\ \cmidrule(lr){1-4}
		 & \textbf{Baseline}     & \textbf{Simple}     & \textbf{Super}      \\ \midrule
		 \textbf{$\Delta\hatpr$} & \textcolor{blue}{$14.5\pm0.3\%$} & \textcolor{blue}{$18.7\pm0.5\%$} & \textcolor{blue}{$15.6\pm0.1\%$}\\ \midrule
		 \textbf{$\hatpr_\text{NW}$} & $45.3\pm1.2\%$ & $43.8\pm1.1\%$ & $44.2\pm0.7\%$\\ \midrule
		 \textbf{$\hatpr_\text{W}$} & $30.8\pm1.5\%$ & $25.1\pm1.6\%$ & $28.6\pm0.6\%$\\ \midrule
		 \textbf{$\Delta\haterr$} & \textcolor{blue}{$0.2\pm0.2\%$} & \textcolor{blue}{$1.1\pm1.5\%$} & \textcolor{blue}{$0.9\pm1.1\%$}\\ \midrule
		 \textbf{$\haterr_\text{NW}$} & $33.0\pm1.3\%$ & $27.9\pm0.9\%$ & $31.0\pm1.0\%$\\ \midrule
		 \textbf{$\haterr_\text{W}$} & $33.2\pm1.1\%$ & $29.0\pm2.4\%$ & $31.9\pm2.1\%$\\ \midrule
		 \textbf{$\Delta\hatfpr$} & \textcolor{blue}{$2.1\pm0.0\%$} & \textcolor{blue}{$3.0\pm0.0\%$} & \textcolor{blue}{$1.8\pm0.2\%$}\\ \midrule
		 \textbf{$\hatfpr_\text{NW}$} & $14.7\pm1.3\%$ & $11.4\pm1.0\%$ & $12.9\pm0.8\%$\\ \midrule
		 \textbf{$\hatfpr_\text{W}$} & $12.6\pm1.3\%$ & $8.4\pm1.0\%$ & $11.1\pm0.6\%$\\ \midrule
		 \textbf{$\Delta\hatfnr$} & \textcolor{blue}{$2.4\pm0.0\%$} & \textcolor{blue}{$4.0\pm1.1\%$} & \textcolor{blue}{$2.8\pm0.8\%$}\\ \midrule
		 \textbf{$\hatfnr_\text{NW}$} & $18.3\pm1.1\%$ & $16.5\pm1.9\%$ & $18.0\pm1.3\%$\\ \midrule
		 \textbf{$\hatfnr_\text{W}$} & $20.7\pm1.1\%$ & $20.5\pm3.0\%$ & $20.8\pm2.1\%$\\ \bottomrule
	\end{tabular}
\end{minipage}%
\caption{\textbf{Logistic regression} on \texttt{COMPAS}\looseness=-1}
\label{fig:compas-lr-all}
\end{figure*}
\vfill
\setlength{\tabcolsep}{6pt}
\begin{figure*}[h!]
\begin{minipage}{.495\linewidth}
\centering
\hspace{-.4cm}
        \includegraphics[width=\linewidth]{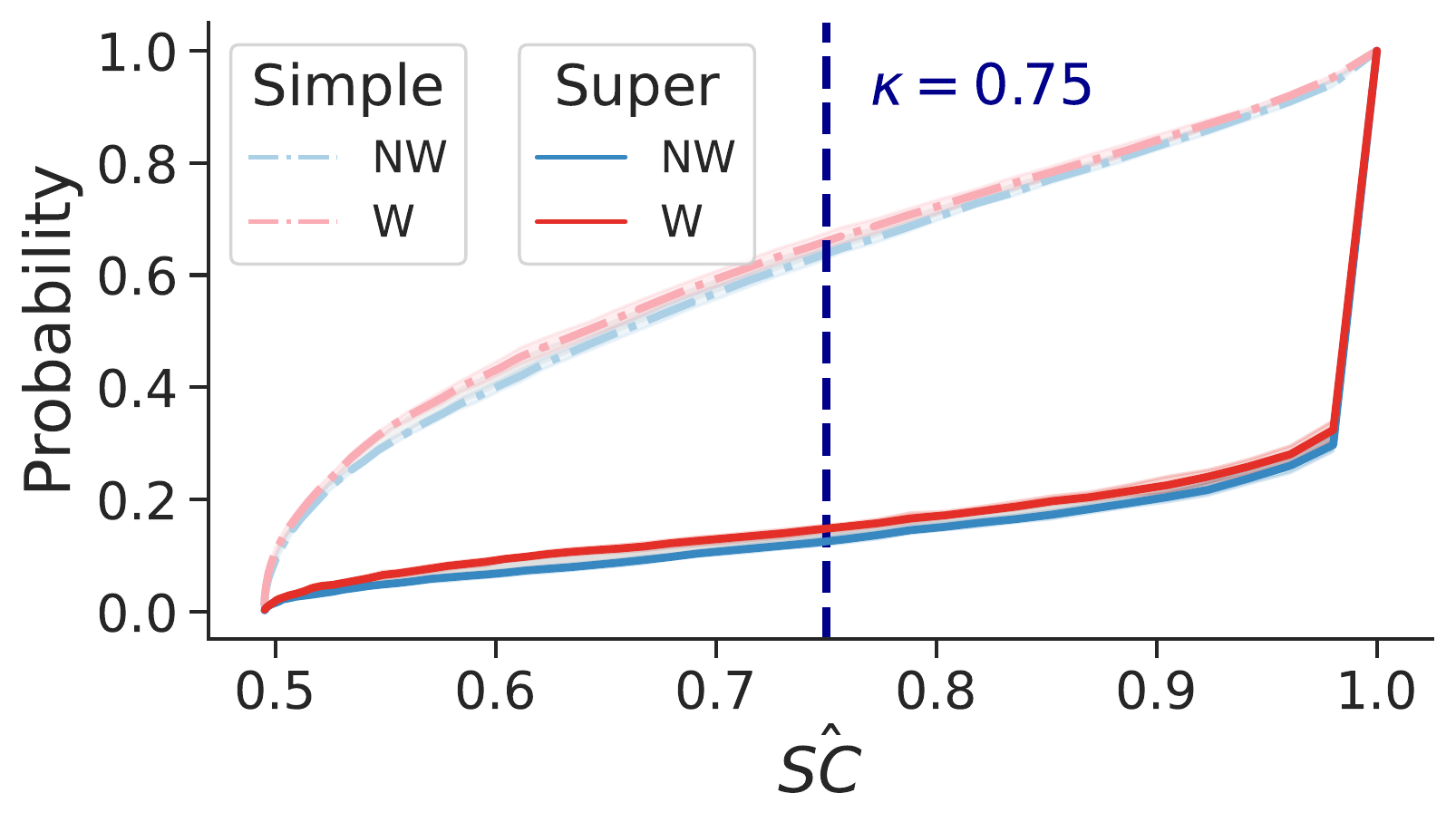}\\
        \vspace{-.1cm}%
	\begin{tabular}{lcc}
		\toprule
		\multicolumn{3}{c}{\textbf{Abstention set metrics}} \\ \cmidrule(lr){1-3}
		 & \textbf{Simple}     & \textbf{Super}      \\ \midrule
		 \textbf{$\Delta\hatar$} & \textcolor{blue}{$1.9\pm1.0\%$} & \textcolor{blue}{$2.3\pm0.1\%$}\\ \midrule
		 \textbf{$\hatar_\text{NW}$} & $62.3\pm1.8\%$ & $12.3\pm0.8\%$\\ \midrule
		 \textbf{$\hatar_\text{W}$} & $64.2\pm2.8\%$ & $14.6\pm0.9\%$\\ \bottomrule
	\end{tabular}
\end{minipage}%
\hspace{.25cm}
\begin{minipage}{.495\linewidth}
\centering
\hspace{-.2cm}
	\begin{tabular}{lccc}
		\toprule
		\multicolumn{4}{c}{\textbf{Decision tree prediction set metrics}} \\ \cmidrule(lr){1-4}
		 & \textbf{Baseline}     & \textbf{Simple}     & \textbf{Super}      \\ \midrule
		 \textbf{$\Delta\hatpr$} & \textcolor{blue}{$10.1\pm0.6\%$} & \textcolor{blue}{$22.9\pm1.7\%$} & \textcolor{blue}{$15.8\pm0.5\%$}\\ \midrule
		 \textbf{$\hatpr_\text{NW}$} & $47.9\pm0.7\%$ & $43.4\pm3.1\%$ & $48.5\pm1.2\%$\\ \midrule
		 \textbf{$\hatpr_\text{W}$} & $37.8\pm1.3\%$ & $20.5\pm1.4\%$ & $32.7\pm1.7\%$\\ \midrule
		 \textbf{$\Delta\haterr$} & \textcolor{blue}{$0.6\pm0.9\%$} & \textcolor{blue}{$1.7\pm0.7\%$} & \textcolor{blue}{$1.2\pm0.8\%$}\\ \midrule
		 \textbf{$\haterr_\text{NW}$} & $38.8\pm0.5\%$ & $24.0\pm0.9\%$ & $32.8\pm0.4\%$\\ \midrule
		 \textbf{$\haterr_\text{W}$} & $38.2\pm1.4\%$ & $22.3\pm1.6\%$ & $31.6\pm1.2\%$\\ \midrule
		 \textbf{$\Delta\hatfpr$} & \textcolor{blue}{$0.2\pm0.4\%$} & \textcolor{blue}{$4.0\pm0.4\%$} & \textcolor{blue}{$2.5\pm0.9\%$}\\ \midrule
		 \textbf{$\hatfpr_\text{NW}$} & $18.8\pm0.8\%$ & $10.4\pm1.8\%$ & $16.1\pm0.9\%$\\ \midrule
		 \textbf{$\hatfpr_\text{W}$} & $18.6\pm1.2\%$ & $6.4\pm1.4\%$ & $13.6\pm1.8\%$\\ \midrule
		 \textbf{$\Delta\hatfnr$} & \textcolor{blue}{$0.3\pm0.3\%$} & \textcolor{blue}{$2.3\pm1.3\%$} & \textcolor{blue}{$1.4\pm0.1\%$}\\ \midrule
		 \textbf{$\hatfnr_\text{NW}$} & $19.9\pm0.7\%$ & $13.6\pm1.0\%$ & $16.6\pm1.3\%$\\ \midrule
		 \textbf{$\hatfnr_\text{W}$} & $19.6\pm1.0\%$ & $15.9\pm2.3\%$ & $18.0\pm1.2\%$\\ \bottomrule
	\end{tabular}
\end{minipage}%
\caption{\textbf{Decision trees} on \texttt{COMPAS}\looseness=-1}
\label{fig:compas-dtc-all}
\end{figure*}

\vspace{.5cm}
\setlength{\tabcolsep}{6pt}
\begin{figure*}[h!]
\begin{minipage}{.495\linewidth}
\centering
\hspace{-.4cm}
        \includegraphics[width=\linewidth]{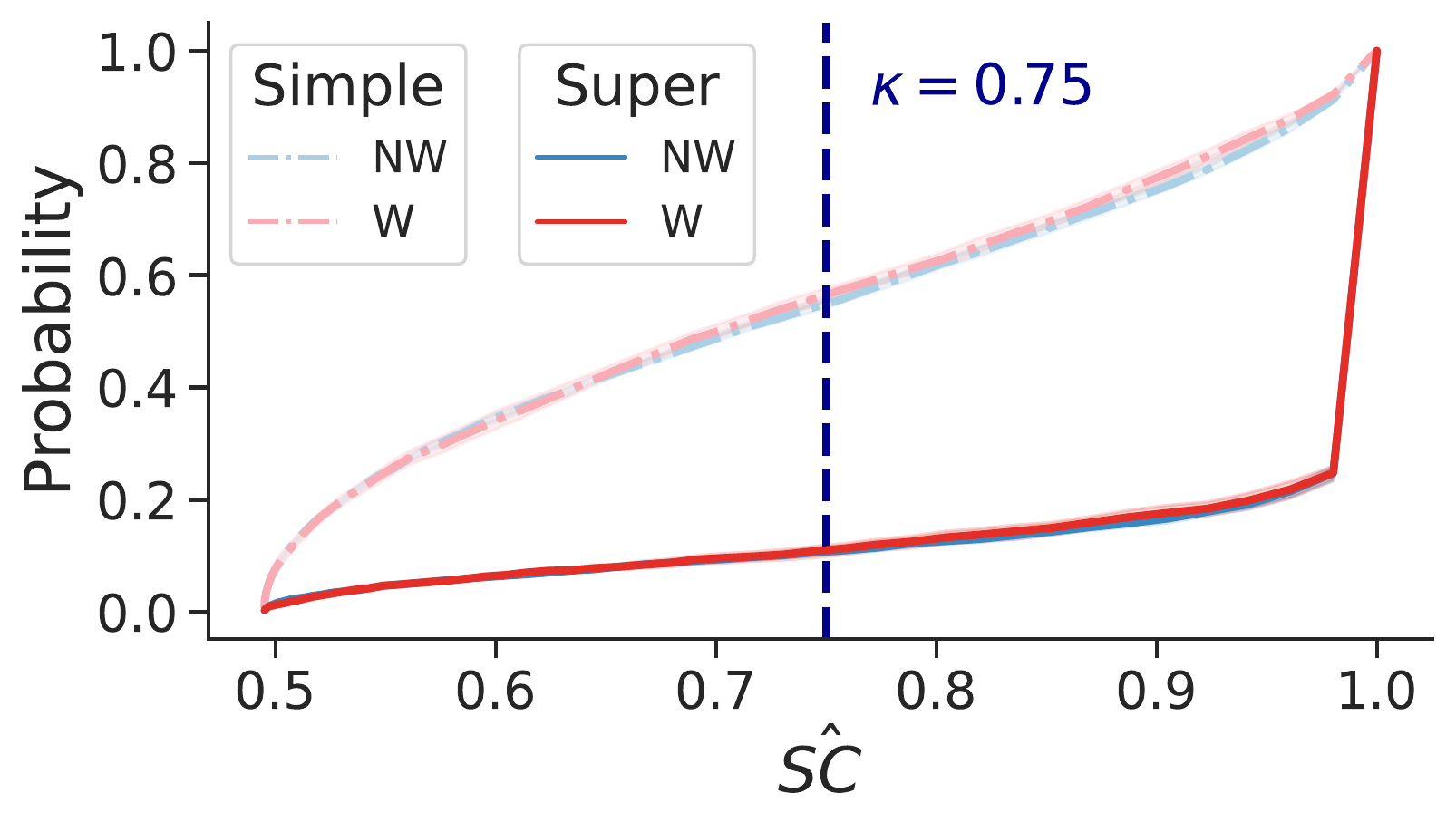}\\
        \vspace{-.1cm}%
	\begin{tabular}{lcc}
		\toprule
		\multicolumn{3}{c}{\textbf{Abstention set metrics}} \\ \cmidrule(lr){1-3}
		 & \textbf{Simple}     & \textbf{Super}      \\ \midrule
		 \textbf{$\Delta\hatar$} & \textcolor{blue}{$0.3\pm0.6\%$} & \textcolor{blue}{$0.2\pm0.7\%$}\\ \midrule
		 \textbf{$\hatar_\text{NW}$} & $53.9\pm1.6\%$ & $10.6\pm0.5\%$\\ \midrule
		 \textbf{$\hatar_\text{W}$} & $53.6\pm2.2\%$ & $10.8\pm1.2\%$\\ \bottomrule
	\end{tabular}
\end{minipage}%
\hspace{.25cm}
\begin{minipage}{.495\linewidth}
\centering
\hspace{-.2cm}
	\begin{tabular}{lccc}
		\toprule
		\multicolumn{4}{c}{\textbf{Random forest prediction set metrics}} \\ \cmidrule(lr){1-4}
		 & \textbf{Baseline}     & \textbf{Simple}     & \textbf{Super}      \\ \midrule
		 \textbf{$\Delta\hatpr$} & \textcolor{blue}{$13.0\pm0.7\%$} & \textcolor{blue}{$24.3\pm0.4\%$} & \textcolor{blue}{$18.6\pm0.5\%$}\\ \midrule
		 \textbf{$\hatpr_\text{NW}$} & $48.0\pm0.6\%$ & $45.6\pm1.7\%$ & $47.8\pm0.9\%$\\ \midrule
		 \textbf{$\hatpr_\text{W}$} & $35.0\pm1.3\%$ & $21.3\pm1.3\%$ & $29.2\pm1.4\%$\\ \midrule
		 \textbf{$\Delta\haterr$} & \textcolor{blue}{$1.0\pm0.8\%$} & \textcolor{blue}{$0.6\pm0.8\%$} & \textcolor{blue}{$2.1\pm1.0\%$}\\ \midrule
		 \textbf{$\haterr_\text{NW}$} & $36.9\pm0.5\%$ & $23.3\pm0.8\%$ & $32.3\pm0.4\%$\\ \midrule
		 \textbf{$\haterr_\text{W}$} & $35.9\pm1.3\%$ & $23.9\pm1.6\%$ & $30.2\pm1.4\%$\\ \midrule
		 \textbf{$\Delta\hatfpr$} & \textcolor{blue}{$2.0\pm0.4\%$} & \textcolor{blue}{$3.2\pm0.0\%$} & \textcolor{blue}{$4.5\pm0.4\%$}\\ \midrule
		 \textbf{$\hatfpr_\text{NW}$} & $18.0\pm0.8\%$ & $10.0\pm1.3\%$ & $15.3\pm1.2\%$\\ \midrule
		 \textbf{$\hatfpr_\text{W}$} & $16.0\pm1.2\%$ & $6.8\pm1.3\%$ & $10.8\pm0.8\%$\\ \midrule
		 \textbf{$\Delta\hatfnr$} & \textcolor{blue}{$0.9\pm0.4\%$} & \textcolor{blue}{$3.7\pm1.2\%$} & \textcolor{blue}{$2.4\pm0.8\%$}\\ \midrule
		 \textbf{$\hatfnr_\text{NW}$} & $19.0\pm0.7\%$ & $13.4\pm1.2\%$ & $16.9\pm1.2\%$\\ \midrule
		 \textbf{$\hatfnr_\text{W}$} & $19.9\pm1.1\%$ & $17.1\pm2.4\%$ & $19.3\pm2.0\%$\\ \bottomrule
	\end{tabular}
\end{minipage}%
\caption{\textbf{Random forests} on \texttt{COMPAS}\looseness=-1}
\label{fig:compas-rfc-all}
\end{figure*}
\FloatBarrier

\subsubsection{\appoldadult}\label{app:sec:adult-algo}

$\hatsc$ CDFs for \texttt{Old Adult} ($\group=\texttt{sex}$) and associated error metrics on the prediction set. \textbf{Baseline} metrics computed with $\boot=101$ models. For \textbf{simple}, $\boot=101$ models; for \textbf{super}, $\boot=101$ ensemble models, each composed of $51$ underlying models. We repeat for $10$ test/train splits. We also report abstention rate $\hatar$. 
\vfill
\setlength{\tabcolsep}{6pt}
\begin{figure*}[h!]
\begin{minipage}{.495\linewidth}
\centering
\hspace{-.4cm}
        \includegraphics[width=\linewidth]{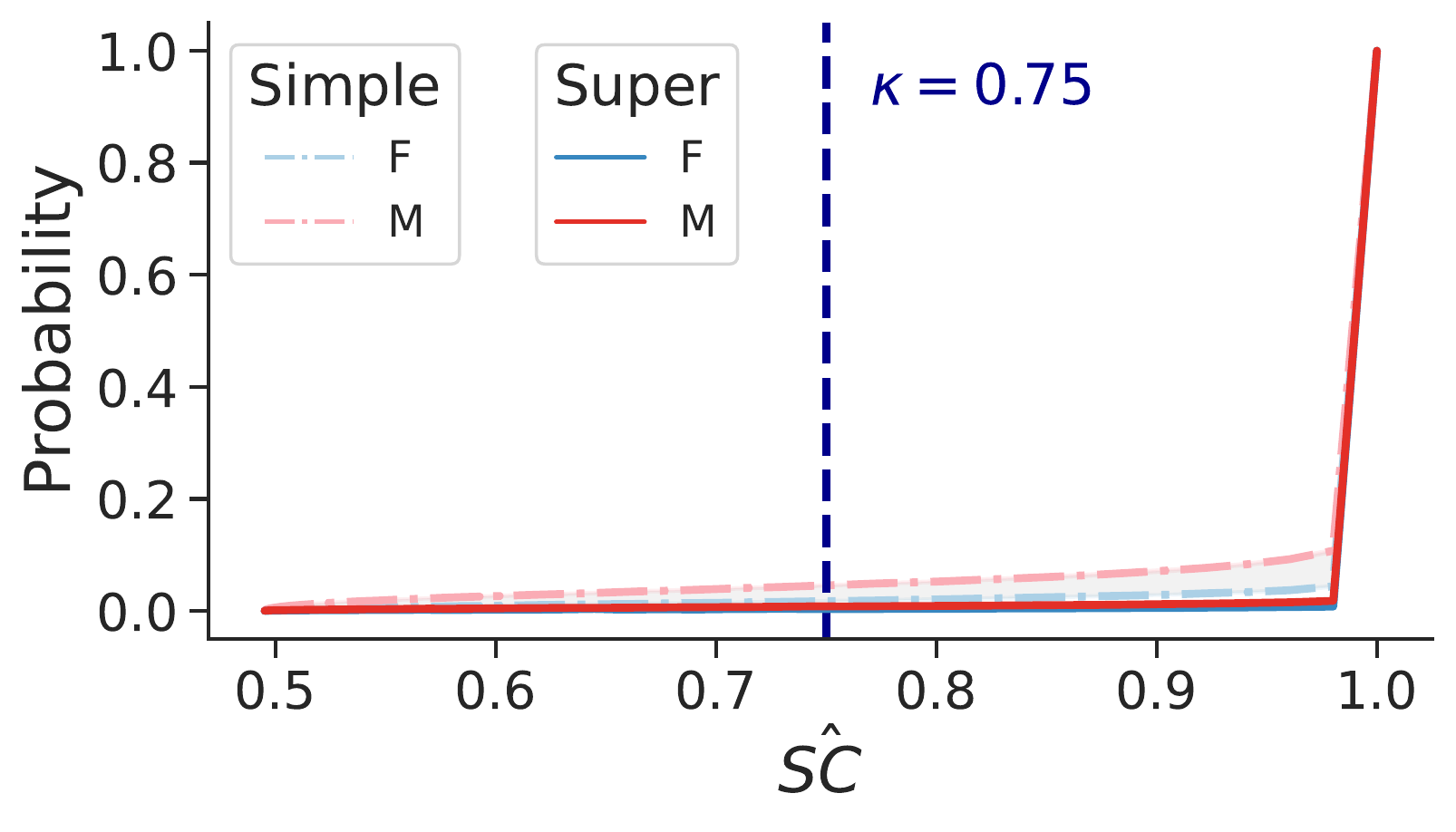}\\
        \vspace{-.1cm}%
	\begin{tabular}{lcc}
		\toprule
		\multicolumn{3}{c}{\textbf{Abstention set metrics}} \\ \cmidrule(lr){1-3}
		 & \textbf{Simple}     & \textbf{Super}      \\ \midrule
		 \textbf{$\Delta\hatar$} & \textcolor{blue}{$2.6\pm0.0\%$} & \textcolor{blue}{$0.5\pm0.0\%$}\\ \midrule
		 \textbf{$\hatar_\text{F}$} & $1.8\pm0.2\%$ & $0.3\pm0.1\%$\\ \midrule
		 \textbf{$\hatar_\text{M}$} & $4.4\pm0.2\%$ & $0.8\pm0.1\%$\\ \bottomrule
	\end{tabular}
\end{minipage}%
\hspace{.25cm}
\begin{minipage}{.495\linewidth}
\centering
\hspace{-.2cm}
	\begin{tabular}{lccc}
		\toprule
		\multicolumn{4}{c}{\textbf{Logistic regression prediction set metrics}} \\ \cmidrule(lr){1-4}
		 & \textbf{Baseline}     & \textbf{Simple}     & \textbf{Super}      \\ \midrule
		 \textbf{$\Delta\hatpr$} & \textcolor{blue}{$18.3\pm0.2\%$} & \textcolor{blue}{$17.8\pm0.1\%$} & \textcolor{blue}{$18.1\pm0.1\%$}\\ \midrule
		 \textbf{$\hatpr_\text{F}$} & $8.2\pm0.3\%$ & $7.1\pm0.4\%$ & $7.6\pm0.4\%$\\ \midrule
		 \textbf{$\hatpr_\text{M}$} & $26.5\pm0.5\%$ & $24.9\pm0.5\%$ & $25.7\pm0.5\%$\\ \midrule
		 \textbf{$\Delta\haterr$} & \textcolor{blue}{$11.3\pm0.1\%$} & \textcolor{blue}{$10.8\pm0.1\%$} & \textcolor{blue}{$11.4\pm0.2\%$}\\ \midrule
		 \textbf{$\haterr_\text{F}$} & $7.8\pm0.4\%$ & $7.0\pm0.3\%$ & $7.5\pm0.2\%$\\ \midrule
		 \textbf{$\haterr_\text{M}$} & $19.1\pm0.3\%$ & $17.8\pm0.4\%$ & $18.9\pm0.4\%$\\ \midrule
		 \textbf{$\Delta\hatfpr$} & \textcolor{blue}{$4.7\pm0.0\%$} & \textcolor{blue}{$4.4\pm0.2\%$} & \textcolor{blue}{$4.8\pm0.2\%$}\\ \midrule
		 \textbf{$\hatfpr_\text{F}$} & $2.3\pm0.3\%$ & $1.6\pm0.1\%$ & $1.8\pm0.1\%$\\ \midrule
		 \textbf{$\hatfpr_\text{M}$} & $7.0\pm0.3\%$ & $6.0\pm0.3\%$ & $6.6\pm0.3\%$\\ \midrule
		 \textbf{$\Delta\hatfnr$} & \textcolor{blue}{$6.7\pm0.1\%$} & \textcolor{blue}{$6.5\pm0.1\%$} & \textcolor{blue}{$6.6\pm0.1\%$}\\ \midrule
		 \textbf{$\hatfnr_\text{F}$} & $5.5\pm0.3\%$ & $5.4\pm0.2\%$ & $5.7\pm0.2\%$\\ \midrule
		 \textbf{$\hatfnr_\text{M}$} & $12.2\pm0.2\%$ & $11.9\pm0.1\%$ & $12.3\pm0.1\%$\\ \bottomrule
	\end{tabular}
\end{minipage}%
\caption{\textbf{Logistic regression} on \texttt{Old Adult}\looseness=-1}
\label{fig:adultold-lr-all}
\end{figure*}
\FloatBarrier
\setlength{\tabcolsep}{6pt}
\begin{figure*}[h!]
\begin{minipage}{.495\linewidth}
\centering
\hspace{-.4cm}
        \includegraphics[width=\linewidth]{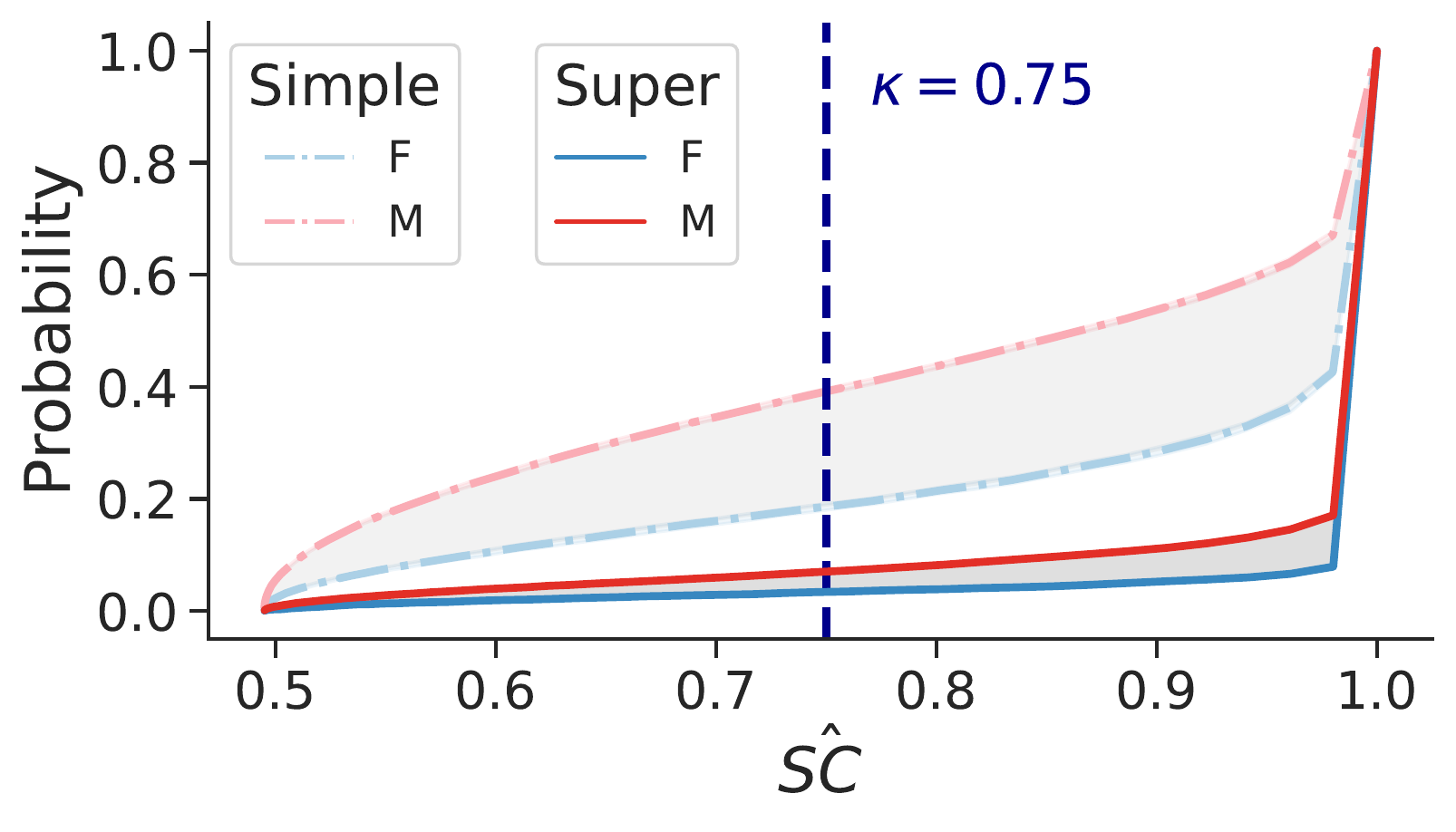}\\
        \vspace{-.1cm}%
	\begin{tabular}{lcc}
		\toprule
		\multicolumn{3}{c}{\textbf{Abstention set metrics}} \\ \cmidrule(lr){1-3}
		 & \textbf{Simple}     & \textbf{Super}      \\ \midrule
		 \textbf{$\Delta\hatar$} & \textcolor{blue}{$19.5\pm0.2\%$} & \textcolor{blue}{$3.5\pm0.1\%$}\\ \midrule
		 \textbf{$\hatar_\text{F}$} & $18.2\pm0.4\%$ & $3.4\pm0.2\%$\\ \midrule
		 \textbf{$\hatar_\text{M}$} & $37.7\pm0.6\%$ & $6.9\pm0.3\%$\\ \bottomrule
	\end{tabular}
\end{minipage}%
\hspace{.25cm}
\begin{minipage}{.495\linewidth}
\centering
\hspace{-.2cm}
	\begin{tabular}{lccc}
		\toprule
		\multicolumn{4}{c}{\textbf{Decision tree prediction set metrics}} \\ \cmidrule(lr){1-4}
		 & \textbf{Baseline}     & \textbf{Simple}     & \textbf{Super}      \\ \midrule
		 \textbf{$\Delta\hatpr$} & \textcolor{blue}{$20.3\pm0.1\%$} & \textcolor{blue}{$18.2\pm0.1\%$} & \textcolor{blue}{$19.9\pm0.1\%$}\\ \midrule
		 \textbf{$\hatpr_\text{F}$} & $12.1\pm0.4\%$ & $4.5\pm0.3\%$ & $7.8\pm0.5\%$\\ \midrule
		 \textbf{$\hatpr_\text{M}$} & $32.4\pm0.5\%$ & $22.7\pm0.2\%$ & $27.7\pm0.4\%$\\ \midrule
		 \textbf{$\Delta\haterr$} & \textcolor{blue}{$12.3\pm0.0\%$} & \textcolor{blue}{$6.0\pm0.1\%$} & \textcolor{blue}{$10.9\pm0.2\%$}\\ \midrule
		 \textbf{$\haterr_\text{F}$} & $10.8\pm0.3\%$ & $3.0\pm0.2\%$ & $6.6\pm0.4\%$\\ \midrule
		 \textbf{$\haterr_\text{M}$} & $23.1\pm0.3\%$ & $9.0\pm0.3\%$ & $17.5\pm0.2\%$\\ \midrule
		 \textbf{$\Delta\hatfpr$} & \textcolor{blue}{$6.2\pm0.1\%$} & \textcolor{blue}{$2.5\pm0.1\%$} & \textcolor{blue}{$5.4\pm0.2\%$}\\ \midrule
		 \textbf{$\hatfpr_\text{F}$} & $5.7\pm0.2\%$ & $0.4\pm0.0\%$ & $1.9\pm0.3\%$\\ \midrule
		 \textbf{$\hatfpr_\text{M}$} & $11.9\pm0.3\%$ & $2.9\pm0.1\%$ & $7.3\pm0.1\%$\\ \midrule
		 \textbf{$\Delta\hatfnr$} & \textcolor{blue}{$6.1\pm0.1\%$} & \textcolor{blue}{$3.4\pm0.0\%$} & \textcolor{blue}{$5.5\pm0.1\%$}\\ \midrule
		 \textbf{$\hatfnr_\text{F}$} & $5.1\pm0.3\%$ & $2.7\pm0.2\%$ & $4.7\pm0.1\%$\\ \midrule
		 \textbf{$\hatfnr_\text{M}$} & $11.2\pm0.2\%$ & $6.1\pm0.2\%$ & $10.2\pm0.2\%$\\ \bottomrule
	\end{tabular}
\end{minipage}%
\caption{\textbf{Decision trees} on \texttt{Old Adult}\looseness=-1}
\label{fig:adultold-dtc-all}
\end{figure*}
\vspace*{1.5cm}
\setlength{\tabcolsep}{6pt}
\begin{figure*}[h!]
	\begin{minipage}{.495\linewidth}
\centering
\hspace{-.4cm}
        \includegraphics[width=\linewidth]{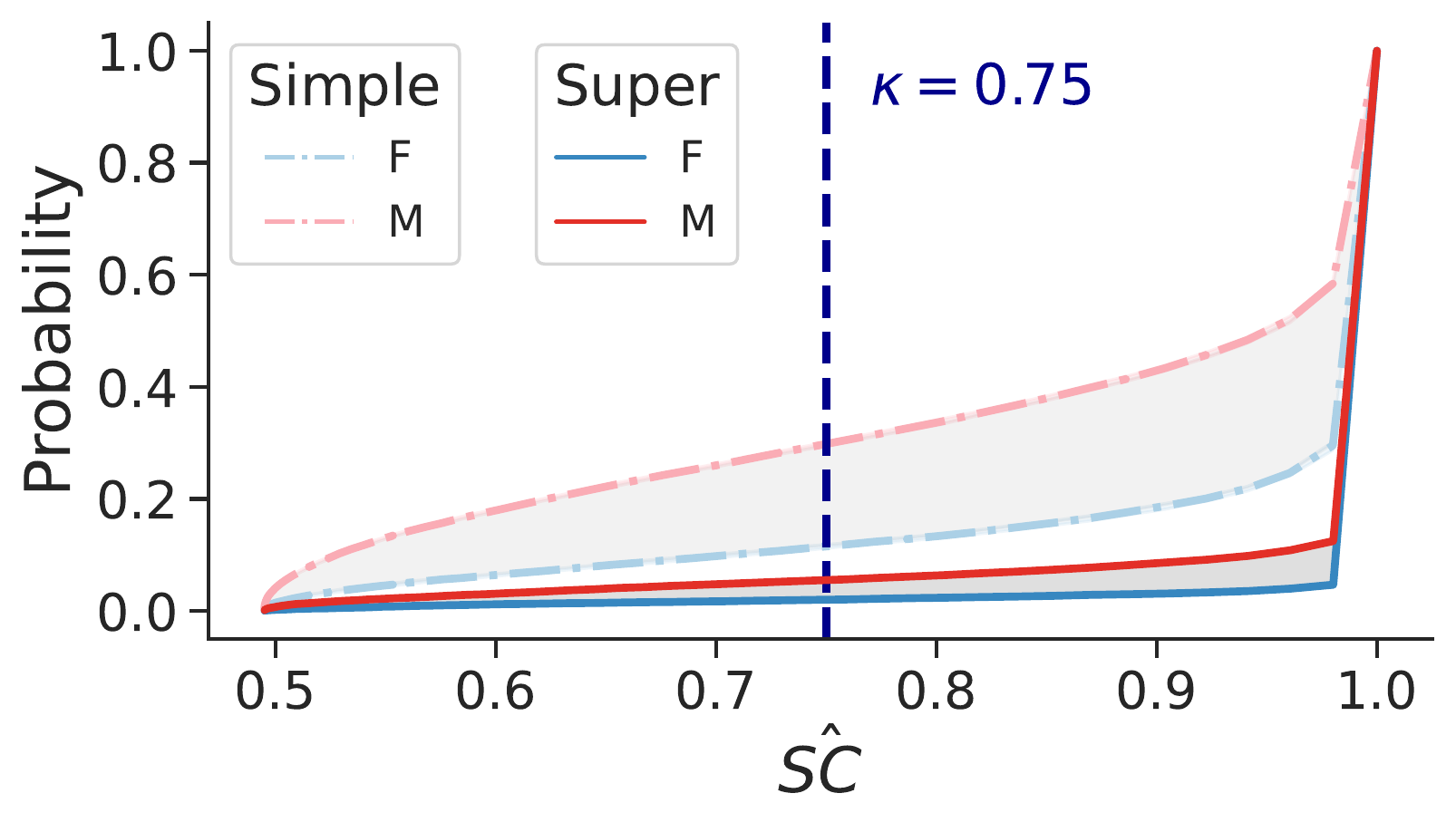}\\
        \vspace{-.1cm}%
	\begin{tabular}{lcc}
		\toprule
		\multicolumn{3}{c}{\textbf{Abstention set metrics}} \\ \cmidrule(lr){1-3}
		 & \textbf{Simple}     & \textbf{Super}      \\ \midrule
		 \textbf{$\Delta\hatar$} & \textcolor{blue}{$17.2\pm0.4\%$} & \textcolor{blue}{$3.4\pm0.1\%$}\\ \midrule
		 \textbf{$\hatar_\text{F}$} & $11.2\pm0.3\%$ & $2.0\pm0.3\%$\\ \midrule
		 \textbf{$\hatar_\text{M}$} & $28.4\pm0.7\%$ & $5.4\pm0.2\%$\\ \bottomrule
	\end{tabular}
\end{minipage}%
\hspace{.25cm}
\begin{minipage}{.495\linewidth}
\centering
\hspace{-.2cm}
	\begin{tabular}{lccc}
		\toprule
		\multicolumn{4}{c}{\textbf{Random forest prediction set metrics}} \\ \cmidrule(lr){1-4}
		 & \textbf{Baseline}     & \textbf{Simple}     & \textbf{Super}      \\ \midrule
		 \textbf{$\Delta\hatpr$} & \textcolor{blue}{$20.0\pm0.2\%$} & \textcolor{blue}{$17.1\pm0.3\%$} & \textcolor{blue}{$19.0\pm0.2\%$}\\ \midrule
		 \textbf{$\hatpr_\text{F}$} & $9.8\pm0.2\%$ & $4.8\pm0.2\%$ & $7.7\pm0.3\%$\\ \midrule
		 \textbf{$\hatpr_\text{M}$} & $29.8\pm0.4\%$ & $21.9\pm0.5\%$ & $26.7\pm0.5\%$\\ \midrule
		 \textbf{$\Delta\haterr$} & \textcolor{blue}{$12.2\pm0.0\%$} & \textcolor{blue}{$6.5\pm0.1\%$} & \textcolor{blue}{$10.7\pm0.0\%$}\\ \midrule
		 \textbf{$\haterr_\text{F}$} & $9.0\pm0.3\%$ & $4.2\pm0.2\%$ & $6.6\pm0.2\%$\\ \midrule
		 \textbf{$\haterr_\text{M}$} & $21.2\pm0.3\%$ & $10.7\pm0.3\%$ & $17.3\pm0.2\%$\\ \midrule
		 \textbf{$\Delta\hatfpr$} & \textcolor{blue}{$6.0\pm0.2\%$} & \textcolor{blue}{$2.5\pm0.1\%$} & \textcolor{blue}{$5.0\pm0.1\%$}\\ \midrule
		 \textbf{$\hatfpr_\text{F}$} & $3.7\pm0.1\%$ & $0.7\pm0.1\%$ & $1.7\pm0.2\%$\\ \midrule
		 \textbf{$\hatfpr_\text{M}$} & $9.7\pm0.3\%$ & $3.2\pm0.2\%$ & $6.7\pm0.3\%$\\ \midrule
		 \textbf{$\Delta\hatfnr$} & \textcolor{blue}{$6.3\pm0.2\%$} & \textcolor{blue}{$4.1\pm0.2\%$} & \textcolor{blue}{$5.8\pm0.1\%$}\\ \midrule
		 \textbf{$\hatfnr_\text{F}$} & $5.3\pm0.3\%$ & $3.5\pm0.1\%$ & $4.9\pm0.2\%$\\ \midrule
		 \textbf{$\hatfnr_\text{M}$} & $11.6\pm0.1\%$ & $7.6\pm0.3\%$ & $10.7\pm0.3\%$\\ \bottomrule
	\end{tabular}
\end{minipage}%
\caption{\textbf{Random forests} on \texttt{Old Adult}\looseness=-1}
\label{fig:adultold-rfc-all}
\end{figure*}


\newpage
\subsubsection{\appgerman}\label{app:sec:german-algo}

$\hatsc$ CDFs for \texttt{German Credit} ($\group=\texttt{sex}$) and associated error metrics on the prediction set. \textbf{Baseline} metrics computed with $\boot=101$ models. For \textbf{simple}, $\boot=101$ models; for \textbf{super}, $\boot=101$ ensemble models, each composed of $51$ underlying models. We repeat for $10$ test/train splits. We also report abstention rate $\hatar$. 

\setlength{\tabcolsep}{6pt}
\begin{figure*}[h!]
\begin{minipage}{.495\linewidth}
\centering
\hspace{-.4cm}
        \includegraphics[width=\linewidth]{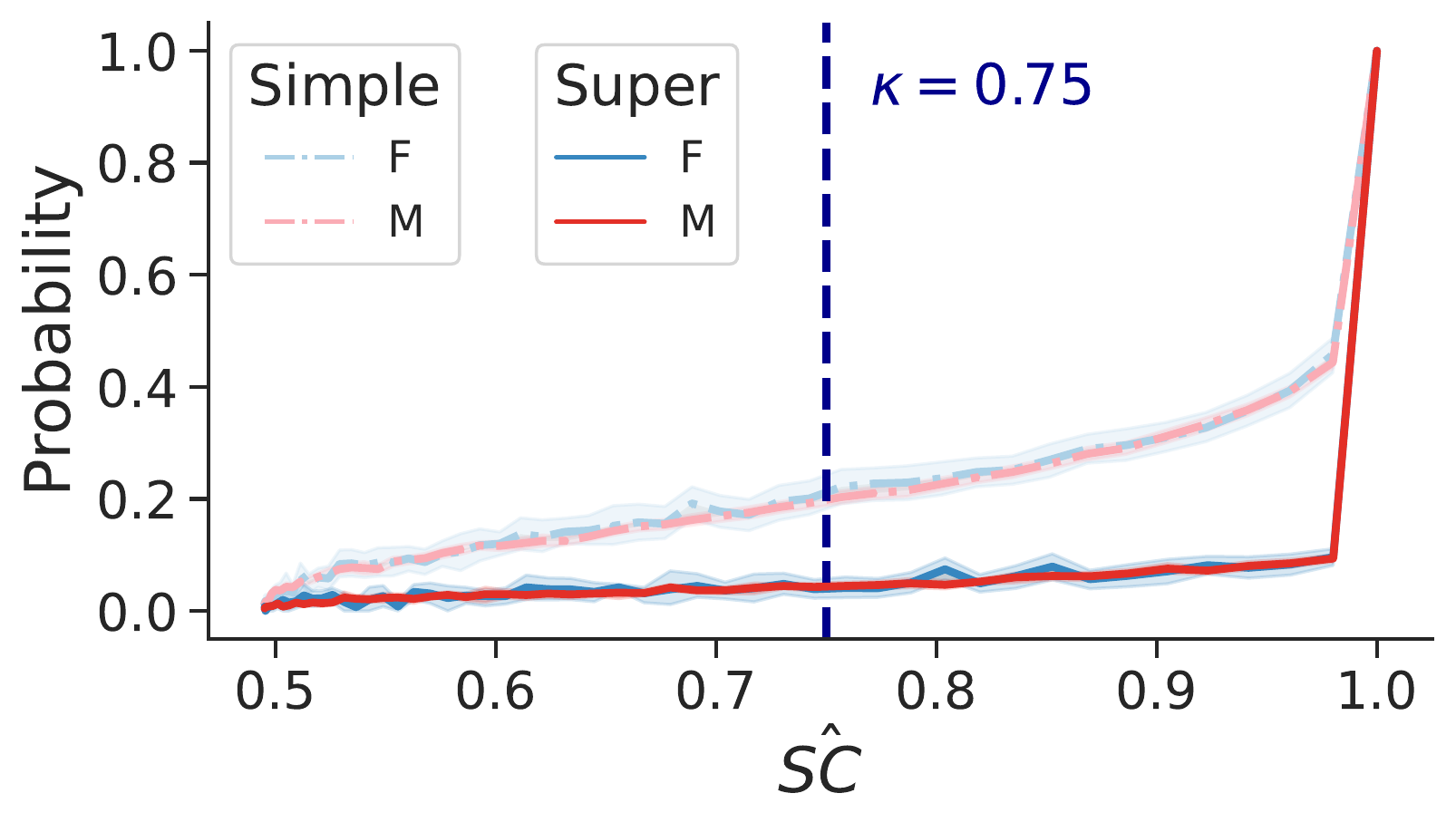}\\
        \vspace{-.1cm}%
	\begin{tabular}{lcc}
		\toprule
		\multicolumn{3}{c}{\textbf{Abstention set metrics}} \\ \cmidrule(lr){1-3}
		 & \textbf{Simple}     & \textbf{Super}      \\ \midrule
		 \textbf{$\Delta\hatar$} & \textcolor{blue}{$0.4\pm3.9\%$} & \textcolor{blue}{$0.1\pm1.8\%$}\\ \midrule
		 \textbf{$\hatar_\text{F}$} & $20.8\pm6.6\%$ & $4.1\pm3.1\%$\\ \midrule
		 \textbf{$\hatar_\text{M}$} & $21.2\pm2.7\%$ & $4.0\pm1.3\%$\\ \bottomrule
	\end{tabular}
\end{minipage}%
\hspace{.25cm}
\begin{minipage}{.495\linewidth}
\centering
\hspace{-.2cm}
	\begin{tabular}{lccc}
		\toprule
		\multicolumn{4}{c}{\textbf{Logistic regression prediction set metrics}} \\ \cmidrule(lr){1-4}
		 & \textbf{Baseline}     & \textbf{Simple}     & \textbf{Super}      \\ \midrule
		 \textbf{$\Delta\hatpr$} & \textcolor{blue}{$8.8\pm1.4\%$} & \textcolor{blue}{$9.1\pm1.2\%$} & \textcolor{blue}{$9.7\pm1.7\%$}\\ \midrule
		 \textbf{$\hatpr_\text{F}$} & $88.8\pm4.7\%$ & $96.0\pm4.1\%$ & $91.7\pm5.0\%$\\ \midrule
		 \textbf{$\hatpr_\text{M}$} & $80.0\pm3.3\%$ & $86.9\pm2.9\%$ & $82.0\pm3.3\%$\\ \midrule
		 \textbf{$\Delta\haterr$} & \textcolor{blue}{$0.9\pm4.0\%$} & \textcolor{blue}{$4.6\pm6.1\%$} & \textcolor{blue}{$3.4\pm4.7\%$}\\ \midrule
		 \textbf{$\haterr_\text{F}$} & $23.3\pm6.9\%$ & $22.8\pm8.7\%$ & $25.5\pm7.4\%$\\ \midrule
		 \textbf{$\haterr_\text{M}$} & $24.2\pm2.9\%$ & $18.2\pm2.6\%$ & $22.1\pm2.7\%$\\ \midrule
		 \textbf{$\Delta\hatfpr$} & \textcolor{blue}{$0.7\pm3.9\%$} & \textcolor{blue}{$5.4\pm5.8\%$} & \textcolor{blue}{$5.5\pm5.4\%$}\\ \midrule
		 \textbf{$\hatfpr_\text{F}$} & $16.2\pm6.2\%$ & $19.6\pm8.4\%$ & $21.1\pm7.8\%$\\ \midrule
		 \textbf{$\hatfpr_\text{M}$} & $15.5\pm2.3\%$ & $14.2\pm2.6\%$ & $15.6\pm2.4\%$\\ \midrule
		 \textbf{$\Delta\hatfnr$} & \textcolor{blue}{$1.6\pm1.1\%$} & \textcolor{blue}{$0.8\pm1.9\%$} & \textcolor{blue}{$2.1\pm1.8\%$}\\ \midrule
		 \textbf{$\hatfnr_\text{F}$} & $7.1\pm3.7\%$ & $3.2\pm3.5\%$ & $4.4\pm3.8\%$\\ \midrule
		 \textbf{$\hatfnr_\text{M}$} & $8.7\pm2.6\%$ & $4.0\pm1.6\%$ & $6.5\pm2.0\%$\\ \bottomrule
	\end{tabular}
\end{minipage}%
\caption{\textbf{Logistic regression} on \texttt{German Credit}\looseness=-1}
\label{fig:german-lr-all}
\end{figure*}
\vfill
\setlength{\tabcolsep}{6pt}
\begin{figure*}[h!]
\begin{minipage}{.495\linewidth}
\centering
\hspace{-.4cm}
        \includegraphics[width=\linewidth]{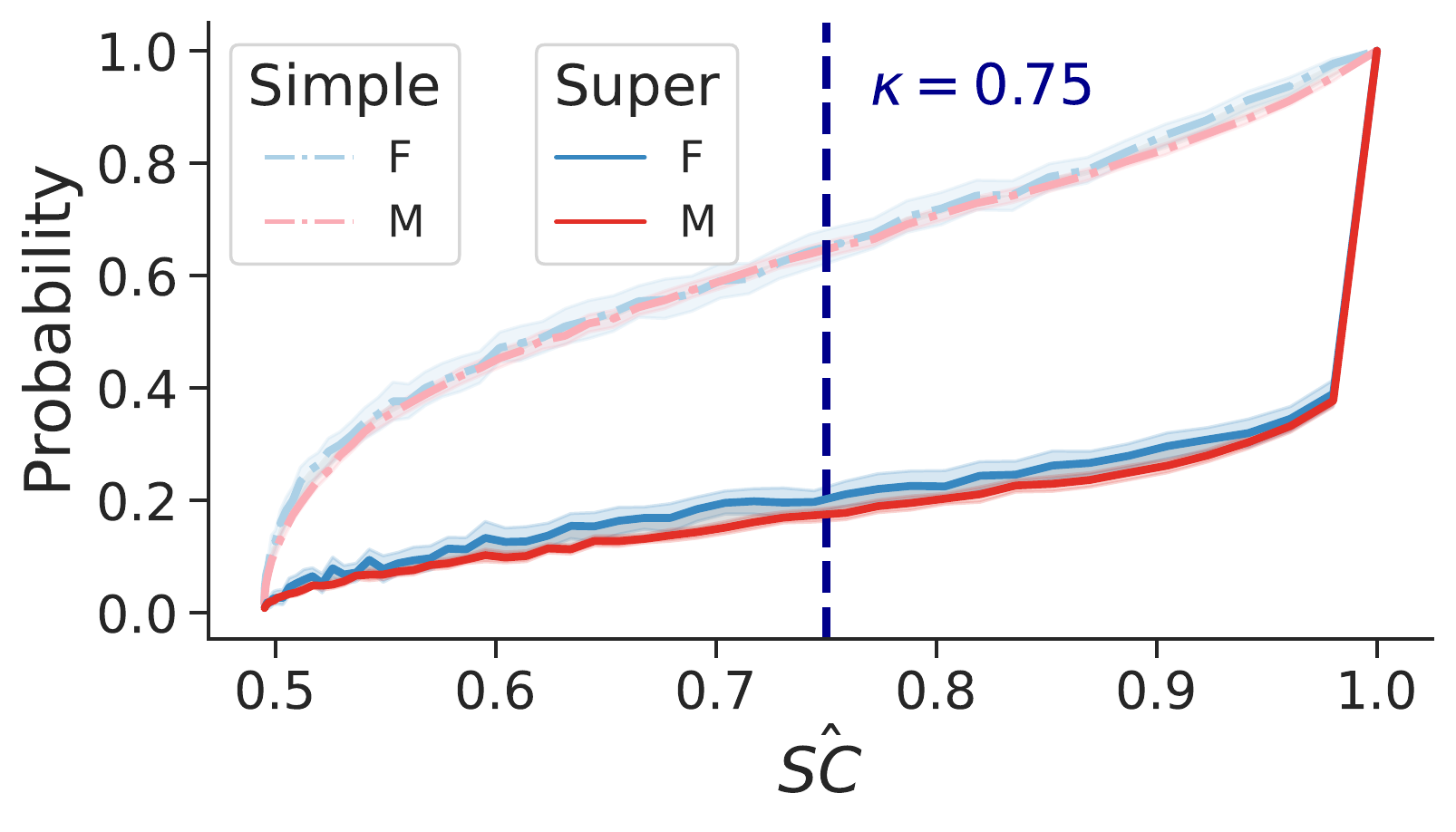}\\
        \vspace{-.1cm}%
	\begin{tabular}{lcc}
		\toprule
		\multicolumn{3}{c}{\textbf{Abstention set metrics}} \\ \cmidrule(lr){1-3}
		 & \textbf{Simple}     & \textbf{Super}      \\ \midrule
		 \textbf{$\Delta\hatar$} & \textcolor{blue}{$0.0\pm2.6\%$} & \textcolor{blue}{$2.8\pm2.6\%$}\\ \midrule
		 \textbf{$\hatar_\text{F}$} & $65.2\pm6.0\%$ & $19.9\pm5.9\%$\\ \midrule
		 \textbf{$\hatar_\text{M}$} & $65.2\pm3.4\%$ & $17.1\pm3.3\%$\\ \bottomrule
	\end{tabular}
\end{minipage}%
\hspace{.25cm}
\begin{minipage}{.495\linewidth}
\centering
\hspace{-.2cm}
	\begin{tabular}{lccc}
		\toprule
		\multicolumn{4}{c}{\textbf{Decision tree prediction set metrics}} \\ \cmidrule(lr){1-4}
		 & \textbf{Baseline}     & \textbf{Simple}     & \textbf{Super}      \\ \midrule
		 \textbf{$\Delta\hatpr$} & \textcolor{blue}{$0.3\pm2.5\%$} & \textcolor{blue}{$1.9\pm0.4\%$} & \textcolor{blue}{$3.7\pm3.7\%$}\\ \midrule
		 \textbf{$\hatpr_\text{F}$} & $71.2\pm4.6\%$ & $99.6\pm0.8\%$ & $87.9\pm6.0\%$\\ \midrule
		 \textbf{$\hatpr_\text{M}$} & $70.9\pm2.1\%$ & $97.7\pm1.2\%$ & $84.2\pm2.3\%$\\ \midrule
		 \textbf{$\Delta\haterr$} & \textcolor{blue}{$1.1\pm2.9\%$} & \textcolor{blue}{$0.3\pm5.2\%$} & \textcolor{blue}{$0.1\pm4.9\%$}\\ \midrule
		 \textbf{$\haterr_\text{F}$} & $33.0\pm4.8\%$ & $9.8\pm8.0\%$ & $20.3\pm8.2\%$\\ \midrule
		 \textbf{$\haterr_\text{M}$} & $31.9\pm1.9\%$ & $9.5\pm2.8\%$ & $20.2\pm3.3\%$\\ \midrule
		 \textbf{$\Delta\hatfpr$} & \textcolor{blue}{$0.7\pm3.5\%$} & \textcolor{blue}{$0.8\pm5.1\%$} & \textcolor{blue}{$0.4\pm3.9\%$}\\ \midrule
		 \textbf{$\hatfpr_\text{F}$} & $15.6\pm5.9\%$ & $9.6\pm7.9\%$ & $15.2\pm7.0\%$\\ \midrule
		 \textbf{$\hatfpr_\text{M}$} & $14.9\pm2.4\%$ & $8.8\pm2.8\%$ & $14.8\pm3.1\%$\\ \midrule
		 \textbf{$\Delta\hatfnr$} & \textcolor{blue}{$0.5\pm2.3\%$} & \textcolor{blue}{$0.4\pm0.0\%$} & \textcolor{blue}{$0.2\pm3.6\%$}\\ \midrule
		 \textbf{$\hatfnr_\text{F}$} & $17.4\pm4.4\%$ & $0.2\pm0.7\%$ & $5.2\pm5.1\%$\\ \midrule
		 \textbf{$\hatfnr_\text{M}$} & $16.9\pm2.1\%$ & $0.6\pm0.7\%$ & $5.4\pm1.5\%$\\ \bottomrule
	\end{tabular}
\end{minipage}%
\caption{\textbf{Decision trees} on \texttt{German Credit}\looseness=-1}
\label{fig:german-dtc-all}
\end{figure*}
\setlength{\tabcolsep}{6pt}
\begin{figure*}[h!]
\begin{minipage}{.495\linewidth}
\centering
\hspace{-.4cm}
        \includegraphics[width=\linewidth]{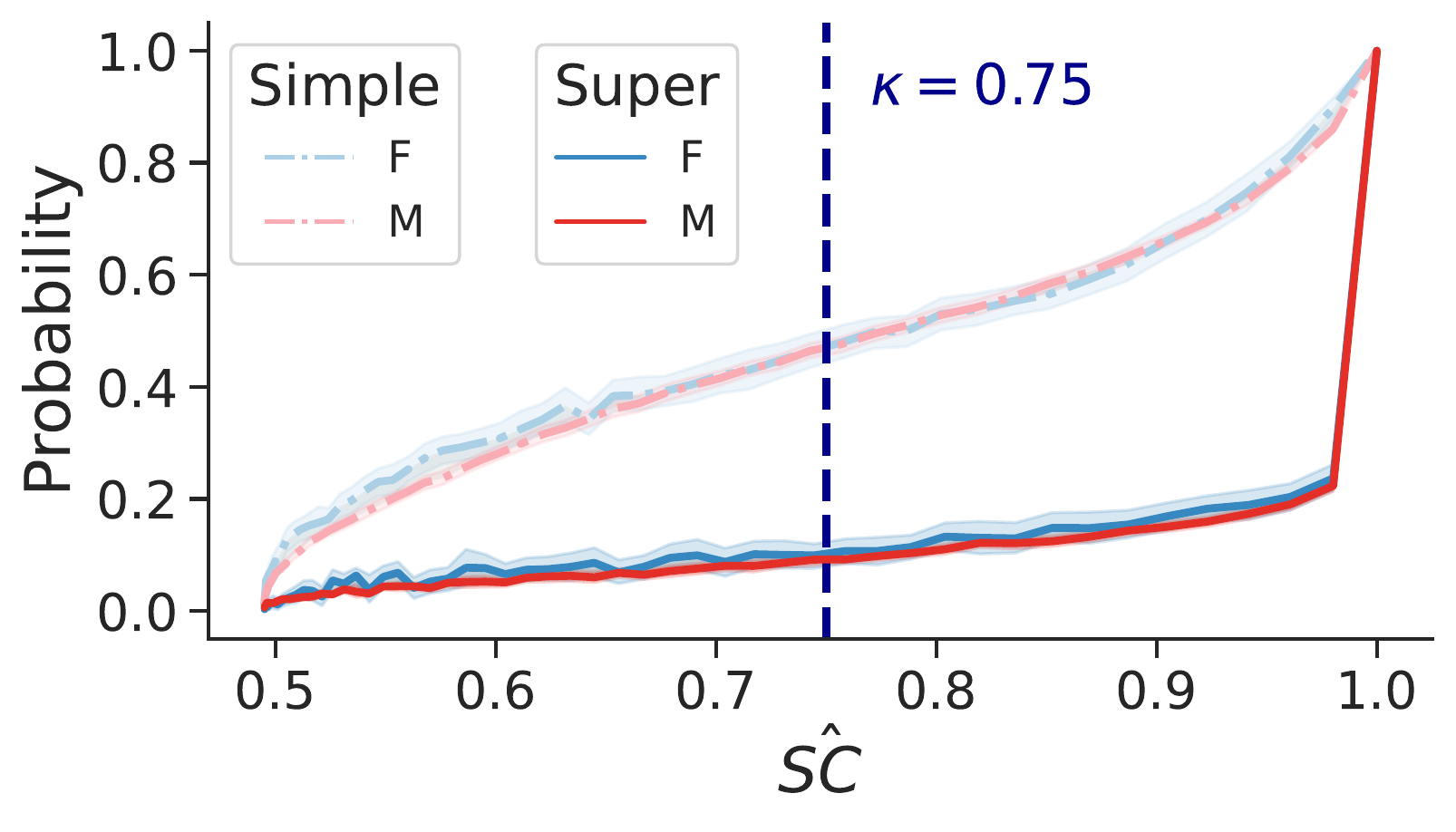}\\
        \vspace{-.1cm}%
	\begin{tabular}{lcc}
		\toprule
		\multicolumn{3}{c}{\textbf{Abstention set metrics}} \\ \cmidrule(lr){1-3}
		 & \textbf{Simple}     & \textbf{Super}      \\ \midrule
		 \textbf{$\Delta\hatar$} & \textcolor{blue}{$0.1\pm2.8\%$} & \textcolor{blue}{$1.2\pm3.3\%$}\\ \midrule
		 \textbf{$\hatar_\text{F}$} & $47.5\pm6.7\%$ & $9.9\pm5.7\%$\\ \midrule
		 \textbf{$\hatar_\text{M}$} & $47.4\pm3.9\%$ & $8.7\pm2.4\%$\\ \bottomrule
	\end{tabular}
\end{minipage}%
\hspace{.25cm}
\begin{minipage}{.495\linewidth}
\centering
\hspace{-.2cm}
	\begin{tabular}{lccc}
		\toprule
		\multicolumn{4}{c}{\textbf{Random forest prediction set metrics}} \\ \cmidrule(lr){1-4}
		 & \textbf{Baseline}     & \textbf{Simple}     & \textbf{Super}      \\ \midrule
		 \textbf{$\Delta\hatpr$} & \textcolor{blue}{$3.9\pm1.6\%$} & \textcolor{blue}{$1.9\pm0.6\%$} & \textcolor{blue}{$4.9\pm2.1\%$}\\ \midrule
		 \textbf{$\hatpr_\text{F}$} & $81.7\pm3.3\%$ & $99.9\pm0.4\%$ & $94.5\pm4.0\%$\\ \midrule
		 \textbf{$\hatpr_\text{M}$} & $77.8\pm1.7\%$ & $98.0\pm1.0\%$ & $89.6\pm1.9\%$\\ \midrule
		 \textbf{$\Delta\haterr$} & \textcolor{blue}{$2.3\pm3.1\%$} & \textcolor{blue}{$0.0\pm4.7\%$} & \textcolor{blue}{$2.4\pm6.7\%$}\\ \midrule
		 \textbf{$\haterr_\text{F}$} & $25.8\pm5.1\%$ & $11.9\pm7.8\%$ & $23.5\pm9.6\%$\\ \midrule
		 \textbf{$\haterr_\text{M}$} & $28.1\pm2.0\%$ & $11.9\pm3.1\%$ & $21.1\pm2.9\%$\\ \midrule
		 \textbf{$\Delta\hatfpr$} & \textcolor{blue}{$2.5\pm3.0\%$} & \textcolor{blue}{$0.3\pm4.7\%$} & \textcolor{blue}{$2.2\pm6.2\%$}\\ \midrule
		 \textbf{$\hatfpr_\text{F}$} & $13.8\pm4.8\%$ & $11.8\pm7.8\%$ & $20.5\pm9.1\%$\\ \midrule
		 \textbf{$\hatfpr_\text{M}$} & $16.3\pm1.8\%$ & $11.5\pm3.1\%$ & $18.3\pm2.9\%$\\ \midrule
		 \textbf{$\Delta\hatfnr$} & \textcolor{blue}{$0.2\pm1.7\%$} & \textcolor{blue}{$0.3\pm0.1\%$} & \textcolor{blue}{$0.1\pm2.3\%$}\\ \midrule
		 \textbf{$\hatfnr_\text{F}$} & $11.9\pm3.1\%$ & $0.1\pm0.3\%$ & $3.0\pm3.5\%$\\ \midrule
		 \textbf{$\hatfnr_\text{M}$} & $11.7\pm1.4\%$ & $0.4\pm0.4\%$ & $2.9\pm1.2\%$\\ \bottomrule
	\end{tabular}
\end{minipage}%
\caption{\textbf{Random forests} on \texttt{German Credit}\looseness=-1}
\label{fig:german-rfc-all}
\end{figure*}

\subsubsection{\apptaiwan}\label{app:sec:taiwan-algo}

$\hatsc$ CDFs for \texttt{Taiwan Credit} ($\group=\texttt{sex}$) and associated error metrics on the prediction set. \textbf{Baseline} metrics computed with $\boot=101$ models. For \textbf{simple}, $\boot=101$ models; for \textbf{super}, $\boot=101$ ensemble models, each composed of $41$ underlying models. We repeat for $10$ test/train splits. We also report abstention rate $\hatar$. 

\setlength{\tabcolsep}{6pt}
\begin{figure*}[h!]
\begin{minipage}{.495\linewidth}
\centering
\hspace{-.4cm}
        \includegraphics[width=\linewidth]{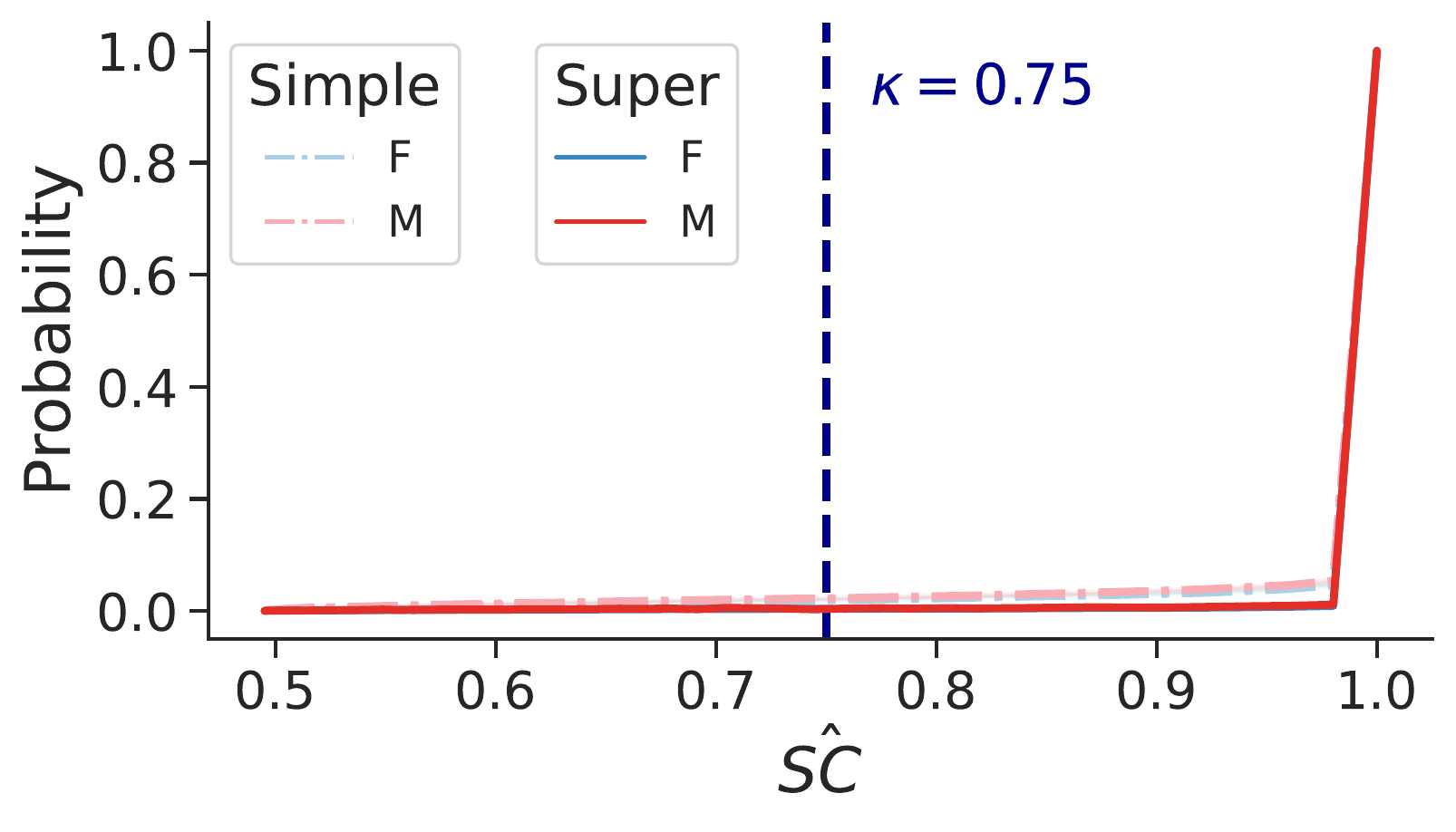}\\
        \vspace{-.1cm}%
	\begin{tabular}{lcc}
		\toprule
		\multicolumn{3}{c}{\textbf{Abstention set metrics}} \\ \cmidrule(lr){1-3}
		 & \textbf{Simple}     & \textbf{Super}      \\ \midrule
		 \textbf{$\Delta\hatar$} & \textcolor{blue}{$0.4\pm0.2\%$} & \textcolor{blue}{$0.0\pm0.2\%$}\\ \midrule
		 \textbf{$\hatar_\text{F}$} & $2.1\pm0.1\%$ & $0.4\pm0.0\%$\\ \midrule
		 \textbf{$\hatar_\text{M}$} & $2.5\pm0.3\%$ & $0.4\pm0.2\%$\\ \bottomrule
	\end{tabular}
\end{minipage}%
\hspace{.25cm}
\begin{minipage}{.495\linewidth}
\centering
\hspace{-.2cm}
	\begin{tabular}{lccc}
		\toprule
		\multicolumn{4}{c}{\textbf{Logistic regression prediction set metrics}} \\ \cmidrule(lr){1-4}
		 & \textbf{Baseline}     & \textbf{Simple}     & \textbf{Super}      \\ \midrule
		 \textbf{$\Delta\hatpr$} & \textcolor{blue}{$1.5\pm0.1\%$} & \textcolor{blue}{$1.0\pm0.1\%$} & \textcolor{blue}{$1.0\pm0.1\%$}\\ \midrule
		 \textbf{$\hatpr_\text{F}$} & $6.7\pm0.3\%$ & $6.2\pm0.1\%$ & $6.9\pm0.1\%$\\ \midrule
		 \textbf{$\hatpr_\text{M}$} & $8.2\pm0.4\%$ & $7.2\pm0.2\%$ & $7.9\pm0.2\%$\\ \midrule
		 \textbf{$\Delta\haterr$} & \textcolor{blue}{$3.1\pm0.1\%$} & \textcolor{blue}{$3.1\pm0.3\%$} & \textcolor{blue}{$3.2\pm0.3\%$}\\ \midrule
		 \textbf{$\haterr_\text{F}$} & $17.8\pm0.5\%$ & $17.0\pm0.2\%$ & $17.5\pm0.3\%$\\ \midrule
		 \textbf{$\haterr_\text{M}$} & $20.9\pm0.4\%$ & $20.1\pm0.5\%$ & $20.7\pm0.6\%$\\ \midrule
		 \textbf{$\Delta\hatfpr$} & \textcolor{blue}{$0.7\pm0.2\%$} & \textcolor{blue}{$0.3\pm0.0\%$} & \textcolor{blue}{$0.3\pm0.1\%$}\\ \midrule
		 \textbf{$\hatfpr_\text{F}$} & $1.8\pm0.1\%$ & $1.7\pm0.1\%$ & $2.0\pm0.1\%$\\ \midrule
		 \textbf{$\hatfpr_\text{M}$} & $2.5\pm0.3\%$ & $2.0\pm0.1\%$ & $2.3\pm0.2\%$\\ \midrule
		 \textbf{$\Delta\hatfnr$} & \textcolor{blue}{$2.4\pm0.2\%$} & \textcolor{blue}{$2.7\pm0.4\%$} & \textcolor{blue}{$2.8\pm0.3\%$}\\ \midrule
		 \textbf{$\hatfnr_\text{F}$} & $16.0\pm0.6\%$ & $15.3\pm0.2\%$ & $15.6\pm0.3\%$\\ \midrule
		 \textbf{$\hatfnr_\text{M}$} & $18.4\pm0.4\%$ & $18.0\pm0.6\%$ & $18.4\pm0.6\%$\\ \bottomrule
	\end{tabular}
\end{minipage}%
\caption{\textbf{Logistic regression} on \texttt{Taiwan Credit}\looseness=-1}
\label{fig:taiwan-lr-all}
\end{figure*}
\vspace{.5cm}
\setlength{\tabcolsep}{6pt}
\begin{figure*}[h!]
\begin{minipage}{.495\linewidth}
\centering
\hspace{-.4cm}
        \includegraphics[width=\linewidth]{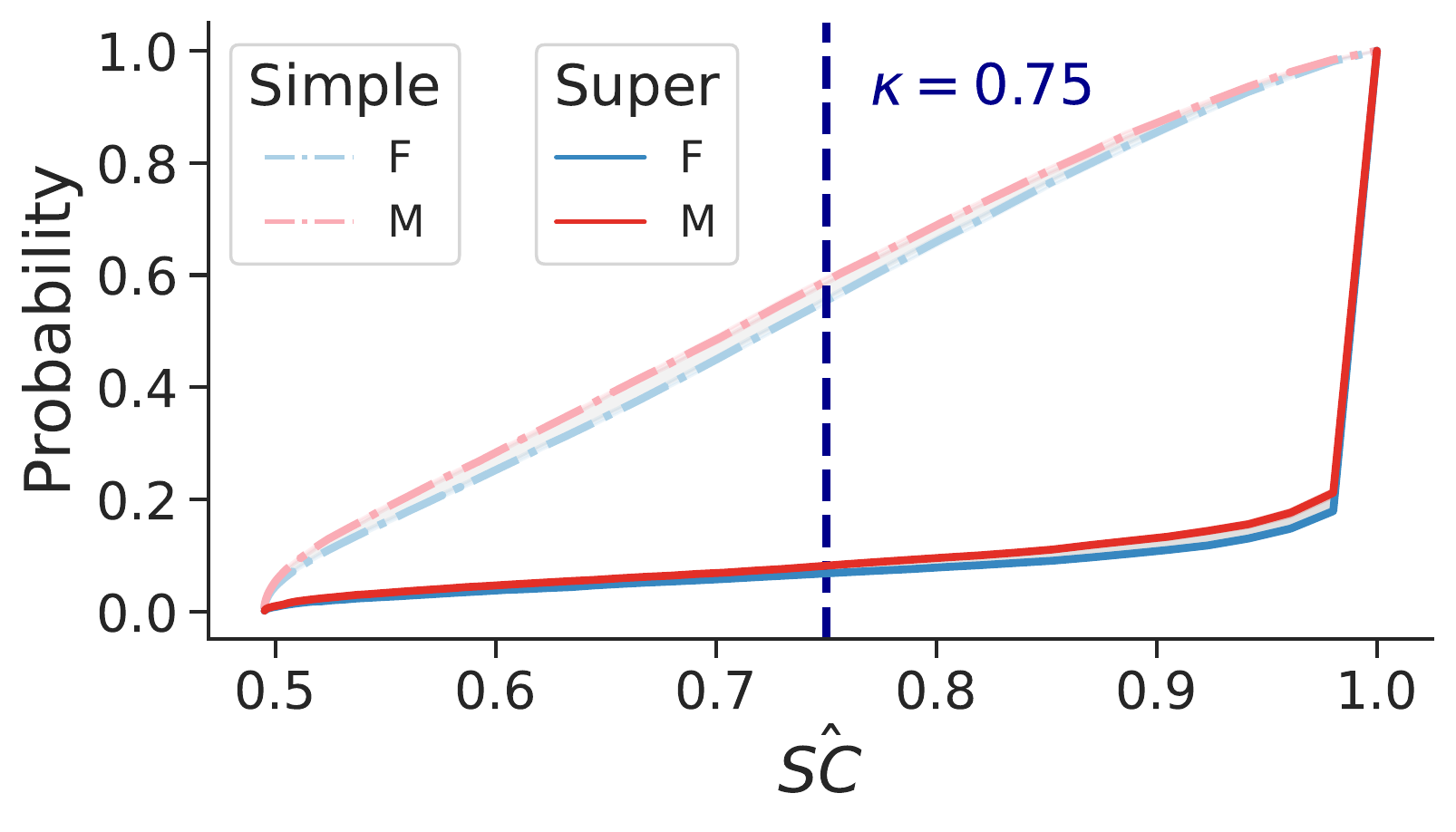}\\
        \vspace{-.1cm}%
	\begin{tabular}{lcc}
		\toprule
		\multicolumn{3}{c}{\textbf{Abstention set metrics}} \\ \cmidrule(lr){1-3}
		 & \textbf{Simple}     & \textbf{Super}      \\ \midrule
		 \textbf{$\Delta\hatar$} & \textcolor{blue}{$3.2\pm0.1\%$} & \textcolor{blue}{$1.3\pm0.1\%$}\\ \midrule
		 \textbf{$\hatar_\text{F}$} & $56.7\pm0.6\%$ & $6.7\pm0.2\%$\\ \midrule
		 \textbf{$\hatar_\text{M}$} & $59.9\pm0.5\%$ & $8.0\pm0.1\%$\\ \bottomrule
	\end{tabular}
\end{minipage}%
\hspace{.25cm}
\begin{minipage}{.495\linewidth}
\centering
\hspace{-.2cm}
	\begin{tabular}{lccc}
		\toprule
		\multicolumn{4}{c}{\textbf{Decision tree prediction set metrics}} \\ \cmidrule(lr){1-4}
		 & \textbf{Baseline}     & \textbf{Simple}     & \textbf{Super}      \\ \midrule
		 \textbf{$\Delta\hatpr$} & \textcolor{blue}{$2.1\pm0.1\%$} & \textcolor{blue}{$1.2\pm0.0\%$} & \textcolor{blue}{$2.0\pm0.2\%$}\\ \midrule
		 \textbf{$\hatpr_\text{F}$} & $22.9\pm0.2\%$ & $3.0\pm0.4\%$ & $9.9\pm0.3\%$\\ \midrule
		 \textbf{$\hatpr_\text{M}$} & $25.0\pm0.3\%$ & $4.2\pm0.4\%$ & $11.9\pm0.5\%$\\ \midrule
		 \textbf{$\Delta\haterr$} & \textcolor{blue}{$2.3\pm0.1\%$} & \textcolor{blue}{$1.6\pm0.0\%$} & \textcolor{blue}{$2.5\pm0.1\%$}\\ \midrule
		 \textbf{$\haterr_\text{F}$} & $26.8\pm0.2\%$ & $9.6\pm0.4\%$ & $15.3\pm0.3\%$\\ \midrule
		 \textbf{$\haterr_\text{M}$} & $29.1\pm0.3\%$ & $11.2\pm0.4\%$ & $17.8\pm0.4\%$\\ \midrule
		 \textbf{$\Delta\hatfpr$} & \textcolor{blue}{$0.6\pm0.1\%$} & \textcolor{blue}{$0.2\pm0.1\%$} & \textcolor{blue}{$0.7\pm0.2\%$}\\ \midrule
		 \textbf{$\hatfpr_\text{F}$} & $14.4\pm0.2\%$ & $0.6\pm0.1\%$ & $3.0\pm0.1\%$\\ \midrule
		 \textbf{$\hatfpr_\text{M}$} & $15.0\pm0.3\%$ & $0.8\pm0.2\%$ & $3.7\pm0.3\%$\\ \midrule
		 \textbf{$\Delta\hatfnr$} & \textcolor{blue}{$1.7\pm0.2\%$} & \textcolor{blue}{$1.3\pm0.1\%$} & \textcolor{blue}{$1.9\pm0.1\%$}\\ \midrule
		 \textbf{$\hatfnr_\text{F}$} & $12.4\pm0.4\%$ & $9.0\pm0.4\%$ & $12.3\pm0.3\%$\\ \midrule
		 \textbf{$\hatfnr_\text{M}$} & $14.1\pm0.2\%$ & $10.3\pm0.5\%$ & $14.2\pm0.4\%$\\ \bottomrule
	\end{tabular}
\end{minipage}%
\caption{\textbf{Decision trees} on \texttt{Taiwan Credit}\looseness=-1}
\label{fig:taiwan-dtc-all}
\vspace{3cm}
\end{figure*}
\setlength{\tabcolsep}{6pt}
\begin{figure*}[h!]
\begin{minipage}{.495\linewidth}
\centering
\hspace{-.4cm}
        \includegraphics[width=\linewidth]{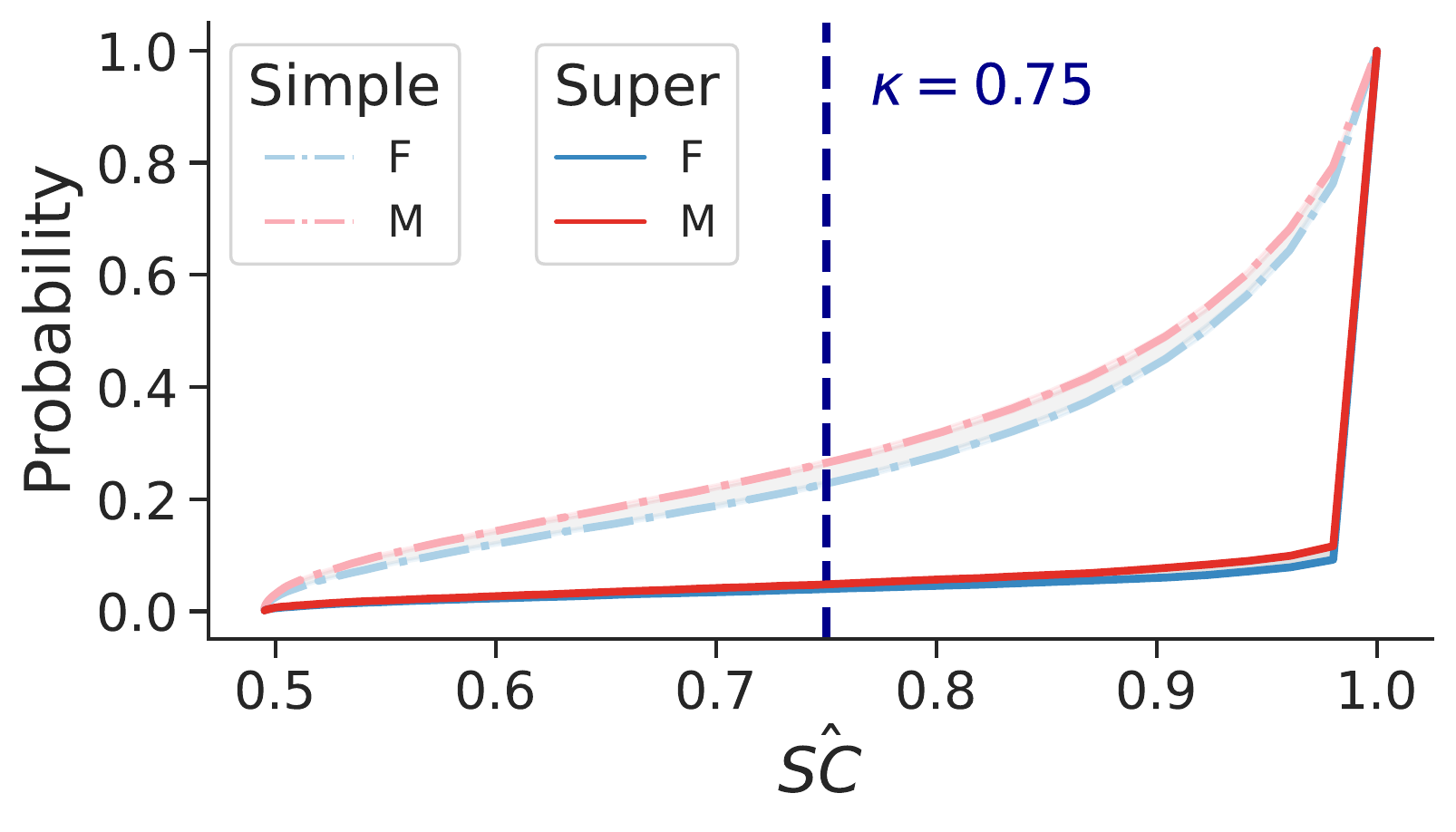}\\
        \vspace{-.1cm}%
	\begin{tabular}{lcc}
		\toprule
		\multicolumn{3}{c}{\textbf{Abstention set metrics}} \\ \cmidrule(lr){1-3}
		 & \textbf{Simple}     & \textbf{Super}      \\ \midrule
		 \textbf{$\Delta\hatar$} & \textcolor{blue}{$4.1\pm0.1\%$} & \textcolor{blue}{$0.8\pm0.0\%$}\\ \midrule
		 \textbf{$\hatar_\text{F}$} & $24.0\pm0.8\%$ & $3.9\pm0.3\%$\\ \midrule
		 \textbf{$\hatar_\text{M}$} & $28.1\pm0.7\%$ & $4.7\pm0.3\%$\\ \bottomrule
	\end{tabular}
\end{minipage}%
\hspace{.25cm}
\begin{minipage}{.495\linewidth}
\centering
\hspace{-.2cm}
	\begin{tabular}{lccc}
		\toprule
		\multicolumn{4}{c}{\textbf{Random forest prediction set metrics}} \\ \cmidrule(lr){1-4}
		 & \textbf{Baseline}     & \textbf{Simple}     & \textbf{Super}      \\ \midrule
		 \textbf{$\Delta\hatpr$} & \textcolor{blue}{$2.5\pm0.1\%$} & \textcolor{blue}{$1.0\pm0.1\%$} & \textcolor{blue}{$2.1\pm0.2\%$}\\ \midrule
		 \textbf{$\hatpr_\text{F}$} & $14.9\pm0.2\%$ & $4.1\pm0.3\%$ & $10.3\pm0.2\%$\\ \midrule
		 \textbf{$\hatpr_\text{M}$} & $17.4\pm0.3\%$ & $5.1\pm0.2\%$ & $12.4\pm0.4\%$\\ \midrule
		 \textbf{$\Delta\haterr$} & \textcolor{blue}{$2.8\pm0.1\%$} & \textcolor{blue}{$1.9\pm0.0\%$} & \textcolor{blue}{$2.5\pm0.1\%$}\\ \midrule
		 \textbf{$\haterr_\text{F}$} & $20.5\pm0.3\%$ & $12.0\pm0.4\%$ & $15.8\pm0.4\%$\\ \midrule
		 \textbf{$\haterr_\text{M}$} & $23.3\pm0.4\%$ & $13.9\pm0.4\%$ & $18.3\pm0.5\%$\\ \midrule
		 \textbf{$\Delta\hatfpr$} & \textcolor{blue}{$1.0\pm0.1\%$} & \textcolor{blue}{$0.3\pm0.1\%$} & \textcolor{blue}{$0.6\pm0.1\%$}\\ \midrule
		 \textbf{$\hatfpr_\text{F}$} & $7.2\pm0.2\%$ & $0.9\pm0.1\%$ & $3.3\pm0.1\%$\\ \midrule
		 \textbf{$\hatfpr_\text{M}$} & $8.2\pm0.3\%$ & $1.2\pm0.2\%$ & $3.9\pm0.2\%$\\ \midrule
		 \textbf{$\Delta\hatfnr$} & \textcolor{blue}{$1.7\pm0.1\%$} & \textcolor{blue}{$1.6\pm0.1\%$} & \textcolor{blue}{$1.8\pm0.0\%$}\\ \midrule
		 \textbf{$\hatfnr_\text{F}$} & $13.3\pm0.4\%$ & $11.0\pm0.3\%$ & $12.6\pm0.4\%$\\ \midrule
		 \textbf{$\hatfnr_\text{M}$} & $15.0\pm0.3\%$ & $12.6\pm0.4\%$ & $14.4\pm0.4\%$\\ \bottomrule
	\end{tabular}
\end{minipage}%
\caption{\textbf{Random forests} on \texttt{Taiwan Credit}\looseness=-1}
\label{fig:taiwan-rfc-all}
\end{figure*}
\FloatBarrier
\vspace*{1cm}

\vspace*{-.75in}
\subsubsection{\appnewadult}\label{app:sec:newadult-algo}

$\hatsc$ CDFs for three tasks (\texttt{Income}, \texttt{Employment}, \texttt{Public Coverage}) in \texttt{New Adult - CA}, using $\group=\texttt{sex}$ and \texttt{race}, and associated error metrics on the prediction set. \textbf{Baseline} metrics computed with $\boot=101$ models. For \textbf{simple}, $\boot=101$ models; for \textbf{super}, $\boot=101$ ensemble models, each composed of $21$ underlying models for \texttt{Income} and \texttt{Public Coverage}; $15$ for \texttt{Employment}. We repeat for $5$ test/train splits. We also report abstention rate $\hatar$. 

\custompar{\texttt{Income} - by \texttt{sex}}

\setlength{\tabcolsep}{6pt}
\begin{figure*}[h!]
\begin{minipage}{.495\linewidth}
\centering
\hspace{-.4cm}
        \includegraphics[width=\linewidth]{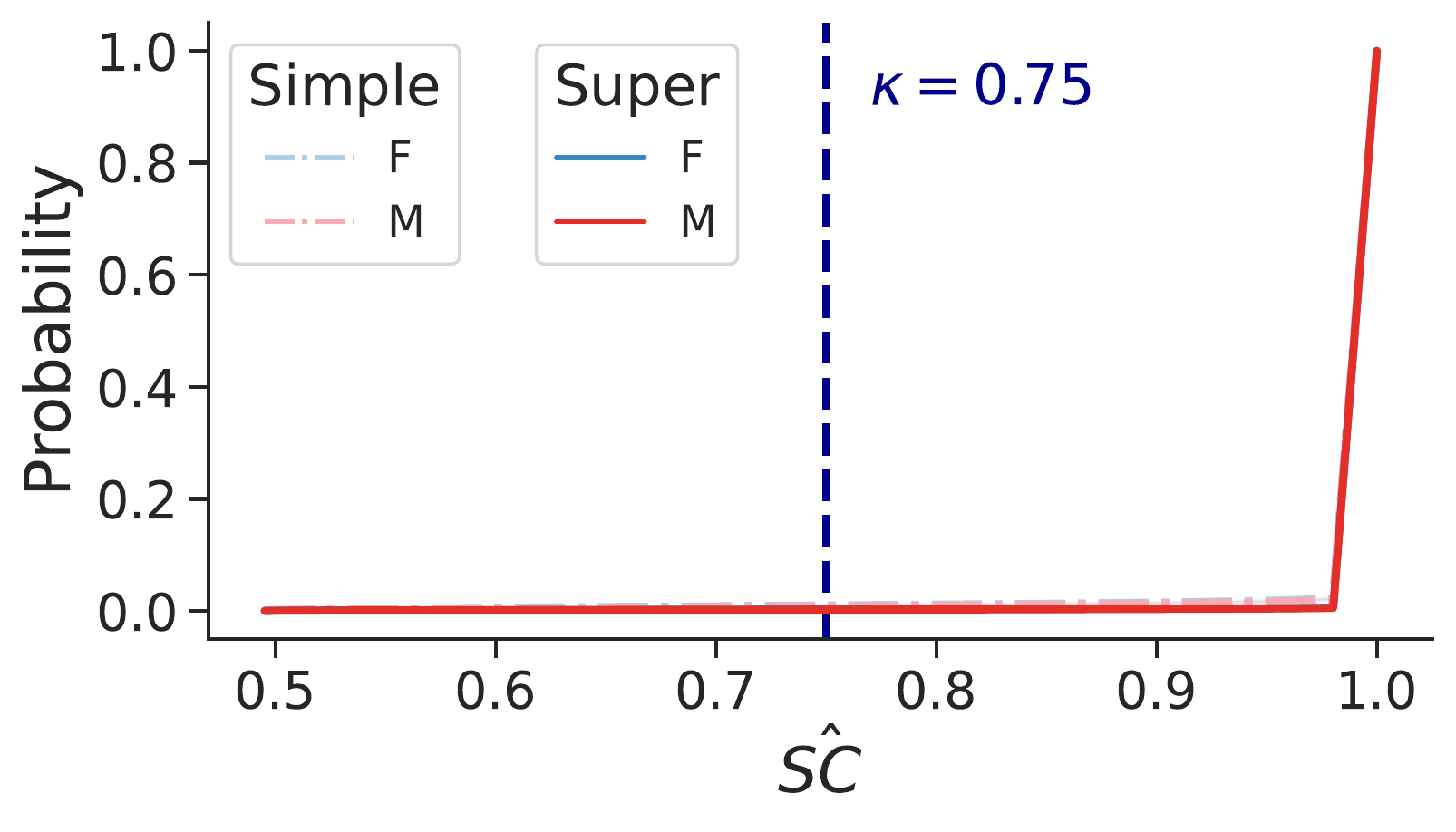}\\
        \vspace{-.1cm}%
	\begin{tabular}{lcc}
		\toprule
		\multicolumn{3}{c}{\textbf{Abstention set metrics}} \\ \cmidrule(lr){1-3}
		 & \textbf{Simple}     & \textbf{Super}      \\ \midrule
		 \textbf{$\Delta\hatar$} & \textcolor{blue}{$0.1\pm0.0\%$} & \textcolor{blue}{$0.1\pm0.0\%$}\\ \midrule
		 \textbf{$\hatar_\text{F}$} & $1.0\pm0.0\%$ & $0.3\pm0.0\%$\\ \midrule
		 \textbf{$\hatar_\text{M}$} & $0.9\pm0.0\%$ & $0.2\pm0.0\%$\\ \bottomrule
	\end{tabular}
\end{minipage}%
\hspace{.25cm}
\begin{minipage}{.495\linewidth}
\centering
\hspace{-.2cm}
	\begin{tabular}{lccc}
		\toprule
		\multicolumn{4}{c}{\textbf{Logistic regression prediction set metrics}} \\ \cmidrule(lr){1-4}
		 & \textbf{Baseline}     & \textbf{Simple}     & \textbf{Super}      \\ \midrule
		 \textbf{$\Delta\hatpr$} & \textcolor{blue}{$2.7\pm0.1\%$} & \textcolor{blue}{$2.9\pm0.1\%$} & \textcolor{blue}{$2.8\pm0.1\%$}\\ \midrule
		 \textbf{$\hatpr_\text{F}$} & $38.4\pm0.2\%$ & $38.1\pm0.2\%$ & $38.2\pm0.2\%$\\ \midrule
		 \textbf{$\hatpr_\text{M}$} & $41.1\pm0.3\%$ & $41.0\pm0.1\%$ & $41.0\pm0.1\%$\\ \midrule
		 \textbf{$\Delta\haterr$} & \textcolor{blue}{$0.9\pm0.0\%$} & \textcolor{blue}{$1.0\pm0.2\%$} & \textcolor{blue}{$1.0\pm0.2\%$}\\ \midrule
		 \textbf{$\haterr_\text{F}$} & $21.5\pm0.2\%$ & $21.1\pm0.3\%$ & $21.3\pm0.3\%$\\ \midrule
		 \textbf{$\haterr_\text{M}$} & $22.4\pm0.2\%$ & $22.1\pm0.1\%$ & $22.3\pm0.1\%$\\ \midrule
		 \textbf{$\Delta\hatfpr$} & \textcolor{blue}{$4.0\pm0.1\%$} & \textcolor{blue}{$3.9\pm0.0\%$} & \textcolor{blue}{$3.9\pm0.0\%$}\\ \midrule
		 \textbf{$\hatfpr_\text{F}$} & $12.5\pm0.2\%$ & $12.2\pm0.1\%$ & $12.3\pm0.1\%$\\ \midrule
		 \textbf{$\hatfpr_\text{M}$} & $8.5\pm0.1\%$ & $8.3\pm0.1\%$ & $8.4\pm0.1\%$\\ \midrule
		 \textbf{$\Delta\hatfnr$} & \textcolor{blue}{$4.9\pm0.0\%$} & \textcolor{blue}{$4.9\pm0.1\%$} & \textcolor{blue}{$4.8\pm0.1\%$}\\ \midrule
		 \textbf{$\hatfnr_\text{F}$} & $9.0\pm0.1\%$ & $8.9\pm0.2\%$ & $9.1\pm0.2\%$\\ \midrule
		 \textbf{$\hatfnr_\text{M}$} & $13.9\pm0.1\%$ & $13.8\pm0.1\%$ & $13.9\pm0.1\%$\\ \bottomrule
	\end{tabular}
\end{minipage}%
\caption{\textbf{Logistic regression} on \texttt{New Adult - CA - Income}, by \texttt{sex}\looseness=-1}
\label{fig:adultnew-income-sex-lr-all}
\vspace{.25cm}
\end{figure*}

\vspace{.1cm}
\setlength{\tabcolsep}{6pt}
\begin{figure*}[h!]
\begin{minipage}{.495\linewidth}
\centering
\hspace{-.4cm}
        \includegraphics[width=\linewidth]{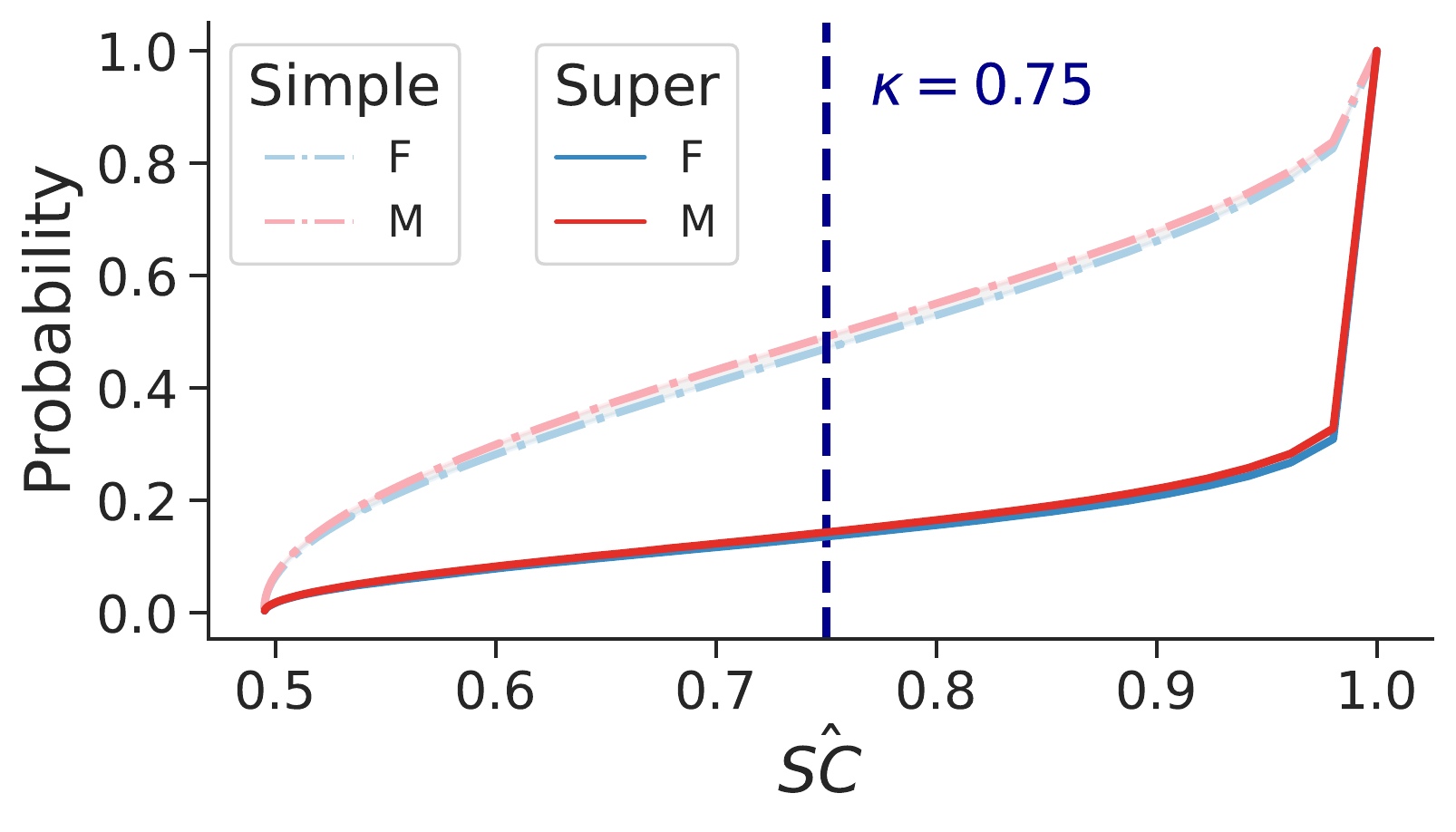}\\
        \vspace{-.1cm}%
	\begin{tabular}{lcc}
		\toprule
		\multicolumn{3}{c}{\textbf{Abstention set metrics}} \\ \cmidrule(lr){1-3}
		 & \textbf{Simple}     & \textbf{Super}      \\ \midrule
		 \textbf{$\Delta\hatar$} & \textcolor{blue}{$2.1\pm0.2\%$} & \textcolor{blue}{$0.8\pm0.0\%$}\\ \midrule
		 \textbf{$\hatar_\text{F}$} & $49.2\pm0.3\%$ & $13.3\pm0.2\%$\\ \midrule
		 \textbf{$\hatar_\text{M}$} & $51.3\pm0.1\%$ & $14.1\pm0.2\%$\\ \bottomrule
	\end{tabular}
\end{minipage}%
\hspace{.25cm}
\begin{minipage}{.495\linewidth}
\centering
\hspace{-.2cm}
	\begin{tabular}{lccc}
		\toprule
		\multicolumn{4}{c}{\textbf{Decision tree prediction set metrics}} \\ \cmidrule(lr){1-4}
		 & \textbf{Baseline}     & \textbf{Simple}     & \textbf{Super}      \\ \midrule
		 \textbf{$\Delta\hatpr$} & \textcolor{blue}{$7.5\pm0.1\%$} & \textcolor{blue}{$12.5\pm0.1\%$} & \textcolor{blue}{$9.7\pm0.1\%$}\\ \midrule
		 \textbf{$\hatpr_\text{F}$} & $37.4\pm0.2\%$ & $26.8\pm0.4\%$ & $34.1\pm0.3\%$\\ \midrule
		 \textbf{$\hatpr_\text{M}$} & $44.9\pm0.1\%$ & $39.3\pm0.3\%$ & $43.8\pm0.2\%$\\ \midrule
		 \textbf{$\Delta\haterr$} & \textcolor{blue}{$1.4\pm0.0\%$} & \textcolor{blue}{$1.0\pm0.0\%$} & \textcolor{blue}{$1.4\pm0.0\%$}\\ \midrule
		 \textbf{$\haterr_\text{F}$} & $24.4\pm0.1\%$ & $6.9\pm0.1\%$ & $14.5\pm0.2\%$\\ \midrule
		 \textbf{$\haterr_\text{M}$} & $25.8\pm0.1\%$ & $7.9\pm0.1\%$ & $15.9\pm0.2\%$\\ \midrule
		 \textbf{$\Delta\hatfpr$} & \textcolor{blue}{$1.4\pm0.0\%$} & \textcolor{blue}{$0.1\pm0.1\%$} & \textcolor{blue}{$0.5\pm0.1\%$}\\ \midrule
		 \textbf{$\hatfpr_\text{F}$} & $13.5\pm0.1\%$ & $3.6\pm0.1\%$ & $7.6\pm0.1\%$\\ \midrule
		 \textbf{$\hatfpr_\text{M}$} & $12.1\pm0.1\%$ & $3.5\pm0.2\%$ & $7.1\pm0.2\%$\\ \midrule
		 \textbf{$\Delta\hatfnr$} & \textcolor{blue}{$2.9\pm0.0\%$} & \textcolor{blue}{$1.1\pm0.0\%$} & \textcolor{blue}{$1.9\pm0.1\%$}\\ \midrule
		 \textbf{$\hatfnr_\text{F}$} & $10.9\pm0.1\%$ & $3.3\pm0.1\%$ & $6.9\pm0.1\%$\\ \midrule
		 \textbf{$\hatfnr_\text{M}$} & $13.8\pm0.1\%$ & $4.4\pm0.1\%$ & $8.8\pm0.2\%$\\ \bottomrule
	\end{tabular}
\end{minipage}%
\caption{\textbf{Decision trees} on \texttt{New Adult - CA - Income}, by \texttt{sex}\looseness=-1}
\label{fig:adultnew-income-sex-dtc-all}
\end{figure*}

\vspace{.5cm}
\setlength{\tabcolsep}{6pt}
\begin{figure*}[h!]
\begin{minipage}{.495\linewidth}
\centering
\hspace{-.4cm}
        \includegraphics[width=\linewidth]{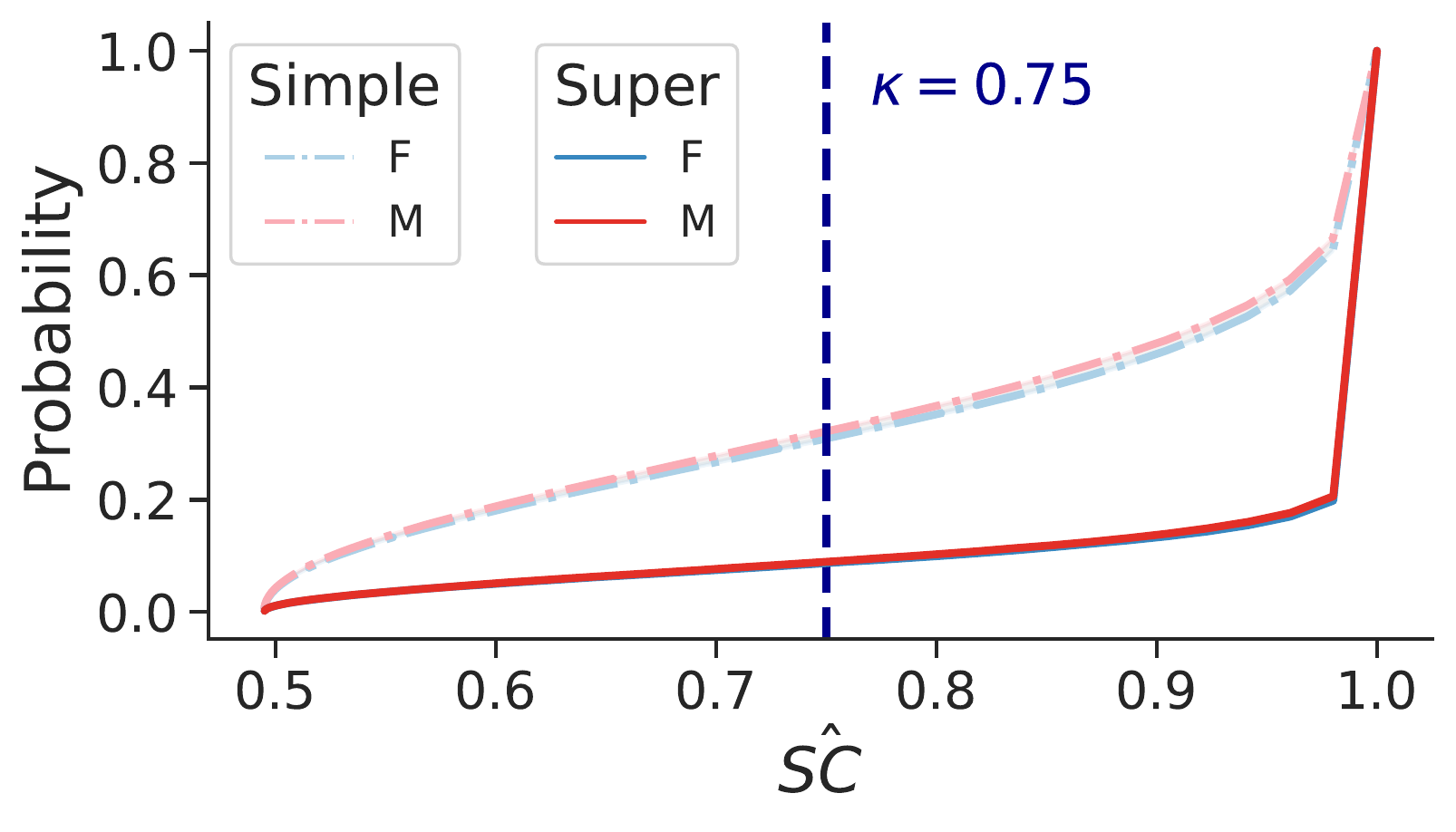}\\
        \vspace{-.1cm}%
	\begin{tabular}{lcc}
		\toprule
		\multicolumn{3}{c}{\textbf{Abstention set metrics}} \\ \cmidrule(lr){1-3}
		 & \textbf{Simple}     & \textbf{Super}      \\ \midrule
		 \textbf{$\Delta\hatar$} & \textcolor{blue}{$1.4\pm0.1\%$} & \textcolor{blue}{$0.2\pm0.0\%$}\\ \midrule
		 \textbf{$\hatar_\text{F}$} & $32.8\pm0.2\%$ & $8.6\pm0.1\%$\\ \midrule
		 \textbf{$\hatar_\text{M}$} & $34.2\pm0.1\%$ & $8.8\pm0.1\%$\\ \bottomrule
	\end{tabular}
\end{minipage}%
\hspace{.25cm}
\begin{minipage}{.495\linewidth}
\centering
\hspace{-.2cm}
	\begin{tabular}{lccc}
		\toprule
		\multicolumn{4}{c}{\textbf{Random forest prediction set metrics}} \\ \cmidrule(lr){1-4}
		 & \textbf{Baseline}     & \textbf{Simple}     & \textbf{Super}      \\ \midrule
		 \textbf{$\Delta\hatpr$} & \textcolor{blue}{$7.4\pm0.1\%$} & \textcolor{blue}{$10.3\pm0.2\%$} & \textcolor{blue}{$8.6\pm0.1\%$}\\ \midrule
		 \textbf{$\hatpr_\text{F}$} & $36.7\pm0.2\%$ & $30.6\pm0.4\%$ & $34.9\pm0.3\%$\\ \midrule
		 \textbf{$\hatpr_\text{M}$} & $44.1\pm0.1\%$ & $40.9\pm0.2\%$ & $43.5\pm0.2\%$\\ \midrule
		 \textbf{$\Delta\haterr$} & \textcolor{blue}{$1.4\pm0.0\%$} & \textcolor{blue}{$1.2\pm0.0\%$} & \textcolor{blue}{$1.4\pm0.1\%$}\\ \midrule
		 \textbf{$\haterr_\text{F}$} & $21.0\pm0.1\%$ & $9.3\pm0.2\%$ & $15.3\pm0.1\%$\\ \midrule
		 \textbf{$\haterr_\text{M}$} & $22.4\pm0.1\%$ & $10.5\pm0.2\%$ & $16.7\pm0.2\%$\\ \midrule
		 \textbf{$\Delta\hatfpr$} & \textcolor{blue}{$1.4\pm0.0\%$} & \textcolor{blue}{$0.5\pm0.0\%$} & \textcolor{blue}{$0.9\pm0.1\%$}\\ \midrule
		 \textbf{$\hatfpr_\text{F}$} & $11.4\pm0.1\%$ & $4.9\pm0.1\%$ & $8.1\pm0.1\%$\\ \midrule
		 \textbf{$\hatfpr_\text{M}$} & $10.0\pm0.1\%$ & $4.4\pm0.1\%$ & $7.2\pm0.2\%$\\ \midrule
		 \textbf{$\Delta\hatfnr$} & \textcolor{blue}{$2.8\pm0.0\%$} & \textcolor{blue}{$1.8\pm0.0\%$} & \textcolor{blue}{$2.3\pm0.0\%$}\\ \midrule
		 \textbf{$\hatfnr_\text{F}$} & $9.6\pm0.1\%$ & $4.4\pm0.1\%$ & $7.2\pm0.1\%$\\ \midrule
		 \textbf{$\hatfnr_\text{M}$} & $12.4\pm0.1\%$ & $6.2\pm0.1\%$ & $9.5\pm0.1\%$\\ \bottomrule
	\end{tabular}
\end{minipage}%
\caption{\textbf{Random forests} on \texttt{New Adult - CA - Income}, by \texttt{sex}\looseness=-1}
\label{fig:adultnew-income-sex-rfc-all}
\end{figure*}
\FloatBarrier

\vspace{1in}
\custompar{\texttt{Income} - by \texttt{race}}

\setlength{\tabcolsep}{6pt}
\begin{figure*}[h!]
\begin{minipage}{.495\linewidth}
\centering
\hspace{-.4cm}
        \includegraphics[width=\linewidth]{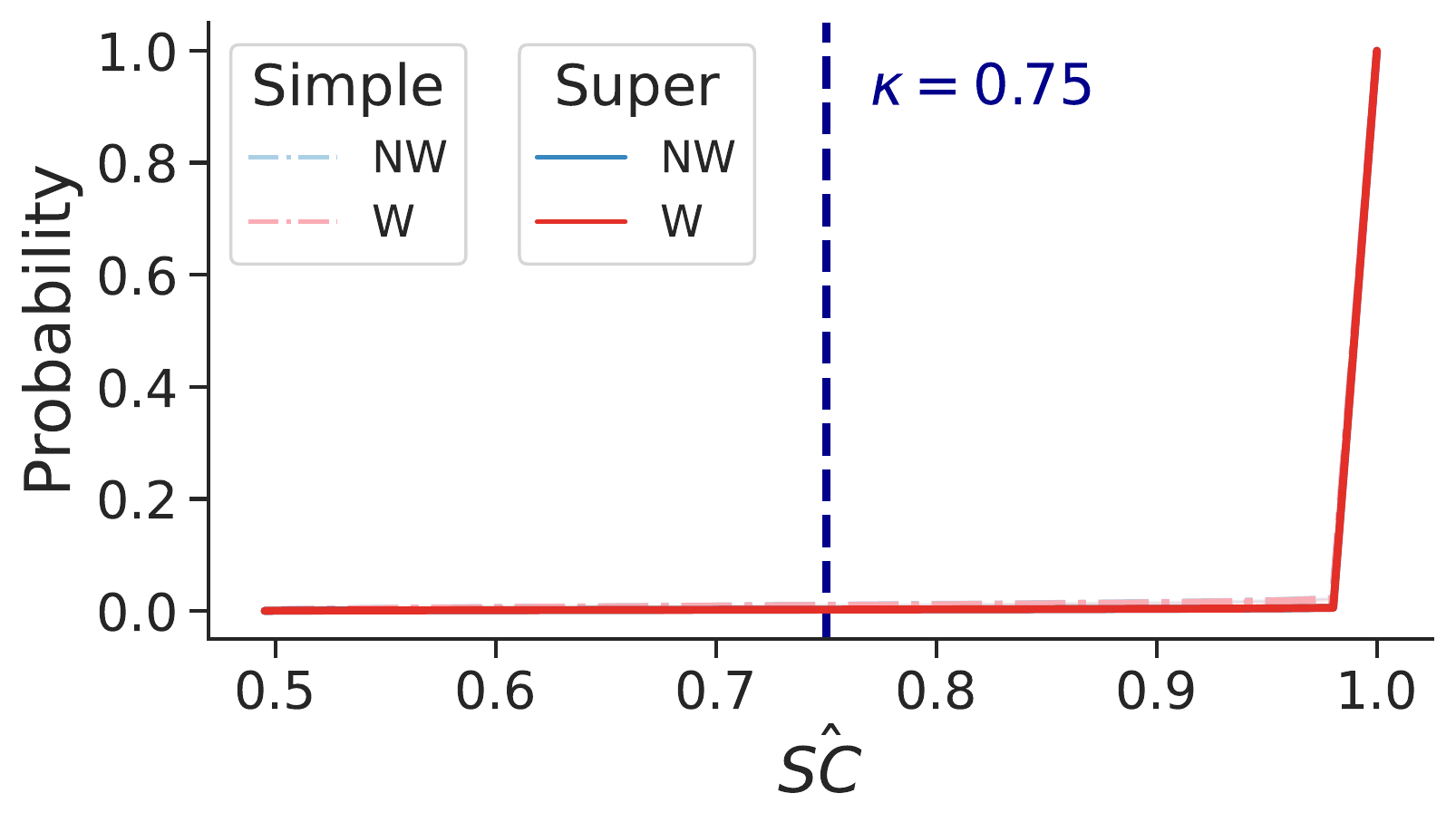}\\
        \vspace{-.1cm}%
	\begin{tabular}{lcc}
		\toprule
		\multicolumn{3}{c}{\textbf{Abstention set metrics}} \\ \cmidrule(lr){1-3}
		 & \textbf{Simple}     & \textbf{Super}      \\ \midrule
		 \textbf{$\Delta\hatar$} & \textcolor{blue}{$0.0\pm0.0\%$} & \textcolor{blue}{$0.0\pm0.0\%$}\\ \midrule
		 \textbf{$\hatar_\text{NW}$} & $1.0\pm0.0\%$ & $0.2\pm0.0\%$\\ \midrule
		 \textbf{$\hatar_\text{W}$} & $1.0\pm0.0\%$ & $0.2\pm0.0\%$\\ \bottomrule
	\end{tabular}
\end{minipage}%
\hspace{.25cm}
\begin{minipage}{.495\linewidth}
\centering
\hspace{-.2cm}
	\begin{tabular}{lccc}
		\toprule
		\multicolumn{4}{c}{\textbf{Logistic regression prediction set metrics}} \\ \cmidrule(lr){1-4}
		 & \textbf{Baseline}     & \textbf{Simple}     & \textbf{Super}      \\ \midrule
		 \textbf{$\Delta\hatpr$} & \textcolor{blue}{$9.2\pm0.1\%$} & \textcolor{blue}{$9.2\pm0.3\%$} & \textcolor{blue}{$9.2\pm0.2\%$}\\ \midrule
		 \textbf{$\hatpr_\text{NW}$} & $34.1\pm0.3\%$ & $33.9\pm0.3\%$ & $34.0\pm0.3\%$\\ \midrule
		 \textbf{$\hatpr_\text{W}$} & $43.3\pm0.2\%$ & $43.1\pm0.0\%$ & $43.2\pm0.1\%$\\ \midrule
		 \textbf{$\Delta\haterr$} & \textcolor{blue}{$0.6\pm0.1\%$} & \textcolor{blue}{$0.4\pm0.1\%$} & \textcolor{blue}{$0.4\pm0.1\%$}\\ \midrule
		 \textbf{$\haterr_\text{NW}$} & $21.6\pm0.2\%$ & $21.4\pm0.2\%$ & $21.6\pm0.2\%$\\ \midrule
		 \textbf{$\haterr_\text{W}$} & $22.2\pm0.1\%$ & $21.8\pm0.1\%$ & $22.0\pm0.1\%$\\ \midrule
		 \textbf{$\Delta\hatfpr$} & \textcolor{blue}{$0.6\pm0.1\%$} & \textcolor{blue}{$0.5\pm0.0\%$} & \textcolor{blue}{$0.5\pm0.0\%$}\\ \midrule
		 \textbf{$\hatfpr_\text{NW}$} & $10.0\pm0.2\%$ & $9.8\pm0.1\%$ & $9.9\pm0.1\%$\\ \midrule
		 \textbf{$\hatfpr_\text{W}$} & $10.6\pm0.1\%$ & $10.3\pm0.1\%$ & $10.4\pm0.1\%$\\ \midrule
		 \textbf{$\Delta\hatfnr$} & \textcolor{blue}{$0.0\pm0.1\%$} & \textcolor{blue}{$0.1\pm0.2\%$} & \textcolor{blue}{$0.1\pm0.3\%$}\\ \midrule
		 \textbf{$\hatfnr_\text{NW}$} & $11.6\pm0.2\%$ & $11.6\pm0.3\%$ & $11.7\pm0.3\%$\\ \midrule
		 \textbf{$\hatfnr_\text{W}$} & $11.6\pm0.1\%$ & $11.5\pm0.1\%$ & $11.6\pm0.0\%$\\ \bottomrule
	\end{tabular}
\end{minipage}%
\caption{\textbf{Logistic regression} on \texttt{New Adult - CA - Income}, by \texttt{race}\looseness=-1}
\label{fig:adultnew-income-race-lr-all}
\end{figure*}

\vspace{.5cm}
\setlength{\tabcolsep}{6pt}
\begin{figure*}[h!]
\begin{minipage}{.495\linewidth}
\centering
\hspace{-.4cm}
        \includegraphics[width=\linewidth]{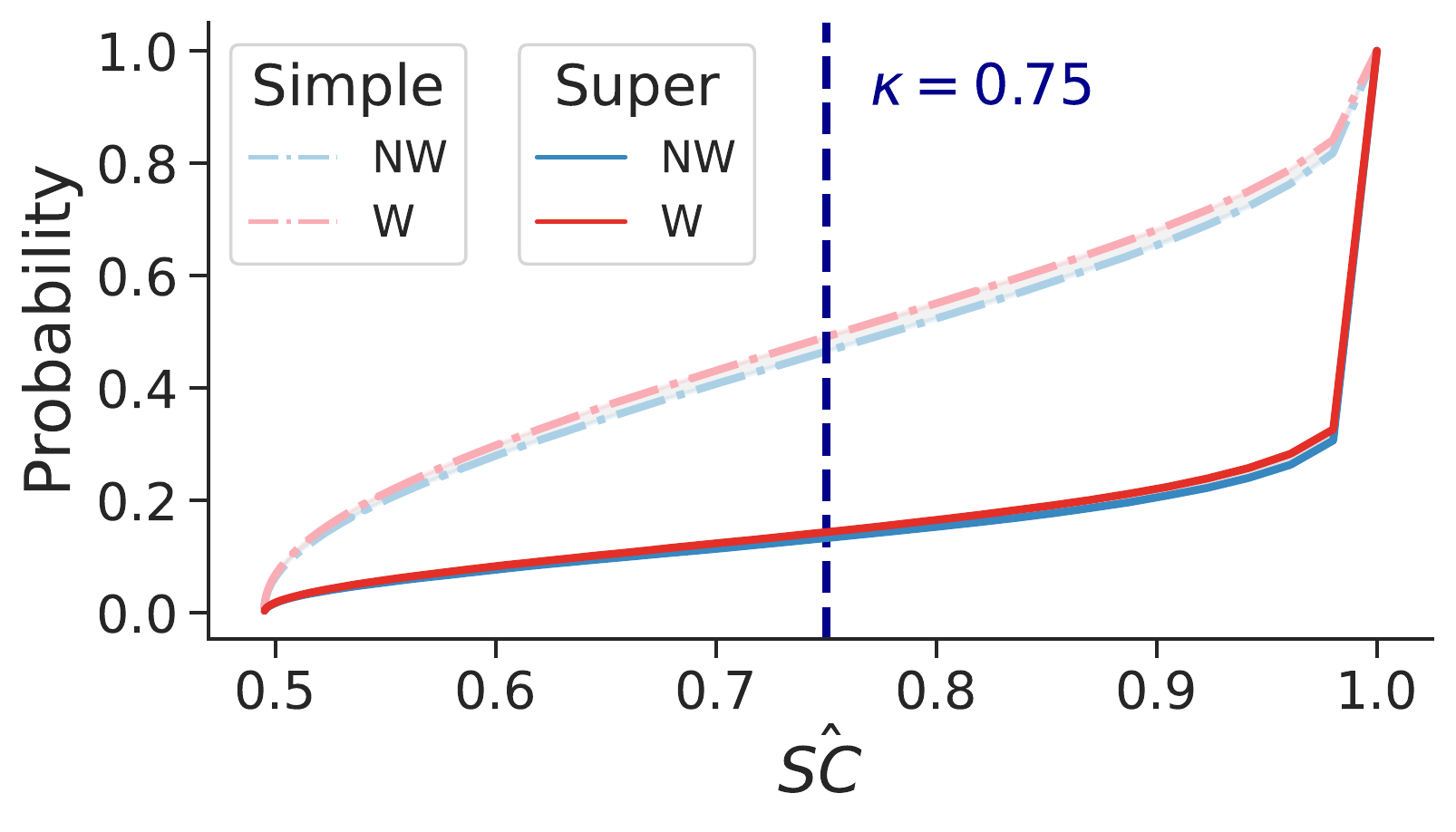}\\
        \vspace{-.1cm}%
	\begin{tabular}{lcc}
		\toprule
		\multicolumn{3}{c}{\textbf{Abstention set metrics}} \\ \cmidrule(lr){1-3}
		 & \textbf{Simple}     & \textbf{Super}      \\ \midrule
		 \textbf{$\Delta\hatar$} & \textcolor{blue}{$2.5\pm0.1\%$} & \textcolor{blue}{$1.0\pm0.1\%$}\\ \midrule
		 \textbf{$\hatar_\text{NW}$} & $48.8\pm0.3\%$ & $13.1\pm0.2\%$\\ \midrule
		 \textbf{$\hatar_\text{W}$} & $51.3\pm0.2\%$ & $14.1\pm0.1\%$\\ \bottomrule
	\end{tabular}
\end{minipage}%
\hspace{.25cm}
\begin{minipage}{.495\linewidth}
\centering
\hspace{-.2cm}
	\begin{tabular}{lccc}
		\toprule
		\multicolumn{4}{c}{\textbf{Decision tree prediction set metrics}} \\ \cmidrule(lr){1-4}
		 & \textbf{Baseline}     & \textbf{Simple}     & \textbf{Super}      \\ \midrule
		 \textbf{$\Delta\hatpr$} & \textcolor{blue}{$7.0\pm0.0\%$} & \textcolor{blue}{$10.3\pm0.0\%$} & \textcolor{blue}{$9.9\pm0.1\%$}\\ \midrule
		 \textbf{$\hatpr_\text{NW}$} & $37.0\pm0.2\%$ & $27.0\pm0.2\%$ & $33.1\pm0.3\%$\\ \midrule
		 \textbf{$\hatpr_\text{W}$} & $44.0\pm0.2\%$ & $37.3\pm0.2\%$ & $43.0\pm0.2\%$\\ \midrule
		 \textbf{$\Delta\haterr$} & \textcolor{blue}{$1.1\pm0.0\%$} & \textcolor{blue}{$0.2\pm0.1\%$} & \textcolor{blue}{$0.6\pm0.0\%$}\\ \midrule
		 \textbf{$\haterr_\text{NW}$} & $24.5\pm0.1\%$ & $7.3\pm0.0\%$ & $14.8\pm0.2\%$\\ \midrule
		 \textbf{$\haterr_\text{W}$} & $25.6\pm0.1\%$ & $7.5\pm0.1\%$ & $15.4\pm0.2\%$\\ \midrule
		 \textbf{$\Delta\hatfpr$} & \textcolor{blue}{$0.2\pm0.0\%$} & \textcolor{blue}{$0.5\pm0.0\%$} & \textcolor{blue}{$0.7\pm0.0\%$}\\ \midrule
		 \textbf{$\hatfpr_\text{NW}$} & $12.9\pm0.1\%$ & $3.2\pm0.1\%$ & $6.9\pm0.1\%$\\ \midrule
		 \textbf{$\hatfpr_\text{W}$} & $12.7\pm0.1\%$ & $3.7\pm0.1\%$ & $7.6\pm0.1\%$\\ \midrule
		 \textbf{$\Delta\hatfnr$} & \textcolor{blue}{$1.4\pm0.0\%$} & \textcolor{blue}{$0.3\pm0.0\%$} & \textcolor{blue}{$0.2\pm0.0\%$}\\ \midrule
		 \textbf{$\hatfnr_\text{NW}$} & $11.6\pm0.1\%$ & $4.1\pm0.1\%$ & $8.0\pm0.1\%$\\ \midrule
		 \textbf{$\hatfnr_\text{W}$} & $13.0\pm0.1\%$ & $3.8\pm0.1\%$ & $7.8\pm0.1\%$\\ \bottomrule
	\end{tabular}
\end{minipage}%
\caption{\textbf{Decision trees} on \texttt{New Adult - CA - Income}, by \texttt{race}\looseness=-1}
\label{fig:adultnew-income-race-dtc-all}
\vspace{2cm}
\end{figure*}

\vspace{.5cm}
\setlength{\tabcolsep}{6pt}
\begin{figure*}[h!]
\begin{minipage}{.495\linewidth}
\centering
\hspace{-.4cm}
        \includegraphics[width=\linewidth]{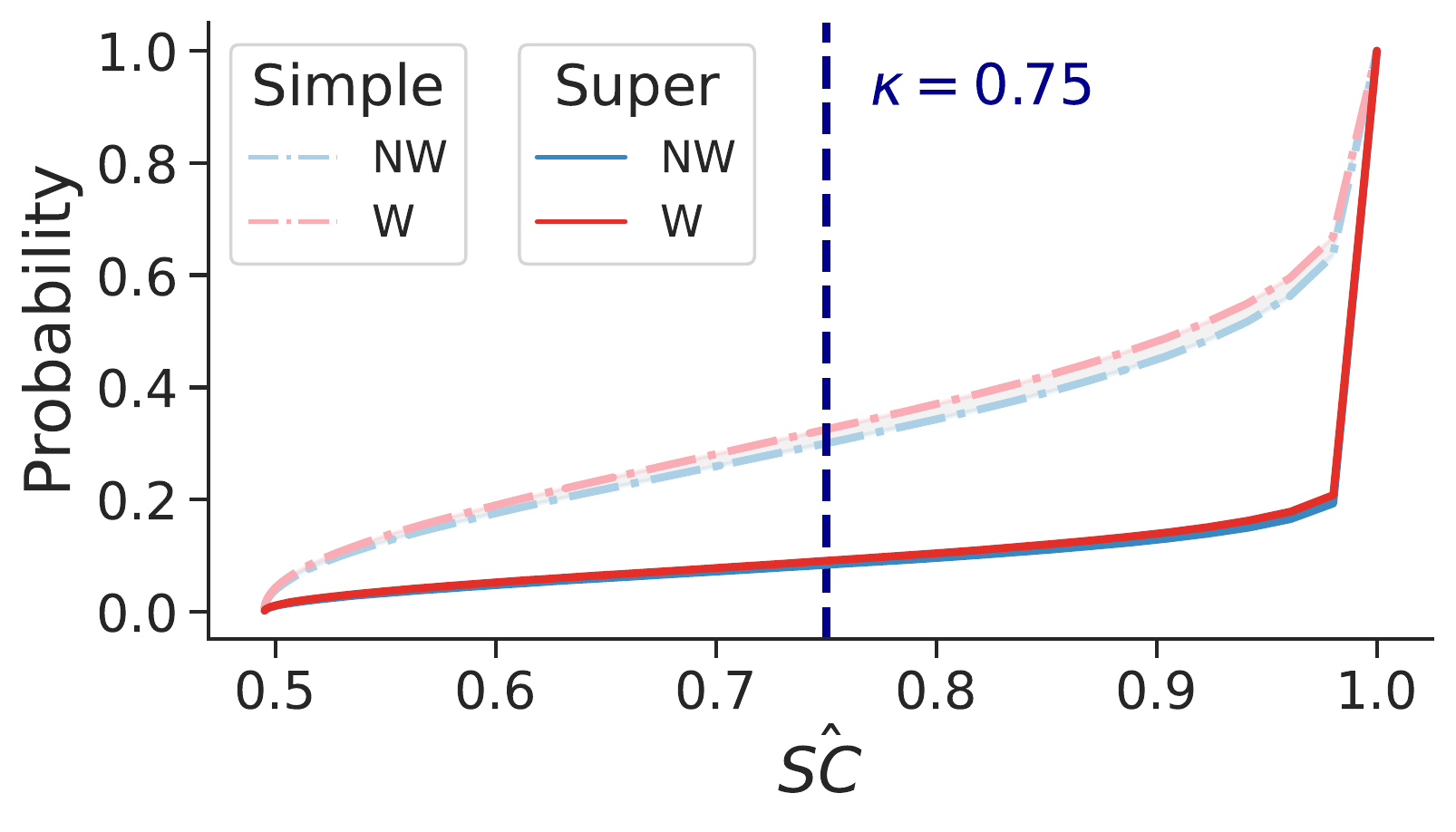}\\
        \vspace{-.1cm}%
	\begin{tabular}{lcc}
		\toprule
		\multicolumn{3}{c}{\textbf{Abstention set metrics}} \\ \cmidrule(lr){1-3}
		 & \textbf{Simple}     & \textbf{Super}      \\ \midrule
		 \textbf{$\Delta\hatar$} & \textcolor{blue}{$2.4\pm0.2\%$} & \textcolor{blue}{$0.8\pm0.0\%$}\\ \midrule
		 \textbf{$\hatar_\text{NW}$} & $32.1\pm0.3\%$ & $8.2\pm0.1\%$\\ \midrule
		 \textbf{$\hatar_\text{W}$} & $34.5\pm0.1\%$ & $9.0\pm0.1\%$\\ \bottomrule
	\end{tabular}
\end{minipage}%
\hspace{.25cm}
\begin{minipage}{.495\linewidth}
\centering
\hspace{-.2cm}
	\begin{tabular}{lccc}
		\toprule
		\multicolumn{4}{c}{\textbf{Random forest prediction set metrics}} \\ \cmidrule(lr){1-4}
		 & \textbf{Baseline}     & \textbf{Simple}     & \textbf{Super}      \\ \midrule
		 \textbf{$\Delta\hatpr$} & \textcolor{blue}{$8.4\pm0.0\%$} & \textcolor{blue}{$11.0\pm0.0\%$} & \textcolor{blue}{$10.1\pm0.2\%$}\\ \midrule
		 \textbf{$\hatpr_\text{NW}$} & $35.4\pm0.2\%$ & $29.3\pm0.2\%$ & $33.2\pm0.3\%$\\ \midrule
		 \textbf{$\hatpr_\text{W}$} & $43.8\pm0.2\%$ & $40.3\pm0.2\%$ & $43.3\pm0.1\%$\\ \midrule
		 \textbf{$\Delta\haterr$} & \textcolor{blue}{$1.2\pm0.0\%$} & \textcolor{blue}{$0.2\pm0.1\%$} & \textcolor{blue}{$0.5\pm0.0\%$}\\ \midrule
		 \textbf{$\haterr_\text{NW}$} & $21.0\pm0.1\%$ & $9.8\pm0.1\%$ & $15.7\pm0.2\%$\\ \midrule
		 \textbf{$\haterr_\text{W}$} & $22.2\pm0.1\%$ & $10.0\pm0.2\%$ & $16.2\pm0.2\%$\\ \midrule
		 \textbf{$\Delta\hatfpr$} & \textcolor{blue}{$0.6\pm0.0\%$} & \textcolor{blue}{$0.7\pm0.0\%$} & \textcolor{blue}{$0.9\pm0.1\%$}\\ \midrule
		 \textbf{$\hatfpr_\text{NW}$} & $10.3\pm0.1\%$ & $4.2\pm0.1\%$ & $7.1\pm0.2\%$\\ \midrule
		 \textbf{$\hatfpr_\text{W}$} & $10.9\pm0.1\%$ & $4.9\pm0.1\%$ & $8.0\pm0.1\%$\\ \midrule
		 \textbf{$\Delta\hatfnr$} & \textcolor{blue}{$0.7\pm0.0\%$} & \textcolor{blue}{$0.5\pm0.0\%$} & \textcolor{blue}{$0.3\pm0.1\%$}\\ \midrule
		 \textbf{$\hatfnr_\text{NW}$} & $10.7\pm0.1\%$ & $5.6\pm0.1\%$ & $8.6\pm0.2\%$\\ \midrule
		 \textbf{$\hatfnr_\text{W}$} & $11.4\pm0.1\%$ & $5.1\pm0.1\%$ & $8.3\pm0.1\%$\\ \bottomrule
	\end{tabular}
\end{minipage}%
\caption{\textbf{Random forests} on \texttt{New Adult - CA - Income}, by \texttt{race}\looseness=-1}
\label{fig:adultnew-income-race-rfc-all}
\end{figure*}
\FloatBarrier


\pagebreak
\custompar{\texttt{Employment} - by \texttt{sex}}

\setlength{\tabcolsep}{6pt}
\begin{figure*}[h!]
\begin{minipage}{.495\linewidth}
\centering
\hspace{-.4cm}
        \includegraphics[width=\linewidth]{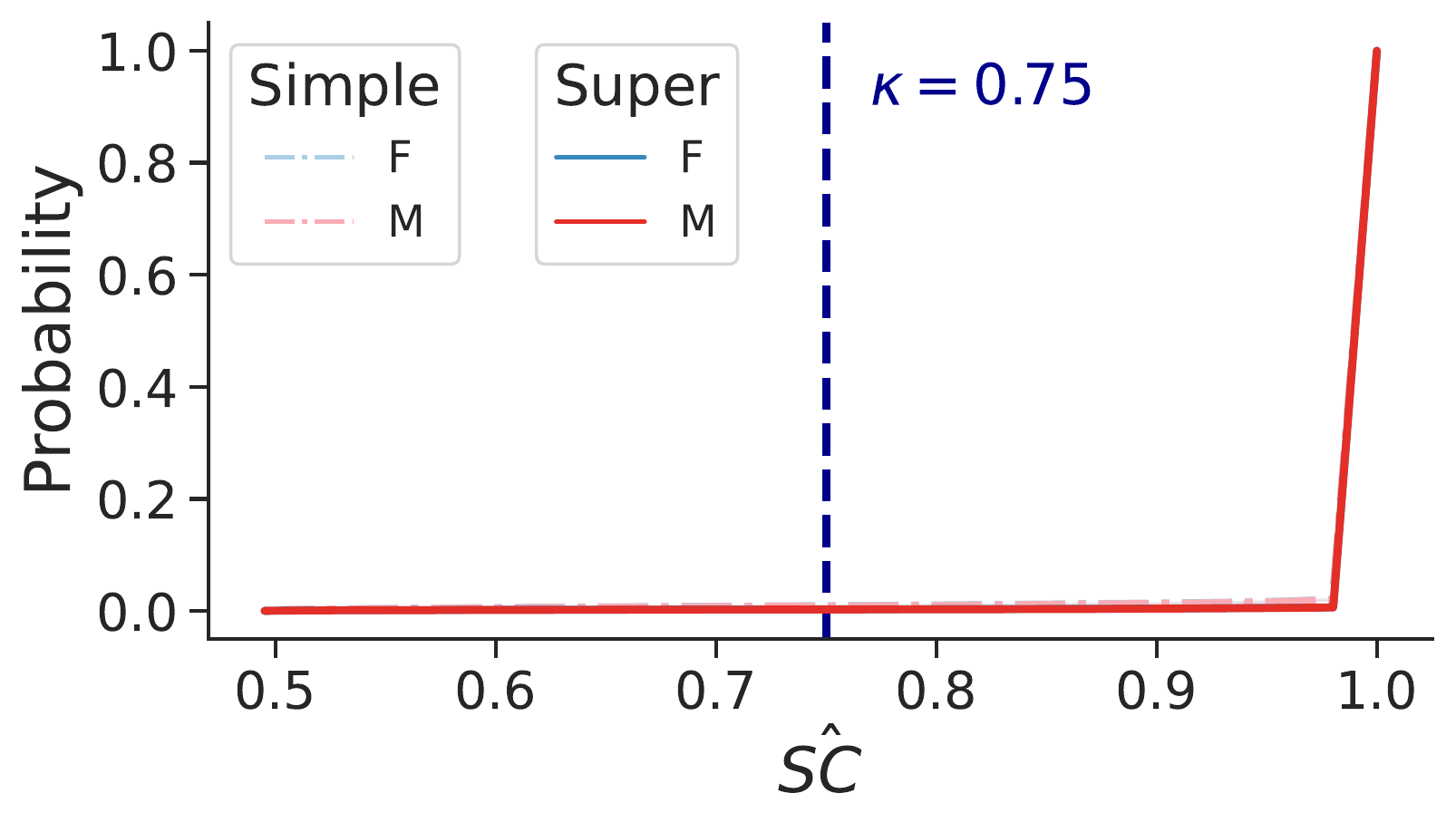}\\
        \vspace{-.1cm}%
	\begin{tabular}{lcc}
		\toprule
		\multicolumn{3}{c}{\textbf{Abstention set metrics}} \\ \cmidrule(lr){1-3}
		 & \textbf{Simple}     & \textbf{Super}      \\ \midrule
		 \textbf{$\Delta\hatar$} & \textcolor{blue}{$0.1\pm0.0\%$} & \textcolor{blue}{$0.0\pm0.0\%$}\\ \midrule
		 \textbf{$\hatar_\text{F}$} & $0.8\pm0.0\%$ & $0.3\pm0.0\%$\\ \midrule
		 \textbf{$\hatar_\text{M}$} & $0.7\pm0.0\%$ & $0.3\pm0.0\%$\\ \bottomrule
	\end{tabular}
\end{minipage}%
\hspace{.25cm}
\begin{minipage}{.495\linewidth}
\centering
\hspace{-.2cm}
	\begin{tabular}{lccc}
		\toprule
		\multicolumn{4}{c}{\textbf{Logistic regression prediction set metrics}} \\ \cmidrule(lr){1-4}
		 & \textbf{Baseline}     & \textbf{Simple}     & \textbf{Super}      \\ \midrule
		 \textbf{$\Delta\hatpr$} & \textcolor{blue}{$4.3\pm0.1\%$} & \textcolor{blue}{$4.5\pm0.0\%$} & \textcolor{blue}{$4.4\pm0.0\%$}\\ \midrule
		 \textbf{$\hatpr_\text{F}$} & $56.6\pm0.1\%$ & $56.8\pm0.1\%$ & $56.7\pm0.1\%$\\ \midrule
		 \textbf{$\hatpr_\text{M}$} & $52.3\pm0.2\%$ & $52.3\pm0.1\%$ & $52.3\pm0.1\%$\\ \midrule
		 \textbf{$\Delta\haterr$} & \textcolor{blue}{$5.0\pm0.0\%$} & \textcolor{blue}{$4.9\pm0.0\%$} & \textcolor{blue}{$4.9\pm0.0\%$}\\ \midrule
		 \textbf{$\haterr_\text{F}$} & $25.8\pm0.1\%$ & $25.5\pm0.1\%$ & $25.6\pm0.1\%$\\ \midrule
		 \textbf{$\haterr_\text{M}$} & $20.8\pm0.1\%$ & $20.6\pm0.1\%$ & $20.7\pm0.1\%$\\ \midrule
		 \textbf{$\Delta\hatfpr$} & \textcolor{blue}{$8.1\pm0.0\%$} & \textcolor{blue}{$8.1\pm0.0\%$} & \textcolor{blue}{$8.1\pm0.0\%$}\\ \midrule
		 \textbf{$\hatfpr_\text{F}$} & $20.1\pm0.1\%$ & $20.1\pm0.0\%$ & $20.1\pm0.0\%$\\ \midrule
		 \textbf{$\hatfpr_\text{M}$} & $12.0\pm0.1\%$ & $12.0\pm0.0\%$ & $12.0\pm0.0\%$\\ \midrule
		 \textbf{$\Delta\hatfnr$} & \textcolor{blue}{$3.1\pm0.0\%$} & \textcolor{blue}{$3.2\pm0.1\%$} & \textcolor{blue}{$3.2\pm0.1\%$}\\ \midrule
		 \textbf{$\hatfnr_\text{F}$} & $5.7\pm0.1\%$ & $5.4\pm0.0\%$ & $5.5\pm0.0\%$\\ \midrule
		 \textbf{$\hatfnr_\text{M}$} & $8.8\pm0.1\%$ & $8.6\pm0.1\%$ & $8.7\pm0.1\%$\\ \bottomrule
	\end{tabular}
\end{minipage}%
\caption{\textbf{Logistic regression} on \texttt{New Adult - CA - Employment}, by \texttt{sex}\looseness=-1}
\label{fig:adultnew-employment-sex-lr-all}
\vspace{2cm}
\end{figure*}

\vspace{.5cm}
\setlength{\tabcolsep}{6pt}
\begin{figure*}[h!]
\begin{minipage}{.495\linewidth}
\centering
\hspace{-.4cm}
        \includegraphics[width=\linewidth]{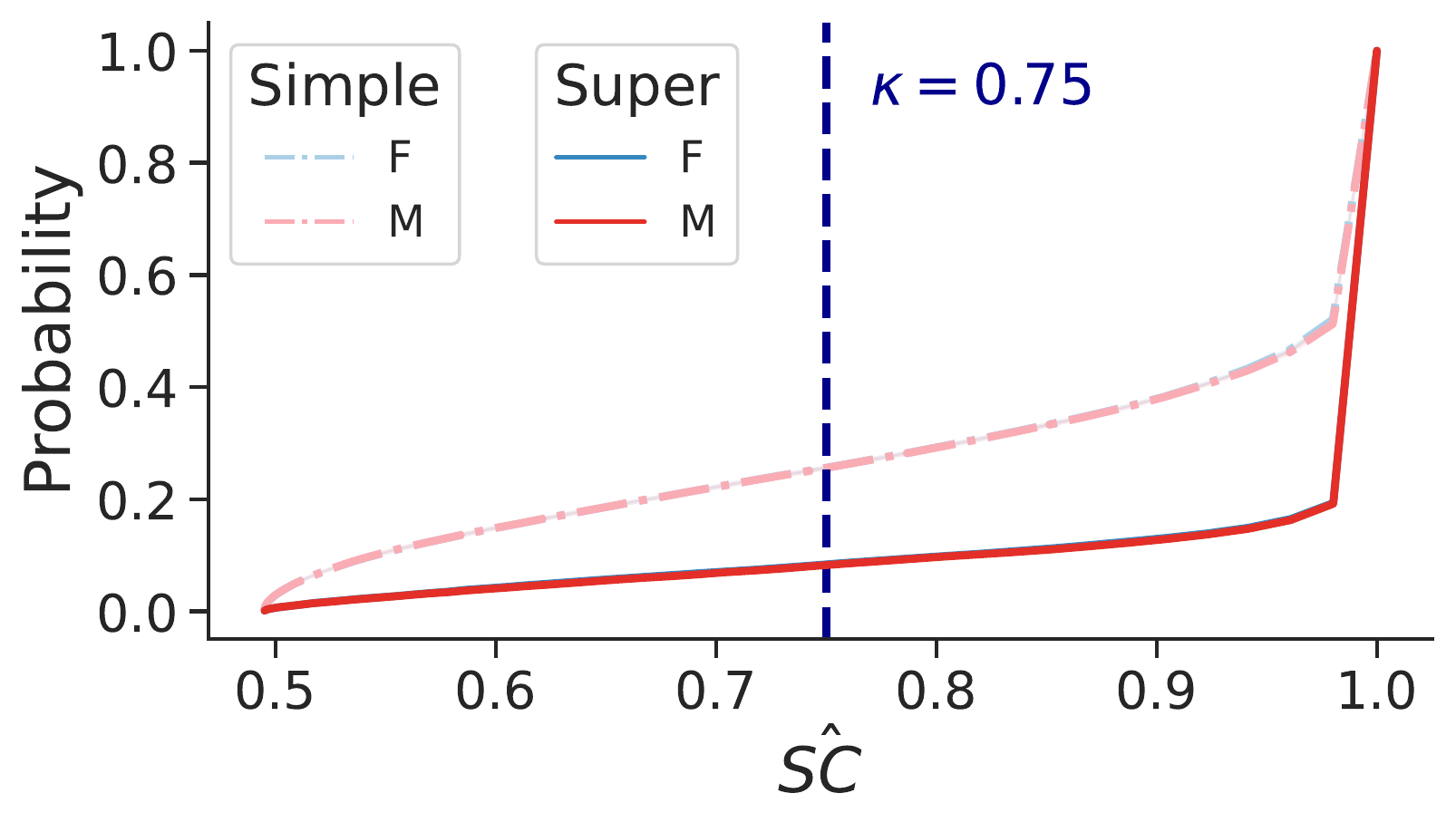}\\
        \vspace{-.1cm}%
	\begin{tabular}{lcc}
		\toprule
		\multicolumn{3}{c}{\textbf{Abstention set metrics}} \\ \cmidrule(lr){1-3}
		 & \textbf{Simple}     & \textbf{Super}      \\ \midrule
		 \textbf{$\Delta\hatar$} & \textcolor{blue}{$0.2\pm0.0\%$} & \textcolor{blue}{$0.1\pm0.1\%$}\\ \midrule
		 \textbf{$\hatar_\text{F}$} & $22.5\pm0.1\%$ & $8.2\pm0.2\%$\\ \midrule
		 \textbf{$\hatar_\text{M}$} & $22.3\pm0.1\%$ & $8.1\pm0.3\%$\\ \bottomrule
	\end{tabular}
\end{minipage}%
\hspace{.25cm}
\begin{minipage}{.495\linewidth}
\centering
\hspace{-.2cm}
	\begin{tabular}{lccc}
		\toprule
		\multicolumn{4}{c}{\textbf{Decision tree prediction set metrics}} \\ \cmidrule(lr){1-4}
		 & \textbf{Baseline}     & \textbf{Simple}     & \textbf{Super}      \\ \midrule
		 \textbf{$\Delta\hatpr$} & \textcolor{blue}{$0.5\pm0.0\%$} & \textcolor{blue}{$0.7\pm0.2\%$} & \textcolor{blue}{$0.6\pm0.1\%$}\\ \midrule
		 \textbf{$\hatpr_\text{F}$} & $50.3\pm0.2\%$ & $50.0\pm0.3\%$ & $51.2\pm0.2\%$\\ \midrule
		 \textbf{$\hatpr_\text{M}$} & $49.8\pm0.2\%$ & $49.3\pm0.1\%$ & $50.6\pm0.1\%$\\ \midrule
		 \textbf{$\Delta\haterr$} & \textcolor{blue}{$4.8\pm0.0\%$} & \textcolor{blue}{$5.8\pm0.1\%$} & \textcolor{blue}{$5.8\pm0.1\%$}\\ \midrule
		 \textbf{$\haterr_\text{F}$} & $24.8\pm0.1\%$ & $17.8\pm0.1\%$ & $20.9\pm0.2\%$\\ \midrule
		 \textbf{$\haterr_\text{M}$} & $20.0\pm0.1\%$ & $12.0\pm0.0\%$ & $15.1\pm0.1\%$\\ \midrule
		 \textbf{$\Delta\hatfpr$} & \textcolor{blue}{$6.1\pm0.0\%$} & \textcolor{blue}{$6.2\pm0.1\%$} & \textcolor{blue}{$6.4\pm0.1\%$}\\ \midrule
		 \textbf{$\hatfpr_\text{F}$} & $16.5\pm0.1\%$ & $13.8\pm0.1\%$ & $15.2\pm0.1\%$\\ \midrule
		 \textbf{$\hatfpr_\text{M}$} & $10.4\pm0.1\%$ & $7.6\pm0.0\%$ & $8.8\pm0.0\%$\\ \midrule
		 \textbf{$\Delta\hatfnr$} & \textcolor{blue}{$1.3\pm0.0\%$} & \textcolor{blue}{$0.3\pm0.0\%$} & \textcolor{blue}{$0.7\pm0.1\%$}\\ \midrule
		 \textbf{$\hatfnr_\text{F}$} & $8.3\pm0.1\%$ & $4.0\pm0.1\%$ & $5.6\pm0.2\%$\\ \midrule
		 \textbf{$\hatfnr_\text{M}$} & $9.6\pm0.1\%$ & $4.3\pm0.1\%$ & $6.3\pm0.1\%$\\ \bottomrule
	\end{tabular}
\end{minipage}%
\caption{\textbf{Decision trees} on \texttt{New Adult - CA - Employment}, by \texttt{sex}\looseness=-1}
\label{fig:adultnew-employment-sex-dtc-all}
\end{figure*}

\vspace{.5cm}
\setlength{\tabcolsep}{6pt}
\begin{figure*}[h!]
\begin{minipage}{.495\linewidth}
\centering
\hspace{-.4cm}
        \includegraphics[width=\linewidth]{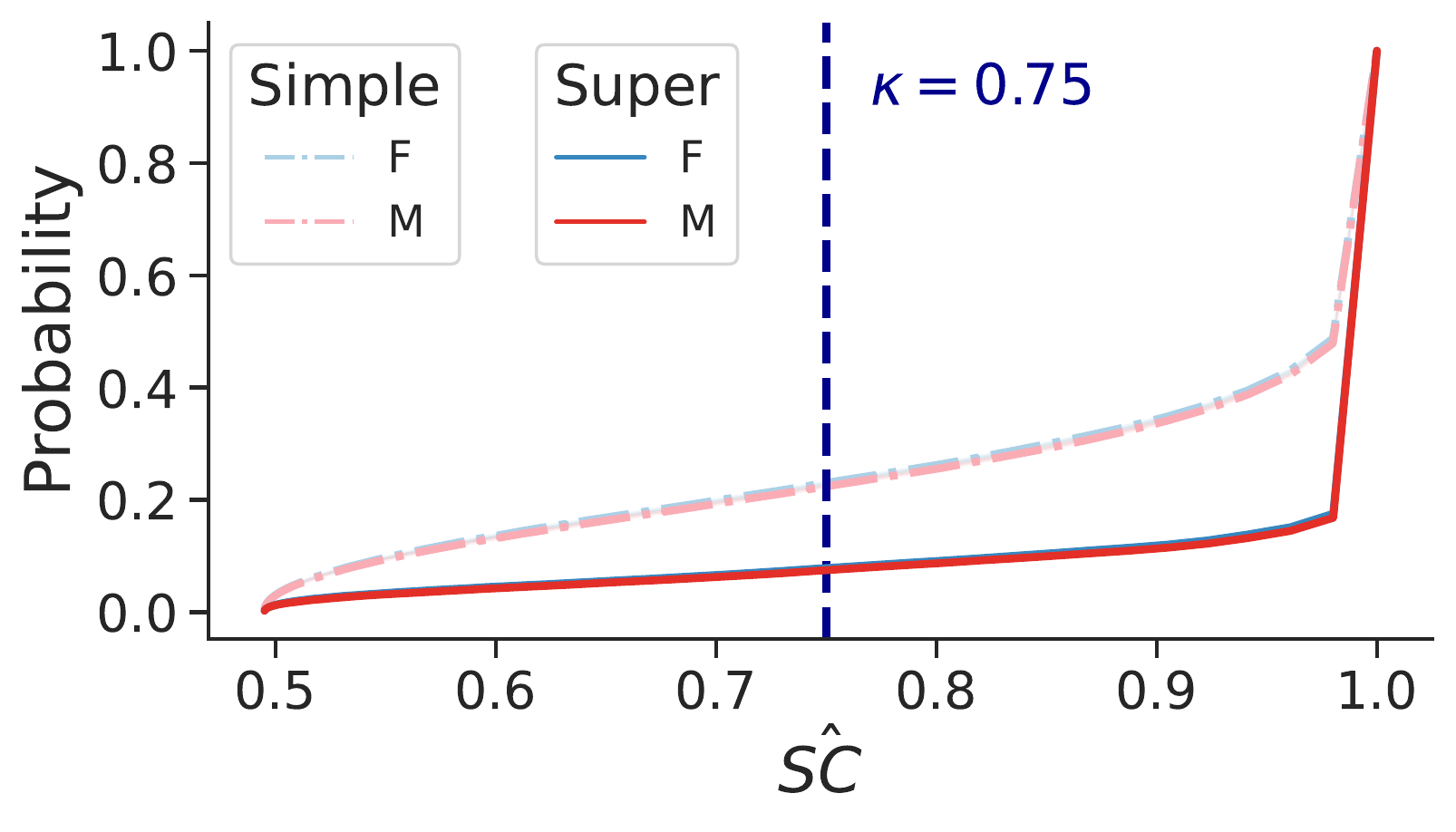}\\
        \vspace{-.1cm}%
	\begin{tabular}{lcc}
		\toprule
		\multicolumn{3}{c}{\textbf{Abstention set metrics}} \\ \cmidrule(lr){1-3}
		 & \textbf{Simple}     & \textbf{Super}      \\ \midrule
		 \textbf{$\Delta\hatar$} & \textcolor{blue}{$0.7\pm0.0\%$} & \textcolor{blue}{$0.4\pm0.0\%$}\\ \midrule
		 \textbf{$\hatar_\text{F}$} & $20.3\pm0.2\%$ & $7.7\pm0.2\%$\\ \midrule
		 \textbf{$\hatar_\text{M}$} & $19.6\pm0.2\%$ & $7.3\pm0.2\%$\\ \bottomrule
	\end{tabular}
\end{minipage}%
\hspace{.25cm}
\begin{minipage}{.495\linewidth}
\centering
\hspace{-.2cm}
	\begin{tabular}{lccc}
		\toprule
		\multicolumn{4}{c}{\textbf{Random forest prediction set metrics}} \\ \cmidrule(lr){1-4}
		 & \textbf{Baseline}     & \textbf{Simple}     & \textbf{Super}      \\ \midrule
		 \textbf{$\Delta\hatpr$} & \textcolor{blue}{$0.3\pm0.1\%$} & \textcolor{blue}{$0.5\pm0.1\%$} & \textcolor{blue}{$0.6\pm0.1\%$}\\ \midrule
		 \textbf{$\hatpr_\text{F}$} & $49.2\pm0.1\%$ & $48.7\pm0.2\%$ & $50.3\pm0.2\%$\\ \midrule
		 \textbf{$\hatpr_\text{M}$} & $48.9\pm0.2\%$ & $48.2\pm0.1\%$ & $49.7\pm0.1\%$\\ \midrule
		 \textbf{$\Delta\haterr$} & \textcolor{blue}{$4.8\pm0.0\%$} & \textcolor{blue}{$5.4\pm0.0\%$} & \textcolor{blue}{$5.5\pm0.2\%$}\\ \midrule
		 \textbf{$\haterr_\text{F}$} & $24.0\pm0.1\%$ & $17.5\pm0.1\%$ & $20.5\pm0.2\%$\\ \midrule
		 \textbf{$\haterr_\text{M}$} & $19.2\pm0.1\%$ & $12.1\pm0.1\%$ & $15.0\pm0.0\%$\\ \midrule
		 \textbf{$\Delta\hatfpr$} & \textcolor{blue}{$6.0\pm0.0\%$} & \textcolor{blue}{$5.9\pm0.1\%$} & \textcolor{blue}{$6.2\pm0.1\%$}\\ \midrule
		 \textbf{$\hatfpr_\text{F}$} & $15.5\pm0.1\%$ & $13.2\pm0.1\%$ & $14.7\pm0.1\%$\\ \midrule
		 \textbf{$\hatfpr_\text{M}$} & $9.5\pm0.1\%$ & $7.3\pm0.0\%$ & $8.5\pm0.0\%$\\ \midrule
		 \textbf{$\Delta\hatfnr$} & \textcolor{blue}{$1.2\pm0.0\%$} & \textcolor{blue}{$0.4\pm0.0\%$} & \textcolor{blue}{$0.8\pm0.0\%$}\\ \midrule
		 \textbf{$\hatfnr_\text{F}$} & $8.5\pm0.1\%$ & $4.3\pm0.1\%$ & $5.8\pm0.1\%$\\ \midrule
		 \textbf{$\hatfnr_\text{M}$} & $9.7\pm0.1\%$ & $4.7\pm0.1\%$ & $6.6\pm0.1\%$\\ \bottomrule
	\end{tabular}
\end{minipage}%
\caption{\textbf{Random forests} on \texttt{New Adult - CA - Employment}, by \texttt{sex}\looseness=-1}
\label{fig:adultnew-employment-sex-rfc-all}
\end{figure*}
\FloatBarrier

\vspace*{.25in}
\custompar{\texttt{Employment} - by \texttt{race}}

\setlength{\tabcolsep}{6pt}
\begin{figure*}[h!]
\begin{minipage}{.495\linewidth}
\centering
\hspace{-.4cm}
        \includegraphics[width=\linewidth]{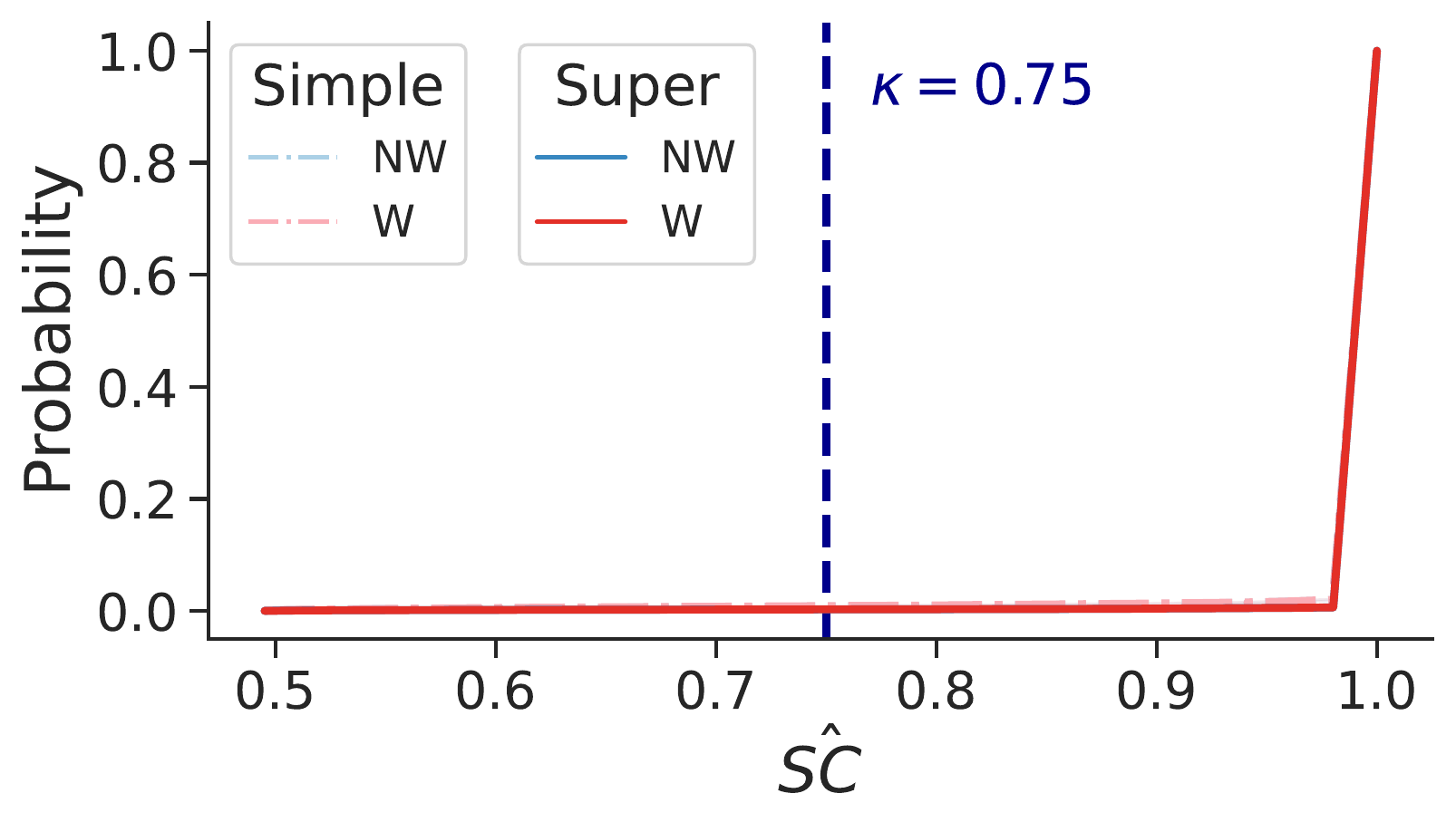}\\
        \vspace{-.1cm}%
	\begin{tabular}{lcc}
		\toprule
		\multicolumn{3}{c}{\textbf{Abstention set metrics}} \\ \cmidrule(lr){1-3}
		 & \textbf{Simple}     & \textbf{Super}      \\ \midrule
		 \textbf{$\Delta\hatar$} & \textcolor{blue}{$0.1\pm0.0\%$} & \textcolor{blue}{$0.0\pm0.0\%$}\\ \midrule
		 \textbf{$\hatar_\text{NW}$} & $0.7\pm0.0\%$ & $0.3\pm0.0\%$\\ \midrule
		 \textbf{$\hatar_\text{W}$} & $0.8\pm0.0\%$ & $0.3\pm0.0\%$\\ \bottomrule
	\end{tabular}
\end{minipage}%
\hspace{.25cm}
\begin{minipage}{.495\linewidth}
\centering
\hspace{-.2cm}
	\begin{tabular}{lccc}
		\toprule
		\multicolumn{4}{c}{\textbf{Logistic regression prediction set metrics}} \\ \cmidrule(lr){1-4}
		 & \textbf{Baseline}     & \textbf{Simple}     & \textbf{Super}      \\ \midrule
		 \textbf{$\Delta\hatpr$} & \textcolor{blue}{$1.1\pm0.2\%$} & \textcolor{blue}{$1.2\pm0.1\%$} & \textcolor{blue}{$1.2\pm0.1\%$}\\ \midrule
		 \textbf{$\hatpr_\text{NW}$} & $55.2\pm0.3\%$ & $55.3\pm0.2\%$ & $55.3\pm0.2\%$\\ \midrule
		 \textbf{$\hatpr_\text{W}$} & $54.1\pm0.1\%$ & $54.1\pm0.1\%$ & $54.1\pm0.1\%$\\ \midrule
		 \textbf{$\Delta\haterr$} & \textcolor{blue}{$0.1\pm0.0\%$} & \textcolor{blue}{$0.2\pm0.1\%$} & \textcolor{blue}{$0.2\pm0.0\%$}\\ \midrule
		 \textbf{$\haterr_\text{NW}$} & $23.3\pm0.1\%$ & $23.0\pm0.0\%$ & $23.1\pm0.1\%$\\ \midrule
		 \textbf{$\haterr_\text{W}$} & $23.4\pm0.1\%$ & $23.2\pm0.1\%$ & $23.3\pm0.1\%$\\ \midrule
		 \textbf{$\Delta\hatfpr$} & \textcolor{blue}{$0.8\pm0.0\%$} & \textcolor{blue}{$0.7\pm0.1\%$} & \textcolor{blue}{$0.7\pm0.1\%$}\\ \midrule
		 \textbf{$\hatfpr_\text{NW}$} & $16.6\pm0.1\%$ & $16.5\pm0.0\%$ & $16.6\pm0.0\%$\\ \midrule
		 \textbf{$\hatfpr_\text{W}$} & $15.8\pm0.1\%$ & $15.8\pm0.1\%$ & $15.9\pm0.1\%$\\ \midrule
		 \textbf{$\Delta\hatfnr$} & \textcolor{blue}{$1.0\pm0.0\%$} & \textcolor{blue}{$1.0\pm0.0\%$} & \textcolor{blue}{$0.9\pm0.0\%$}\\ \midrule
		 \textbf{$\hatfnr_\text{NW}$} & $6.6\pm0.1\%$ & $6.4\pm0.1\%$ & $6.5\pm0.1\%$\\ \midrule
		 \textbf{$\hatfnr_\text{W}$} & $7.6\pm0.1\%$ & $7.4\pm0.1\%$ & $7.4\pm0.1\%$\\ \bottomrule
	\end{tabular}
\end{minipage}%
\caption{\textbf{Logistic regression} on \texttt{New Adult - CA - Employment}, by \texttt{race}\looseness=-1}
\label{fig:adultnew-employment-race-lr-all}
\end{figure*}

\vspace{.1cm}
\setlength{\tabcolsep}{6pt}
\begin{figure*}[h!]
\begin{minipage}{.495\linewidth}
\centering
\hspace{-.4cm}
        \includegraphics[width=\linewidth]{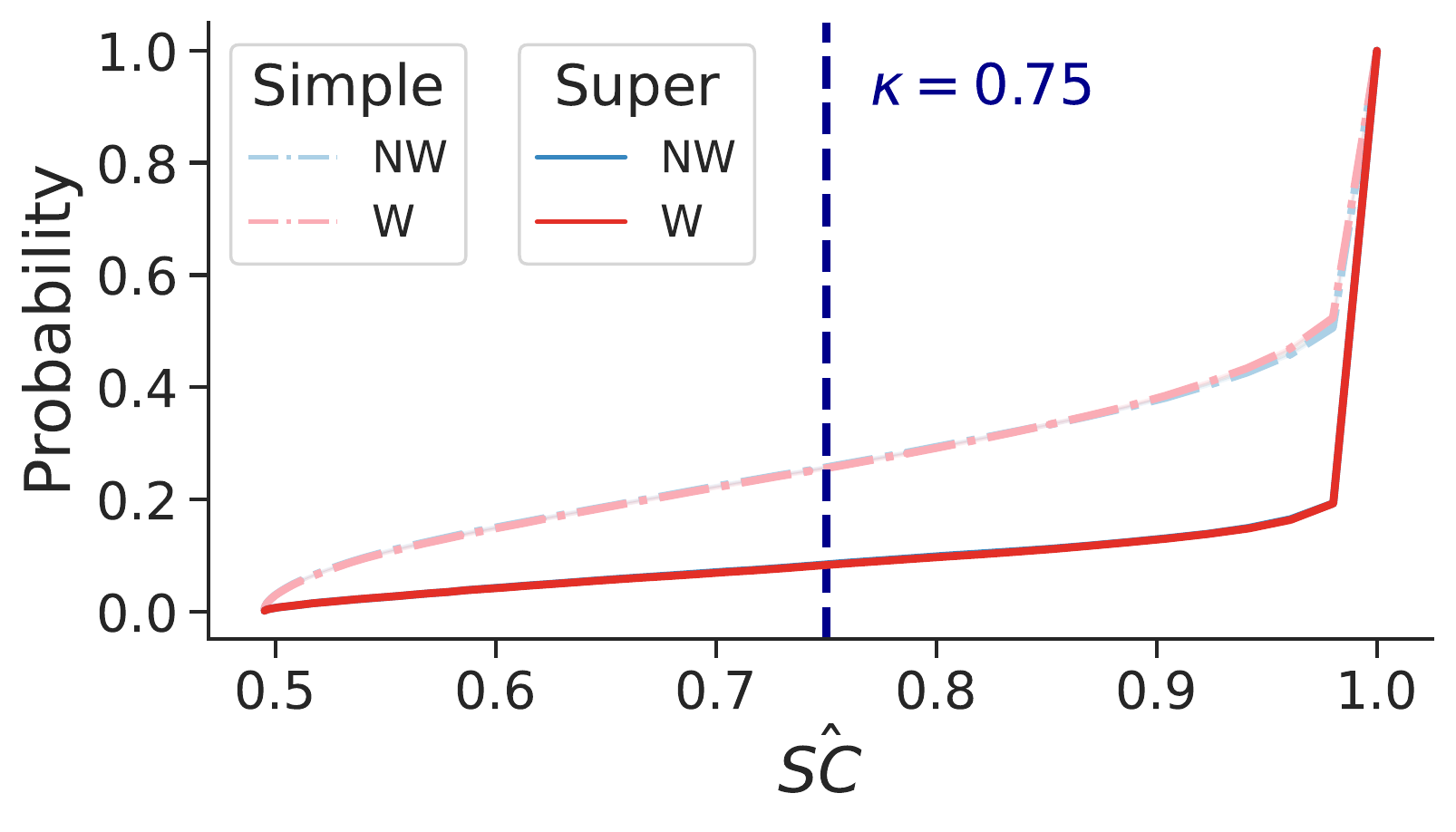}\\
        \vspace{-.1cm}%
	\begin{tabular}{lcc}
		\toprule
		\multicolumn{3}{c}{\textbf{Abstention set metrics}} \\ \cmidrule(lr){1-3}
		 & \textbf{Simple}     & \textbf{Super}      \\ \midrule
		 \textbf{$\Delta\hatar$} & \textcolor{blue}{$0.0\pm0.1\%$} & \textcolor{blue}{$0.1\pm0.1\%$}\\ \midrule
		 \textbf{$\hatar_\text{NW}$} & $22.4\pm0.2\%$ & $8.2\pm0.2\%$\\ \midrule
		 \textbf{$\hatar_\text{W}$} & $22.4\pm0.1\%$ & $8.1\pm0.3\%$\\ \bottomrule
	\end{tabular}
\end{minipage}%
\hspace{.25cm}
\begin{minipage}{.495\linewidth}
\centering
\hspace{-.2cm}
	\begin{tabular}{lccc}
		\toprule
		\multicolumn{4}{c}{\textbf{Decision tree prediction set metrics}} \\ \cmidrule(lr){1-4}
		 & \textbf{Baseline}     & \textbf{Simple}     & \textbf{Super}      \\ \midrule
		 \textbf{$\Delta\hatpr$} & \textcolor{blue}{$0.2\pm0.2\%$} & \textcolor{blue}{$0.6\pm0.0\%$} & \textcolor{blue}{$1.2\pm0.1\%$}\\ \midrule
		 \textbf{$\hatpr_\text{NW}$} & $50.2\pm0.3\%$ & $50.0\pm0.1\%$ & $51.6\pm0.0\%$\\ \midrule
		 \textbf{$\hatpr_\text{W}$} & $50.0\pm0.1\%$ & $49.4\pm0.1\%$ & $50.4\pm0.1\%$\\ \midrule
		 \textbf{$\Delta\haterr$} & \textcolor{blue}{$0.6\pm0.0\%$} & \textcolor{blue}{$0.7\pm0.0\%$} & \textcolor{blue}{$0.6\pm0.0\%$}\\ \midrule
		 \textbf{$\haterr_\text{NW}$} & $22.1\pm0.1\%$ & $14.5\pm0.1\%$ & $17.7\pm0.1\%$\\ \midrule
		 \textbf{$\haterr_\text{W}$} & $22.7\pm0.1\%$ & $15.2\pm0.1\%$ & $18.3\pm0.1\%$\\ \midrule
		 \textbf{$\Delta\hatfpr$} & \textcolor{blue}{$0.1\pm0.0\%$} & \textcolor{blue}{$0.6\pm0.1\%$} & \textcolor{blue}{$0.7\pm0.1\%$}\\ \midrule
		 \textbf{$\hatfpr_\text{NW}$} & $13.5\pm0.1\%$ & $11.1\pm0.0\%$ & $12.5\pm0.0\%$\\ \midrule
		 \textbf{$\hatfpr_\text{W}$} & $13.4\pm0.1\%$ & $10.5\pm0.1\%$ & $11.8\pm0.1\%$\\ \midrule
		 \textbf{$\Delta\hatfnr$} & \textcolor{blue}{$0.6\pm0.0\%$} & \textcolor{blue}{$1.3\pm0.0\%$} & \textcolor{blue}{$1.3\pm0.0\%$}\\ \midrule
		 \textbf{$\hatfnr_\text{NW}$} & $8.6\pm0.1\%$ & $3.4\pm0.1\%$ & $5.2\pm0.1\%$\\ \midrule
		 \textbf{$\hatfnr_\text{W}$} & $9.2\pm0.1\%$ & $4.7\pm0.1\%$ & $6.5\pm0.1\%$\\ \bottomrule
	\end{tabular}
\end{minipage}%
\caption{\textbf{Decision trees} on \texttt{New Adult - CA - Employment}, by \texttt{race}\looseness=-1}
\label{fig:adultnew-employment-race-dtc-all}
\vspace{2cm}
\end{figure*}

\vspace{.75cm}
\setlength{\tabcolsep}{6pt}
\begin{figure*}[h!]
\begin{minipage}{.495\linewidth}
\centering
\hspace{-.4cm}
        \includegraphics[width=\linewidth]{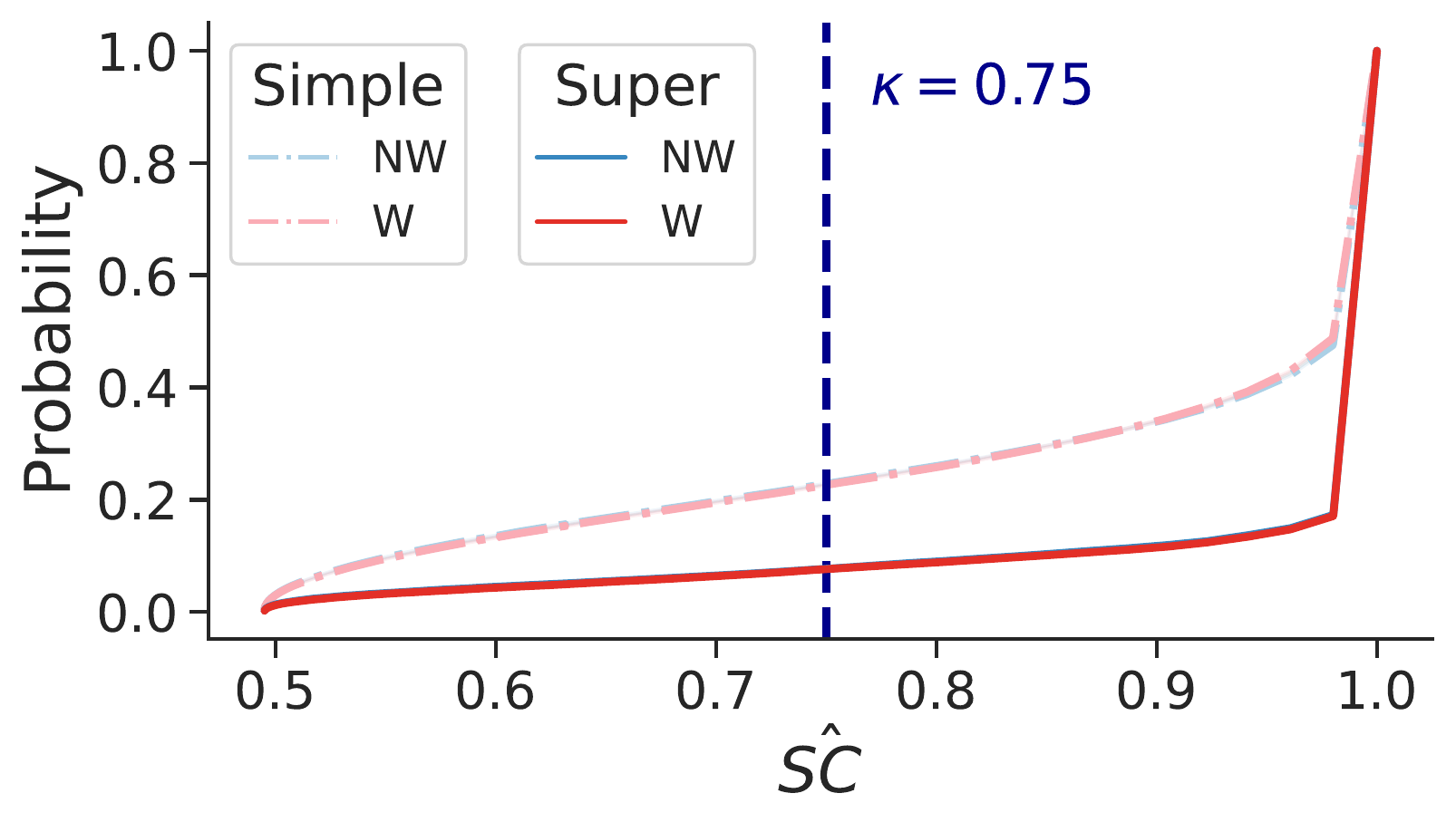}\\
        \vspace{-.1cm}%
	\begin{tabular}{lcc}
		\toprule
		\multicolumn{3}{c}{\textbf{Abstention set metrics}} \\ \cmidrule(lr){1-3}
		 & \textbf{Simple}     & \textbf{Super}      \\ \midrule
		 \textbf{$\Delta\hatar$} & \textcolor{blue}{$0.3\pm0.1\%$} & \textcolor{blue}{$0.1\pm0.1\%$}\\ \midrule
		 \textbf{$\hatar_\text{NW}$} & $20.1\pm0.1\%$ & $7.6\pm0.1\%$\\ \midrule
		 \textbf{$\hatar_\text{W}$} & $19.8\pm0.2\%$ & $7.5\pm0.2\%$\\ \bottomrule
	\end{tabular}
\end{minipage}%
\hspace{.25cm}
\begin{minipage}{.495\linewidth}
\centering
\hspace{-.2cm}
	\begin{tabular}{lccc}
		\toprule
		\multicolumn{4}{c}{\textbf{Random forest prediction set metrics}} \\ \cmidrule(lr){1-4}
		 & \textbf{Baseline}     & \textbf{Simple}     & \textbf{Super}      \\ \midrule
		 \textbf{$\Delta\hatpr$} & \textcolor{blue}{$0.6\pm0.2\%$} & \textcolor{blue}{$1.0\pm0.0\%$} & \textcolor{blue}{$1.4\pm0.1\%$}\\ \midrule
		 \textbf{$\hatpr_\text{NW}$} & $49.4\pm0.3\%$ & $49.1\pm0.1\%$ & $50.9\pm0.0\%$\\ \midrule
		 \textbf{$\hatpr_\text{W}$} & $48.8\pm0.1\%$ & $48.1\pm0.1\%$ & $49.5\pm0.1\%$\\ \midrule
		 \textbf{$\Delta\haterr$} & \textcolor{blue}{$0.5\pm0.0\%$} & \textcolor{blue}{$0.7\pm0.0\%$} & \textcolor{blue}{$0.5\pm0.0\%$}\\ \midrule
		 \textbf{$\haterr_\text{NW}$} & $21.3\pm0.1\%$ & $14.4\pm0.1\%$ & $17.5\pm0.1\%$\\ \midrule
		 \textbf{$\haterr_\text{W}$} & $21.8\pm0.1\%$ & $15.1\pm0.1\%$ & $18.0\pm0.1\%$\\ \midrule
		 \textbf{$\Delta\hatfpr$} & \textcolor{blue}{$0.3\pm0.1\%$} & \textcolor{blue}{$0.6\pm0.1\%$} & \textcolor{blue}{$0.8\pm0.1\%$}\\ \midrule
		 \textbf{$\hatfpr_\text{NW}$} & $12.7\pm0.2\%$ & $10.7\pm0.0\%$ & $12.1\pm0.0\%$\\ \midrule
		 \textbf{$\hatfpr_\text{W}$} & $12.4\pm0.1\%$ & $10.1\pm0.1\%$ & $11.3\pm0.1\%$\\ \midrule
		 \textbf{$\Delta\hatfnr$} & \textcolor{blue}{$0.8\pm0.0\%$} & \textcolor{blue}{$1.3\pm0.0\%$} & \textcolor{blue}{$1.3\pm0.0\%$}\\ \midrule
		 \textbf{$\hatfnr_\text{NW}$} & $8.6\pm0.1\%$ & $3.7\pm0.1\%$ & $5.4\pm0.1\%$\\ \midrule
		 \textbf{$\hatfnr_\text{W}$} & $9.4\pm0.1\%$ & $5.0\pm0.1\%$ & $6.7\pm0.1\%$\\ \bottomrule
	\end{tabular}
\end{minipage}%
\caption{\textbf{Random forests} on \texttt{New Adult - CA - Employment}, by \texttt{race}\looseness=-1}
\label{fig:adultnew-employment-race-rfc-all}
\end{figure*}
\vspace*{2in}
\FloatBarrier


\pagebreak
\custompar{\texttt{Public Coverage} - by \texttt{sex}}

\setlength{\tabcolsep}{6pt}
\begin{figure*}[h!]
\begin{minipage}{.495\linewidth}
\centering
\hspace{-.4cm}
        \includegraphics[width=\linewidth]{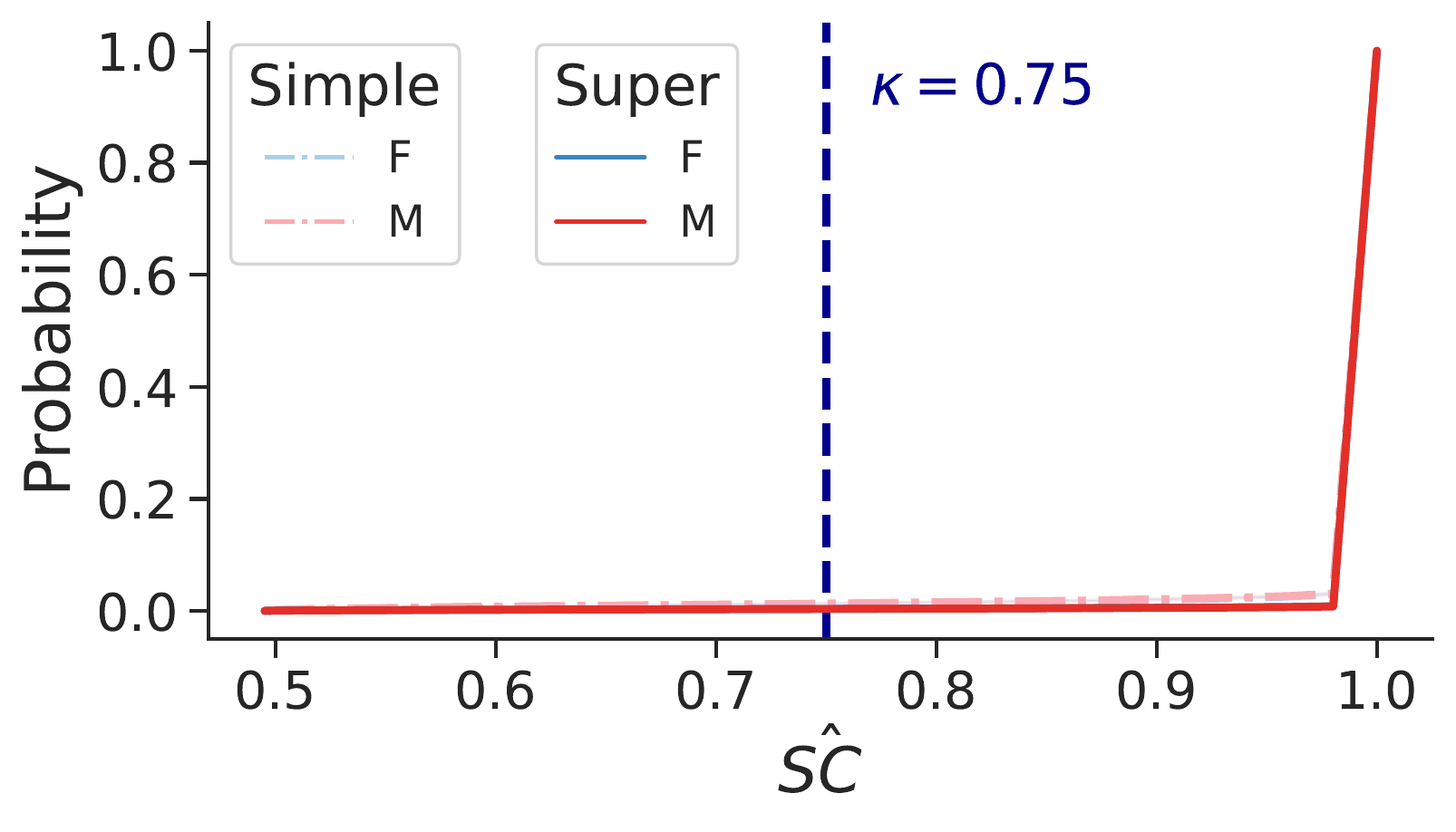}\\
        \vspace{-.1cm}%
	\begin{tabular}{lcc}
		\toprule
		\multicolumn{3}{c}{\textbf{Abstention set metrics}} \\ \cmidrule(lr){1-3}
		 & \textbf{Simple}     & \textbf{Super}      \\ \midrule
		 \textbf{$\Delta\hatar$} & \textcolor{blue}{$0.0\pm0.1\%$} & \textcolor{blue}{$0.0\pm0.0\%$}\\ \midrule
		 \textbf{$\hatar_\text{F}$} & $1.4\pm0.1\%$ & $0.4\pm0.0\%$\\ \midrule
		 \textbf{$\hatar_\text{M}$} & $1.4\pm0.0\%$ & $0.4\pm0.0\%$\\ \bottomrule
	\end{tabular}
\end{minipage}%
\hspace{.25cm}
\begin{minipage}{.495\linewidth}
\centering
\hspace{-.2cm}
	\begin{tabular}{lccc}
		\toprule
		\multicolumn{4}{c}{\textbf{Logistic regression prediction set metrics}} \\ \cmidrule(lr){1-4}
		 & \textbf{Baseline}     & \textbf{Simple}     & \textbf{Super}      \\ \midrule
		 \textbf{$\Delta\hatpr$} & \textcolor{blue}{$2.6\pm0.1\%$} & \textcolor{blue}{$2.6\pm0.1\%$} & \textcolor{blue}{$2.6\pm0.1\%$}\\ \midrule
		 \textbf{$\hatpr_\text{F}$} & $15.1\pm0.2\%$ & $14.7\pm0.2\%$ & $15.1\pm0.2\%$\\ \midrule
		 \textbf{$\hatpr_\text{M}$} & $17.7\pm0.3\%$ & $17.3\pm0.1\%$ & $17.7\pm0.1\%$\\ \midrule
		 \textbf{$\Delta\haterr$} & \textcolor{blue}{$0.9\pm0.1\%$} & \textcolor{blue}{$0.5\pm0.0\%$} & \textcolor{blue}{$0.5\pm0.0\%$}\\ \midrule
		 \textbf{$\haterr_\text{F}$} & $31.2\pm0.3\%$ & $30.8\pm0.2\%$ & $31.0\pm0.2\%$\\ \midrule
		 \textbf{$\haterr_\text{M}$} & $32.1\pm0.2\%$ & $31.3\pm0.2\%$ & $31.5\pm0.2\%$\\ \midrule
		 \textbf{$\Delta\hatfpr$} & \textcolor{blue}{$0.0\pm0.0\%$} & \textcolor{blue}{$0.0\pm0.0\%$} & \textcolor{blue}{$0.0\pm0.1\%$}\\ \midrule
		 \textbf{$\hatfpr_\text{F}$} & $5.5\pm0.1\%$ & $5.1\pm0.1\%$ & $5.3\pm0.1\%$\\ \midrule
		 \textbf{$\hatfpr_\text{M}$} & $5.5\pm0.1\%$ & $5.1\pm0.1\%$ & $5.3\pm0.2\%$\\ \midrule
		 \textbf{$\Delta\hatfnr$} & \textcolor{blue}{$0.9\pm0.1\%$} & \textcolor{blue}{$0.5\pm0.0\%$} & \textcolor{blue}{$0.5\pm0.0\%$}\\ \midrule
		 \textbf{$\hatfnr_\text{F}$} & $25.7\pm0.3\%$ & $25.7\pm0.2\%$ & $25.7\pm0.2\%$\\ \midrule
		 \textbf{$\hatfnr_\text{M}$} & $26.6\pm0.2\%$ & $26.2\pm0.2\%$ & $26.2\pm0.2\%$\\ \bottomrule
	\end{tabular}
\end{minipage}%
\caption{\textbf{Logistic regression} on \texttt{New Adult - CA - Public Coverage}, by \texttt{sex}\looseness=-1}
\label{fig:adultnew-public-sex-lr-all}
\vspace{2cm}
\end{figure*}

\vspace{.5cm}
\setlength{\tabcolsep}{6pt}
\begin{figure*}[h!]
\begin{minipage}{.495\linewidth}
\centering
\hspace{-.4cm}
        \includegraphics[width=\linewidth]{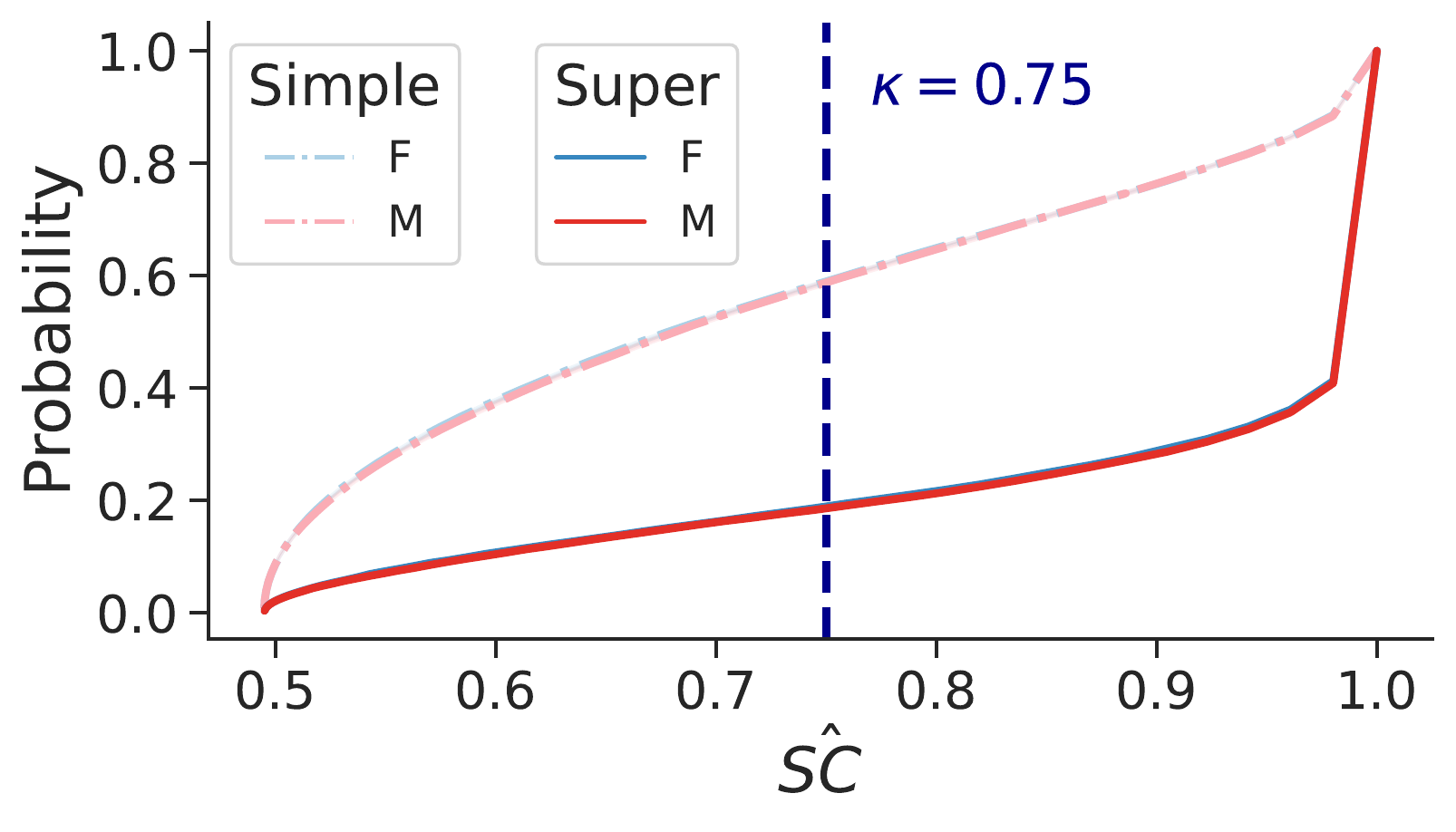}\\
        \vspace{-.1cm}%
	\begin{tabular}{lcc}
		\toprule
		\multicolumn{3}{c}{\textbf{Abstention set metrics}} \\ \cmidrule(lr){1-3}
		 & \textbf{Simple}     & \textbf{Super}      \\ \midrule
		 \textbf{$\Delta\hatar$} & \textcolor{blue}{$0.1\pm0.0\%$} & \textcolor{blue}{$0.3\pm0.1\%$}\\ \midrule
		 \textbf{$\hatar_\text{F}$} & $60.7\pm0.4\%$ & $18.6\pm0.4\%$\\ \midrule
		 \textbf{$\hatar_\text{M}$} & $60.8\pm0.4\%$ & $18.3\pm0.3\%$\\ \bottomrule
	\end{tabular}
\end{minipage}%
\hspace{.25cm}
\begin{minipage}{.495\linewidth}
\centering
\hspace{-.2cm}
	\begin{tabular}{lccc}
		\toprule
		\multicolumn{4}{c}{\textbf{Decision tree prediction set metrics}} \\ \cmidrule(lr){1-4}
		 & \textbf{Baseline}     & \textbf{Simple}     & \textbf{Super}      \\ \midrule
		 \textbf{$\Delta\hatpr$} & \textcolor{blue}{$2.6\pm0.2\%$} & \textcolor{blue}{$6.8\pm0.1\%$} & \textcolor{blue}{$3.0\pm0.0\%$}\\ \midrule
		 \textbf{$\hatpr_\text{F}$} & $35.5\pm0.2\%$ & $20.5\pm0.4\%$ & $27.5\pm0.3\%$\\ \midrule
		 \textbf{$\hatpr_\text{M}$} & $38.1\pm0.4\%$ & $27.3\pm0.3\%$ & $30.5\pm0.3\%$\\ \midrule
		 \textbf{$\Delta\haterr$} & \textcolor{blue}{$0.1\pm0.1\%$} & \textcolor{blue}{$0.5\pm0.1\%$} & \textcolor{blue}{$0.2\pm0.1\%$}\\ \midrule
		 \textbf{$\haterr_\text{F}$} & $35.1\pm0.2\%$ & $18.8\pm0.3\%$ & $26.7\pm0.3\%$\\ \midrule
		 \textbf{$\haterr_\text{M}$} & $35.2\pm0.1\%$ & $19.3\pm0.2\%$ & $26.9\pm0.4\%$\\ \midrule
		 \textbf{$\Delta\hatfpr$} & \textcolor{blue}{$0.4\pm0.1\%$} & \textcolor{blue}{$0.1\pm0.0\%$} & \textcolor{blue}{$0.3\pm0.1\%$}\\ \midrule
		 \textbf{$\hatfpr_\text{F}$} & $17.6\pm0.1\%$ & $4.6\pm0.2\%$ & $10.0\pm0.3\%$\\ \midrule
		 \textbf{$\hatfpr_\text{M}$} & $17.2\pm0.2\%$ & $4.5\pm0.2\%$ & $9.7\pm0.2\%$\\ \midrule
		 \textbf{$\Delta\hatfnr$} & \textcolor{blue}{$0.6\pm0.0\%$} & \textcolor{blue}{$0.6\pm0.2\%$} & \textcolor{blue}{$0.5\pm0.3\%$}\\ \midrule
		 \textbf{$\hatfnr_\text{F}$} & $17.4\pm0.2\%$ & $14.2\pm0.3\%$ & $16.7\pm0.1\%$\\ \midrule
		 \textbf{$\hatfnr_\text{M}$} & $18.0\pm0.2\%$ & $14.8\pm0.1\%$ & $17.2\pm0.4\%$\\ \bottomrule
	\end{tabular}
\end{minipage}%
\caption{\textbf{Decision trees} on \texttt{New Adult - CA - Public Coverage}, by \texttt{sex}\looseness=-1}
\label{fig:adultnew-coverage-sex-dtc-all}
\end{figure*}

\vspace{.5cm}
\setlength{\tabcolsep}{6pt}
\begin{figure*}[h!]
\begin{minipage}{.495\linewidth}
\centering
\hspace{-.4cm}
        \includegraphics[width=\linewidth]{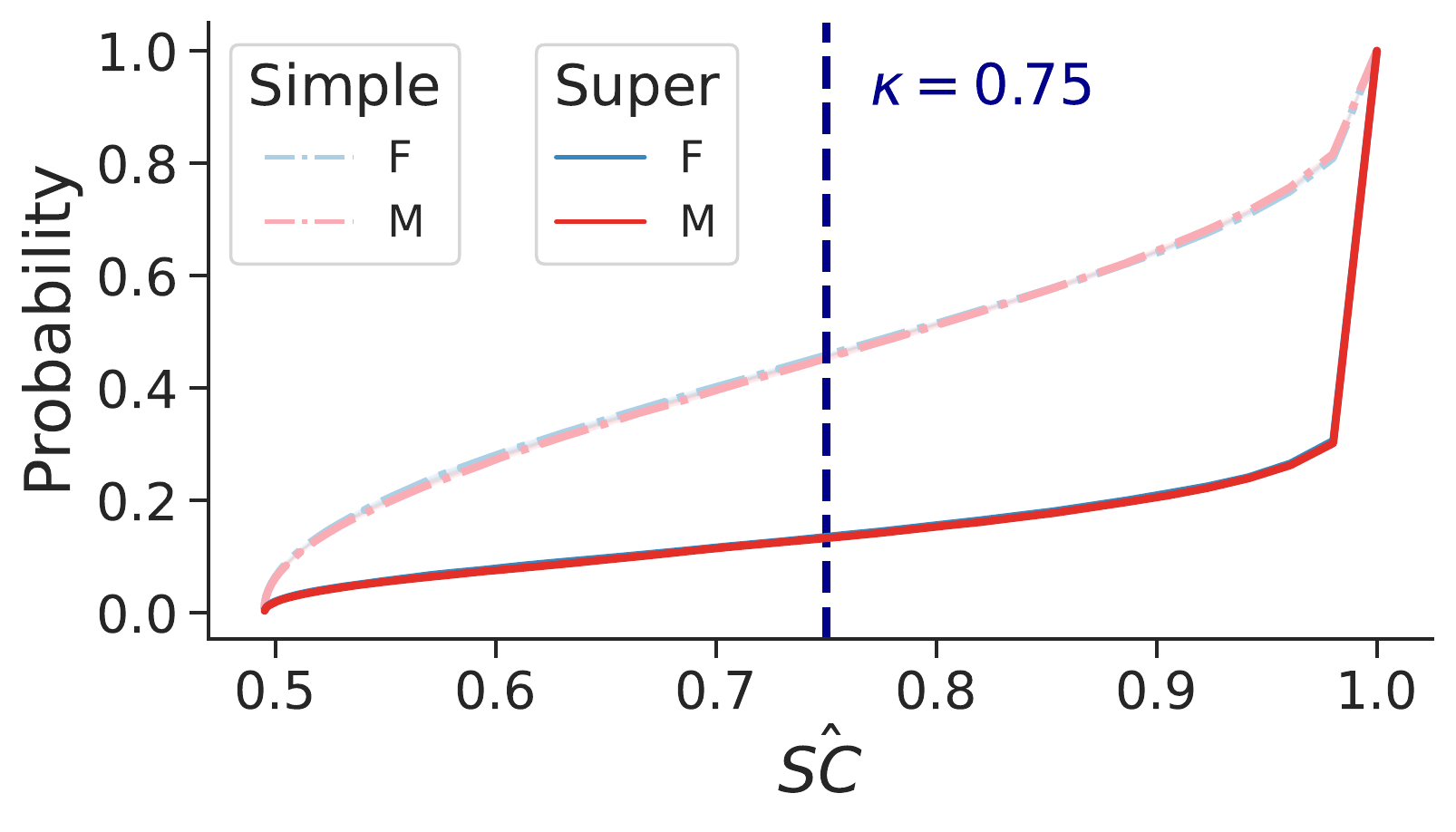}\\
        \vspace{-.1cm}%
	\begin{tabular}{lcc}
		\toprule
		\multicolumn{3}{c}{\textbf{Abstention set metrics}} \\ \cmidrule(lr){1-3}
		 & \textbf{Simple}     & \textbf{Super}      \\ \midrule
		 \textbf{$\Delta\hatar$} & \textcolor{blue}{$0.2\pm0.0\%$} & \textcolor{blue}{$0.2\pm0.2\%$}\\ \midrule
		 \textbf{$\hatar_\text{F}$} & $48.1\pm0.3\%$ & $13.2\pm0.1\%$\\ \midrule
		 \textbf{$\hatar_\text{M}$} & $47.9\pm0.3\%$ & $13.0\pm0.3\%$\\ \bottomrule
	\end{tabular}
\end{minipage}%
\hspace{.25cm}
\begin{minipage}{.495\linewidth}
\centering
\hspace{-.2cm}
	\begin{tabular}{lccc}
		\toprule
		\multicolumn{4}{c}{\textbf{Random forest prediction set metrics}} \\ \cmidrule(lr){1-4}
		 & \textbf{Baseline}     & \textbf{Simple}     & \textbf{Super}      \\ \midrule
		 \textbf{$\Delta\hatpr$} & \textcolor{blue}{$2.5\pm0.1\%$} & \textcolor{blue}{$5.3\pm0.0\%$} & \textcolor{blue}{$2.6\pm0.1\%$}\\ \midrule
		 \textbf{$\hatpr_\text{F}$} & $31.9\pm0.3\%$ & $19.5\pm0.3\%$ & $25.7\pm0.4\%$\\ \midrule
		 \textbf{$\hatpr_\text{M}$} & $34.4\pm0.4\%$ & $24.8\pm0.3\%$ & $28.3\pm0.3\%$\\ \midrule
		 \textbf{$\Delta\haterr$} & \textcolor{blue}{$0.4\pm0.1\%$} & \textcolor{blue}{$1.0\pm0.2\%$} & \textcolor{blue}{$0.3\pm0.1\%$}\\ \midrule
		 \textbf{$\haterr_\text{F}$} & $32.3\pm0.2\%$ & $19.3\pm0.3\%$ & $26.3\pm0.2\%$\\ \midrule
		 \textbf{$\haterr_\text{M}$} & $32.7\pm0.1\%$ & $20.3\pm0.1\%$ & $26.6\pm0.3\%$\\ \midrule
		 \textbf{$\Delta\hatfpr$} & \textcolor{blue}{$0.2\pm0.1\%$} & \textcolor{blue}{$0.2\pm0.1\%$} & \textcolor{blue}{$0.3\pm0.1\%$}\\ \midrule
		 \textbf{$\hatfpr_\text{F}$} & $14.4\pm0.1\%$ & $4.1\pm0.2\%$ & $8.7\pm0.3\%$\\ \midrule
		 \textbf{$\hatfpr_\text{M}$} & $14.2\pm0.2\%$ & $4.3\pm0.1\%$ & $8.4\pm0.2\%$\\ \midrule
		 \textbf{$\Delta\hatfnr$} & \textcolor{blue}{$0.7\pm0.1\%$} & \textcolor{blue}{$0.8\pm0.0\%$} & \textcolor{blue}{$0.7\pm0.1\%$}\\ \midrule
		 \textbf{$\hatfnr_\text{F}$} & $17.9\pm0.2\%$ & $15.2\pm0.2\%$ & $17.5\pm0.2\%$\\ \midrule
		 \textbf{$\hatfnr_\text{M}$} & $18.6\pm0.3\%$ & $16.0\pm0.2\%$ & $18.2\pm0.3\%$\\ \bottomrule
	\end{tabular}
\end{minipage}%
\caption{\textbf{Random forests} on \texttt{New Adult - CA - Public Coverage}, by \texttt{sex}\looseness=-1}
\label{fig:adultnew-coverage-sex-rfc-all}
\end{figure*}
\FloatBarrier

\vspace*{1in}
\custompar{\texttt{Public Coverage} - by \texttt{race}}

\setlength{\tabcolsep}{6pt}
\begin{figure*}[h!]
\begin{minipage}{.495\linewidth}
\centering
\hspace{-.4cm}
        \includegraphics[width=\linewidth]{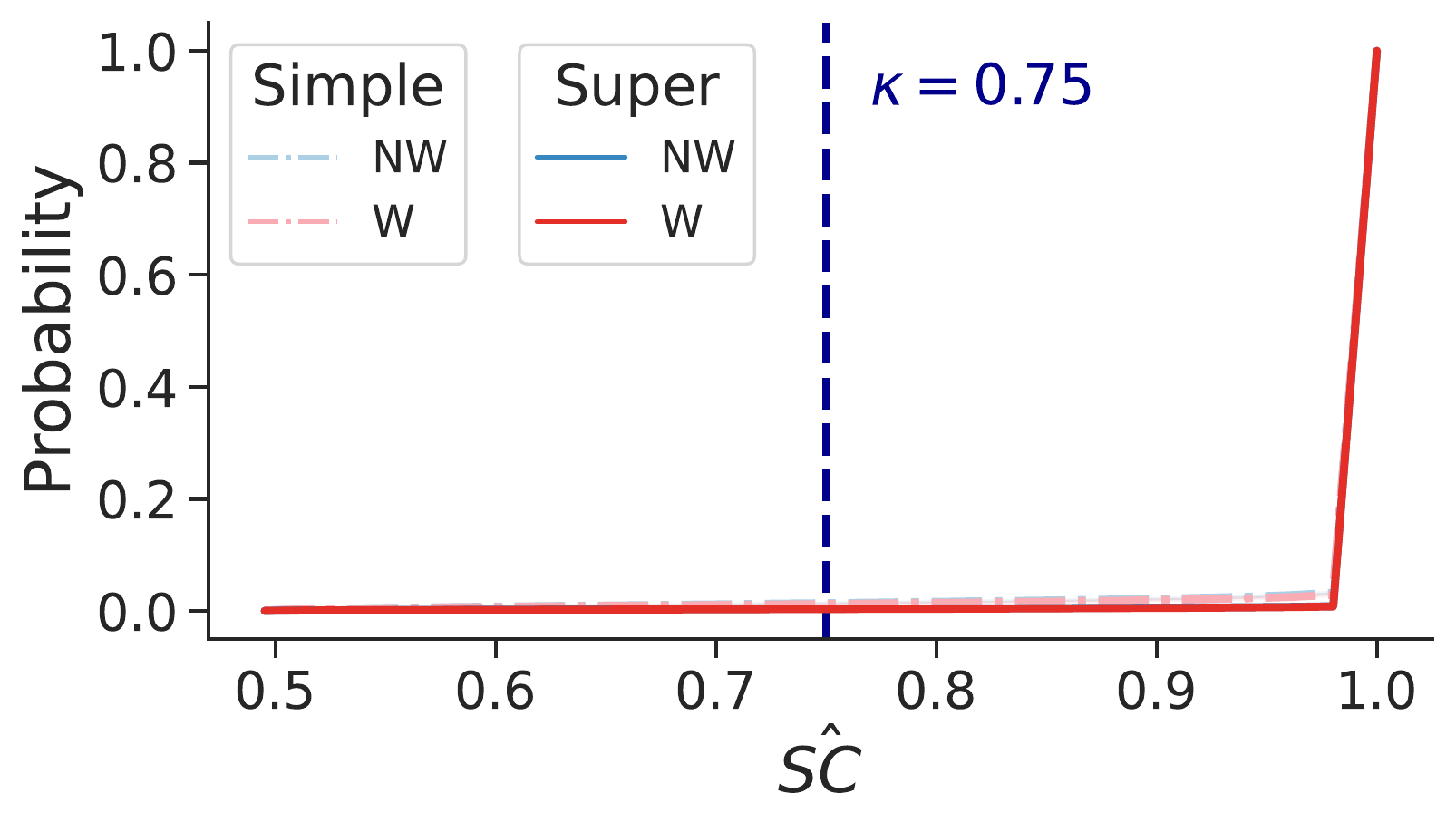}\\
        \vspace{-.1cm}%
	\begin{tabular}{lcc}
		\toprule
		\multicolumn{3}{c}{\textbf{Abstention set metrics}} \\ \cmidrule(lr){1-3}
		 & \textbf{Simple}     & \textbf{Super}      \\ \midrule
		 \textbf{$\Delta\hatar$} & \textcolor{blue}{$0.1\pm0.0\%$} & \textcolor{blue}{$0.1\pm0.0\%$}\\ \midrule
		 \textbf{$\hatar_\text{NW}$} & $1.5\pm0.1\%$ & $0.4\pm0.0\%$\\ \midrule
		 \textbf{$\hatar_\text{W}$} & $1.4\pm0.1\%$ & $0.3\pm0.0\%$\\ \bottomrule
	\end{tabular}
\end{minipage}%
\hspace{.25cm}
\begin{minipage}{.495\linewidth}
\centering
\hspace{-.2cm}
	\begin{tabular}{lccc}
		\toprule
		\multicolumn{4}{c}{\textbf{Logistic regression prediction set metrics}} \\ \cmidrule(lr){1-4}
		 & \textbf{Baseline}     & \textbf{Simple}     & \textbf{Super}      \\ \midrule
		 \textbf{$\Delta\hatpr$} & \textcolor{blue}{$0.1\pm0.1\%$} & \textcolor{blue}{$0.1\pm0.1\%$} & \textcolor{blue}{$0.1\pm0.1\%$}\\ \midrule
		 \textbf{$\hatpr_\text{NW}$} & $16.3\pm0.3\%$ & $15.8\pm0.3\%$ & $16.2\pm0.3\%$\\ \midrule
		 \textbf{$\hatpr_\text{W}$} & $16.2\pm0.2\%$ & $15.9\pm0.2\%$ & $16.3\pm0.2\%$\\ \midrule
		 \textbf{$\Delta\haterr$} & \textcolor{blue}{$3.2\pm0.0\%$} & \textcolor{blue}{$2.7\pm0.1\%$} & \textcolor{blue}{$2.8\pm0.1\%$}\\ \midrule
		 \textbf{$\haterr_\text{NW}$} & $33.4\pm0.3\%$ & $32.6\pm0.3\%$ & $32.8\pm0.3\%$\\ \midrule
		 \textbf{$\haterr_\text{W}$} & $30.2\pm0.3\%$ & $29.9\pm0.2\%$ & $30.0\pm0.2\%$\\ \midrule
		 \textbf{$\Delta\hatfpr$} & \textcolor{blue}{$0.2\pm0.0\%$} & \textcolor{blue}{$0.1\pm0.0\%$} & \textcolor{blue}{$0.1\pm0.1\%$}\\ \midrule
		 \textbf{$\hatfpr_\text{NW}$} & $5.6\pm0.1\%$ & $5.2\pm0.1\%$ & $5.4\pm0.2\%$\\ \midrule
		 \textbf{$\hatfpr_\text{W}$} & $5.4\pm0.1\%$ & $5.1\pm0.1\%$ & $5.3\pm0.1\%$\\ \midrule
		 \textbf{$\Delta\hatfnr$} & \textcolor{blue}{$3.0\pm0.1\%$} & \textcolor{blue}{$2.6\pm0.1\%$} & \textcolor{blue}{$2.7\pm0.1\%$}\\ \midrule
		 \textbf{$\hatfnr_\text{NW}$} & $27.8\pm0.3\%$ & $27.4\pm0.3\%$ & $27.4\pm0.3\%$\\ \midrule
		 \textbf{$\hatfnr_\text{W}$} & $24.8\pm0.2\%$ & $24.8\pm0.2\%$ & $24.7\pm0.2\%$\\ \bottomrule
	\end{tabular}
\end{minipage}%
\caption{\textbf{Logistic regression} on \texttt{New Adult - CA - Public Coverage}, by \texttt{race}\looseness=-1}
\label{fig:adultnew-coverage-race-lr-all}
\end{figure*}

\vspace{.1cm}
\setlength{\tabcolsep}{6pt}
\begin{figure*}[h!]
\begin{minipage}{.495\linewidth}
\centering
\hspace{-.4cm}
        \includegraphics[width=\linewidth]{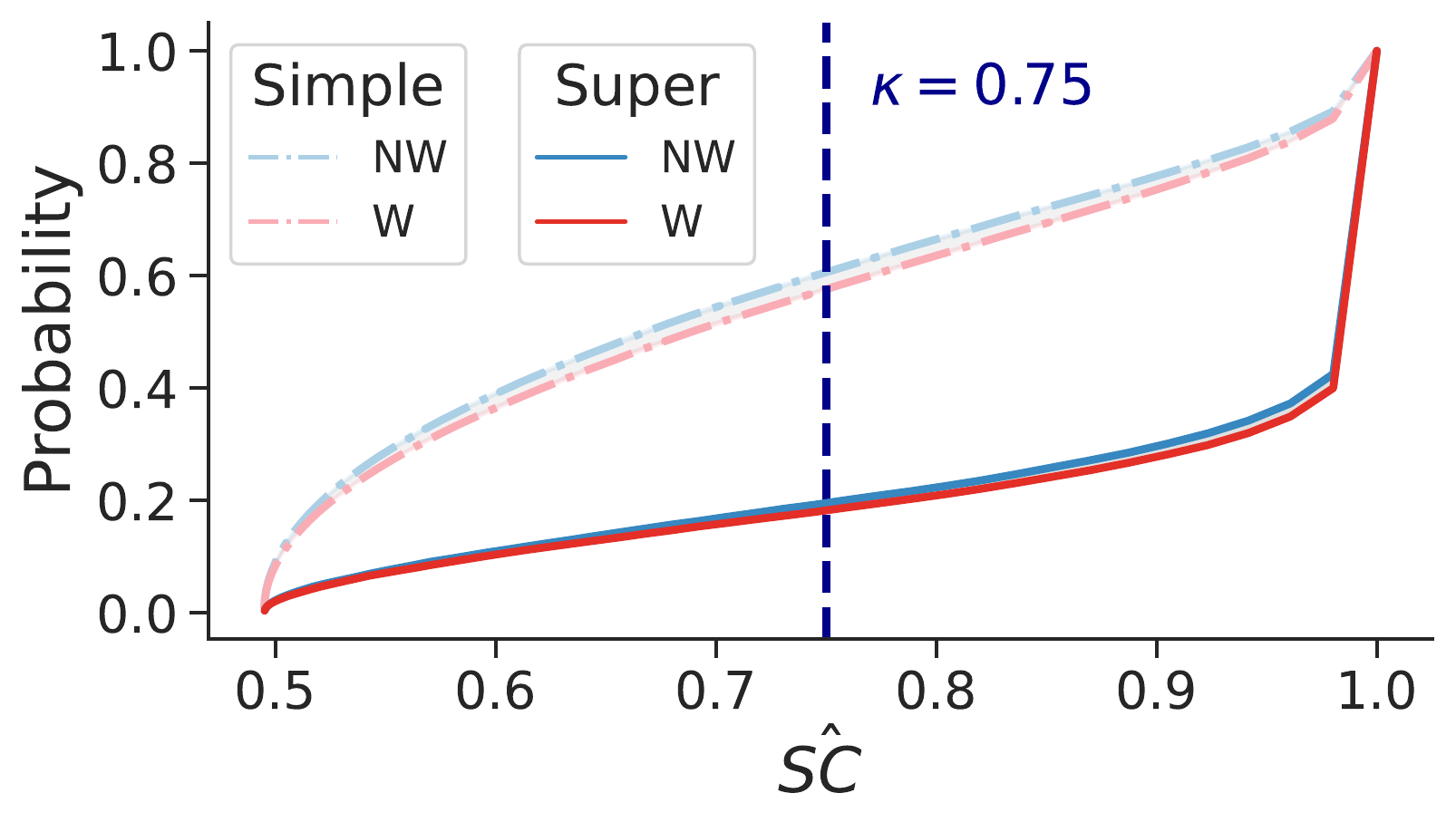}\\
        \vspace{-.1cm}%
	\begin{tabular}{lcc}
		\toprule
		\multicolumn{3}{c}{\textbf{Abstention set metrics}} \\ \cmidrule(lr){1-3}
		 & \textbf{Simple}     & \textbf{Super}      \\ \midrule
		 \textbf{$\Delta\hatar$} & \textcolor{blue}{$2.8\pm0.0\%$} & \textcolor{blue}{$1.3\pm0.1\%$}\\ \midrule
		 \textbf{$\hatar_\text{NW}$} & $62.3\pm0.3\%$ & $19.2\pm0.3\%$\\ \midrule
		 \textbf{$\hatar_\text{W}$} & $59.5\pm0.3\%$ & $17.9\pm0.4\%$\\ \bottomrule
	\end{tabular}
\end{minipage}%
\hspace{.25cm}
\begin{minipage}{.495\linewidth}
\centering
\hspace{-.2cm}
	\begin{tabular}{lccc}
		\toprule
		\multicolumn{4}{c}{\textbf{Decision tree prediction set metrics}} \\ \cmidrule(lr){1-4}
		 & \textbf{Baseline}     & \textbf{Simple}     & \textbf{Super}      \\ \midrule
		 \textbf{$\Delta\hatpr$} & \textcolor{blue}{$1.5\pm0.0\%$} & \textcolor{blue}{$2.2\pm0.2\%$} & \textcolor{blue}{$1.4\pm0.1\%$}\\ \midrule
		 \textbf{$\hatpr_\text{NW}$} & $37.5\pm0.2\%$ & $24.8\pm0.3\%$ & $29.6\pm0.4\%$\\ \midrule
		 \textbf{$\hatpr_\text{W}$} & $36.0\pm0.2\%$ & $22.6\pm0.5\%$ & $28.2\pm0.3\%$\\ \midrule
		 \textbf{$\Delta\haterr$} & \textcolor{blue}{$2.2\pm0.0\%$} & \textcolor{blue}{$3.0\pm0.1\%$} & \textcolor{blue}{$2.7\pm0.1\%$}\\ \midrule
		 \textbf{$\haterr_\text{NW}$} & $36.4\pm0.1\%$ & $20.8\pm0.3\%$ & $28.3\pm0.2\%$\\ \midrule
		 \textbf{$\haterr_\text{W}$} & $34.2\pm0.1\%$ & $17.8\pm0.4\%$ & $25.6\pm0.3\%$\\ \midrule
		 \textbf{$\Delta\hatfpr$} & \textcolor{blue}{$0.4\pm0.1\%$} & \textcolor{blue}{$0.7\pm0.0\%$} & \textcolor{blue}{$0.6\pm0.0\%$}\\ \midrule
		 \textbf{$\hatfpr_\text{NW}$} & $17.7\pm0.1\%$ & $5.0\pm0.2\%$ & $10.2\pm0.2\%$\\ \midrule
		 \textbf{$\hatfpr_\text{W}$} & $17.3\pm0.2\%$ & $4.3\pm0.2\%$ & $9.6\pm0.2\%$\\ \midrule
		 \textbf{$\Delta\hatfnr$} & \textcolor{blue}{$1.8\pm0.0\%$} & \textcolor{blue}{$2.3\pm0.1\%$} & \textcolor{blue}{$2.1\pm0.1\%$}\\ \midrule
		 \textbf{$\hatfnr_\text{NW}$} & $18.7\pm0.2\%$ & $15.8\pm0.2\%$ & $18.1\pm0.3\%$\\ \midrule
		 \textbf{$\hatfnr_\text{W}$} & $16.9\pm0.2\%$ & $13.5\pm0.3\%$ & $16.0\pm0.2\%$\\ \bottomrule
	\end{tabular}
\end{minipage}%
\caption{\textbf{Decision trees} on \texttt{New Adult - CA - Public Coverage}, by \texttt{race}\looseness=-1}
\label{fig:adultnew-coverage-race-dtc-all}
\vspace{2cm}
\end{figure*}

\setlength{\tabcolsep}{6pt}
\begin{figure*}[h!]
\begin{minipage}{.495\linewidth}
\centering
\hspace{-.4cm}
        \includegraphics[width=\linewidth]{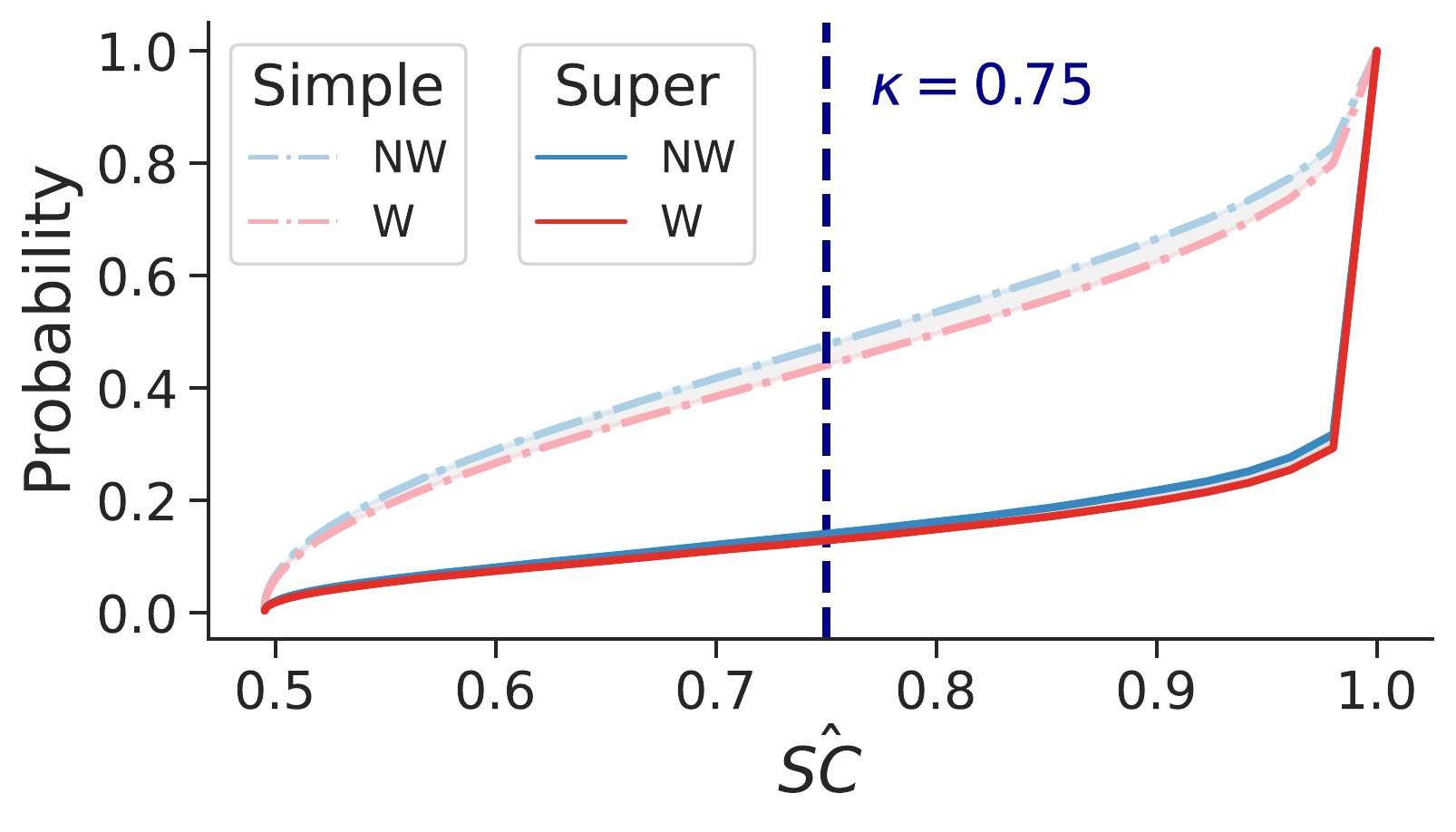}\\
        \vspace{-.1cm}%
	\begin{tabular}{lcc}
		\toprule
		\multicolumn{3}{c}{\textbf{Abstention set metrics}} \\ \cmidrule(lr){1-3}
		 & \textbf{Simple}     & \textbf{Super}      \\ \midrule
		 \textbf{$\Delta\hatar$} & \textcolor{blue}{$3.6\pm0.1\%$} & \textcolor{blue}{$1.2\pm0.0\%$}\\ \midrule
		 \textbf{$\hatar_\text{NW}$} & $50.0\pm0.2\%$ & $13.8\pm0.2\%$\\ \midrule
		 \textbf{$\hatar_\text{W}$} & $46.4\pm0.3\%$ & $12.6\pm0.2\%$\\ \bottomrule
	\end{tabular}
\end{minipage}%
\hspace{.25cm}
\begin{minipage}{.495\linewidth}
\centering
\hspace{-.2cm}
	\begin{tabular}{lccc}
		\toprule
		\multicolumn{4}{c}{\textbf{Random forest prediction set metrics}} \\ \cmidrule(lr){1-4}
		 & \textbf{Baseline}     & \textbf{Simple}     & \textbf{Super}      \\ \midrule
		 \textbf{$\Delta\hatpr$} & \textcolor{blue}{$1.2\pm0.0\%$} & \textcolor{blue}{$1.0\pm0.1\%$} & \textcolor{blue}{$0.7\pm0.0\%$}\\ \midrule
		 \textbf{$\hatpr_\text{NW}$} & $33.7\pm0.3\%$ & $22.4\pm0.4\%$ & $27.3\pm0.4\%$\\ \midrule
		 \textbf{$\hatpr_\text{W}$} & $32.5\pm0.3\%$ & $21.4\pm0.3\%$ & $26.6\pm0.4\%$\\ \midrule
		 \textbf{$\Delta\haterr$} & \textcolor{blue}{$2.7\pm0.1\%$} & \textcolor{blue}{$2.9\pm0.0\%$} & \textcolor{blue}{$2.6\pm0.1\%$}\\ \midrule
		 \textbf{$\haterr_\text{NW}$} & $34.0\pm0.2\%$ & $21.4\pm0.3\%$ & $27.9\pm0.3\%$\\ \midrule
		 \textbf{$\haterr_\text{W}$} & $31.3\pm0.1\%$ & $18.5\pm0.3\%$ & $25.3\pm0.2\%$\\ \midrule
		 \textbf{$\Delta\hatfpr$} & \textcolor{blue}{$0.5\pm0.0\%$} & \textcolor{blue}{$0.4\pm0.0\%$} & \textcolor{blue}{$0.5\pm0.0\%$}\\ \midrule
		 \textbf{$\hatfpr_\text{NW}$} & $14.6\pm0.2\%$ & $4.4\pm0.2\%$ & $8.9\pm0.2\%$\\ \midrule
		 \textbf{$\hatfpr_\text{W}$} & $14.1\pm0.2\%$ & $4.0\pm0.2\%$ & $8.4\pm0.2\%$\\ \midrule
		 \textbf{$\Delta\hatfnr$} & \textcolor{blue}{$2.2\pm0.0\%$} & \textcolor{blue}{$2.5\pm0.0\%$} & \textcolor{blue}{$2.3\pm0.1\%$}\\ \midrule
		 \textbf{$\hatfnr_\text{NW}$} & $19.4\pm0.2\%$ & $17.0\pm0.2\%$ & $19.1\pm0.3\%$\\ \midrule
		 \textbf{$\hatfnr_\text{W}$} & $17.2\pm0.2\%$ & $14.5\pm0.2\%$ & $16.8\pm0.2\%$\\ \bottomrule
	\end{tabular}
\end{minipage}%
\caption{\textbf{Random forests} on \texttt{New Adult - CA - Public Coverage}, by \texttt{race}\looseness=-1}
\label{fig:adultnew-coverage-race-rfc-all}
\end{figure*}
\FloatBarrier
\vspace*{1cm}

\newpage
\subsubsection{\apphmdaillustrative}\label{app:sec:hmda-algo} 

$\hatsc$ CDFs for two states (\texttt{NY}, \texttt{TX}) in \texttt{HMDA -2017}, using $\group=\texttt{ethnicity}, \;\texttt{race},$ and \texttt{sex}, and associated error metrics on the prediction set. \textbf{Baseline} metrics computed with $\boot=101$ models. For \textbf{simple}, $\boot=101$ models; for \textbf{super}, $\boot=101$ ensemble models, each composed of $21$ underlying models for \texttt{NY}; $15$ for \texttt{TX}. We repeat for $5$ test/train splits. We also report abstention rate $\hatar$. 

\custompar{\texttt{NY - 2017} - by \texttt{ethnicity}}

\setlength{\tabcolsep}{6pt}
\begin{figure*}[h!]
\begin{minipage}{.495\linewidth}
\centering
\hspace{-.4cm}
        \includegraphics[width=\linewidth]{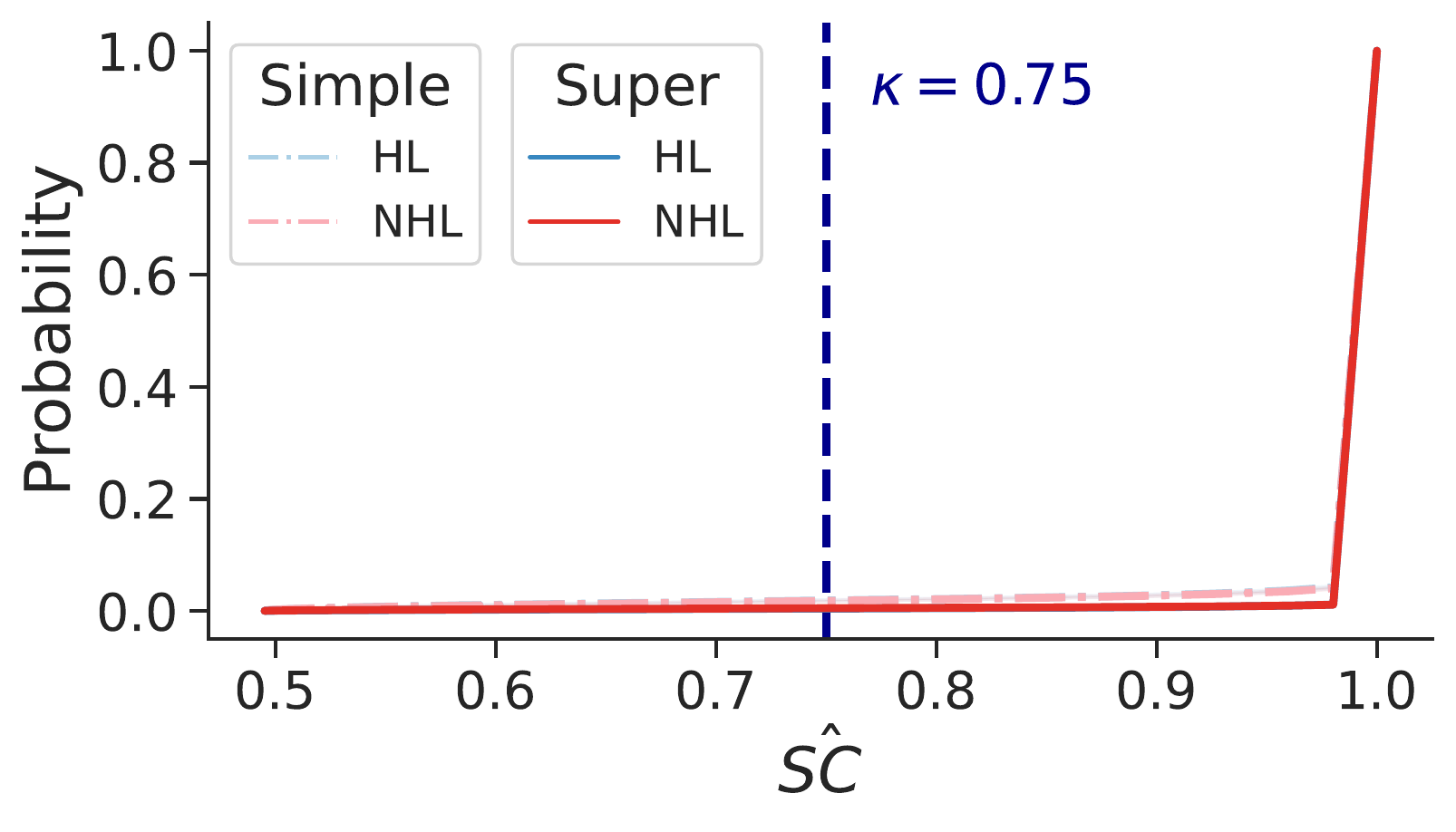}\\
        \vspace{-.1cm}%
	\begin{tabular}{lcc}
		\toprule
		\multicolumn{3}{c}{\textbf{Abstention set metrics}} \\ \cmidrule(lr){1-3}
		 & \textbf{Simple}     & \textbf{Super}      \\ \midrule
		 \textbf{$\Delta\hatar$} & \textcolor{blue}{$0.0\pm0.1\%$} & \textcolor{blue}{$0.1\pm0.1\%$}\\ \midrule
		 \textbf{$\hatar_\text{HL}$} & $1.9\pm0.2\%$ & $0.4\pm0.1\%$\\ \midrule
		 \textbf{$\hatar_\text{NHL}$} & $1.9\pm0.1\%$ & $0.5\pm0.0\%$\\ \bottomrule
	\end{tabular}
\end{minipage}%
\hspace{.25cm}
\begin{minipage}{.495\linewidth}
\centering
\hspace{-.2cm}
	\begin{tabular}{lccc}
		\toprule
		\multicolumn{4}{c}{\textbf{Logistic regression prediction set metrics}} \\ \cmidrule(lr){1-4}
		 & \textbf{Baseline}     & \textbf{Simple}     & \textbf{Super}      \\ \midrule
		 \textbf{$\Delta\hatpr$} & \textcolor{blue}{$9.7\pm0.2\%$} & \textcolor{blue}{$10.5\pm0.5\%$} & \textcolor{blue}{$10.4\pm0.5\%$}\\ \midrule
		 \textbf{$\hatpr_\text{HL}$} & $73.5\pm0.3\%$ & $73.5\pm0.6\%$ & $73.1\pm0.6\%$\\ \midrule
		 \textbf{$\hatpr_\text{NHL}$} & $83.2\pm0.1\%$ & $84.0\pm0.1\%$ & $83.5\pm0.1\%$\\ \midrule
		 \textbf{$\Delta\haterr$} & \textcolor{blue}{$1.3\pm0.2\%$} & \textcolor{blue}{$1.7\pm0.3\%$} & \textcolor{blue}{$1.8\pm0.4\%$}\\ \midrule
		 \textbf{$\haterr_\text{HL}$} & $18.7\pm0.3\%$ & $18.4\pm0.4\%$ & $18.9\pm0.5\%$\\ \midrule
		 \textbf{$\haterr_\text{NHL}$} & $17.4\pm0.1\%$ & $16.7\pm0.1\%$ & $17.1\pm0.1\%$\\ \midrule
		 \textbf{$\Delta\hatfpr$} & \textcolor{blue}{$0.8\pm0.1\%$} & \textcolor{blue}{$1.0\pm0.1\%$} & \textcolor{blue}{$0.9\pm0.1\%$}\\ \midrule
		 \textbf{$\hatfpr_\text{HL}$} & $10.7\pm0.2\%$ & $10.2\pm0.2\%$ & $10.5\pm0.2\%$\\ \midrule
		 \textbf{$\hatfpr_\text{NHL}$} & $11.5\pm0.1\%$ & $11.2\pm0.1\%$ & $11.4\pm0.1\%$\\ \midrule
		 \textbf{$\Delta\hatfnr$} & \textcolor{blue}{$2.2\pm0.2\%$} & \textcolor{blue}{$2.7\pm0.3\%$} & \textcolor{blue}{$2.7\pm0.4\%$}\\ \midrule
		 \textbf{$\hatfnr_\text{HL}$} & $8.0\pm0.3\%$ & $8.2\pm0.4\%$ & $8.4\pm0.5\%$\\ \midrule
		 \textbf{$\hatfnr_\text{NHL}$} & $5.8\pm0.1\%$ & $5.5\pm0.1\%$ & $5.7\pm0.1\%$\\ \bottomrule
	\end{tabular}
\end{minipage}%
\caption{\textbf{Logistic regression} on \texttt{HMDA - 2017 - NY}, by \texttt{ethnicity}\looseness=-1}
\label{fig:hmda-ny-eth-lr-all}
\end{figure*}

\vspace{.5cm}
\setlength{\tabcolsep}{6pt}
\begin{figure*}[h!]
\begin{minipage}{.495\linewidth}
\centering
\hspace{-.4cm}
        \includegraphics[width=\linewidth]{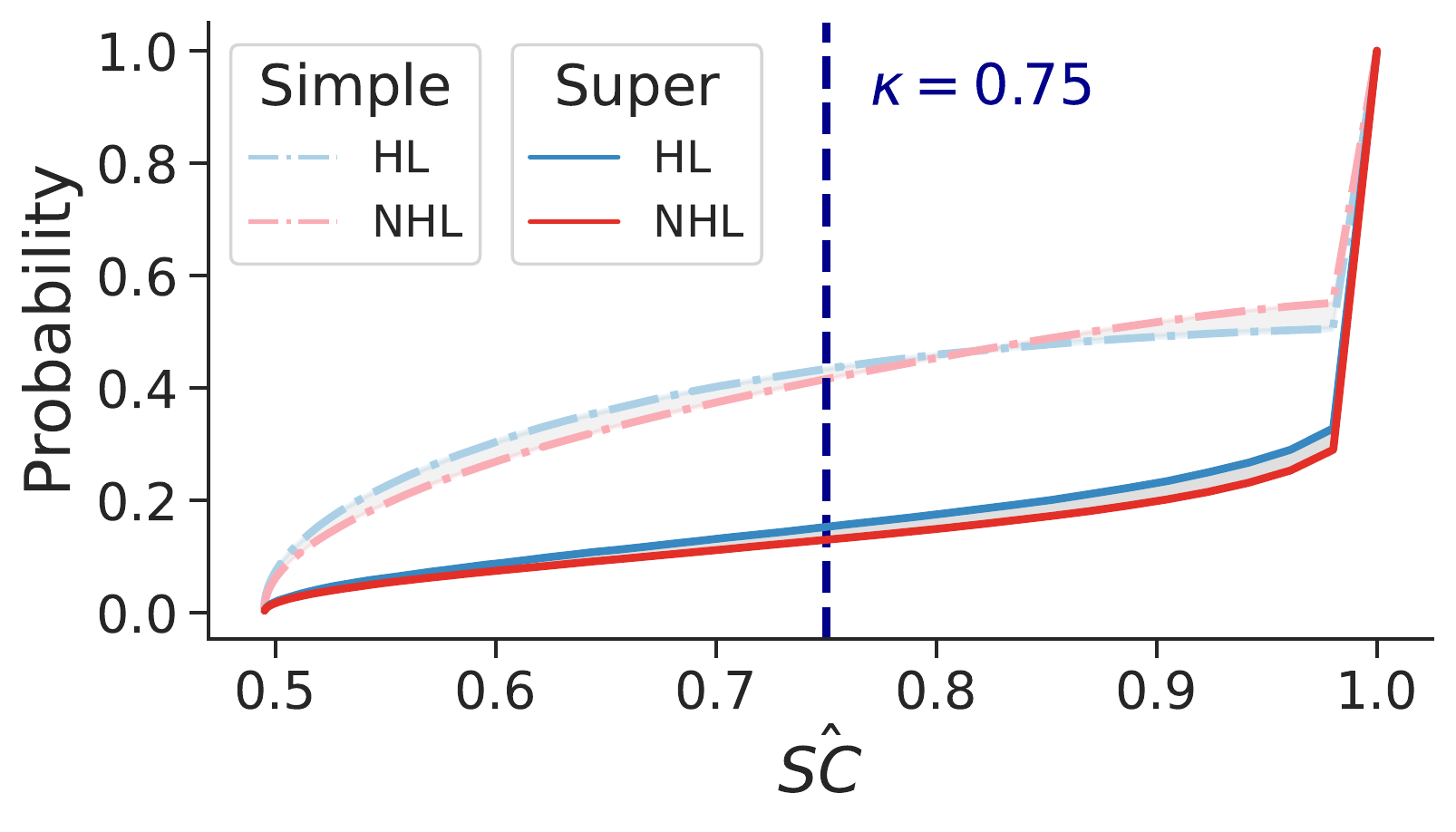}\\
        \vspace{-.1cm}%
	\begin{tabular}{lcc}
		\toprule
		\multicolumn{3}{c}{\textbf{Abstention set metrics}} \\ \cmidrule(lr){1-3}
		 & \textbf{Simple}     & \textbf{Super}      \\ \midrule
		 \textbf{$\Delta\hatar$} & \textcolor{blue}{$1.2\pm0.2\%$} & \textcolor{blue}{$2.3\pm0.1\%$}\\ \midrule
		 \textbf{$\hatar_\text{HL}$} & $43.3\pm0.4\%$ & $15.0\pm0.3\%$\\ \midrule
		 \textbf{$\hatar_\text{NHL}$} & $42.1\pm0.2\%$ & $12.7\pm0.2\%$\\ \bottomrule
	\end{tabular}
\end{minipage}%
\hspace{.25cm}
\begin{minipage}{.495\linewidth}
\centering
\hspace{-.2cm}
	\begin{tabular}{lccc}
		\toprule
		\multicolumn{4}{c}{\textbf{Decision tree prediction set metrics}} \\ \cmidrule(lr){1-4}
		 & \textbf{Baseline}     & \textbf{Simple}     & \textbf{Super}      \\ \midrule
		 \textbf{$\Delta\hatpr$} & \textcolor{blue}{$4.0\pm0.2\%$} & \textcolor{blue}{$1.5\pm0.2\%$} & \textcolor{blue}{$8.2\pm0.5\%$}\\ \midrule
		 \textbf{$\hatpr_\text{HL}$} & $73.1\pm0.3\%$ & $94.0\pm0.3\%$ & $74.1\pm0.6\%$\\ \midrule
		 \textbf{$\hatpr_\text{NHL}$} & $77.1\pm0.1\%$ & $95.5\pm0.1\%$ & $82.3\pm0.1\%$\\ \midrule
		 \textbf{$\Delta\haterr$} & \textcolor{blue}{$0.4\pm0.1\%$} & \textcolor{blue}{$0.7\pm0.1\%$} & \textcolor{blue}{$0.6\pm0.2\%$}\\ \midrule
		 \textbf{$\haterr_\text{HL}$} & $20.6\pm0.2\%$ & $2.7\pm0.2\%$ & $12.2\pm0.4\%$\\ \midrule
		 \textbf{$\haterr_\text{NHL}$} & $20.2\pm0.1\%$ & $3.4\pm0.1\%$ & $11.6\pm0.2\%$\\ \midrule
		 \textbf{$\Delta\hatfpr$} & \textcolor{blue}{$1.6\pm0.1\%$} & \textcolor{blue}{$1.1\pm0.0\%$} & \textcolor{blue}{$1.1\pm0.1\%$}\\ \midrule
		 \textbf{$\hatfpr_\text{HL}$} & $11.5\pm0.2\%$ & $1.4\pm0.1\%$ & $5.1\pm0.2\%$\\ \midrule
		 \textbf{$\hatfpr_\text{NHL}$} & $9.9\pm0.1\%$ & $2.5\pm0.1\%$ & $6.2\pm0.1\%$\\ \midrule
		 \textbf{$\Delta\hatfnr$} & \textcolor{blue}{$1.2\pm0.1\%$} & \textcolor{blue}{$0.4\pm0.1\%$} & \textcolor{blue}{$1.6\pm0.3\%$}\\ \midrule
		 \textbf{$\hatfnr_\text{HL}$} & $9.1\pm0.2\%$ & $1.3\pm0.2\%$ & $7.0\pm0.4\%$\\ \midrule
		 \textbf{$\hatfnr_\text{NHL}$} & $10.3\pm0.1\%$ & $0.9\pm0.1\%$ & $5.4\pm0.1\%$\\ \bottomrule
	\end{tabular}
\end{minipage}%
\caption{\textbf{Decision trees} on \texttt{HMDA - 2017 - NY}, by \texttt{ethnicity}\looseness=-1}
\label{fig:hmda-ny-eth-dtc-all}
\end{figure*}

\vspace{.5cm}
\setlength{\tabcolsep}{6pt}
\begin{figure*}[h!]
\begin{minipage}{.495\linewidth}
\centering
\hspace{-.4cm}
        \includegraphics[width=\linewidth]{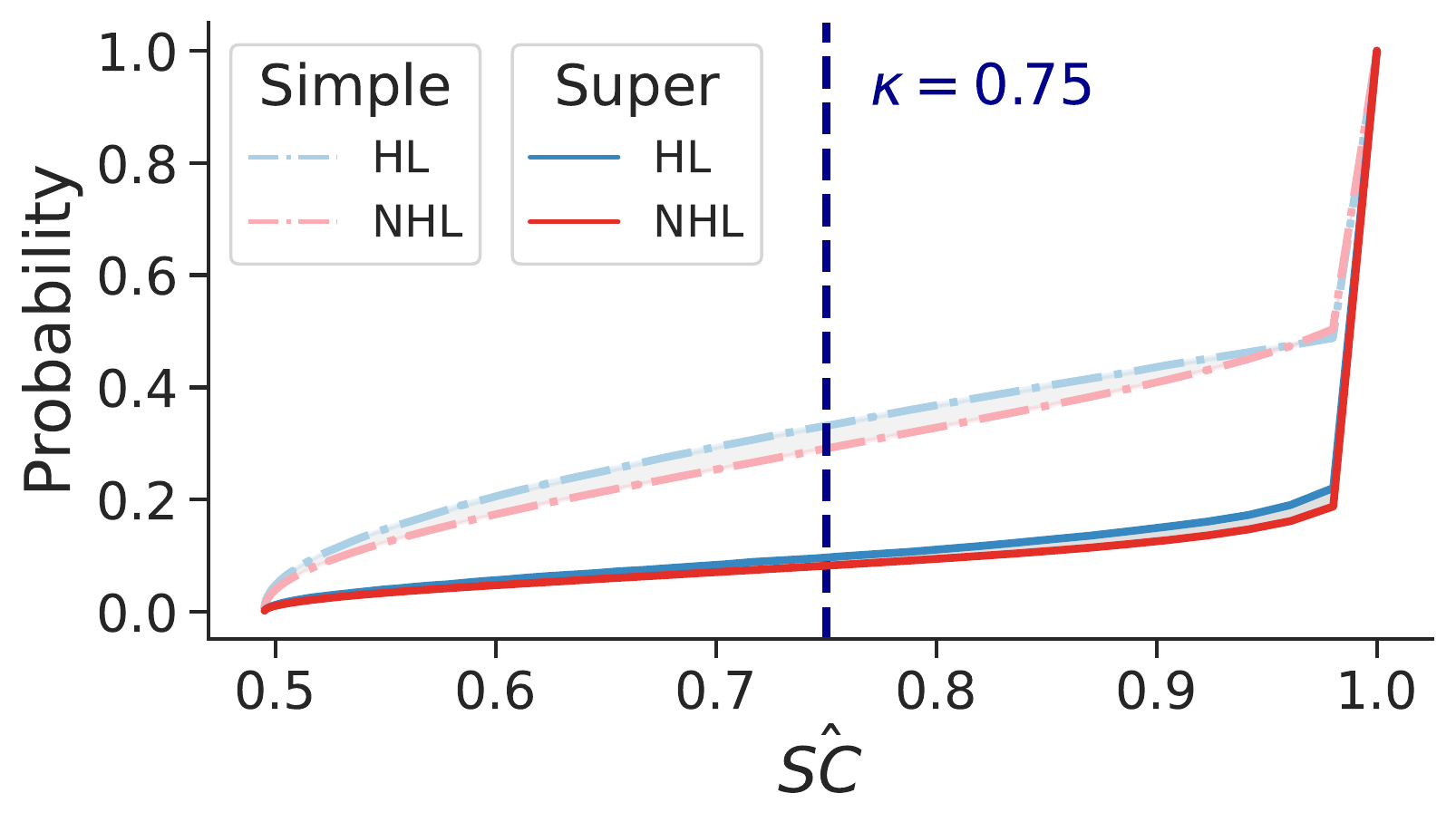}\\
        \vspace{-.1cm}%
	\begin{tabular}{lcc}
		\toprule
		\multicolumn{3}{c}{\textbf{Abstention set metrics}} \\ \cmidrule(lr){1-3}
		 & \textbf{Simple}     & \textbf{Super}      \\ \midrule
		 \textbf{$\Delta\hatar$} & \textcolor{blue}{$3.6\pm0.2\%$} & \textcolor{blue}{$1.4\pm0.3\%$}\\ \midrule
		 \textbf{$\hatar_\text{HL}$} & $33.9\pm0.3\%$ & $9.5\pm0.4\%$\\ \midrule
		 \textbf{$\hatar_\text{NHL}$} & $30.3\pm0.1\%$ & $8.1\pm0.1\%$\\ \bottomrule
	\end{tabular}
\end{minipage}%
\hspace{.25cm}
\begin{minipage}{.495\linewidth}
\centering
\hspace{-.2cm}
	\begin{tabular}{lccc}
		\toprule
		\multicolumn{4}{c}{\textbf{Random forest prediction set metrics}} \\ \cmidrule(lr){1-4}
		 & \textbf{Baseline}     & \textbf{Simple}     & \textbf{Super}      \\ \midrule
		 \textbf{$\Delta\hatpr$} & \textcolor{blue}{$6.3\pm0.3\%$} & \textcolor{blue}{$5.1\pm0.6\%$} & \textcolor{blue}{$9.4\pm0.4\%$}\\ \midrule
		 \textbf{$\hatpr_\text{HL}$} & $71.9\pm0.4\%$ & $85.8\pm0.7\%$ & $71.6\pm0.5\%$\\ \midrule
		 \textbf{$\hatpr_\text{NHL}$} & $78.2\pm0.1\%$ & $90.9\pm0.1\%$ & $81.0\pm0.1\%$\\ \midrule
		 \textbf{$\Delta\haterr$} & \textcolor{blue}{$1.1\pm0.1\%$} & \textcolor{blue}{$0.5\pm0.3\%$} & \textcolor{blue}{$1.0\pm0.2\%$}\\ \midrule
		 \textbf{$\haterr_\text{HL}$} & $19.1\pm0.2\%$ & $5.8\pm0.4\%$ & $13.7\pm0.4\%$\\ \midrule
		 \textbf{$\haterr_\text{NHL}$} & $18.0\pm0.1\%$ & $6.3\pm0.1\%$ & $12.7\pm0.2\%$\\ \midrule
		 \textbf{$\Delta\hatfpr$} & \textcolor{blue}{$0.7\pm0.1\%$} & \textcolor{blue}{$1.5\pm0.1\%$} & \textcolor{blue}{$1.2\pm0.1\%$}\\ \midrule
		 \textbf{$\hatfpr_\text{HL}$} & $10.1\pm0.2\%$ & $2.6\pm0.1\%$ & $5.7\pm0.2\%$\\ \midrule
		 \textbf{$\hatfpr_\text{NHL}$} & $9.4\pm0.1\%$ & $4.1\pm0.0\%$ & $6.9\pm0.1\%$\\ \midrule
		 \textbf{$\Delta\hatfnr$} & \textcolor{blue}{$0.4\pm0.2\%$} & \textcolor{blue}{$1.1\pm0.2\%$} & \textcolor{blue}{$2.2\pm0.2\%$}\\ \midrule
		 \textbf{$\hatfnr_\text{HL}$} & $9.0\pm0.3\%$ & $3.3\pm0.3\%$ & $8.0\pm0.3\%$\\ \midrule
		 \textbf{$\hatfnr_\text{NHL}$} & $8.6\pm0.1\%$ & $2.2\pm0.1\%$ & $5.8\pm0.1\%$\\ \bottomrule
	\end{tabular}
\end{minipage}%
\caption{\textbf{Random forests} on \texttt{HMDA - 2017 - NY}, by \texttt{ethnicity}\looseness=-1}
\label{fig:hmda-ny-eth-rfc-all}
\end{figure*}
\FloatBarrier


\custompar{\texttt{NY - 2017} - by \texttt{race}}

\setlength{\tabcolsep}{6pt}
\begin{figure*}[h!]
\begin{minipage}{.495\linewidth}
\centering
\hspace{-.4cm}
        \includegraphics[width=\linewidth]{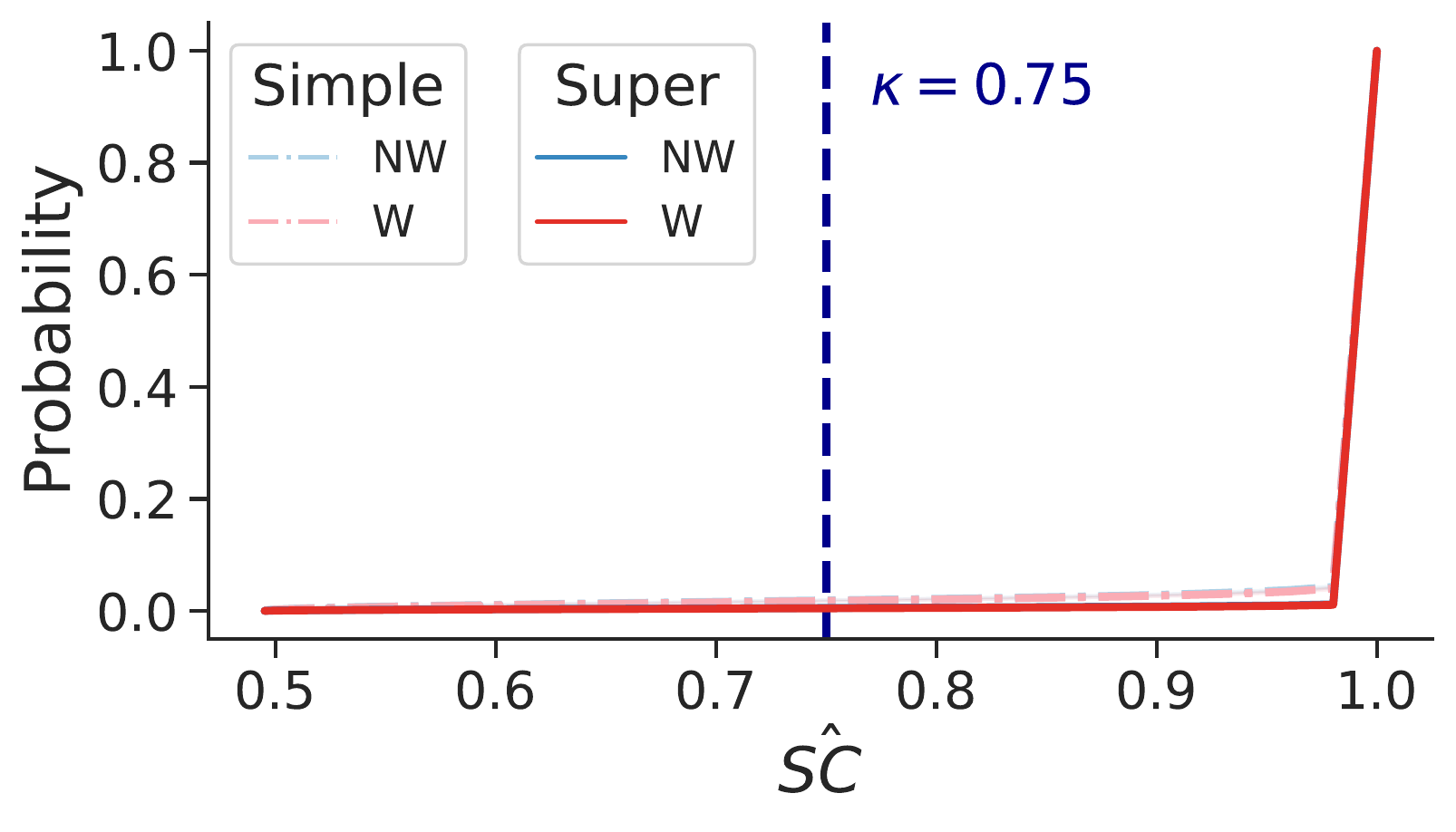}\\
        \vspace{-.1cm}%
	\begin{tabular}{lcc}
		\toprule
		\multicolumn{3}{c}{\textbf{Abstention set metrics}} \\ \cmidrule(lr){1-3}
		 & \textbf{Simple}     & \textbf{Super}      \\ \midrule
		 \textbf{$\Delta\hatar$} & \textcolor{blue}{$0.1\pm0.0\%$} & \textcolor{blue}{$0.0\pm0.1\%$}\\ \midrule
		 \textbf{$\hatar_\text{NW}$} & $2.0\pm0.1\%$ & $0.5\pm0.1\%$\\ \midrule
		 \textbf{$\hatar_\text{W}$} & $1.9\pm0.1\%$ & $0.5\pm0.0\%$\\ \bottomrule
	\end{tabular}
\end{minipage}%
\hspace{.25cm}
\begin{minipage}{.495\linewidth}
\centering
\hspace{-.2cm}
	\begin{tabular}{lccc}
		\toprule
		\multicolumn{4}{c}{\textbf{Logistic regression prediction set metrics}} \\ \cmidrule(lr){1-4}
		 & \textbf{Baseline}     & \textbf{Simple}     & \textbf{Super}      \\ \midrule
		 \textbf{$\Delta\hatpr$} & \textcolor{blue}{$11.5\pm0.2\%$} & \textcolor{blue}{$11.6\pm0.3\%$} & \textcolor{blue}{$11.5\pm0.3\%$}\\ \midrule
		 \textbf{$\hatpr_\text{NW}$} & $73.3\pm0.3\%$ & $73.9\pm0.4\%$ & $73.5\pm0.4\%$\\ \midrule
		 \textbf{$\hatpr_\text{W}$} & $84.8\pm0.1\%$ & $85.5\pm0.1\%$ & $85.0\pm0.1\%$\\ \midrule
		 \textbf{$\Delta\haterr$} & \textcolor{blue}{$2.8\pm0.1\%$} & \textcolor{blue}{$2.9\pm0.1\%$} & \textcolor{blue}{$3.0\pm0.0\%$}\\ \midrule
		 \textbf{$\haterr_\text{NW}$} & $19.7\pm0.2\%$ & $19.1\pm0.2\%$ & $19.6\pm0.1\%$\\ \midrule
		 \textbf{$\haterr_\text{W}$} & $16.9\pm0.1\%$ & $16.2\pm0.1\%$ & $16.6\pm0.1\%$\\ \midrule
		 \textbf{$\Delta\hatfpr$} & \textcolor{blue}{$0.2\pm0.1\%$} & \textcolor{blue}{$0.2\pm0.1\%$} & \textcolor{blue}{$0.1\pm0.1\%$}\\ \midrule
		 \textbf{$\hatfpr_\text{NW}$} & $11.3\pm0.2\%$ & $11.0\pm0.2\%$ & $11.3\pm0.2\%$\\ \midrule
		 \textbf{$\hatfpr_\text{W}$} & $11.5\pm0.1\%$ & $11.2\pm0.1\%$ & $11.4\pm0.1\%$\\ \midrule
		 \textbf{$\Delta\hatfnr$} & \textcolor{blue}{$3.0\pm0.1\%$} & \textcolor{blue}{$3.0\pm0.1\%$} & \textcolor{blue}{$3.0\pm0.1\%$}\\ \midrule
		 \textbf{$\hatfnr_\text{NW}$} & $8.4\pm0.2\%$ & $8.1\pm0.2\%$ & $8.3\pm0.2\%$\\ \midrule
		 \textbf{$\hatfnr_\text{W}$} & $5.4\pm0.1\%$ & $5.1\pm0.1\%$ & $5.3\pm0.1\%$\\ \bottomrule
	\end{tabular}
\end{minipage}%
\caption{\textbf{Logistic regression} on \texttt{HMDA - 2017 - NY}, by \texttt{race}\looseness=-1}
\label{fig:hmda-ny-race-lr-all}
\end{figure*}

\vspace{.5cm}
\setlength{\tabcolsep}{6pt}
\begin{figure*}[h!]
\begin{minipage}{.495\linewidth}
\centering
\hspace{-.4cm}
        \includegraphics[width=\linewidth]{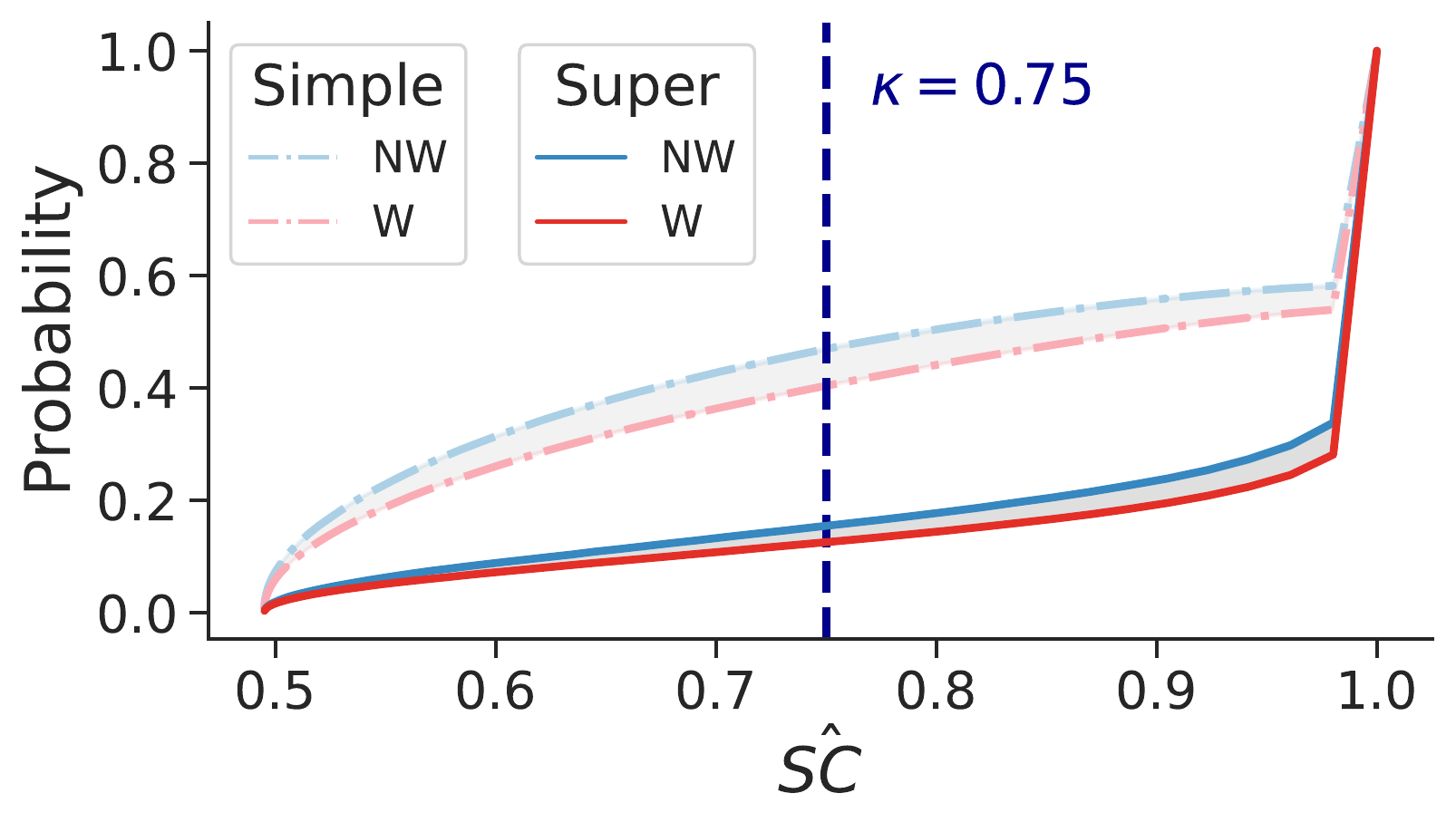}\\
        \vspace{-.1cm}%
	\begin{tabular}{lcc}
		\toprule
		\multicolumn{3}{c}{\textbf{Abstention set metrics}} \\ \cmidrule(lr){1-3}
		 & \textbf{Simple}     & \textbf{Super}      \\ \midrule
		 \textbf{$\Delta\hatar$} & \textcolor{blue}{$6.2\pm0.2\%$} & \textcolor{blue}{$2.9\pm0.0\%$}\\ \midrule
		 \textbf{$\hatar_\text{NW}$} & $47.1\pm0.4\%$ & $15.2\pm0.2\%$\\ \midrule
		 \textbf{$\hatar_\text{W}$} & $40.9\pm0.2\%$ & $12.3\pm0.2\%$\\ \bottomrule
	\end{tabular}
\end{minipage}%
\hspace{.25cm}
\begin{minipage}{.495\linewidth}
\centering
\hspace{-.2cm}
	\begin{tabular}{lccc}
		\toprule
		\multicolumn{4}{c}{\textbf{Decision tree prediction set metrics}} \\ \cmidrule(lr){1-4}
		 & \textbf{Baseline}     & \textbf{Simple}     & \textbf{Super}      \\ \midrule
		 \textbf{$\Delta\hatpr$} & \textcolor{blue}{$6.2\pm0.2\%$} & \textcolor{blue}{$3.0\pm0.1\%$} & \textcolor{blue}{$9.1\pm0.1\%$}\\ \midrule
		 \textbf{$\hatpr_\text{NW}$} & $71.8\pm0.3\%$ & $93.0\pm0.2\%$ & $74.4\pm0.3\%$\\ \midrule
		 \textbf{$\hatpr_\text{W}$} & $78.0\pm0.1\%$ & $96.0\pm0.1\%$ & $83.5\pm0.2\%$\\ \midrule
		 \textbf{$\Delta\haterr$} & \textcolor{blue}{$3.1\pm0.0\%$} & \textcolor{blue}{$0.7\pm0.1\%$} & \textcolor{blue}{$2.3\pm0.0\%$}\\ \midrule
		 \textbf{$\haterr_\text{NW}$} & $22.7\pm0.1\%$ & $3.9\pm0.2\%$ & $13.5\pm0.2\%$\\ \midrule
		 \textbf{$\haterr_\text{W}$} & $19.6\pm0.1\%$ & $3.2\pm0.1\%$ & $11.2\pm0.2\%$\\ \midrule
		 \textbf{$\Delta\hatfpr$} & \textcolor{blue}{$2.5\pm0.1\%$} & \textcolor{blue}{$0.2\pm0.0\%$} & \textcolor{blue}{$0.6\pm0.1\%$}\\ \midrule
		 \textbf{$\hatfpr_\text{NW}$} & $12.0\pm0.2\%$ & $2.5\pm0.1\%$ & $6.6\pm0.2\%$\\ \midrule
		 \textbf{$\hatfpr_\text{W}$} & $9.5\pm0.1\%$ & $2.3\pm0.1\%$ & $6.0\pm0.1\%$\\ \midrule
		 \textbf{$\Delta\hatfnr$} & \textcolor{blue}{$0.6\pm0.0\%$} & \textcolor{blue}{$0.5\pm0.1\%$} & \textcolor{blue}{$1.8\pm0.1\%$}\\ \midrule
		 \textbf{$\hatfnr_\text{NW}$} & $10.7\pm0.1\%$ & $1.4\pm0.1\%$ & $6.9\pm0.2\%$\\ \midrule
		 \textbf{$\hatfnr_\text{W}$} & $10.1\pm0.1\%$ & $0.9\pm0.0\%$ & $5.1\pm0.1\%$\\ \bottomrule
	\end{tabular}
\end{minipage}%
\caption{\textbf{Decision trees} on \texttt{HMDA - 2017 - NY}, by \texttt{race}\looseness=-1}
\label{fig:hmda-ny-race-dtc-all}
\vspace{2cm}
\end{figure*}

\vspace{.5cm}
\setlength{\tabcolsep}{6pt}
\begin{figure*}[h!]
\begin{minipage}{.495\linewidth}
\centering
\hspace{-.4cm}
        \includegraphics[width=\linewidth]{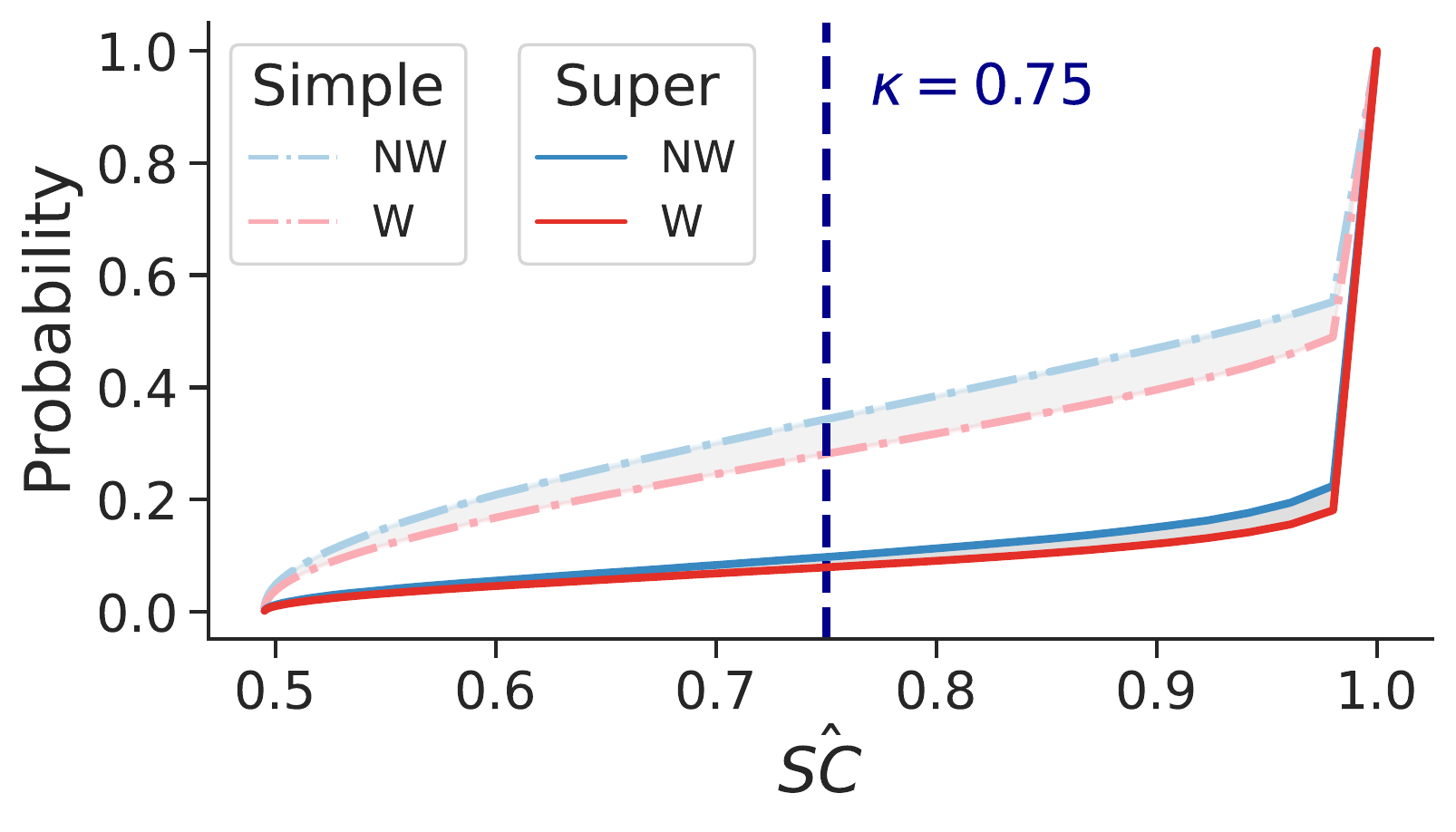}\\
        \vspace{-.1cm}%
	\begin{tabular}{lcc}
		\toprule
		\multicolumn{3}{c}{\textbf{Abstention set metrics}} \\ \cmidrule(lr){1-3}
		 & \textbf{Simple}     & \textbf{Super}      \\ \midrule
		 \textbf{$\Delta\hatar$} & \textcolor{blue}{$6.3\pm0.3\%$} & \textcolor{blue}{$1.8\pm0.2\%$}\\ \midrule
		 \textbf{$\hatar_\text{NW}$} & $35.6\pm0.4\%$ & $9.6\pm0.3\%$\\ \midrule
		 \textbf{$\hatar_\text{W}$} & $29.3\pm0.1\%$ & $7.8\pm0.1\%$\\ \bottomrule
	\end{tabular}
\end{minipage}%
\hspace{.25cm}
\begin{minipage}{.495\linewidth}
\centering
\hspace{-.2cm}
	\begin{tabular}{lccc}
		\toprule
		\multicolumn{4}{c}{\textbf{Random forest prediction set metrics}} \\ \cmidrule(lr){1-4}
		 & \textbf{Baseline}     & \textbf{Simple}     & \textbf{Super}      \\ \midrule
		 \textbf{$\Delta\hatpr$} & \textcolor{blue}{$7.9\pm0.2\%$} & \textcolor{blue}{$6.5\pm0.1\%$} & \textcolor{blue}{$9.8\pm0.1\%$}\\ \midrule
		 \textbf{$\hatpr_\text{NW}$} & $71.5\pm0.3\%$ & $85.3\pm0.2\%$ & $72.4\pm0.3\%$\\ \midrule
		 \textbf{$\hatpr_\text{W}$} & $79.4\pm0.1\%$ & $91.8\pm0.1\%$ & $82.2\pm0.2\%$\\ \midrule
		 \textbf{$\Delta\haterr$} & \textcolor{blue}{$3.3\pm0.0\%$} & \textcolor{blue}{$1.3\pm0.0\%$} & \textcolor{blue}{$2.7\pm0.1\%$}\\ \midrule
		 \textbf{$\haterr_\text{NW}$} & $20.7\pm0.1\%$ & $7.3\pm0.1\%$ & $14.9\pm0.1\%$\\ \midrule
		 \textbf{$\haterr_\text{W}$} & $17.4\pm0.1\%$ & $6.0\pm0.1\%$ & $12.2\pm0.2\%$\\ \midrule
		 \textbf{$\Delta\hatfpr$} & \textcolor{blue}{$1.8\pm0.1\%$} & \textcolor{blue}{$0.0\pm0.0\%$} & \textcolor{blue}{$0.5\pm0.0\%$}\\ \midrule
		 \textbf{$\hatfpr_\text{NW}$} & $10.9\pm0.2\%$ & $4.0\pm0.1\%$ & $7.2\pm0.1\%$\\ \midrule
		 \textbf{$\hatfpr_\text{W}$} & $9.1\pm0.1\%$ & $4.0\pm0.1\%$ & $6.7\pm0.1\%$\\ \midrule
		 \textbf{$\Delta\hatfnr$} & \textcolor{blue}{$1.4\pm0.0\%$} & \textcolor{blue}{$1.3\pm0.0\%$} & \textcolor{blue}{$2.1\pm0.1\%$}\\ \midrule
		 \textbf{$\hatfnr_\text{NW}$} & $9.8\pm0.1\%$ & $3.3\pm0.1\%$ & $7.7\pm0.1\%$\\ \midrule
		 \textbf{$\hatfnr_\text{W}$} & $8.4\pm0.1\%$ & $2.0\pm0.1\%$ & $5.6\pm0.2\%$\\ \bottomrule
	\end{tabular}
\end{minipage}%
\caption{\textbf{Random forests} on \texttt{HMDA - 2017 - NY}, by \texttt{race}\looseness=-1}
\label{fig:hmda-ny-race-rfc-all}
\end{figure*}
\vspace*{2in}
\FloatBarrier

\custompar{\texttt{NY - 2017} - by \texttt{sex}}

\setlength{\tabcolsep}{6pt}
\begin{figure*}[h!]
\begin{minipage}{.495\linewidth}
\centering
\hspace{-.4cm}
        \includegraphics[width=\linewidth]{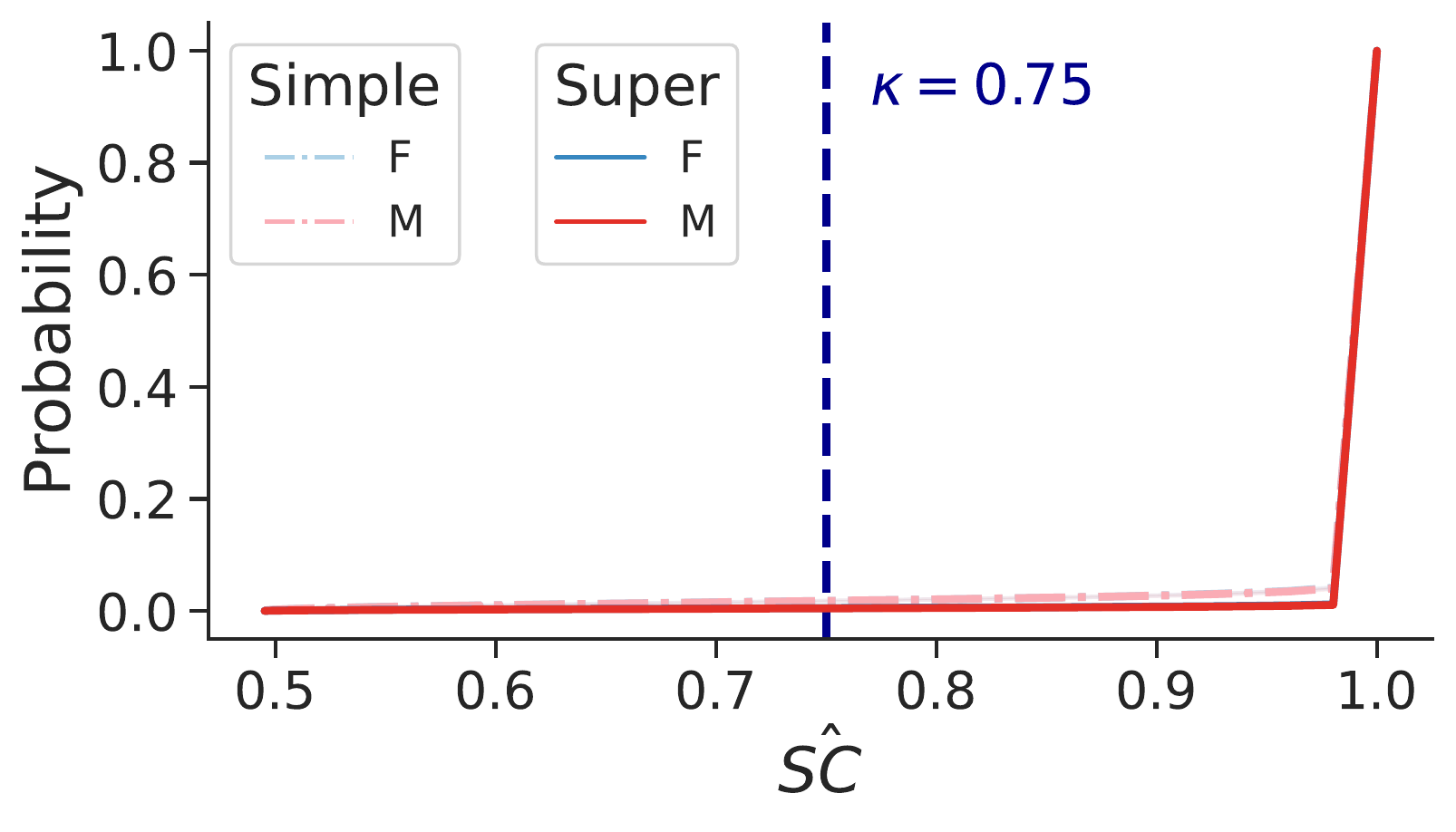}\\
        \vspace{-.1cm}%
	\begin{tabular}{lcc}
		\toprule
		\multicolumn{3}{c}{\textbf{Abstention set metrics}} \\ \cmidrule(lr){1-3}
		 & \textbf{Simple}     & \textbf{Super}      \\ \midrule
		 \textbf{$\Delta\hatar$} & \textcolor{blue}{$0.1\pm0.0\%$} & \textcolor{blue}{$0.0\pm0.0\%$}\\ \midrule
		 \textbf{$\hatar_\text{F}$} & $2.0\pm0.1\%$ & $0.5\pm0.0\%$\\ \midrule
		 \textbf{$\hatar_\text{M}$} & $1.9\pm0.1\%$ & $0.5\pm0.0\%$\\ \bottomrule
	\end{tabular}
\end{minipage}%
\hspace{.25cm}
\begin{minipage}{.495\linewidth}
\centering
\hspace{-.2cm}
	\begin{tabular}{lccc}
		\toprule
		\multicolumn{4}{c}{\textbf{Logistic regression prediction set metrics}} \\ \cmidrule(lr){1-4}
		 & \textbf{Baseline}     & \textbf{Simple}     & \textbf{Super}      \\ \midrule
		 \textbf{$\Delta\hatpr$} & \textcolor{blue}{$1.5\pm0.0\%$} & \textcolor{blue}{$1.5\pm0.2\%$} & \textcolor{blue}{$1.5\pm0.2\%$}\\ \midrule
		 \textbf{$\hatpr_\text{F}$} & $81.5\pm0.1\%$ & $82.2\pm0.3\%$ & $81.7\pm0.3\%$\\ \midrule
		 \textbf{$\hatpr_\text{M}$} & $83.0\pm0.1\%$ & $83.7\pm0.1\%$ & $83.2\pm0.1\%$\\ \midrule
		 \textbf{$\Delta\haterr$} & \textcolor{blue}{$0.5\pm0.1\%$} & \textcolor{blue}{$0.6\pm0.1\%$} & \textcolor{blue}{$0.6\pm0.1\%$}\\ \midrule
		 \textbf{$\haterr_\text{F}$} & $17.8\pm0.2\%$ & $17.2\pm0.2\%$ & $17.6\pm0.2\%$\\ \midrule
		 \textbf{$\haterr_\text{M}$} & $17.3\pm0.1\%$ & $16.6\pm0.1\%$ & $17.0\pm0.1\%$\\ \midrule
		 \textbf{$\Delta\hatfpr$} & \textcolor{blue}{$0.2\pm0.1\%$} & \textcolor{blue}{$0.3\pm0.1\%$} & \textcolor{blue}{$0.2\pm0.1\%$}\\ \midrule
		 \textbf{$\hatfpr_\text{F}$} & $11.6\pm0.2\%$ & $11.3\pm0.2\%$ & $11.5\pm0.2\%$\\ \midrule
		 \textbf{$\hatfpr_\text{M}$} & $11.4\pm0.1\%$ & $11.0\pm0.1\%$ & $11.3\pm0.1\%$\\ \midrule
		 \textbf{$\Delta\hatfnr$} & \textcolor{blue}{$0.3\pm0.0\%$} & \textcolor{blue}{$0.3\pm0.0\%$} & \textcolor{blue}{$0.3\pm0.1\%$}\\ \midrule
		 \textbf{$\hatfnr_\text{F}$} & $6.2\pm0.1\%$ & $5.9\pm0.1\%$ & $6.1\pm0.2\%$\\ \midrule
		 \textbf{$\hatfnr_\text{M}$} & $5.9\pm0.1\%$ & $5.6\pm0.1\%$ & $5.8\pm0.1\%$\\ \bottomrule
	\end{tabular}
\end{minipage}%
\caption{\textbf{Logistic regression} on \texttt{HMDA - 2017 - NY}, by \texttt{sex}\looseness=-1}
\label{fig:hmda-ny-sex-lr-all}
\vspace{1cm}
\end{figure*}

\vspace{.5cm}
\setlength{\tabcolsep}{6pt}
\begin{figure*}[h!]
\begin{minipage}{.495\linewidth}
\centering
\hspace{-.4cm}
        \includegraphics[width=\linewidth]{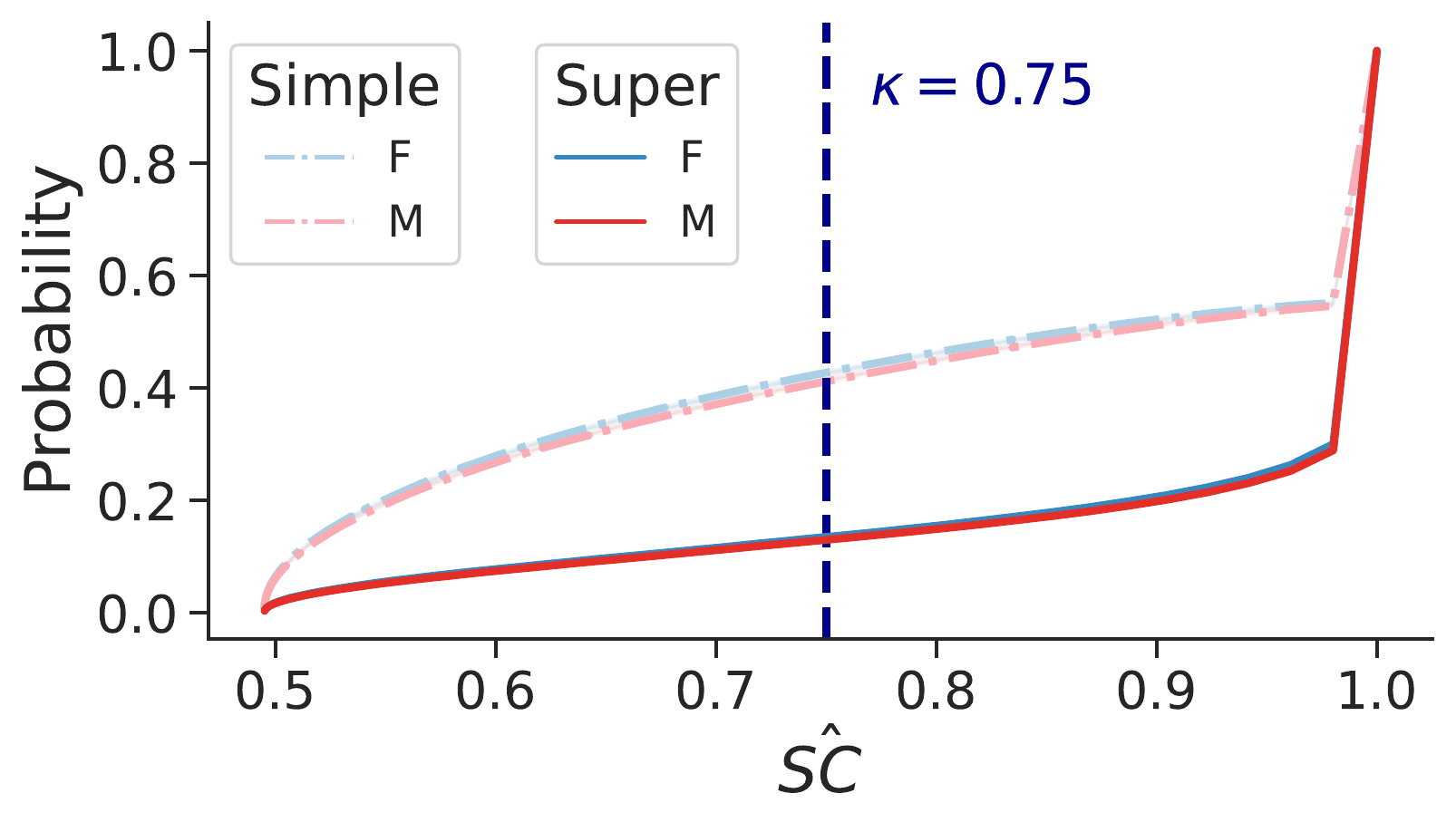}\\
        \vspace{-.1cm}%
	\begin{tabular}{lcc}
		\toprule
		\multicolumn{3}{c}{\textbf{Abstention set metrics}} \\ \cmidrule(lr){1-3}
		 & \textbf{Simple}     & \textbf{Super}      \\ \midrule
		 \textbf{$\Delta\hatar$} & \textcolor{blue}{$1.4\pm0.0\%$} & \textcolor{blue}{$0.5\pm0.0\%$}\\ \midrule
		 \textbf{$\hatar_\text{F}$} & $43.1\pm0.2\%$ & $13.2\pm0.2\%$\\ \midrule
		 \textbf{$\hatar_\text{M}$} & $41.7\pm0.2\%$ & $12.7\pm0.2\%$\\ \bottomrule
	\end{tabular}
\end{minipage}%
\hspace{.25cm}
\begin{minipage}{.495\linewidth}
\centering
\hspace{-.2cm}
	\begin{tabular}{lccc}
		\toprule
		\multicolumn{4}{c}{\textbf{Decision tree prediction set metrics}} \\ \cmidrule(lr){1-4}
		 & \textbf{Baseline}     & \textbf{Simple}     & \textbf{Super}      \\ \midrule
		 \textbf{$\Delta\hatpr$} & \textcolor{blue}{$1.5\pm0.0\%$} & \textcolor{blue}{$0.9\pm0.0\%$} & \textcolor{blue}{$1.9\pm0.1\%$}\\ \midrule
		 \textbf{$\hatpr_\text{F}$} & $75.8\pm0.1\%$ & $94.8\pm0.1\%$ & $80.5\pm0.2\%$\\ \midrule
		 \textbf{$\hatpr_\text{M}$} & $77.3\pm0.1\%$ & $95.7\pm0.1\%$ & $82.4\pm0.1\%$\\ \midrule
		 \textbf{$\Delta\haterr$} & \textcolor{blue}{$0.4\pm0.1\%$} & \textcolor{blue}{$0.1\pm0.0\%$} & \textcolor{blue}{$0.3\pm0.0\%$}\\ \midrule
		 \textbf{$\haterr_\text{F}$} & $20.5\pm0.2\%$ & $3.3\pm0.1\%$ & $11.8\pm0.2\%$\\ \midrule
		 \textbf{$\haterr_\text{M}$} & $20.1\pm0.1\%$ & $3.4\pm0.1\%$ & $11.5\pm0.2\%$\\ \midrule
		 \textbf{$\Delta\hatfpr$} & \textcolor{blue}{$0.2\pm0.1\%$} & \textcolor{blue}{$0.2\pm0.0\%$} & \textcolor{blue}{$0.1\pm0.0\%$}\\ \midrule
		 \textbf{$\hatfpr_\text{F}$} & $10.1\pm0.2\%$ & $2.2\pm0.1\%$ & $6.1\pm0.1\%$\\ \midrule
		 \textbf{$\hatfpr_\text{M}$} & $9.9\pm0.1\%$ & $2.4\pm0.1\%$ & $6.2\pm0.1\%$\\ \midrule
		 \textbf{$\Delta\hatfnr$} & \textcolor{blue}{$0.3\pm0.0\%$} & \textcolor{blue}{$0.2\pm0.0\%$} & \textcolor{blue}{$0.4\pm0.1\%$}\\ \midrule
		 \textbf{$\hatfnr_\text{F}$} & $10.4\pm0.1\%$ & $1.1\pm0.1\%$ & $5.7\pm0.2\%$\\ \midrule
		 \textbf{$\hatfnr_\text{M}$} & $10.1\pm0.1\%$ & $0.9\pm0.1\%$ & $5.3\pm0.1\%$\\ \bottomrule
	\end{tabular}
\end{minipage}%
\caption{\textbf{Decision trees} on \texttt{HMDA - 2017 - NY}, by \texttt{sex}\looseness=-1}
\label{fig:hmda-ny-sex-dtc-all}
\end{figure*}

\vspace{.5cm}
\setlength{\tabcolsep}{6pt}
\begin{figure*}[h!]
\begin{minipage}{.495\linewidth}
\centering
\hspace{-.4cm}
        \includegraphics[width=\linewidth]{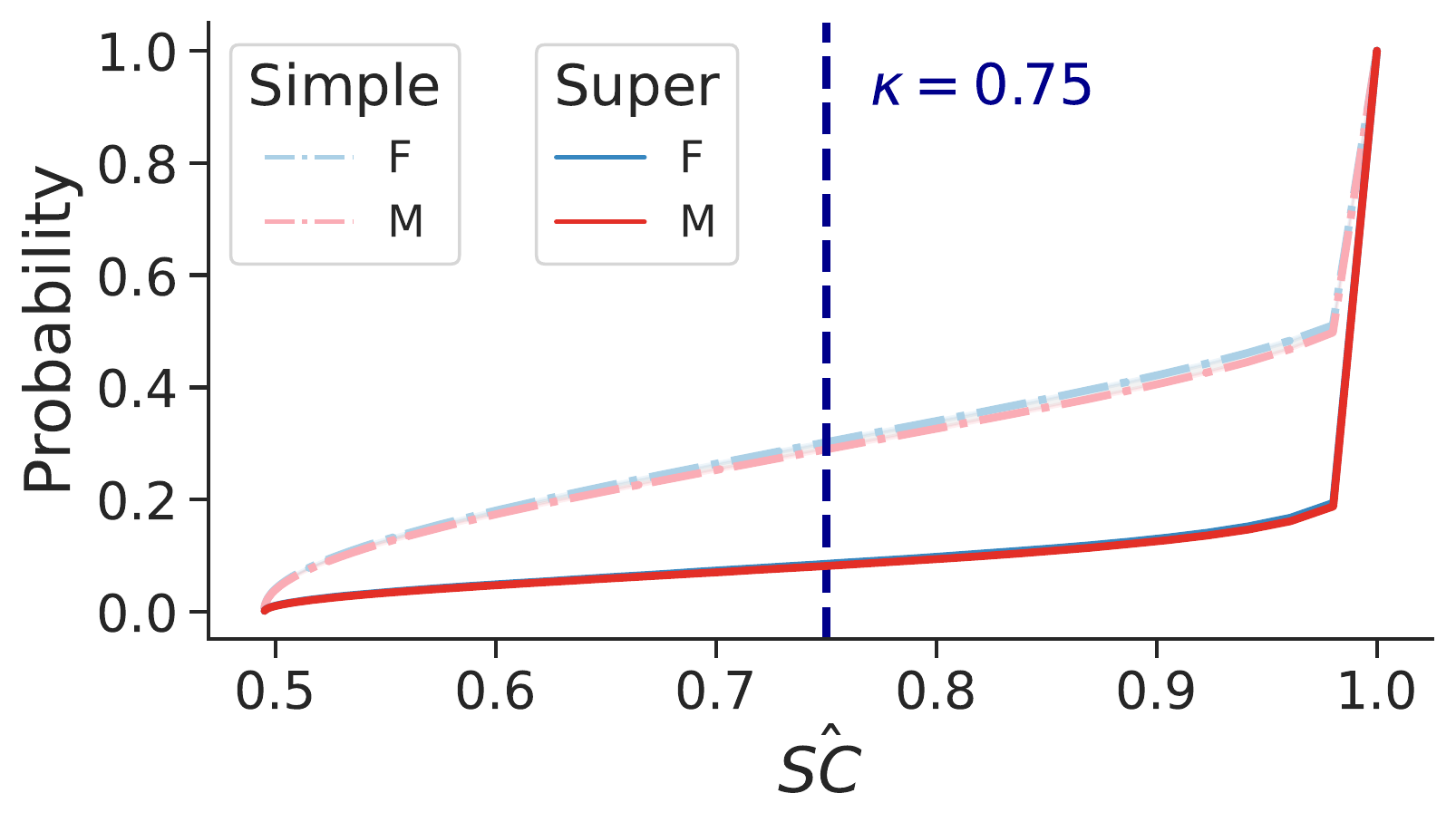}\\
        \vspace{-.1cm}%
	\begin{tabular}{lcc}
		\toprule
		\multicolumn{3}{c}{\textbf{Abstention set metrics}} \\ \cmidrule(lr){1-3}
		 & \textbf{Simple}     & \textbf{Super}      \\ \midrule
		 \textbf{$\Delta\hatar$} & \textcolor{blue}{$1.2\pm0.0\%$} & \textcolor{blue}{$0.3\pm0.0\%$}\\ \midrule
		 \textbf{$\hatar_\text{F}$} & $31.4\pm0.1\%$ & $8.4\pm0.1\%$\\ \midrule
		 \textbf{$\hatar_\text{M}$} & $30.2\pm0.1\%$ & $8.1\pm0.1\%$\\ \bottomrule
	\end{tabular}
\end{minipage}%
\hspace{.25cm}
\begin{minipage}{.495\linewidth}
\centering
\hspace{-.2cm}
	\begin{tabular}{lccc}
		\toprule
		\multicolumn{4}{c}{\textbf{Random forest prediction set metrics}} \\ \cmidrule(lr){1-4}
		 & \textbf{Baseline}     & \textbf{Simple}     & \textbf{Super}      \\ \midrule
		 \textbf{$\Delta\hatpr$} & \textcolor{blue}{$1.8\pm0.0\%$} & \textcolor{blue}{$1.7\pm0.1\%$} & \textcolor{blue}{$2.0\pm0.1\%$}\\ \midrule
		 \textbf{$\hatpr_\text{F}$} & $76.6\pm0.1\%$ & $89.4\pm0.2\%$ & $79.0\pm0.2\%$\\ \midrule
		 \textbf{$\hatpr_\text{M}$} & $78.4\pm0.1\%$ & $91.1\pm0.1\%$ & $81.0\pm0.1\%$\\ \midrule
		 \textbf{$\Delta\haterr$} & \textcolor{blue}{$0.4\pm0.1\%$} & \textcolor{blue}{$0.1\pm0.1\%$} & \textcolor{blue}{$0.4\pm0.0\%$}\\ \midrule
		 \textbf{$\haterr_\text{F}$} & $18.4\pm0.2\%$ & $6.3\pm0.2\%$ & $13.0\pm0.2\%$\\ \midrule
		 \textbf{$\haterr_\text{M}$} & $18.0\pm0.1\%$ & $6.2\pm0.1\%$ & $12.6\pm0.2\%$\\ \midrule
		 \textbf{$\Delta\hatfpr$} & \textcolor{blue}{$0.0\pm0.1\%$} & \textcolor{blue}{$0.3\pm0.0\%$} & \textcolor{blue}{$0.1\pm0.1\%$}\\ \midrule
		 \textbf{$\hatfpr_\text{F}$} & $9.4\pm0.2\%$ & $3.8\pm0.1\%$ & $6.7\pm0.2\%$\\ \midrule
		 \textbf{$\hatfpr_\text{M}$} & $9.4\pm0.1\%$ & $4.1\pm0.1\%$ & $6.8\pm0.1\%$\\ \midrule
		 \textbf{$\Delta\hatfnr$} & \textcolor{blue}{$0.4\pm0.0\%$} & \textcolor{blue}{$0.4\pm0.1\%$} & \textcolor{blue}{$0.5\pm0.1\%$}\\ \midrule
		 \textbf{$\hatfnr_\text{F}$} & $8.9\pm0.1\%$ & $2.5\pm0.1\%$ & $6.3\pm0.2\%$\\ \midrule
		 \textbf{$\hatfnr_\text{M}$} & $8.5\pm0.1\%$ & $2.1\pm0.0\%$ & $5.8\pm0.1\%$\\ \bottomrule
	\end{tabular}
\end{minipage}%
\caption{\textbf{Random forests} on \texttt{HMDA - 2017 - NY}, by \texttt{sex}\looseness=-1}
\label{fig:hmda-ny-sex-rfc-all}
\end{figure*}
\FloatBarrier

\vspace*{1cm}
\custompar{\texttt{TX - 2017} - by \texttt{ethnicity}}

\setlength{\tabcolsep}{6pt}
\begin{figure*}[h!]
\begin{minipage}{.495\linewidth}
\centering
\hspace{-.4cm}
        \includegraphics[width=\linewidth]{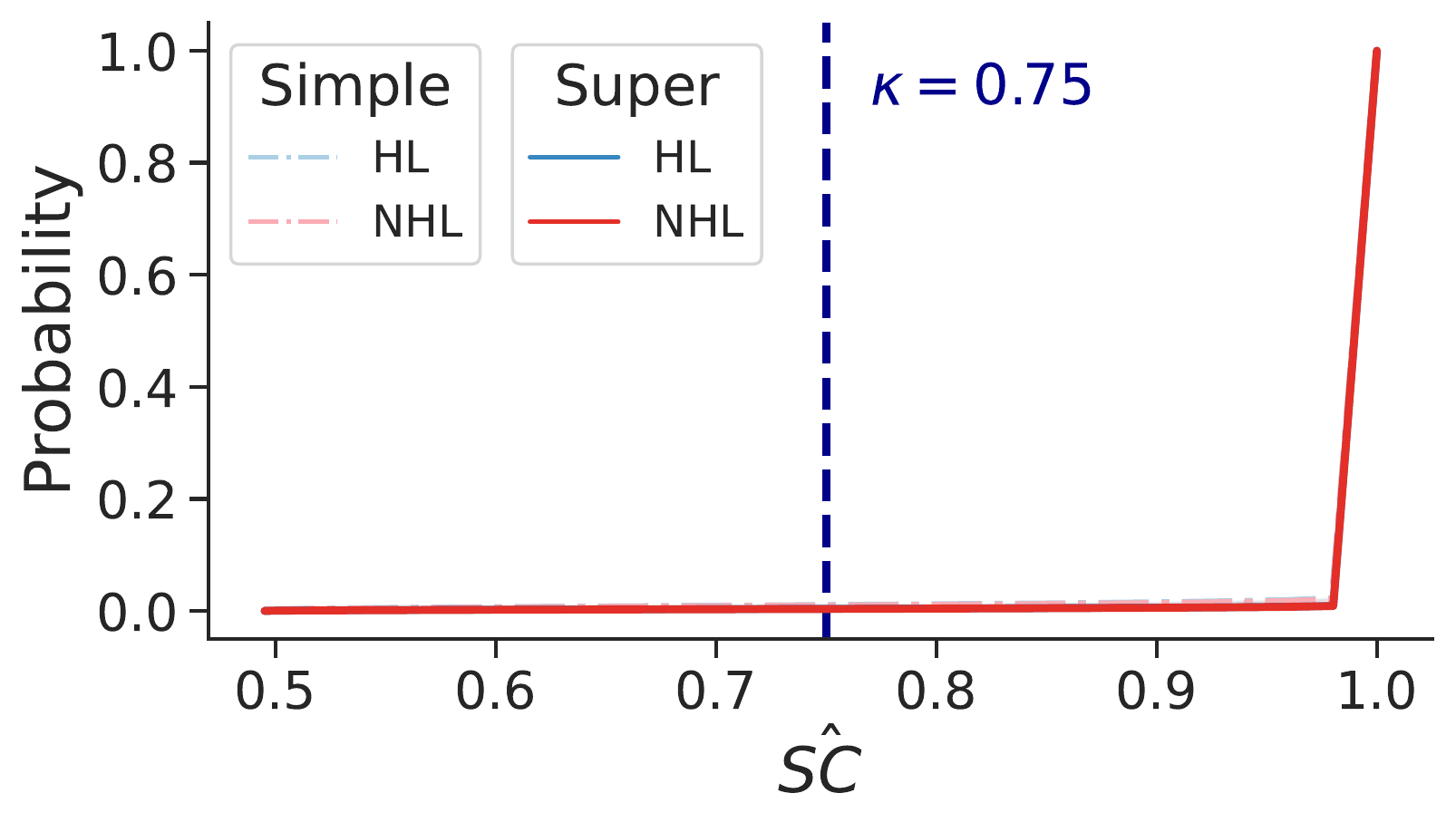}\\
        \vspace{-.1cm}%
	\begin{tabular}{lcc}
		\toprule
		\multicolumn{3}{c}{\textbf{Abstention set metrics}} \\ \cmidrule(lr){1-3}
		 & \textbf{Simple}     & \textbf{Super}      \\ \midrule
		 \textbf{$\Delta\hatar$} & \textcolor{blue}{$0.1\pm0.0\%$} & \textcolor{blue}{$0.0\pm0.0\%$}\\ \midrule
		 \textbf{$\hatar_\text{HL}$} & $1.1\pm0.0\%$ & $0.4\pm0.0\%$\\ \midrule
		 \textbf{$\hatar_\text{NHL}$} & $1.0\pm0.0\%$ & $0.4\pm0.0\%$\\ \bottomrule
	\end{tabular}
\end{minipage}%
\hspace{.25cm}
\begin{minipage}{.495\linewidth}
\centering
	\begin{tabular}{lccc}
		\toprule
		\multicolumn{4}{c}{\textbf{Logistic regression prediction set metrics}} \\ \cmidrule(lr){1-4}
		 & \textbf{Baseline}     & \textbf{Simple}     & \textbf{Super}      \\ \midrule
		 \textbf{$\Delta\hatpr$} & \textcolor{blue}{$13.3\pm0.1\%$} & \textcolor{blue}{$13.1\pm0.0\%$} & \textcolor{blue}{$13.0\pm0.0\%$}\\ \midrule
		 \textbf{$\hatpr_\text{HL}$} & $64.7\pm0.2\%$ & $65.1\pm0.1\%$ & $65.0\pm0.1\%$\\ \midrule
		 \textbf{$\hatpr_\text{NHL}$} & $78.0\pm0.1\%$ & $78.2\pm0.1\%$ & $78.0\pm0.1\%$\\ \midrule
		 \textbf{$\Delta\haterr$} & \textcolor{blue}{$3.4\pm0.1\%$} & \textcolor{blue}{$3.3\pm0.1\%$} & \textcolor{blue}{$3.3\pm0.2\%$}\\ \midrule
		 \textbf{$\haterr_\text{HL}$} & $17.5\pm0.2\%$ & $17.1\pm0.2\%$ & $17.3\pm0.2\%$\\ \midrule
		 \textbf{$\haterr_\text{NHL}$} & $14.1\pm0.1\%$ & $13.8\pm0.1\%$ & $14.0\pm0.0\%$\\ \midrule
		 \textbf{$\Delta\hatfpr$} & \textcolor{blue}{$0.6\pm0.0\%$} & \textcolor{blue}{$0.6\pm0.0\%$} & \textcolor{blue}{$0.6\pm0.0\%$}\\ \midrule
		 \textbf{$\hatfpr_\text{HL}$} & $5.6\pm0.1\%$ & $5.4\pm0.0\%$ & $5.6\pm0.0\%$\\ \midrule
		 \textbf{$\hatfpr_\text{NHL}$} & $6.2\pm0.1\%$ & $6.0\pm0.0\%$ & $6.2\pm0.0\%$\\ \midrule
		 \textbf{$\Delta\hatfnr$} & \textcolor{blue}{$4.0\pm0.1\%$} & \textcolor{blue}{$3.9\pm0.1\%$} & \textcolor{blue}{$3.9\pm0.2\%$}\\ \midrule
		 \textbf{$\hatfnr_\text{HL}$} & $11.9\pm0.2\%$ & $11.6\pm0.2\%$ & $11.7\pm0.2\%$\\ \midrule
		 \textbf{$\hatfnr_\text{NHL}$} & $7.9\pm0.1\%$ & $7.7\pm0.1\%$ & $7.8\pm0.0\%$\\ \bottomrule
	\end{tabular}
\end{minipage}%
\caption{\textbf{Logistic regression} on \texttt{HMDA - 2017 - TX}, by \texttt{ethnicity}\looseness=-1}
\label{fig:hmda-tx-eth-lr-all}
\end{figure*}

\vspace{.5cm}
\setlength{\tabcolsep}{6pt}
\begin{figure*}[h!]
\begin{minipage}{.495\linewidth}
\centering
\hspace{-.4cm}
        \includegraphics[width=\linewidth]{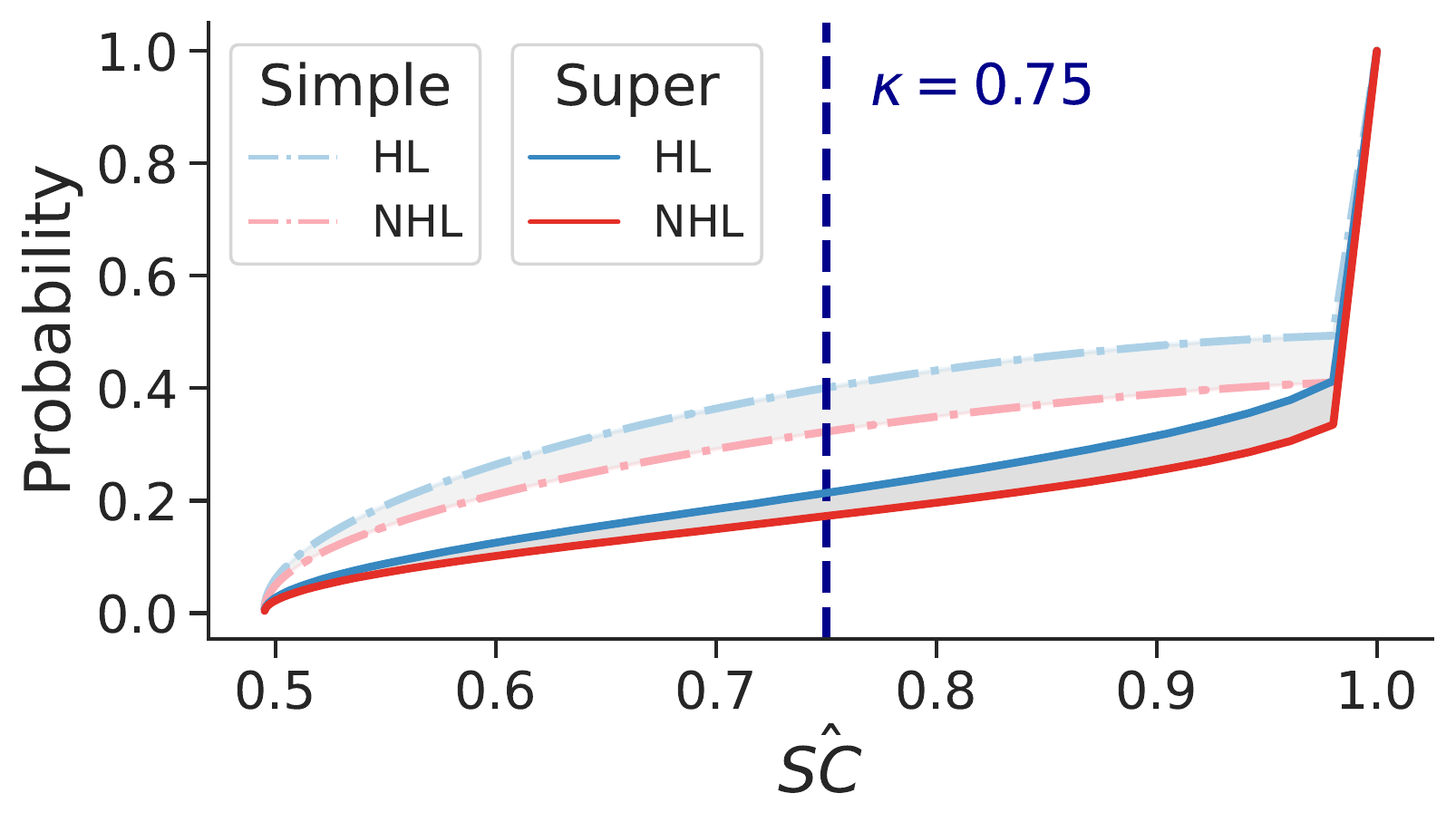}\\
        \vspace{-.1cm}%
	\begin{tabular}{lcc}
		\toprule
		\multicolumn{3}{c}{\textbf{Abstention set metrics}} \\ \cmidrule(lr){1-3}
		 & \textbf{Simple}     & \textbf{Super}      \\ \midrule
		 \textbf{$\Delta\hatar$} & \textcolor{blue}{$7.3\pm0.0\%$} & \textcolor{blue}{$4.1\pm0.1\%$}\\ \midrule
		 \textbf{$\hatar_\text{HL}$} & $40.3\pm0.1\%$ & $21.1\pm0.0\%$\\ \midrule
		 \textbf{$\hatar_\text{NHL}$} & $33.0\pm0.1\%$ & $17.0\pm0.1\%$\\ \bottomrule
	\end{tabular}
\end{minipage}%
\hspace{.25cm}
\begin{minipage}{.495\linewidth}
\centering
\hspace{-.2cm}
	\begin{tabular}{lccc}
		\toprule
		\multicolumn{4}{c}{\textbf{Decision tree prediction set metrics}} \\ \cmidrule(lr){1-4}
		 & \textbf{Baseline}     & \textbf{Simple}     & \textbf{Super}      \\ \midrule
		 \textbf{$\Delta\hatpr$} & \textcolor{blue}{$7.4\pm0.2\%$} & \textcolor{blue}{$4.1\pm0.1\%$} & \textcolor{blue}{$8.5\pm0.0\%$}\\ \midrule
		 \textbf{$\hatpr_\text{HL}$} & $72.0\pm0.2\%$ & $90.2\pm0.1\%$ & $76.7\pm0.1\%$\\ \midrule
		 \textbf{$\hatpr_\text{NHL}$} & $79.4\pm0.0\%$ & $94.3\pm0.0\%$ & $85.2\pm0.1\%$\\ \midrule
		 \textbf{$\Delta\haterr$} & \textcolor{blue}{$3.7\pm0.0\%$} & \textcolor{blue}{$0.7\pm0.1\%$} & \textcolor{blue}{$2.0\pm0.1\%$}\\ \midrule
		 \textbf{$\haterr_\text{HL}$} & $19.2\pm0.1\%$ & $3.0\pm0.1\%$ & $8.4\pm0.1\%$\\ \midrule
		 \textbf{$\haterr_\text{NHL}$} & $15.5\pm0.1\%$ & $2.3\pm0.0\%$ & $6.4\pm0.0\%$\\ \midrule
		 \textbf{$\Delta\hatfpr$} & \textcolor{blue}{$2.4\pm0.1\%$} & \textcolor{blue}{$0.1\pm0.0\%$} & \textcolor{blue}{$0.4\pm0.0\%$}\\ \midrule
		 \textbf{$\hatfpr_\text{HL}$} & $10.1\pm0.1\%$ & $1.2\pm0.0\%$ & $3.3\pm0.1\%$\\ \midrule
		 \textbf{$\hatfpr_\text{NHL}$} & $7.7\pm0.0\%$ & $1.1\pm0.0\%$ & $2.9\pm0.1\%$\\ \midrule
		 \textbf{$\Delta\hatfnr$} & \textcolor{blue}{$1.2\pm0.1\%$} & \textcolor{blue}{$0.6\pm0.0\%$} & \textcolor{blue}{$1.5\pm0.0\%$}\\ \midrule
		 \textbf{$\hatfnr_\text{HL}$} & $9.1\pm0.1\%$ & $1.8\pm0.0\%$ & $5.0\pm0.0\%$\\ \midrule
		 \textbf{$\hatfnr_\text{NHL}$} & $7.9\pm0.0\%$ & $1.2\pm0.0\%$ & $3.5\pm0.0\%$\\ \bottomrule
	\end{tabular}
\end{minipage}%
\caption{\textbf{Decision trees} on \texttt{HMDA - 2017 - TX}, by \texttt{ethnicity}\looseness=-1}
\label{fig:hmda-tx-eth-dtc-all}
\vspace{2cm}
\end{figure*}

\vspace*{.5cm}
\setlength{\tabcolsep}{6pt}
\begin{figure*}[h!]
\begin{minipage}{.495\linewidth}
\centering
\hspace{-.4cm}
        \includegraphics[width=\linewidth]{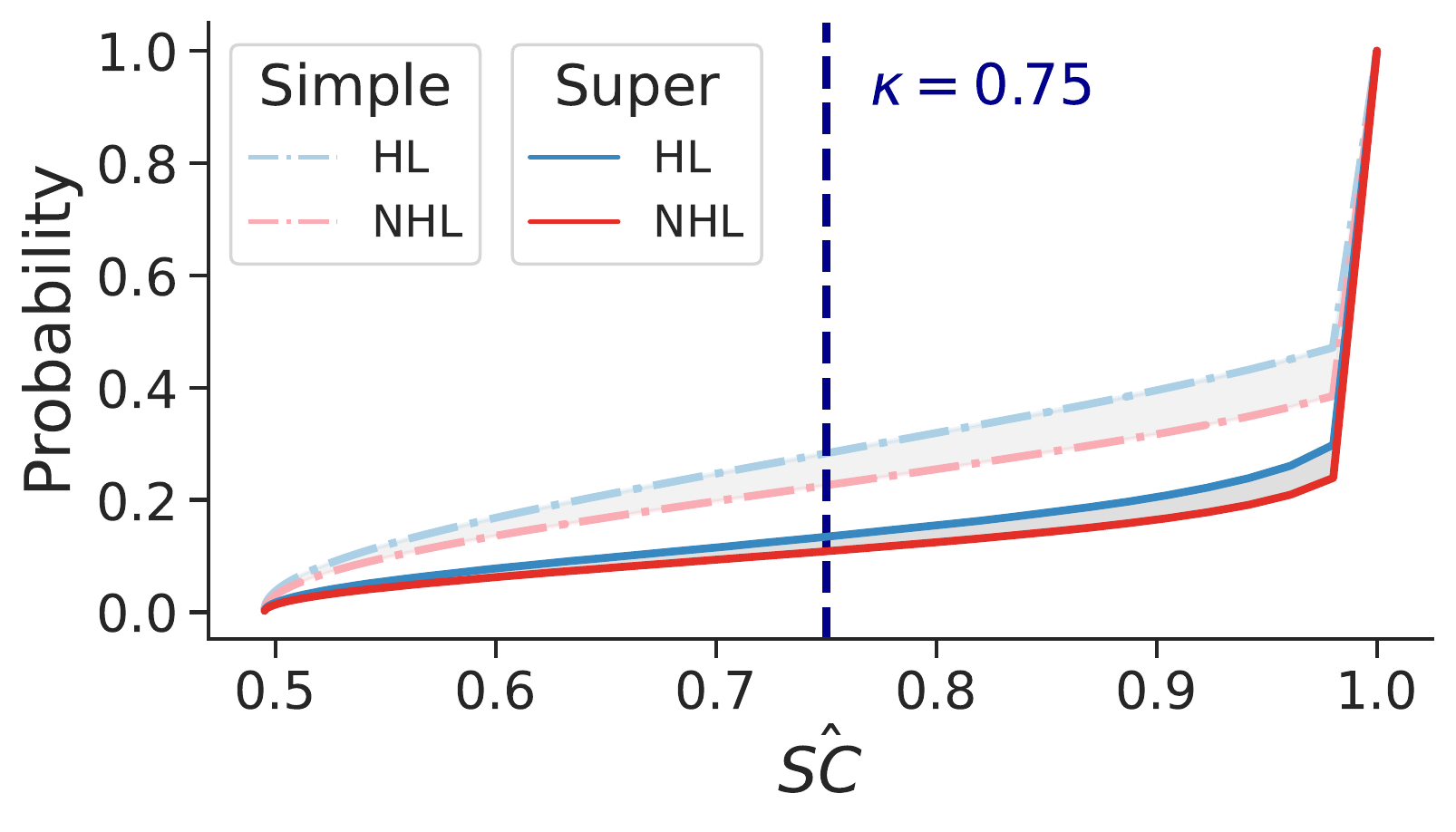}\\
        \vspace{-.1cm}%
	\begin{tabular}{lcc}
		\toprule
		\multicolumn{3}{c}{\textbf{Abstention set metrics}} \\ \cmidrule(lr){1-3}
		 & \textbf{Simple}     & \textbf{Super}      \\ \midrule
		 \textbf{$\Delta\hatar$} & \textcolor{blue}{$6.2\pm0.1\%$} & \textcolor{blue}{$2.5\pm0.1\%$}\\ \midrule
		 \textbf{$\hatar_\text{HL}$} & $31.9\pm0.0\%$ & $13.2\pm0.1\%$\\ \midrule
		 \textbf{$\hatar_\text{NHL}$} & $25.7\pm0.1\%$ & $10.7\pm0.0\%$\\ \bottomrule
	\end{tabular}
\end{minipage}%
\hspace{.25cm}
\begin{minipage}{.495\linewidth}
\centering
\hspace{-.2cm}
	\begin{tabular}{lccc}
		\toprule
		\multicolumn{4}{c}{\textbf{Random forest prediction set metrics}} \\ \cmidrule(lr){1-4}
		 & \textbf{Baseline}     & \textbf{Simple}     & \textbf{Super}      \\ \midrule
		 \textbf{$\Delta\hatpr$} & \textcolor{blue}{$8.7\pm0.1\%$} & \textcolor{blue}{$6.9\pm0.1\%$} & \textcolor{blue}{$9.6\pm0.1\%$}\\ \midrule
		 \textbf{$\hatpr_\text{HL}$} & $70.6\pm0.2\%$ & $83.0\pm0.2\%$ & $72.7\pm0.2\%$\\ \midrule
		 \textbf{$\hatpr_\text{NHL}$} & $79.3\pm0.1\%$ & $89.9\pm0.1\%$ & $82.3\pm0.1\%$\\ \midrule
		 \textbf{$\Delta\haterr$} & \textcolor{blue}{$3.3\pm0.0\%$} & \textcolor{blue}{$1.2\pm0.1\%$} & \textcolor{blue}{$2.2\pm0.1\%$}\\ \midrule
		 \textbf{$\haterr_\text{HL}$} & $17.3\pm0.1\%$ & $5.3\pm0.1\%$ & $10.5\pm0.1\%$\\ \midrule
		 \textbf{$\haterr_\text{NHL}$} & $14.0\pm0.1\%$ & $4.1\pm0.0\%$ & $8.3\pm0.0\%$\\ \midrule
		 \textbf{$\Delta\hatfpr$} & \textcolor{blue}{$1.7\pm0.1\%$} & \textcolor{blue}{$0.1\pm0.0\%$} & \textcolor{blue}{$0.4\pm0.1\%$}\\ \midrule
		 \textbf{$\hatfpr_\text{HL}$} & $8.5\pm0.1\%$ & $2.0\pm0.0\%$ & $4.1\pm0.0\%$\\ \midrule
		 \textbf{$\hatfpr_\text{NHL}$} & $6.8\pm0.0\%$ & $1.9\pm0.0\%$ & $3.7\pm0.1\%$\\ \midrule
		 \textbf{$\Delta\hatfnr$} & \textcolor{blue}{$1.7\pm0.1\%$} & \textcolor{blue}{$1.1\pm0.1\%$} & \textcolor{blue}{$1.8\pm0.1\%$}\\ \midrule
		 \textbf{$\hatfnr_\text{HL}$} & $8.9\pm0.1\%$ & $3.3\pm0.1\%$ & $6.4\pm0.1\%$\\ \midrule
		 \textbf{$\hatfnr_\text{NHL}$} & $7.2\pm0.0\%$ & $2.2\pm0.0\%$ & $4.6\pm0.0\%$\\ \bottomrule
	\end{tabular}
\end{minipage}%
\caption{\textbf{Random forests} on \texttt{HMDA - 2017 - TX}, by \texttt{ethnicity}\looseness=-1}
\label{fig:hmda-tx-eth-rfc-all}
\end{figure*}
\vspace*{2in}
\FloatBarrier


\custompar{\texttt{TX - 2017} - by \texttt{race}}

\setlength{\tabcolsep}{6pt}
\begin{figure*}[h!]
\begin{minipage}{.495\linewidth}
\centering
\hspace{-.4cm}
        \includegraphics[width=\linewidth]{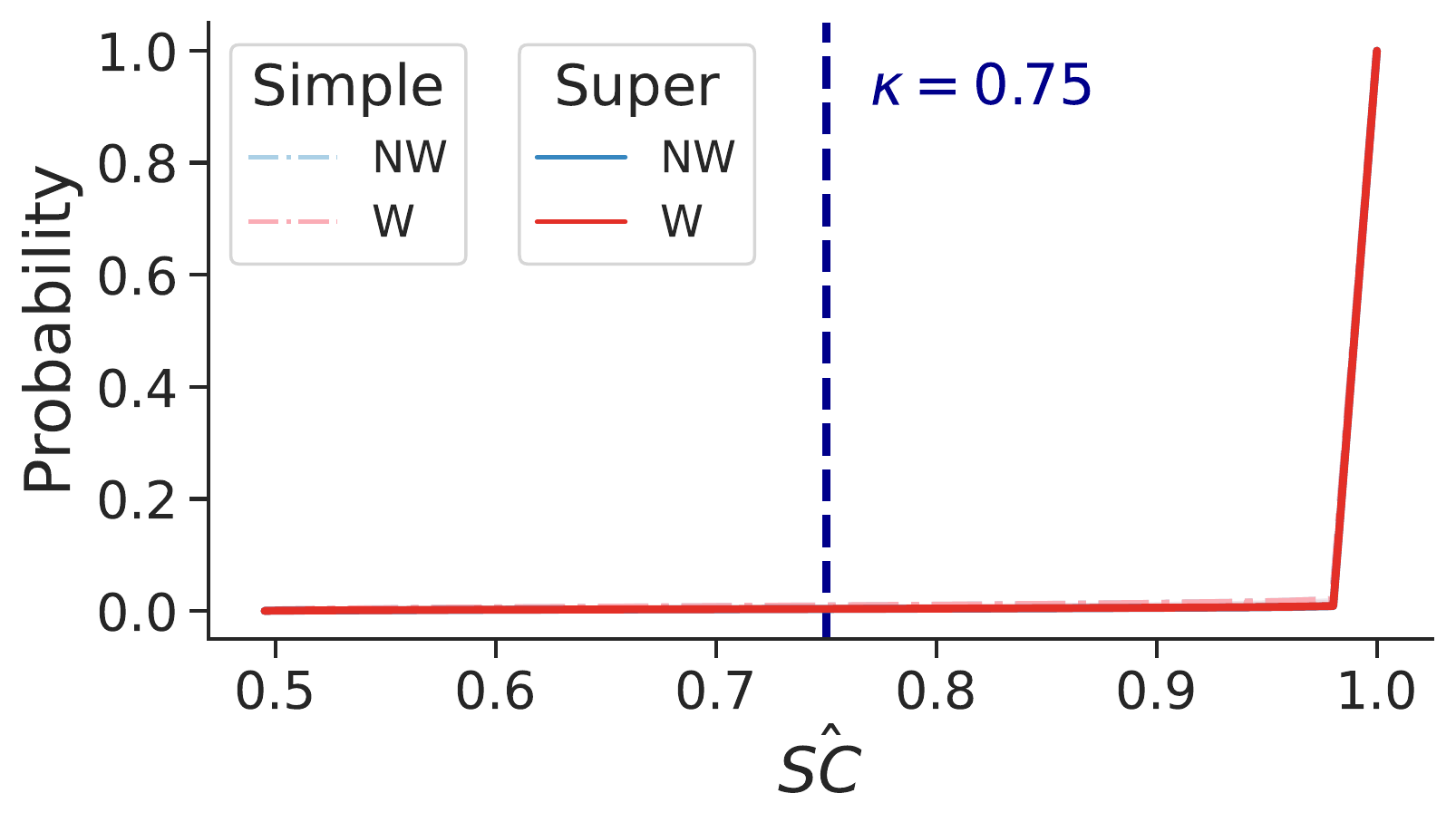}\\
        \vspace{-.1cm}%
	\begin{tabular}{lcc}
		\toprule
		\multicolumn{3}{c}{\textbf{Abstention set metrics}} \\ \cmidrule(lr){1-3}
		 & \textbf{Simple}     & \textbf{Super}      \\ \midrule
		 \textbf{$\Delta\hatar$} & \textcolor{blue}{$0.0\pm0.0\%$} & \textcolor{blue}{$0.0\pm0.0\%$}\\ \midrule
		 \textbf{$\hatar_\text{NW}$} & $1.0\pm0.0\%$ & $0.4\pm0.0\%$\\ \midrule
		 \textbf{$\hatar_\text{W}$} & $1.0\pm0.0\%$ & $0.4\pm0.0\%$\\ \bottomrule
	\end{tabular}
\end{minipage}%
\hspace{.25cm}
\begin{minipage}{.495\linewidth}
\centering
\hspace{-.2cm}
	\begin{tabular}{lccc}
		\toprule
		\multicolumn{4}{c}{\textbf{Logistic regression prediction set metrics}} \\ \cmidrule(lr){1-4}
		 & \textbf{Baseline}     & \textbf{Simple}     & \textbf{Super}      \\ \midrule
		 \textbf{$\Delta\hatpr$} & \textcolor{blue}{$2.6\pm0.1\%$} & \textcolor{blue}{$2.7\pm0.1\%$} & \textcolor{blue}{$2.7\pm0.1\%$}\\ \midrule
		 \textbf{$\hatpr_\text{NW}$} & $72.6\pm0.2\%$ & $72.8\pm0.0\%$ & $72.6\pm0.0\%$\\ \midrule
		 \textbf{$\hatpr_\text{W}$} & $75.2\pm0.1\%$ & $75.5\pm0.1\%$ & $75.3\pm0.1\%$\\ \midrule
		 \textbf{$\Delta\haterr$} & \textcolor{blue}{$0.0\pm0.1\%$} & \textcolor{blue}{$0.2\pm0.0\%$} & \textcolor{blue}{$0.1\pm0.0\%$}\\ \midrule
		 \textbf{$\haterr_\text{NW}$} & $14.9\pm0.2\%$ & $14.4\pm0.1\%$ & $14.7\pm0.1\%$\\ \midrule
		 \textbf{$\haterr_\text{W}$} & $14.9\pm0.1\%$ & $14.6\pm0.1\%$ & $14.8\pm0.1\%$\\ \midrule
		 \textbf{$\Delta\hatfpr$} & \textcolor{blue}{$1.1\pm0.1\%$} & \textcolor{blue}{$0.9\pm0.1\%$} & \textcolor{blue}{$1.0\pm0.1\%$}\\ \midrule
		 \textbf{$\hatfpr_\text{NW}$} & $7.0\pm0.2\%$ & $6.6\pm0.1\%$ & $6.8\pm0.1\%$\\ \midrule
		 \textbf{$\hatfpr_\text{W}$} & $5.9\pm0.1\%$ & $5.7\pm0.0\%$ & $5.8\pm0.0\%$\\ \midrule
		 \textbf{$\Delta\hatfnr$} & \textcolor{blue}{$1.2\pm0.0\%$} & \textcolor{blue}{$1.1\pm0.1\%$} & \textcolor{blue}{$1.1\pm0.0\%$}\\ \midrule
		 \textbf{$\hatfnr_\text{NW}$} & $7.9\pm0.1\%$ & $7.8\pm0.0\%$ & $7.9\pm0.1\%$\\ \midrule
		 \textbf{$\hatfnr_\text{W}$} & $9.1\pm0.1\%$ & $8.9\pm0.1\%$ & $9.0\pm0.1\%$\\ \bottomrule
	\end{tabular}
\end{minipage}%
\caption{\textbf{Logistic regression} on \texttt{HMDA - 2017 - TX}, by \texttt{race}\looseness=-1}
\label{fig:hmda-tx-race-lr-all}
\vspace{1cm}
\end{figure*}

\vspace{.5cm}
\setlength{\tabcolsep}{6pt}
\begin{figure*}[h!]
\begin{minipage}{.495\linewidth}
\centering
\hspace{-.4cm}
        \includegraphics[width=\linewidth]{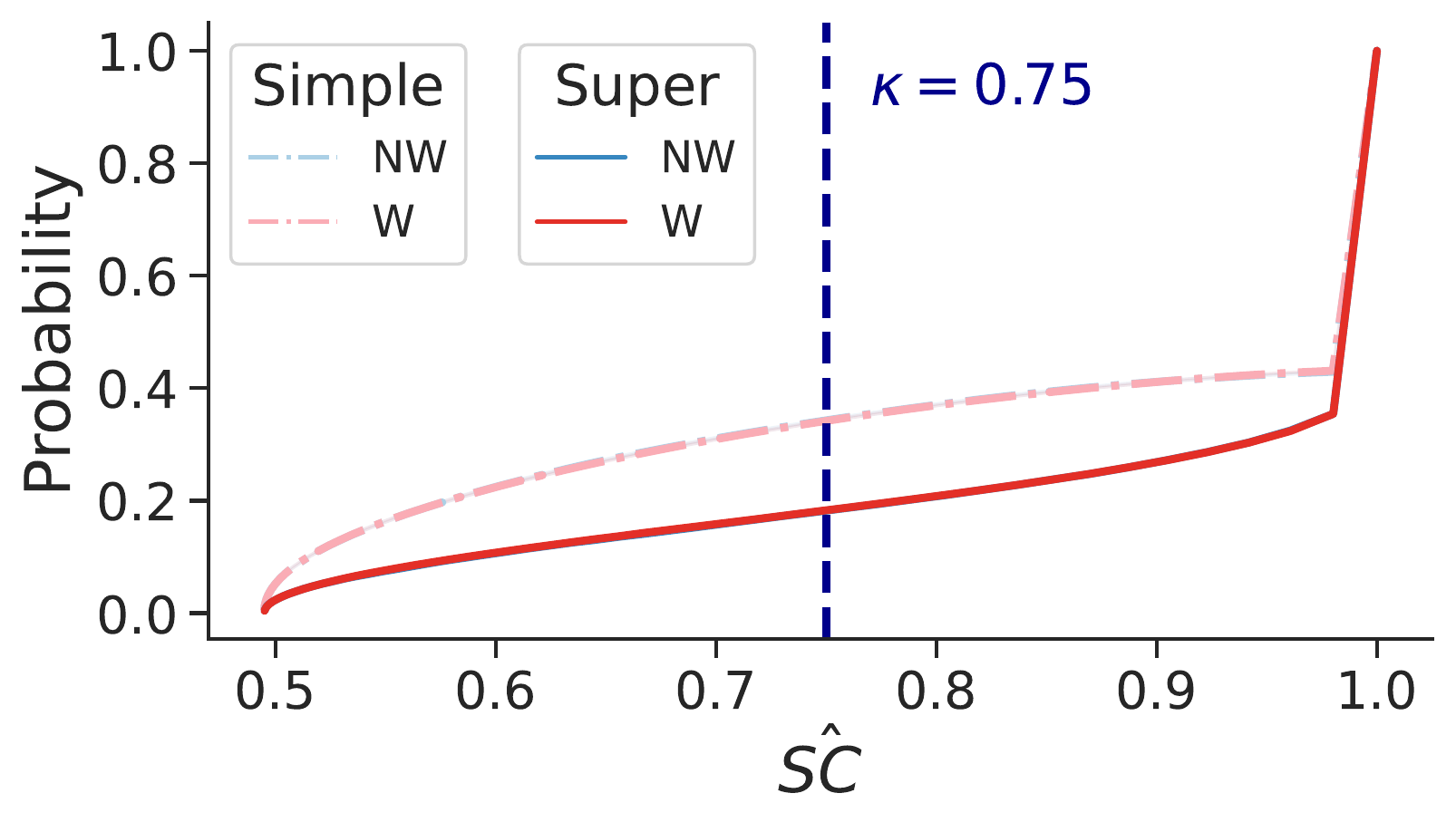}\\
        \vspace{-.1cm}%
	\begin{tabular}{lcc}
		\toprule
		\multicolumn{3}{c}{\textbf{Abstention set metrics}} \\ \cmidrule(lr){1-3}
		 & \textbf{Simple}     & \textbf{Super}      \\ \midrule
		 \textbf{$\Delta\hatar$} & \textcolor{blue}{$0.1\pm0.1\%$} & \textcolor{blue}{$0.0\pm0.3\%$}\\ \midrule
		 \textbf{$\hatar_\text{NW}$} & $34.7\pm0.2\%$ & $18.0\pm0.3\%$\\ \midrule
		 \textbf{$\hatar_\text{W}$} & $34.8\pm0.1\%$ & $18.0\pm0.0\%$\\ \bottomrule
	\end{tabular}
\end{minipage}%
\hspace{.25cm}
\begin{minipage}{.495\linewidth}
\centering
\hspace{-.2cm}
	\begin{tabular}{lccc}
		\toprule
		\multicolumn{4}{c}{\textbf{Decision tree prediction set metrics}} \\ \cmidrule(lr){1-4}
		 & \textbf{Baseline}     & \textbf{Simple}     & \textbf{Super}      \\ \midrule
		 \textbf{$\Delta\hatpr$} & \textcolor{blue}{$2.5\pm0.0\%$} & \textcolor{blue}{$1.8\pm0.0\%$} & \textcolor{blue}{$4.1\pm0.2\%$}\\ \midrule
		 \textbf{$\hatpr_\text{NW}$} & $75.6\pm0.1\%$ & $91.9\pm0.1\%$ & $79.8\pm0.3\%$\\ \midrule
		 \textbf{$\hatpr_\text{W}$} & $78.1\pm0.1\%$ & $93.7\pm0.1\%$ & $83.9\pm0.1\%$\\ \midrule
		 \textbf{$\Delta\haterr$} & \textcolor{blue}{$0.0\pm0.1\%$} & \textcolor{blue}{$0.0\pm0.0\%$} & \textcolor{blue}{$0.1\pm0.0\%$}\\ \midrule
		 \textbf{$\haterr_\text{NW}$} & $16.4\pm0.1\%$ & $2.4\pm0.0\%$ & $6.8\pm0.1\%$\\ \midrule
		 \textbf{$\haterr_\text{W}$} & $16.4\pm0.0\%$ & $2.4\pm0.0\%$ & $6.9\pm0.1\%$\\ \midrule
		 \textbf{$\Delta\hatfpr$} & \textcolor{blue}{$1.2\pm0.0\%$} & \textcolor{blue}{$0.0\pm0.0\%$} & \textcolor{blue}{$0.1\pm0.0\%$}\\ \midrule
		 \textbf{$\hatfpr_\text{NW}$} & $9.2\pm0.1\%$ & $1.1\pm0.0\%$ & $2.9\pm0.1\%$\\ \midrule
		 \textbf{$\hatfpr_\text{W}$} & $8.0\pm0.1\%$ & $1.1\pm0.0\%$ & $3.0\pm0.1\%$\\ \midrule
		 \textbf{$\Delta\hatfnr$} & \textcolor{blue}{$1.2\pm0.0\%$} & \textcolor{blue}{$0.1\pm0.0\%$} & \textcolor{blue}{$0.1\pm0.1\%$}\\ \midrule
		 \textbf{$\hatfnr_\text{NW}$} & $7.2\pm0.1\%$ & $1.4\pm0.0\%$ & $3.9\pm0.1\%$\\ \midrule
		 \textbf{$\hatfnr_\text{W}$} & $8.4\pm0.1\%$ & $1.3\pm0.0\%$ & $3.8\pm0.0\%$\\ \bottomrule
	\end{tabular}
\end{minipage}%
\caption{\textbf{Decision trees} on \texttt{HMDA - 2017 - TX}, by \texttt{race}\looseness=-1}
\label{fig:hmda-tx-race-dtc-all}
\end{figure*}

\vspace{.5cm}
\setlength{\tabcolsep}{6pt}
\begin{figure*}[h!]
\begin{minipage}{.495\linewidth}
\centering
\hspace{-.4cm}
        \includegraphics[width=\linewidth]{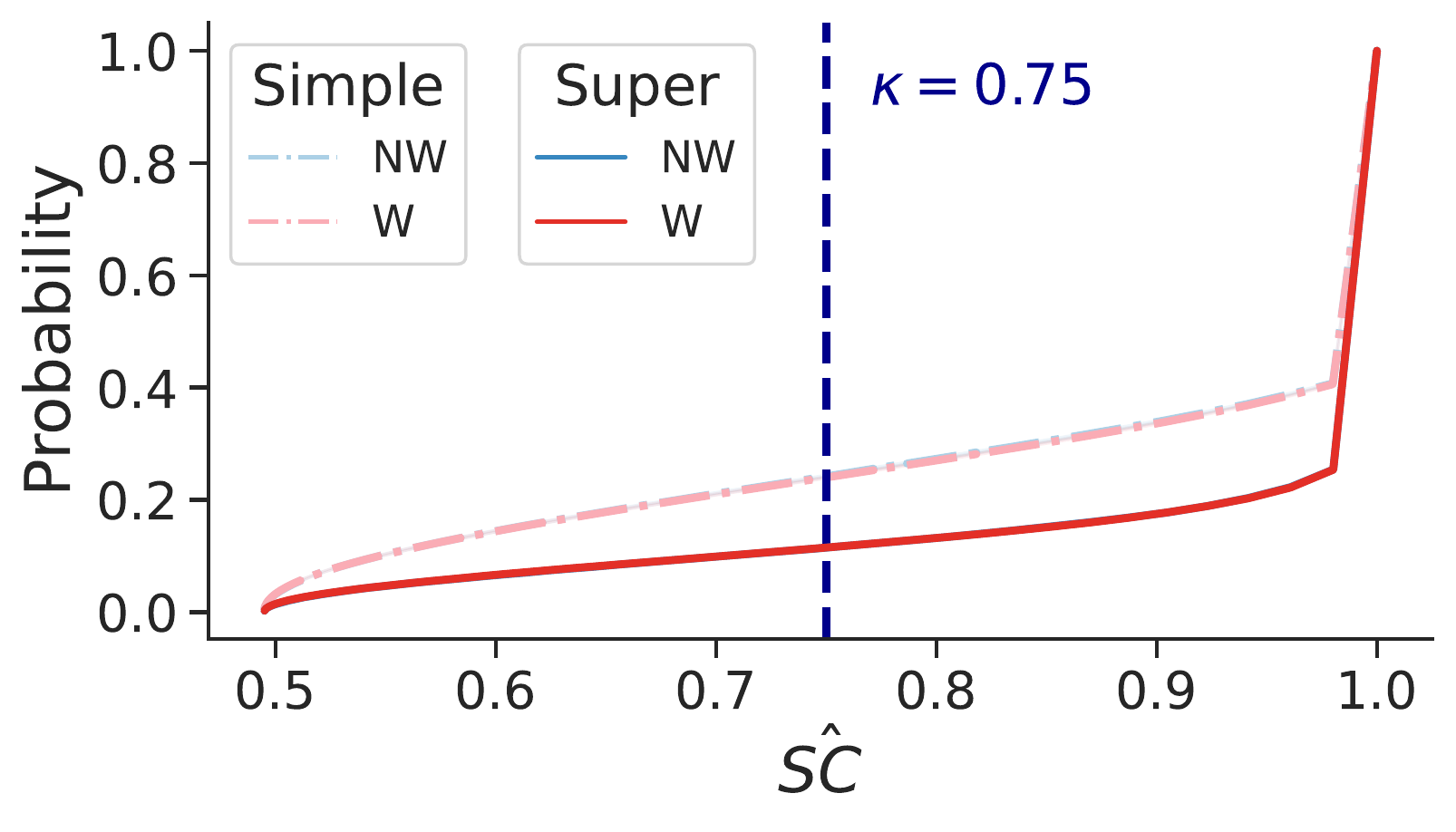}\\
        \vspace{-.1cm}%
	\begin{tabular}{lcc}
		\toprule
		\multicolumn{3}{c}{\textbf{Abstention set metrics}} \\ \cmidrule(lr){1-3}
		 & \textbf{Simple}     & \textbf{Super}      \\ \midrule
		 \textbf{$\Delta\hatar$} & \textcolor{blue}{$0.1\pm0.1\%$} & \textcolor{blue}{$0.0\pm0.0\%$}\\ \midrule
		 \textbf{$\hatar_\text{NW}$} & $27.3\pm0.1\%$ & $11.3\pm0.1\%$\\ \midrule
		 \textbf{$\hatar_\text{W}$} & $27.2\pm0.0\%$ & $11.3\pm0.1\%$\\ \bottomrule
	\end{tabular}
\end{minipage}%
\hspace{.25cm}
\begin{minipage}{.495\linewidth}
\centering
\hspace{-.2cm}
	\begin{tabular}{lccc}
		\toprule
		\multicolumn{4}{c}{\textbf{Random forest prediction set metrics}} \\ \cmidrule(lr){1-4}
		 & \textbf{Baseline}     & \textbf{Simple}     & \textbf{Super}      \\ \midrule
		 \textbf{$\Delta\hatpr$} & \textcolor{blue}{$3.4\pm0.1\%$} & \textcolor{blue}{$3.3\pm0.0\%$} & \textcolor{blue}{$4.5\pm0.0\%$}\\ \midrule
		 \textbf{$\hatpr_\text{NW}$} & $74.4\pm0.2\%$ & $85.6\pm0.1\%$ & $76.3\pm0.1\%$\\ \midrule
		 \textbf{$\hatpr_\text{W}$} & $77.8\pm0.1\%$ & $88.9\pm0.1\%$ & $80.8\pm0.1\%$\\ \midrule
		 \textbf{$\Delta\haterr$} & \textcolor{blue}{$0.0\pm0.0\%$} & \textcolor{blue}{$0.1\pm0.0\%$} & \textcolor{blue}{$0.1\pm0.0\%$}\\ \midrule
		 \textbf{$\haterr_\text{NW}$} & $14.8\pm0.1\%$ & $4.3\pm0.0\%$ & $8.7\pm0.1\%$\\ \midrule
		 \textbf{$\haterr_\text{W}$} & $14.8\pm0.1\%$ & $4.4\pm0.0\%$ & $8.8\pm0.1\%$\\ \midrule
		 \textbf{$\Delta\hatfpr$} & \textcolor{blue}{$0.7\pm0.0\%$} & \textcolor{blue}{$0.1\pm0.1\%$} & \textcolor{blue}{$0.1\pm0.1\%$}\\ \midrule
		 \textbf{$\hatfpr_\text{NW}$} & $7.8\pm0.1\%$ & $1.8\pm0.1\%$ & $3.7\pm0.1\%$\\ \midrule
		 \textbf{$\hatfpr_\text{W}$} & $7.1\pm0.1\%$ & $1.9\pm0.0\%$ & $3.8\pm0.0\%$\\ \midrule
		 \textbf{$\Delta\hatfnr$} & \textcolor{blue}{$0.8\pm0.0\%$} & \textcolor{blue}{$0.2\pm0.1\%$} & \textcolor{blue}{$0.1\pm0.0\%$}\\ \midrule
		 \textbf{$\hatfnr_\text{NW}$} & $6.9\pm0.1\%$ & $2.6\pm0.1\%$ & $5.1\pm0.0\%$\\ \midrule
		 \textbf{$\hatfnr_\text{W}$} & $7.7\pm0.1\%$ & $2.4\pm0.0\%$ & $5.0\pm0.0\%$\\ \bottomrule
	\end{tabular}
\end{minipage}%
\caption{\textbf{Random forests} on \texttt{HMDA - 2017 - TX}, by \texttt{race}\looseness=-1}
\label{fig:hmda-tx-race-rfc-all}
\end{figure*}
\FloatBarrier


\vspace*{1cm}
\custompar{\texttt{TX - 2017} - by \texttt{sex}}

\setlength{\tabcolsep}{6pt}
\begin{figure*}[h!]
\begin{minipage}{.495\linewidth}
\centering
\hspace{-.4cm}
        \includegraphics[width=\linewidth]{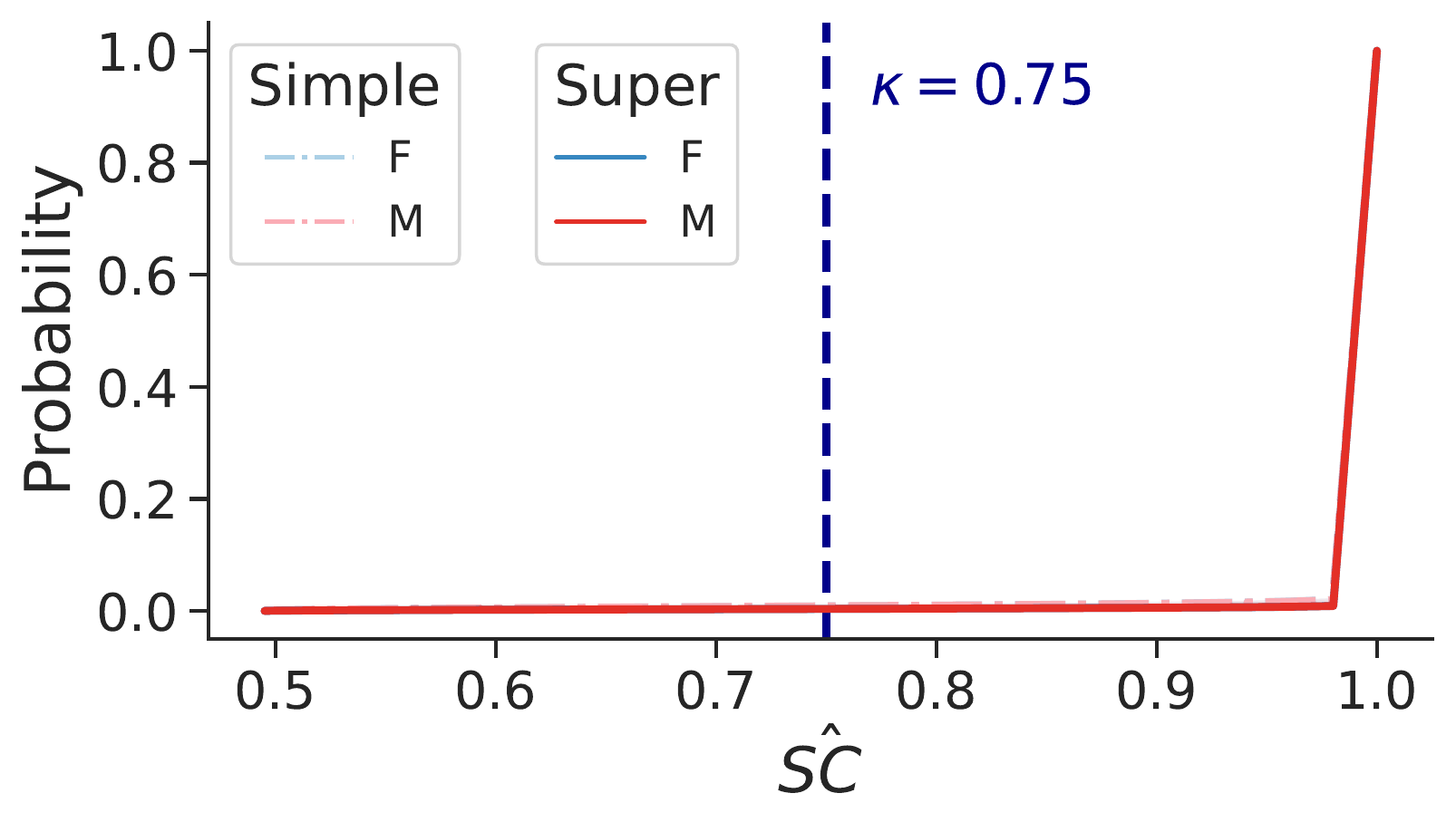}\\
        \vspace{-.1cm}%
	\begin{tabular}{lcc}
		\toprule
		\multicolumn{3}{c}{\textbf{Abstention set metrics}} \\ \cmidrule(lr){1-3}
		 & \textbf{Simple}     & \textbf{Super}      \\ \midrule
		 \textbf{$\Delta\hatar$} & \textcolor{blue}{$0.0\pm0.0\%$} & \textcolor{blue}{$0.0\pm0.0\%$}\\ \midrule
		 \textbf{$\hatar_\text{F}$} & $1.0\pm0.0\%$ & $0.4\pm0.0\%$\\ \midrule
		 \textbf{$\hatar_\text{M}$} & $1.0\pm0.0\%$ & $0.4\pm0.0\%$\\ \bottomrule
	\end{tabular}
\end{minipage}%
\hspace{.25cm}
\begin{minipage}{.495\linewidth}
\centering
\hspace{-.2cm}
	\begin{tabular}{lccc}
		\toprule
		\multicolumn{4}{c}{\textbf{Logistic regression prediction set metrics}} \\ \cmidrule(lr){1-4}
		 & \textbf{Baseline}     & \textbf{Simple}     & \textbf{Super}      \\ \midrule
		 \textbf{$\Delta\hatpr$} & \textcolor{blue}{$5.7\pm0.1\%$} & \textcolor{blue}{$5.6\pm0.0\%$} & \textcolor{blue}{$5.6\pm0.0\%$}\\ \midrule
		 \textbf{$\hatpr_\text{F}$} & $70.8\pm0.2\%$ & $71.1\pm0.2\%$ & $70.9\pm0.2\%$\\ \midrule
		 \textbf{$\hatpr_\text{M}$} & $76.5\pm0.1\%$ & $76.7\pm0.2\%$ & $76.5\pm0.2\%$\\ \midrule
		 \textbf{$\Delta\haterr$} & \textcolor{blue}{$1.1\pm0.2\%$} & \textcolor{blue}{$1.0\pm0.1\%$} & \textcolor{blue}{$1.0\pm0.1\%$}\\ \midrule
		 \textbf{$\haterr_\text{F}$} & $15.7\pm0.2\%$ & $15.3\pm0.0\%$ & $15.5\pm0.0\%$\\ \midrule
		 \textbf{$\haterr_\text{M}$} & $14.6\pm0.0\%$ & $14.3\pm0.1\%$ & $14.5\pm0.1\%$\\ \midrule
		 \textbf{$\Delta\hatfpr$} & \textcolor{blue}{$0.4\pm0.0\%$} & \textcolor{blue}{$0.4\pm0.1\%$} & \textcolor{blue}{$0.4\pm0.0\%$}\\ \midrule
		 \textbf{$\hatfpr_\text{F}$} & $5.8\pm0.1\%$ & $5.6\pm0.1\%$ & $5.7\pm0.0\%$\\ \midrule
		 \textbf{$\hatfpr_\text{M}$} & $6.2\pm0.1\%$ & $6.0\pm0.0\%$ & $6.1\pm0.0\%$\\ \midrule
		 \textbf{$\Delta\hatfnr$} & \textcolor{blue}{$1.4\pm0.1\%$} & \textcolor{blue}{$1.3\pm0.1\%$} & \textcolor{blue}{$1.3\pm0.1\%$}\\ \midrule
		 \textbf{$\hatfnr_\text{F}$} & $9.8\pm0.2\%$ & $9.6\pm0.0\%$ & $9.7\pm0.0\%$\\ \midrule
		 \textbf{$\hatfnr_\text{M}$} & $8.4\pm0.1\%$ & $8.3\pm0.1\%$ & $8.4\pm0.1\%$\\ \bottomrule
	\end{tabular}
\end{minipage}%
\caption{\textbf{Logistic regression} on \texttt{HMDA - 2017 - TX}, by \texttt{sex}\looseness=-1}
\label{fig:hmda-tx-sex-lr-all}
\end{figure*}

\vspace{.5cm}
\setlength{\tabcolsep}{6pt}
\begin{figure*}[h!]
\begin{minipage}{.495\linewidth}
\centering
\hspace{-.4cm}
        \includegraphics[width=\linewidth]{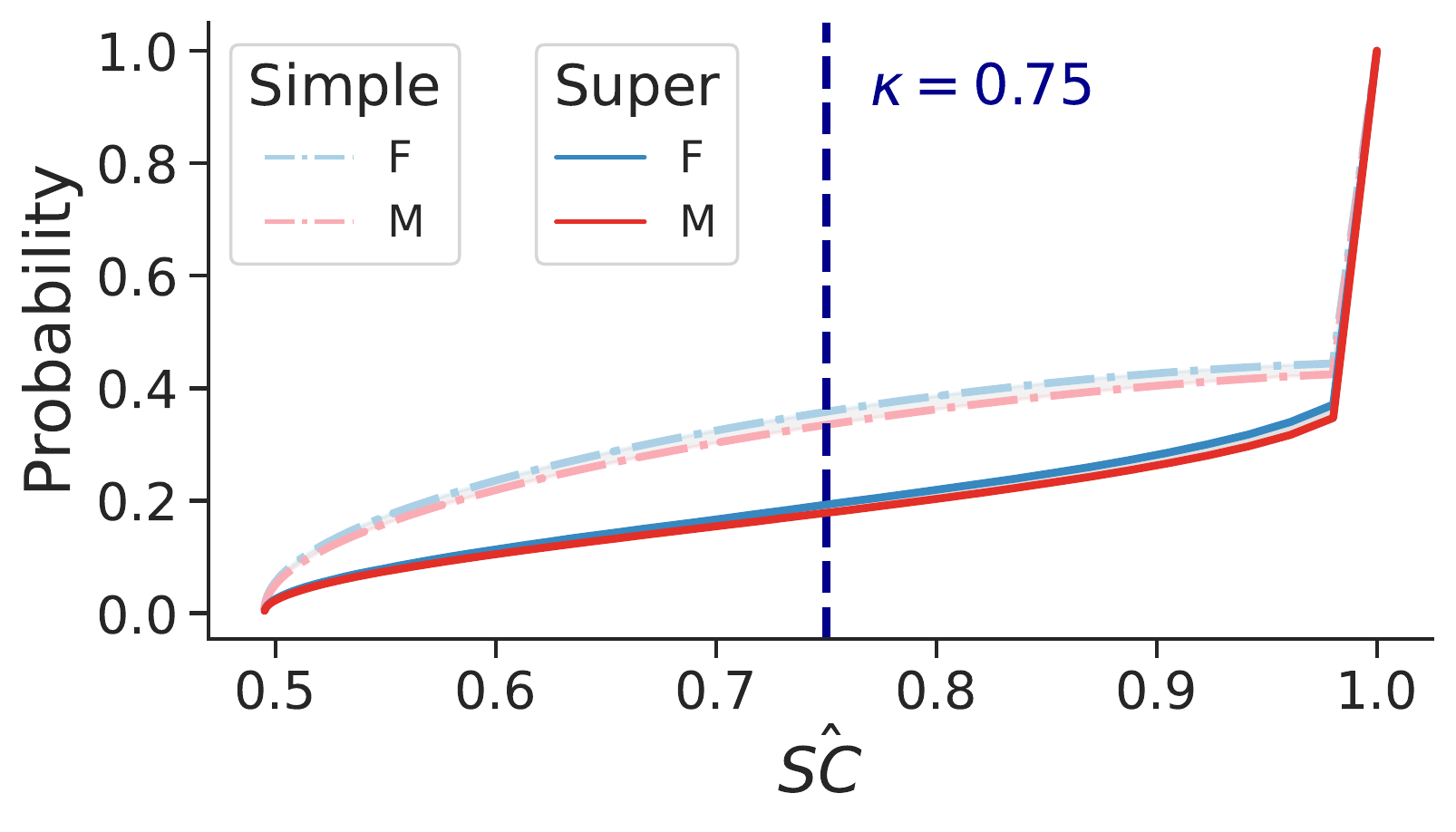}\\
        \vspace{-.1cm}%
	\begin{tabular}{lcc}
		\toprule
		\multicolumn{3}{c}{\textbf{Abstention set metrics}} \\ \cmidrule(lr){1-3}
		 & \textbf{Simple}     & \textbf{Super}      \\ \midrule
		 \textbf{$\Delta\hatar$} & \textcolor{blue}{$2.0\pm0.1\%$} & \textcolor{blue}{$1.4\pm0.0\%$}\\ \midrule
		 \textbf{$\hatar_\text{F}$} & $36.2\pm0.2\%$ & $19.0\pm0.1\%$\\ \midrule
		 \textbf{$\hatar_\text{M}$} & $34.2\pm0.1\%$ & $17.6\pm0.1\%$\\ \bottomrule
	\end{tabular}
\end{minipage}%
\hspace{.25cm}
\begin{minipage}{.495\linewidth}
\centering
\hspace{-.2cm}
	\begin{tabular}{lccc}
		\toprule
		\multicolumn{4}{c}{\textbf{Decision tree prediction set metrics}} \\ \cmidrule(lr){1-4}
		 & \textbf{Baseline}     & \textbf{Simple}     & \textbf{Super}      \\ \midrule
		 \textbf{$\Delta\hatpr$} & \textcolor{blue}{$3.5\pm0.0\%$} & \textcolor{blue}{$2.2\pm0.1\%$} & \textcolor{blue}{$4.8\pm0.0\%$}\\ \midrule
		 \textbf{$\hatpr_\text{F}$} & $75.2\pm0.1\%$ & $91.8\pm0.0\%$ & $79.8\pm0.1\%$\\ \midrule
		 \textbf{$\hatpr_\text{M}$} & $78.7\pm0.1\%$ & $94.0\pm0.1\%$ & $84.6\pm0.1\%$\\ \midrule
		 \textbf{$\Delta\haterr$} & \textcolor{blue}{$1.1\pm0.1\%$} & \textcolor{blue}{$0.2\pm0.0\%$} & \textcolor{blue}{$0.5\pm0.0\%$}\\ \midrule
		 \textbf{$\haterr_\text{F}$} & $17.2\pm0.1\%$ & $2.6\pm0.0\%$ & $7.2\pm0.1\%$\\ \midrule
		 \textbf{$\haterr_\text{M}$} & $16.1\pm0.0\%$ & $2.4\pm0.0\%$ & $6.7\pm0.1\%$\\ \midrule
		 \textbf{$\Delta\hatfpr$} & \textcolor{blue}{$0.8\pm0.1\%$} & \textcolor{blue}{$0.2\pm0.0\%$} & \textcolor{blue}{$0.3\pm0.1\%$}\\ \midrule
		 \textbf{$\hatfpr_\text{F}$} & $8.8\pm0.1\%$ & $1.0\pm0.0\%$ & $2.8\pm0.0\%$\\ \midrule
		 \textbf{$\hatfpr_\text{M}$} & $8.0\pm0.0\%$ & $1.2\pm0.0\%$ & $3.1\pm0.1\%$\\ \midrule
		 \textbf{$\Delta\hatfnr$} & \textcolor{blue}{$0.3\pm0.1\%$} & \textcolor{blue}{$0.4\pm0.0\%$} & \textcolor{blue}{$0.9\pm0.1\%$}\\ \midrule
		 \textbf{$\hatfnr_\text{F}$} & $8.4\pm0.1\%$ & $1.6\pm0.0\%$ & $4.5\pm0.1\%$\\ \midrule
		 \textbf{$\hatfnr_\text{M}$} & $8.1\pm0.0\%$ & $1.2\pm0.0\%$ & $3.6\pm0.0\%$\\ \bottomrule
	\end{tabular}
\end{minipage}%
\caption{\textbf{Decision trees} on \texttt{HMDA - 2017 - TX}, by \texttt{sex}\looseness=-1}
\label{fig:hmda-tx-sex-dtc-all}
\vspace{2cm}
\end{figure*}

\vspace{.5cm}
\setlength{\tabcolsep}{6pt}
\begin{figure*}[h!]
\begin{minipage}{.495\linewidth}
\centering
\hspace{-.4cm}
        \includegraphics[width=\linewidth]{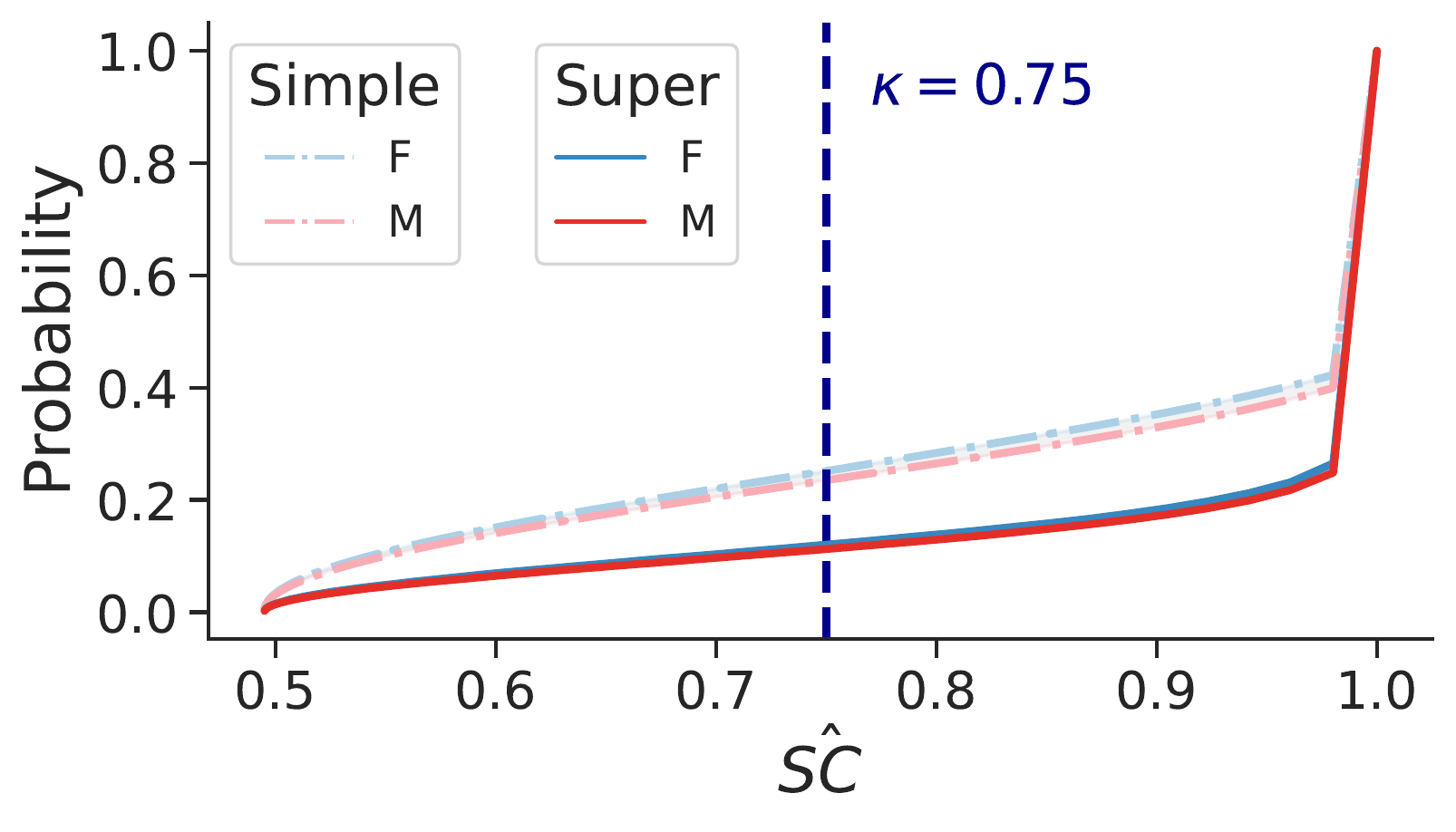}\\
        \vspace{-.1cm}%
	\begin{tabular}{lcc}
		\toprule
		\multicolumn{3}{c}{\textbf{Abstention set metrics}} \\ \cmidrule(lr){1-3}
		 & \textbf{Simple}     & \textbf{Super}      \\ \midrule
		 \textbf{$\Delta\hatar$} & \textcolor{blue}{$1.7\pm0.0\%$} & \textcolor{blue}{$0.8\pm0.2\%$}\\ \midrule
		 \textbf{$\hatar_\text{F}$} & $28.4\pm0.1\%$ & $11.9\pm0.2\%$\\ \midrule
		 \textbf{$\hatar_\text{M}$} & $26.7\pm0.1\%$ & $11.1\pm0.0\%$\\ \bottomrule
	\end{tabular}
\end{minipage}%
\hspace{.25cm}
\begin{minipage}{.495\linewidth}
\centering
\hspace{-.2cm}
	\begin{tabular}{lccc}
		\toprule
		\multicolumn{4}{c}{\textbf{Random forest prediction set metrics}} \\ \cmidrule(lr){1-4}
		 & \textbf{Baseline}     & \textbf{Simple}     & \textbf{Super}      \\ \midrule
		 \textbf{$\Delta\hatpr$} & \textcolor{blue}{$4.4\pm0.0\%$} & \textcolor{blue}{$4.0\pm0.0\%$} & \textcolor{blue}{$5.2\pm0.1\%$}\\ \midrule
		 \textbf{$\hatpr_\text{F}$} & $74.1\pm0.1\%$ & $85.5\pm0.1\%$ & $76.3\pm0.2\%$\\ \midrule
		 \textbf{$\hatpr_\text{M}$} & $78.5\pm0.1\%$ & $89.5\pm0.1\%$ & $81.5\pm0.1\%$\\ \midrule
		 \textbf{$\Delta\haterr$} & \textcolor{blue}{$1.0\pm0.1\%$} & \textcolor{blue}{$0.4\pm0.0\%$} & \textcolor{blue}{$0.7\pm0.1\%$}\\ \midrule
		 \textbf{$\haterr_\text{F}$} & $15.5\pm0.1\%$ & $4.6\pm0.0\%$ & $9.3\pm0.0\%$\\ \midrule
		 \textbf{$\haterr_\text{M}$} & $14.5\pm0.0\%$ & $4.2\pm0.0\%$ & $8.6\pm0.1\%$\\ \midrule
		 \textbf{$\Delta\hatfpr$} & \textcolor{blue}{$0.3\pm0.1\%$} & \textcolor{blue}{$0.3\pm0.0\%$} & \textcolor{blue}{$0.4\pm0.1\%$}\\ \midrule
		 \textbf{$\hatfpr_\text{F}$} & $7.4\pm0.1\%$ & $1.7\pm0.0\%$ & $3.5\pm0.1\%$\\ \midrule
		 \textbf{$\hatfpr_\text{M}$} & $7.1\pm0.0\%$ & $2.0\pm0.0\%$ & $3.9\pm0.0\%$\\ \midrule
		 \textbf{$\Delta\hatfnr$} & \textcolor{blue}{$0.7\pm0.0\%$} & \textcolor{blue}{$0.8\pm0.0\%$} & \textcolor{blue}{$1.1\pm0.1\%$}\\ \midrule
		 \textbf{$\hatfnr_\text{F}$} & $8.1\pm0.1\%$ & $3.0\pm0.0\%$ & $5.8\pm0.1\%$\\ \midrule
		 \textbf{$\hatfnr_\text{M}$} & $7.4\pm0.1\%$ & $2.2\pm0.0\%$ & $4.7\pm0.0\%$\\ \bottomrule
	\end{tabular}
\end{minipage}%
\caption{\textbf{Random forests} on \texttt{HMDA - 2017 - TX}, by \texttt{sex}\looseness=-1}
\label{fig:hmda-tx-sex-rfc-all}
\end{figure*}
\vspace*{2in}
\FloatBarrier
\subsubsection{\appalgodiscussion}\label{app:sec:algo-discussion}

Overall, our results support that examining self-consistency and error together provide a much richer picture of model behavior, both with respect to arbitrariness and fairness metric disparities. Particularly in smaller datasets, the learning process produces models with a large degree of variance. As a result, ensembling with confidence can lead to huge abstention rates.\looseness=-1 

Improving self-consistency by doing a round of variance reduction first and then ensembling with confidence (i.e., super-ensembling) can lead to improvements in error over baselines while having a lower abstention rate. These improvements are typically shared across subgroups, but may not be symmetric; some subgroups may benefit more than others. As a result, even though accuracy increases absolutely for both groups, relative fairness metrics can decrease. This is a different instantiation of the fairness-accuracy trade-off than is often written about, which posits a necessary decrease in accuracy for one subgroup to improve fairness between binarized groups. Our results suggest that it is worth first tuning for accuracy, and then seeing how fairness interventions can balance the benefits across subgroups. Of course, it is possible that doing this could lead to injecting variance back into the model outputs, thereby reducing self-consistency and inducing arbitrariness. We leave this investigation to future work. 



Additionally, our results reify that choice of model matters a lot. While overall error rates across model types may be similar, the sources of that error are not necessarily the same. This is an obvious point, relating to bias and variance. 
However, a lot of fair classification work describes similar performance across logistic regression, decision trees, random forests, SVMs, and MLPs (e.g., \citet{chen2018tradeoff}). Looking at self-consistency confirms that this is not the case, with decision trees and random forests in particular exhibiting higher variance, and thus being more amenable to variance reduction and improvements in overall error. As fair classification research transitions to larger benchmarks, it will likely be fruitful to investigate more complex model classes. 



We provide run times on our cluster environment in Table~\ref{app:table:summary-algo-runtime}. We did not select for CPUs with any particular features, and thus the run times are quite variable. 

\setlength{\tabcolsep}{4pt}

\begin{table}[b]
\vspace{-.5cm}
\centering
\caption{These times are recorded for our cluster environment (hh:mm:ss), described in Appendix~\ref{app:sec:cluster} for our Algorithm~\ref{algo:bagging-confidently} experiments. At the time of running, due to time constraints, the authors had not yet parallelized this part of the code.}
\label{app:table:summary-algo-runtime}
\footnotesize
\begin{tabular}{lllll}
\toprule
 \textbf{Dataset} & \textbf{$g$} & \textbf{Logistic regression} & \textbf{Decision trees} & \textbf{Random forests}  \\ \midrule
\textbf{\texttt{South German Credit}}   & $\texttt{sex}$  & 00:42:50 &  00:25:28 &  00:34:42 \\ \midrule
\textbf{\texttt{COMPAS}}   & $\texttt{race}$ & 00:57:05 & 00:39:24 & 00:31:47 \\ \midrule
\textbf{\texttt{Old Adult}}  & $\texttt{sex}$ & 01:08:37  &  01:23:39  &  00:57:11  \\ \midrule
\textbf{\texttt{Taiwan Credit}}   & $\texttt{sex}$  & 00:31:35 & 01:34:57 & 01:53:33 \\ \midrule
\textbf{\texttt{New Adult - CA}}  & &  &  &    \\
 \;\;\;\textbf{\texttt{Income}} & $\texttt{sex}, \texttt{race}$ & 01:39:53 & 02:51:13 &  04:59:07 \\ \cmidrule(lr){2-5}
 \;\;\;\textbf{\texttt{Employment}} & $\texttt{sex}, \texttt{race}$ & 02:20:15 & 02:18:16 &  03:00:15 \\ \cmidrule(lr){2-5}
\;\;\;\textbf{\texttt{Public Coverage}} & $\texttt{sex}, \texttt{race}$ & 01:13:33 & 02:02:57  & 02:24:08   \\ \midrule
\textbf{\texttt{HMDA - 2017}}  & &  &  &  \\
 \;\;\;\textbf{\texttt{NY}} & $\texttt{sex, race, ethnicity}$ & 03:50:52  & 05:00:19 & 05:39:44  \\ \cmidrule {2-5}
 \;\;\;\textbf{\texttt{TX}} & $\texttt{sex, race, ethnicity}$ & 05:18:59  & 04:10:34 & 04:18:59    \\\bottomrule
\end{tabular}
\vspace{-.5cm}
\end{table}

We also provide details on systematic arbitrariness in Tables~\ref{app:table:wass} and~\ref{app:table:wass-kappa}, which we measure using the $\hat{\mathcal{W}_1}$. As noted in Appendix~\ref{app:sec:consistency}, since this metric is an average, its measure necessarily changes if we compute it over a different set of levels $\hat{\sK}$. To make distances across interventions comparable, we treat CDF values below $\kappa$ as $0$, so that all of the probability mass is on $\hatsc\geq\kappa$. We therefore report two versions of these results, those for no abstention and those that account for abstention at values $<\kappa$. $\Delta$ is the difference between $\hat{\mathcal{W}_1}_\text{Simple} - \hat{\mathcal{W}_1}_\text{Super}$. Positive differences indicate cases for which the super-ensembling method decreases the Wasserstein-1 distance between subgroup CDFs; negative differences indicate increases. While in some cases there is an increase, it is worth noting that this aligns with cases for which the $\hat{\mathcal{W}_1}$ distance is very close to 0. 
\texttt{Old Adult}, highlighted below, is the only dataset that exhibts large amounts of systematic arbitrariness (for decision tress and random forests, in particular; it exhibits the highest amount for logistic regression, but is overall low). \texttt{Old Adult} and \texttt{New Adult - Employment} (by \texttt{sex}) are two of the only tasks that show any fairness disparities that are $>3\%$.

\begin{figure}

\begin{table}[H]
\centering
\caption{Empirical Wasserstein-1 ($\hat{\mathcal{W}_1}$) measurements without abstention}\vspace{.25cm}
\label{app:table:wass}
\footnotesize
\begin{tabular}{lllllllllll}
\toprule
 \textbf{Dataset} & \textbf{$g$} & \multicolumn{3}{c}{\textbf{Logistic regression}} & \multicolumn{3}{c}{\textbf{Decision trees}} & \multicolumn{3}{c}{\textbf{Random forests}}  \\ \cmidrule{3-11}
 & & \textbf{Simple} & \textbf{Super} & $\Delta$ & \textbf{Simple} & \textbf{Super} & $\Delta$ & \textbf{Simple} & \textbf{Super} & $\Delta$\\ \midrule 
\textbf{\texttt{German Credit}}   & $\texttt{sex}$  & 0.0181 &  0.0101 &  0.0079 & 0.0162 & 0.0244 & -0.0082 & 0.0181 & 0.0175 & 0.0006 \\ \midrule
\textbf{\texttt{COMPAS}}   & $\texttt{race}$ & 0.0073 & 0.0043  & 0.0030 & 0.0189 & 0.0170 & 0.0019 & 0.0073 & 0.0031 & 0.0043  \\ \midrule
\textbf{\texttt{Old Adult}}  & $\texttt{sex}$ & 0.0206 & 0.0033 & 0.0173 & 0.1386 & 0.0273 & 0.1112 & 0.1266 & 0.0255 & 0.1011 \\ \midrule
\textbf{\texttt{Taiwan Credit}} & $\texttt{sex}$  & 0.0028 & 0.0007 & 0.0020 & 0.0223 & 0.0108 & 0.0115 & 0.0240 & 0.0065 & 0.0175 \\ \midrule
\textbf{\texttt{New Adult - CA}}  & &  &  &    \\
 \;\;\;\textbf{\texttt{Income}} & $\texttt{sex}$ & 0.0009 & 0.0003 & 0.0006 & 0.0138 & 0.0055 & 0.0083 & 0.0089 & 0.0018 & 0.0071
 \\ \cmidrule(lr){2-11}
& $\texttt{race}$ & 0.0003 & 0.0001 & 0.0002 & 0.0170 & 0.0073 & 0.0096 & 0.0163 & 0.0055 & 0.0108 \\ \midrule
 \;\;\;\textbf{\texttt{Employment}} & $\texttt{sex}$ & 0.0004 & 0.0002 & 0.0003 & 0.0010 & 0.0011 & -0.0001 & 0.0043 & 0.0031 & 0.0013 \\ \cmidrule(lr){2-11}
& $\texttt{race}$ & 0.0004 & 0.0001 & 0.0004 & 0.0020 & 0.0007 & 0.0013 & 0.0021 & 0.0015 & 0.0006   \\ \midrule
\;\;\;\textbf{\texttt{Public Coverage}} & $\texttt{sex}$ & 0.0004 & 0.0001 & 0.0003  & 0.0029 & 0.0024 & 0.0005 & 0.0045 & 0.0023 & 0.0024 \\ \cmidrule{2-11}
 & $\texttt{race}$ & 0.0010 & 0.0003 & 0.0007 & 0.0200 & 0.0089 & 0.0113 & 0.0235 & 0.0089 & 0.0147 \\ \midrule
\textbf{\texttt{HMDA - 2017}}  & &  &  &  \\
 \;\;\;\textbf{\texttt{NY}} & $\texttt{sex}$ & 0.0002 & 0.0004 & -0.0002 & 0.0096 & 0.0039 & 0.0056 & 0.0080 & 0.0023 & 0.0056 \\ \cmidrule {2-11}
   & $\texttt{race}$ & 0.0009 & 0.0005 & 0.0005 & 0.0433 & 0.0203 & 0.0231 & 0.0409 & 0.0133 & 0.0276 \\ \cmidrule {2-11}
   & $\texttt{ethnicity}$ & 0.0005 & 0.0005 & 0.0000 &  0.0229 & 0.0156 & 0.0073 & 0.0248 & 0.0108 & 0.0140 \\ \midrule
 \;\;\;\textbf{\texttt{TX}} & $\texttt{sex}$ & 0.0001 & 0.0001 & 0.0000 & 0.0153 & 0.0097 & 0.0055 & 0.0113 & 0.0054 & 0.0058\\ \cmidrule{2-11}
 & $\texttt{race}$ & 0.0001 & 0.0001 & 0.0000 & 0.0010 & 0.0012 & -0.0002 & 0.0013 & 0.0007 & 0.0006 \\ \cmidrule {2-11}
   & $\texttt{ethnicity}$ & 0.0007 & 0.0002 & 0.0004 & 0.0509 & 0.0291 & 0.0219 & 0.0379 & 0.0190 & 0.0188 \\ \bottomrule
\end{tabular}
\end{table}

\vspace*{-.5cm}
\begin{table}[H]
\centering
\caption{Empirical Wasserstein-1 ($\hat{\mathcal{W}_1}$) measurements with abstention using $\kappa \geq .75$}\vspace{.25cm}
\label{app:table:wass-kappa}
\footnotesize
\begin{tabular}{lllllllllll}
\toprule
 \textbf{Dataset} & \textbf{$g$} & \multicolumn{3}{c}{\textbf{Logistic regression}} & \multicolumn{3}{c}{\textbf{Decision trees}} & \multicolumn{3}{c}{\textbf{Random forests}}  \\ \cmidrule{3-11}
 & & \textbf{Simple} & \textbf{Super} & $\Delta$ & \textbf{Simple} & \textbf{Super} & $\Delta$ & \textbf{Simple} & \textbf{Super} & $\Delta$\\ \midrule 
\textbf{\texttt{German Credit}}   & $\texttt{sex}$  & 0.0113 & 0.0080 & 0.0034 & 0.0090 & 0.0094 & -0.0004 & 0.0084 & 0.0132 & -0.0048 \\ \midrule
\textbf{\texttt{COMPAS}}   & $\texttt{race}$ & 0.0035 & 0.0019 & 0.0017 & 0.0039 & 0.0060 & -0.0021 & 0.0041 & 0.0019 & 0.0022 \\ \midrule
\textbf{\texttt{Old Adult}}  & $\texttt{sex}$ & 0.0110 & 0.0020 & 0.0090 & 0.0654 & 0.0155 & 0.0500 & 0.0634 & 0.0139 & 0.0494 \\ \midrule
\textbf{\texttt{Taiwan Credit}}   & $\texttt{sex}$  & 0.0014 & 0.0005 & 0.0009 & 0.0057 & 0.0059 & -0.0002 & 0.0107 & 0.0040 & 0.0067 \\ \midrule
\textbf{\texttt{New Adult - CA}}  & &  &  &    \\
 \;\;\;\textbf{\texttt{Income}} & $\texttt{sex}$ & 0.0005 & 0.0002 & 0.0004 & 0.0051 & 0.0032 & 0.0019 & 0.0047 & 0.0012 & 0.0035
 \\ \cmidrule(lr){2-11}
& $\texttt{race}$ & 0.0002 & 0.0000 & 0.0002 & 0.0073 & 0.0040 & 0.0033 & 0.0082 & 0.0028 & 0.0053 \\ \midrule
 \;\;\;\textbf{\texttt{Employment}} & $\texttt{sex}$ & 0.0002 & 0.0001 & 0.0001 & 0.0005 & 0.0005 & 0.0001 & 0.0020 & 0.0014 & 0.0006  \\ \cmidrule(lr){2-11}
& $\texttt{race}$ & 0.0002 & 0.0000 & 0.0002 & 0.0012 & 0.0003 & 0.0008 & 0.0008 & 0.0005 & 0.0003 \\ \midrule
\;\;\;\textbf{\texttt{Public Coverage}} & $\texttt{sex}$ & 0.0002 & 0.0001 & 0.0001 & 0.0006 & 0.0012 & -0.0006 & 0.0011 & 0.0009 & 0.0002 \\ \cmidrule{2-11}
 & $\texttt{race}$  & 0.0006 & 0.0001 & 0.0005 & 0.0068 & 0.0049 & 0.0019 & 0.0106 & 0.0047 & 0.0059  \\ \midrule
\textbf{\texttt{HMDA - 2017}}  & &  &  &  \\
 \;\;\;\textbf{\texttt{NY}} & $\texttt{sex}$ & 0.0001 & 0.0002 & -0.0001 & 0.0033 & 0.0020 & 0.0012 & 0.0040 & 0.0013 & 0.0028 \\ \cmidrule {2-11}
   & $\texttt{race}$ & 0.0004 & 0.0002 & 0.0002 & 0.0155 & 0.0111 & 0.0044 & 0.0190 & 0.0076 & 0.0114 \\ \cmidrule {2-11}
   & $\texttt{ethnicity}$ & 0.0002 & 0.0002 & 0.0001 & 0.0055 & 0.0083 & -0.0028 & 0.0081 & 0.0059 & 0.0022 \\ \midrule
 \;\;\;\textbf{\texttt{TX}} & $\texttt{sex}$ & 0.0000 & 0.0000 & 0.0000 & 0.0061 & 0.0050 & 0.0011 & 0.0058 & 0.0029 & 0.0028  \\ \cmidrule{2-11}
 & $\texttt{race}$ & 0.0000 & 0.0001 & 0.0000 & 0.0004 & 0.0005 & -0.0002 & 0.0007 & 0.0004 & 0.0003 \\ \cmidrule {2-11}
   & $\texttt{ethnicity}$ & 0.0003 & 0.0001 & 0.0003 & 0.0229 & 0.0159 & 0.0070 & 0.01200 & 0.0104 & 0.0095 \\ \bottomrule
\end{tabular}
\end{table}
\end{figure}
\FloatBarrier

\subsection{\appfair}\label{app:sec:fair}

Even before we apply our intervention to improve self-consistency, our results in Section~\ref{sec:significance} show close-to-parity $\haterr$, $\hatfpr$, and $\hatfnr$ across subgroups in \texttt{COMPAS} (and similarly for \texttt{South German Credit}, below). These results are surprising. We run $\boot=101$ models to produce estimates of variance and self-consistency, but of course doing this also has the effect of estimating the expected error more generally (with variance representing a portion of that error). Our estimates of expected error for these tasks indicate that the average model produced training on \texttt{COMPAS} and \texttt{South German Credit}, with respect to popular fairness definitions like Equality of Opportunity and Equalized Odds~\cite{barocas2019textbook,hardt2016eo} \emph{are in fact baseline close to parity, with \textbf{no} fairness intervention applied}. We found this across model types for both datasets, though the story becomes more complicated when we apply techniques to improve self-consistency (see discussion at the end of Appendix~\ref{app:sec:experiments-algo}). 

Of course, we did not expect this result, as these are two of the \textit{de facto} standard benchmark datasets in algorithmic fairness. They are used in countless other studies to probe and verify algorithmic fairness interventions~\cite{fabris2022datasets}. As a result, we initially thought that our results must be incorrect. We therefore looked at the underlying models in our bootstrap runs to see the error of the underlying models.

\begin{figure*}[!h]
    \centering
    \begin{minipage}{0.33\textwidth}
        \centering
        \includegraphics[width=0.95\linewidth]{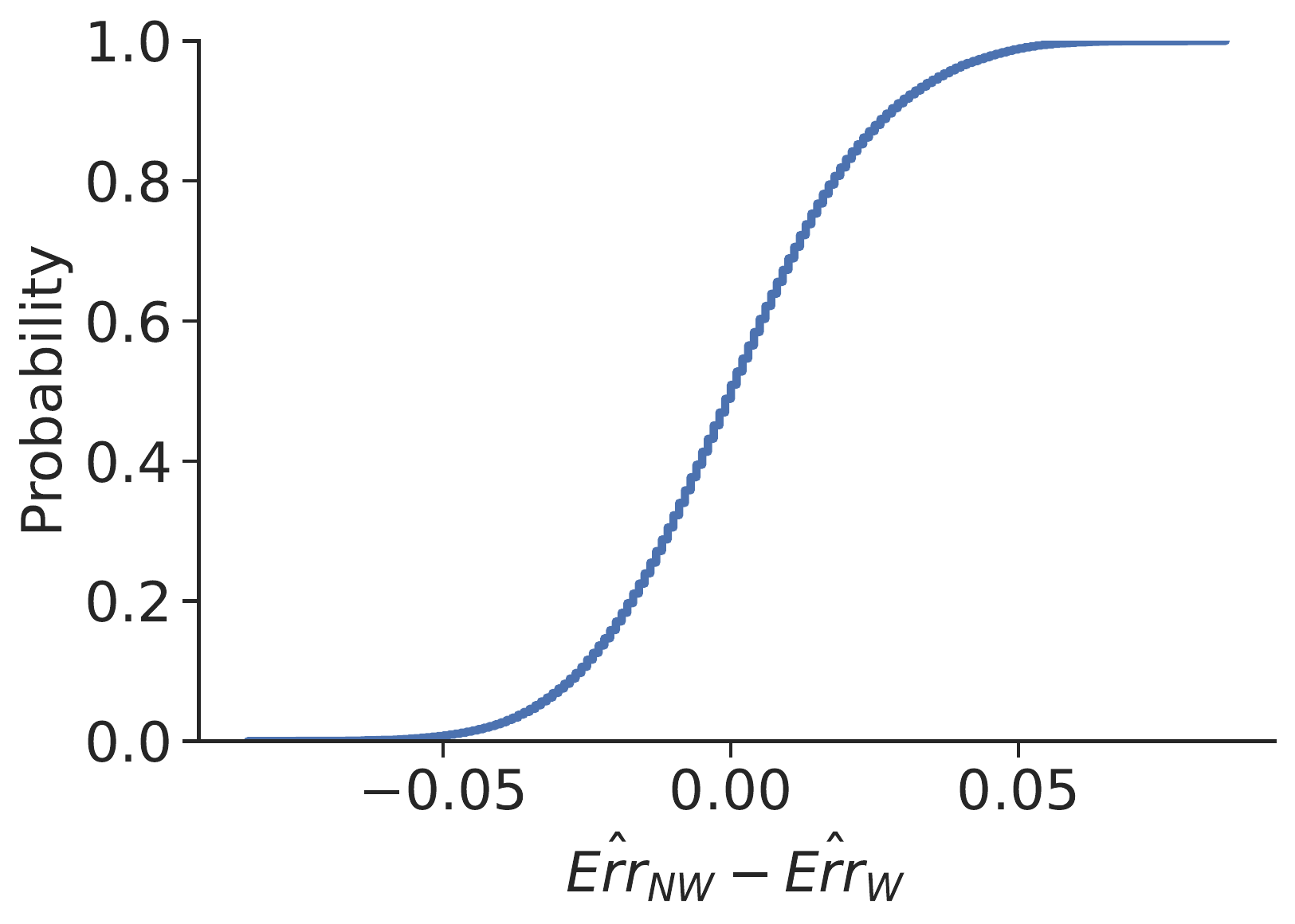}
        \subcaption{$\haterr$ disparity}
    \end{minipage}%
    \begin{minipage}{0.33\textwidth}
        \centering
        \includegraphics[width=0.95\linewidth]{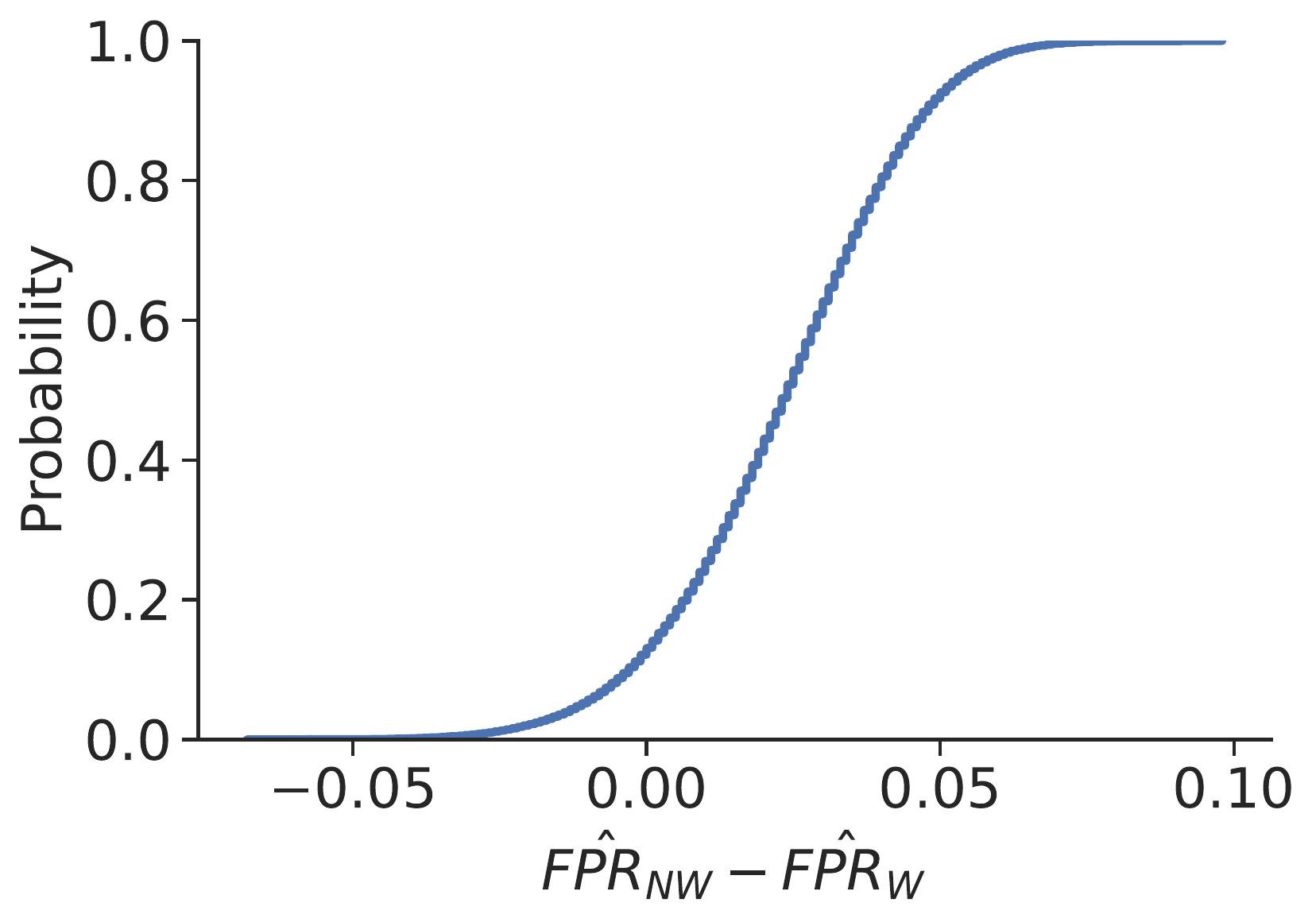}
        \subcaption{$\hatfpr$ disparity}
    \end{minipage}%
    \begin{minipage}{0.33\textwidth}
        \centering
        \includegraphics[width=0.95\linewidth]{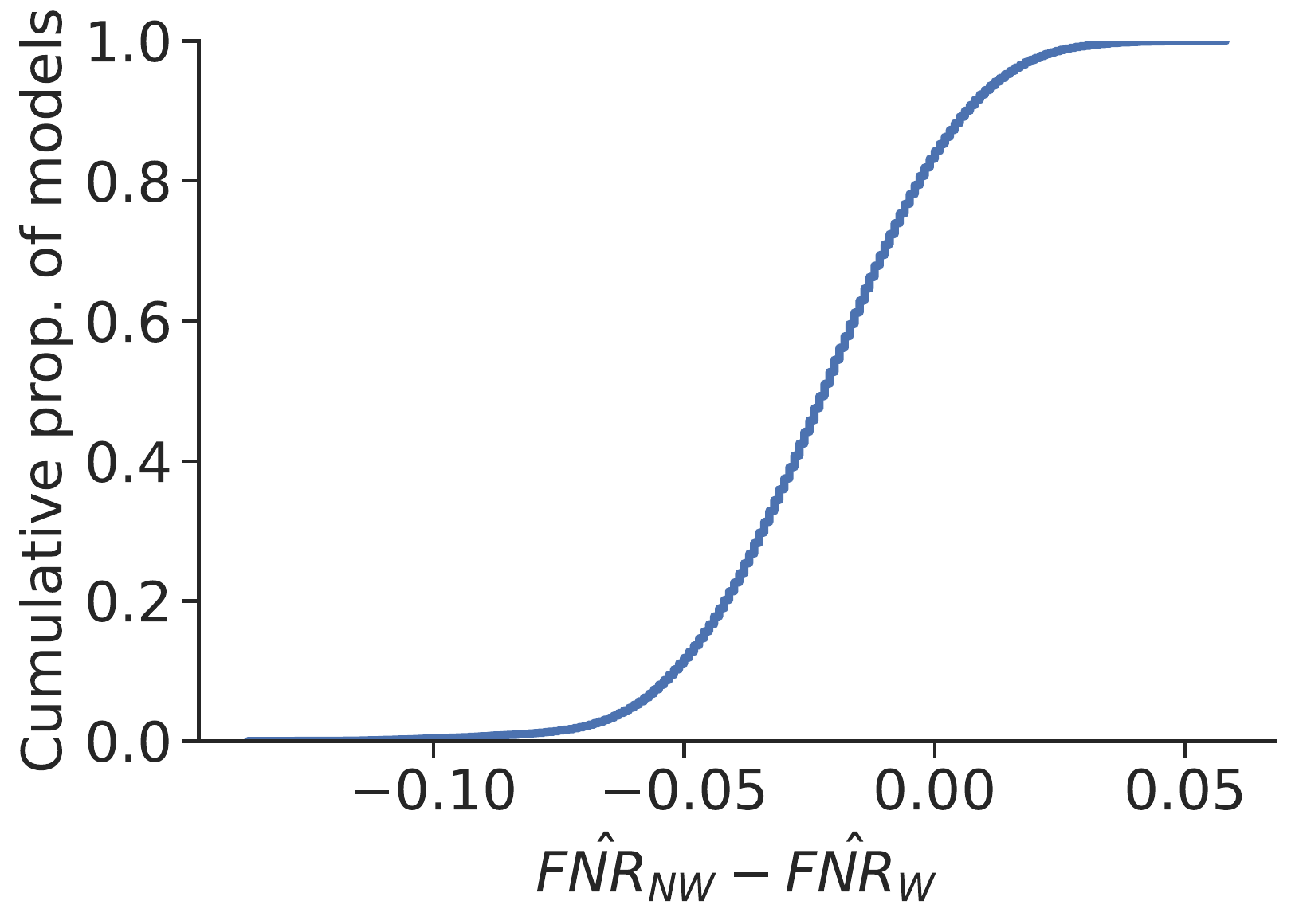}
        \subcaption{$\hatfnr $ disparity}
    \end{minipage}
    \caption{Cumulative distribution of error disparity across $100,100$ logistic regression models trained on \texttt{COMPAS}.}
    \label{app:fig:compas-disparity-ecdf}
\end{figure*}

We re-ran our baseline experiments with $\boot=1001$ and for $100$ test/train splits for logistic regression. In Figure~\ref{app:fig:compas-disparity-ecdf}, we plot the ($100,100$) bootstrap models that went into these results. For another view on analogous information, in Table~\ref{app:table:compas-model-runs-diffs}, we provide an excerpt of the results for \texttt{COMPAS} regarding the underlying $1010$ random forest classifiers used to produce Figure~\ref{subfig:compas-cdf-rfc}.\looseness=-1

Overall, we can see that there is a wide range of error disparities that trend in both directions, with a skew toward higher $\hatfpr$ for $\group=\text{NW}$. These results support our claim that training many models is necessary to get an accurate picture of expected error, with implications both for reproducibility of experiments that just train and analyze a small handful of models and for generalizability. There are models that exhibit worse degrees of unfairness in both directions, but they are more unlikely than models that exhibit smaller disparities.

We subset the above results to the $100$ models that produce the lowest $\haterr$, as this is often the selection critera for picking models to post-process. We plot these results below. These top-performing models in fact exhibit (on average) closer-to-parity for $\hatfpr$ and $\hatfnr$. 

\begin{figure*}[!h]
    \centering
    \begin{minipage}{0.33\textwidth}
        \centering
        \includegraphics[width=0.95\linewidth]{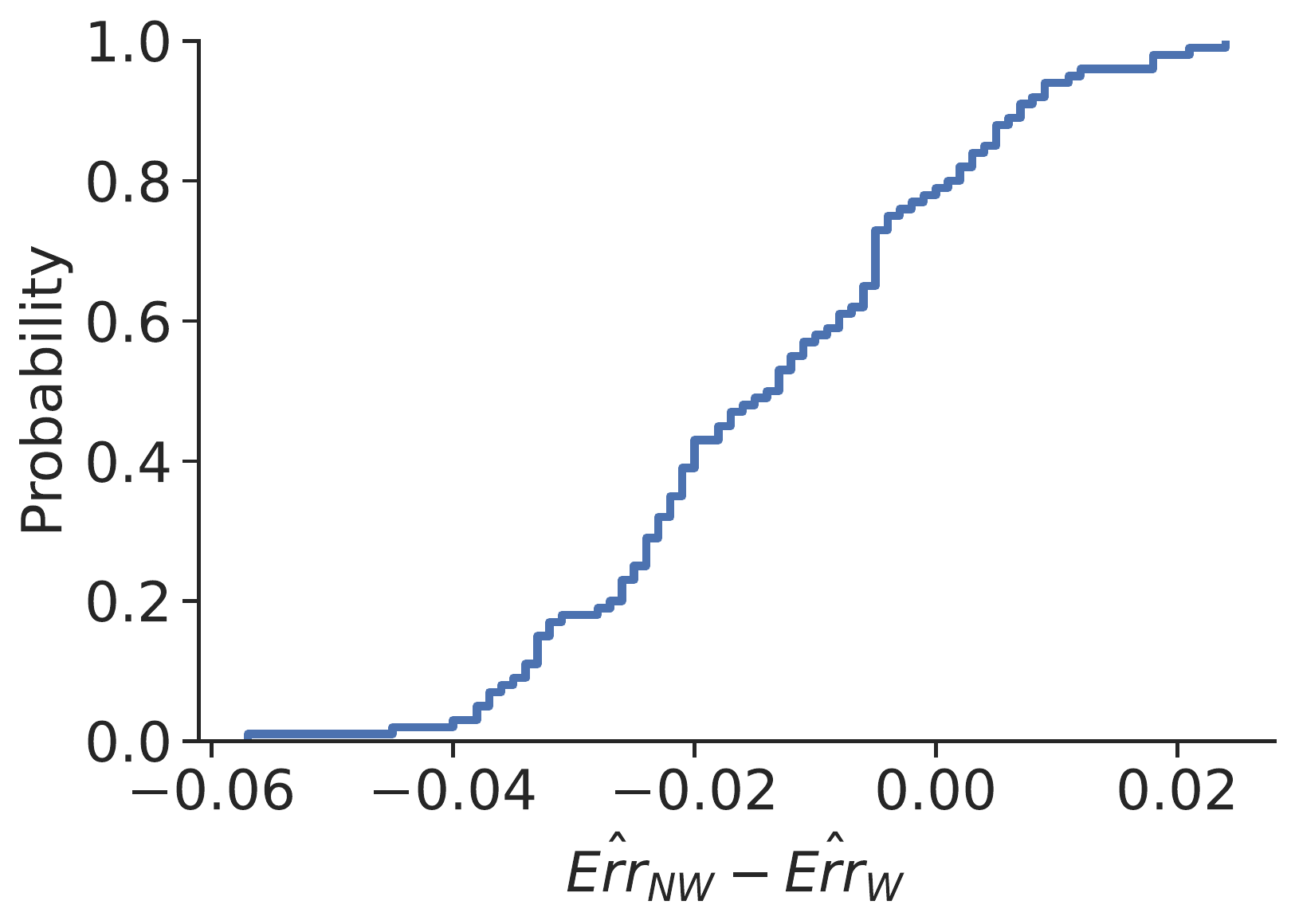}
        \subcaption{$\haterr$ disparity}
    \end{minipage}%
    \begin{minipage}{0.33\textwidth}
        \centering
        \includegraphics[width=0.95\linewidth]{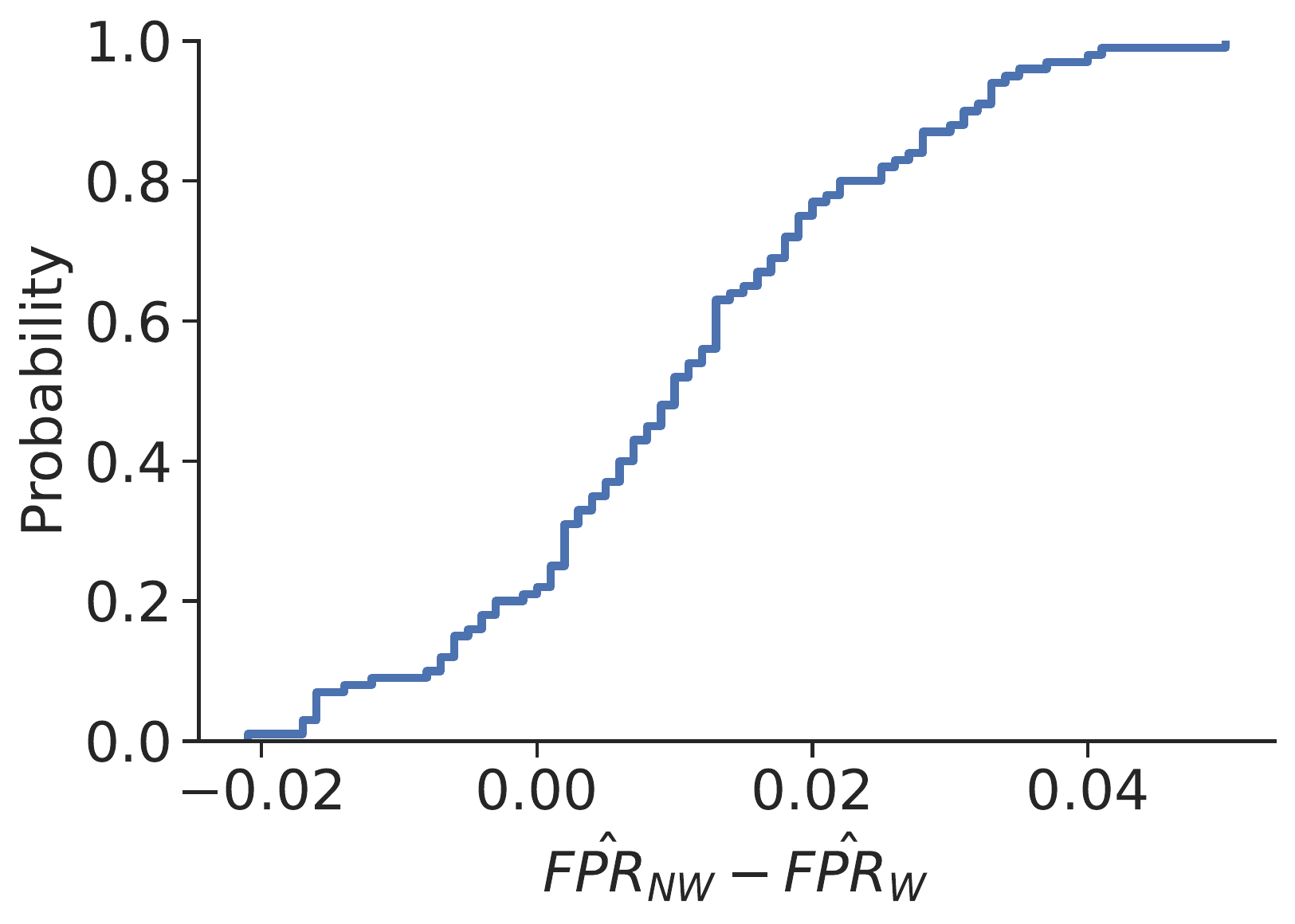}
        \subcaption{$\hatfpr$ disparity}
    \end{minipage}%
    \begin{minipage}{0.33\textwidth}
        \centering
        \includegraphics[width=0.95\linewidth]{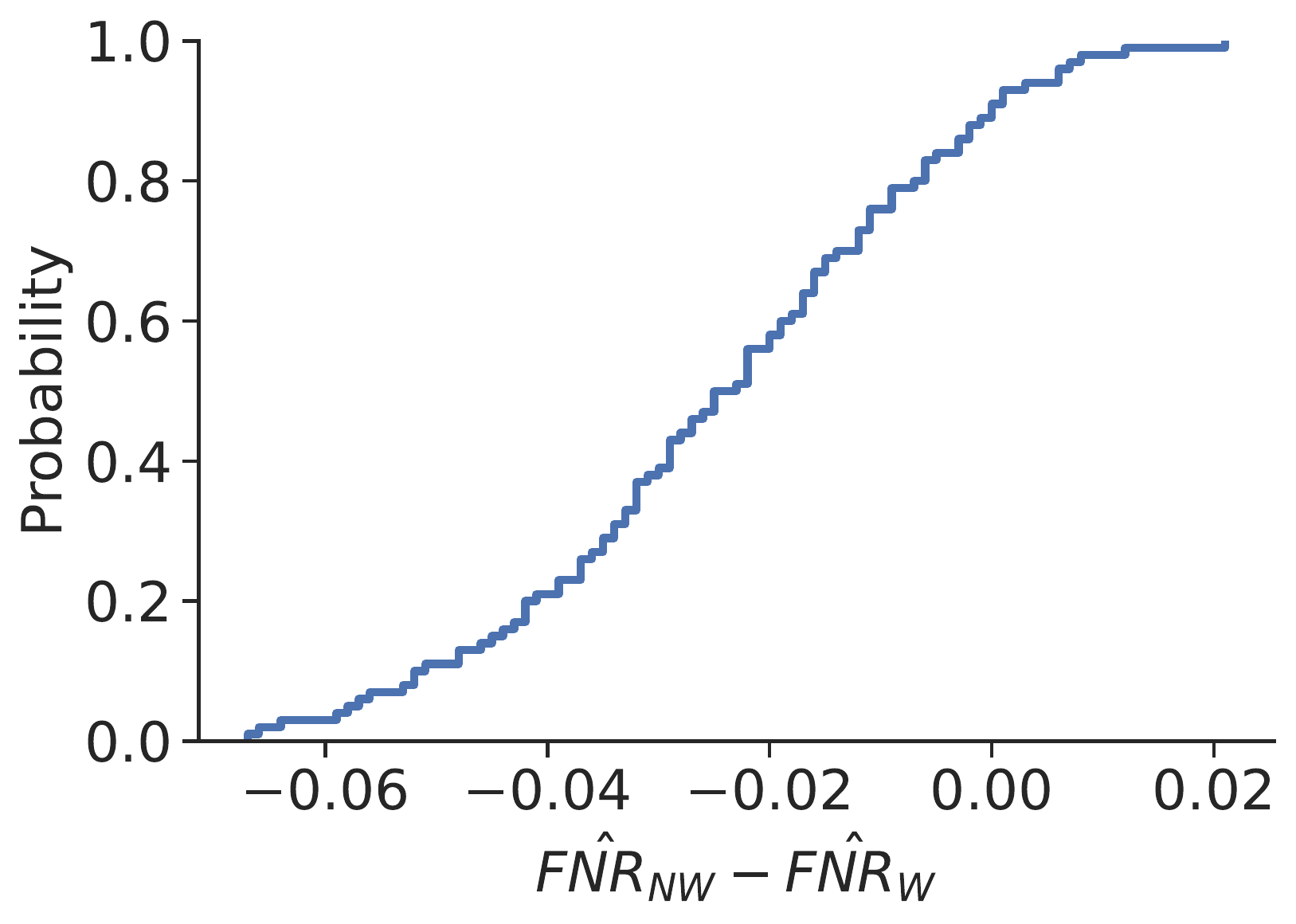}
        \subcaption{$\hatfnr $ disparity}
    \end{minipage}
    \caption{CDF of error disparity across the top $100$ logistic regression models (of the $100,100$ models) trained on \texttt{COMPAS}.}
    \label{app:fig:compas-disparity-ecdf-100}
\end{figure*}

\begin{table}
\centering
\caption{Comparing subgroup error rates in \texttt{COMPAS} for different random forest classifiers trained to produce Figure~\ref{subfig:compas-cdf-rfc}. Each table looks at the top-$3$ highest differences between subgroups for the specified metric: (a) $\haterr_\text{NW} - \haterr_\text{W}$, when $\haterr_\text{NW} > \haterr_\text{W}$; (b) $\haterr_\text{W} - \haterr_\text{NW}$, when $\haterr_\text{W} > \haterr_\text{NW}$; (c) $\hatfpr_\text{NW} - \hatfpr_\text{W}$, when $\hatfpr_\text{NW} > \hatfpr_\text{W}$; (d) $\hatfpr_\text{W} - \hatfpr_\text{NW}$, when $\hatfpr_\text{W} > \hatfpr_\text{NW}$; (e) $\hatfnr_\text{NW} - \hatfnr_\text{W}$, when $\hatfnr_\text{NW} > \hatfnr_\text{W}$; and, (f) (e) $\hatfnr_\text{W} - \hatfnr_\text{NW}$, when $\hatfnr_\text{W} > \hatfnr_\text{NW}$. We highlight the overall error metric in \textcolor{black!30}{gray}, the larger metric (being subtracted from) in \textcolor{blue!50}{blue}, the smaller metric (being subtracted) in \textcolor{red!50}{red}, and the difference in the metric between subgroups in \textcolor{purple!50}{purple}. Note that run 757 appears twice, which we mark in \textcolor{orange!50}{orange}.\looseness=-1}
  \begin{subtable}[h]{\linewidth}
    \centering
    \subcaption{The top-$3$ most unfair models by subgroup-specific $\haterr$, when  $\haterr_\text{NW} > \haterr_\text{W}$ (i.e., unfair toward NW).\looseness=-1}
    \begin{tabular}{lllcccccccccc}
      \toprule
      {Run \#} & $s$ & $b$ & $\haterr$ & $\hatfpr$ & $\hatfnr$ & $\haterr_\text{NW}$ & $\hatfpr_\text{NW}$  & $\hatfnr_\text{NW}$ & $\haterr_\text{W}$  & $\hatfpr_\text{W}$ & $\hatfnr_\text{W}$ & $\haterr_\text{NW} - \haterr_\text{W}$ \\
      \midrule
      762 & $8$ & $504$ & \cellcolor{black!20}$0.374$ &	$0.179$ &	$0.196$	& \cellcolor{blue!33}$0.405$	& $0.204$ &	$0.201$	& \cellcolor{red!33}$0.315$	& $0.13$ &	$0.186$ &	\cellcolor{purple!33}$0.09$\\
      \cellcolor{orange!33}757	& $8$ &	$464$ &	\cellcolor{black!20}$0.369$ &	$0.167$	& $0.202$ &	\cellcolor{blue!33}$0.395$	& $0.201$ &	$0.193$ &	\cellcolor{red!33}$0.318$ &	$0.101$ &	$0.218$ &	\cellcolor{purple!33}$0.077$\\
      328 &	$4$ &	$116$ &	\cellcolor{black!20}$0.371$ &	$0.165$ &	$0.206$ &	\cellcolor{blue!33}$0.395$ &	$0.181$ &	$0.214$ &	\cellcolor{red!33}$0.323$ &	$0.134$ &	$0.189$ &	\cellcolor{purple!33}$0.072$\\
      \bottomrule
    \end{tabular}
    \label{app:subtable:errNW-minus-errW}
    \end{subtable}\\

    \vspace{0.5cm}
    \begin{subtable}[h]{\linewidth}
    \centering
    \subcaption{The top-$3$ most unfair models by subgroup-specific $\haterr$, when  $\haterr_\text{W} > \haterr_\text{NW}$ (i.e., unfair toward W).\looseness=-1}
    \begin{tabular}{lllcccccccccc}
      \toprule
      {Run \#} & $s$ & $b$ & $\haterr$ & $\hatfpr$ & $\hatfnr$ & $\haterr_\text{NW}$ & $\hatfpr_\text{NW}$  & $\hatfnr_\text{NW}$ & $\haterr_\text{W}$  & $\hatfpr_\text{W}$ & $\hatfnr_\text{W}$ & $\haterr_\text{W} - \haterr_\text{NW}$ \\
      \midrule
      414	& $5$ &	$75$ &	\cellcolor{black!20}$0.376$ &	$0.167$ & $0.209$ &	\cellcolor{red!33}$0.352$ &	$0.158$ &	$0.194$ &	\cellcolor{blue!33}$0.422$ &	$0.186$ &	$0.236$ &	\cellcolor{purple!33}$0.07$ \\
      435	& $5$ &	$180$ &	\cellcolor{black!20}$0.376$ &	$0.199$ & $0.177$ & \cellcolor{red!33}$0.355$ &	$0.189$ &	$0.166$ &	\cellcolor{blue!33}$0.416$	& $0.217$ &	$0.198$	& \cellcolor{purple!33}$0.061$\\
      413	& $5$ &	$70$ &	\cellcolor{black!20}$0.378$ &	$0.189$ &	$0.189$ &	\cellcolor{red!33}$0.359$ &	$0.188$ &	$0.171$ &	\cellcolor{blue!33}$0.413$ &	$0.191$ &	$0.222$ & \cellcolor{purple!33}$0.054$\\
      \bottomrule
    \end{tabular}
    \label{app:subtable:errW-minus-errNW}
    \end{subtable}

    \vspace{0.5cm}
    \begin{subtable}[h]{\linewidth}
    \centering
    \subcaption{The top-$3$ most unfair models by subgroup-specific $\hatfpr$, when  $\hatfpr_\text{NW} > \hatfpr_\text{W}$ (i.e., unfair toward NW).}
    \begin{tabular}{lllcccccccccc}
      \toprule
      {Run \#} & $s$ & $b$ & $\haterr$ & $\hatfpr$ & $\hatfnr$ & $\haterr_\text{NW}$ & $\hatfpr_\text{NW}$  & $\hatfnr_\text{NW}$ & $\haterr_\text{W}$  & $\hatfpr_\text{W}$ & $\hatfnr_\text{W}$ & $\hatfpr_\text{NW} - \hatfpr_\text{W}$ \\
      \midrule
      \cellcolor{orange!33}757	 & $8$ &	$464$ &	$0.369$ &	\cellcolor{black!20} $0.167$ &	$0.202$ &	$0.395$ &	\cellcolor{blue!33}$0.201$ &	$0.193$ &	$0.318$ &	\cellcolor{red!33}$0.101$ &	$0.218$ &	\cellcolor{purple!33}$0.1$ \\
      729	& $8$ &	$240$	& $0.358$ &	\cellcolor{black!20}$0.162$ & 	$0.197$ &	$0.376$ &	\cellcolor{blue!33}$0.189$ &	$0.187$ &	$0.323$ &	\cellcolor{red!33}$0.107$ &	$0.216$ &	\cellcolor{purple!33}$0.082$\\
      791	& $8$ & $736$	& $0.377$	& \cellcolor{black!20} $0.171$	& $0.205$ &	$0.395$ &	\cellcolor{blue!33}$0.198$ &	$0.197$ &	$0.341$ &	\cellcolor{red!33}$0.118$ &	$0.222$ &	\cellcolor{purple!33}$0.08$\\
      \bottomrule
    \end{tabular}
    \label{app:subtable:fprNW-minus-fprW}
    \end{subtable}

    \vspace{0.5cm}
    \begin{subtable}[h]{\linewidth}
    \centering
    \subcaption{The top-$3$ most unfair models by subgroup-specific $\hatfpr$, when  $\hatfpr_\text{W} > \hatfpr_\text{NW}$ (i.e., unfair toward W).}
    \begin{tabular}{lllcccccccccc}
      \toprule
      {Run \#} & $s$ & $b$ & $\haterr$ & $\hatfpr$ & $\hatfnr$ & $\haterr_\text{NW}$ & $\hatfpr_\text{NW}$  & $\hatfnr_\text{NW}$ & $\haterr_\text{W}$  & $\hatfpr_\text{W}$ & $\hatfnr_\text{W}$ & $\hatfpr_\text{W} - \hatfpr_\text{NW}$ \\
      \midrule
      639 &	$7$ &	$280$ &	$0.36$ &	\cellcolor{black!20}$0.187$ &	$0.173$ &	$0.352$ &	\cellcolor{red!33}$0.174$ &	$0.178$ &	$0.376$ &	\cellcolor{blue!33}$0.212$ &	$0.164$ &	\cellcolor{purple!33}$0.038$ \\
      807	& $9$ &	$72$ 	& $0.381$ &	\cellcolor{black!20}$0.191$ &	$0.19$ &	$0.372$ &	\cellcolor{red!33}$0.179$ &	$0.192$ &	$0.398$ &	\cellcolor{blue!33}$0.214$ &	$0.184$ &	\cellcolor{purple!33}$0.035$\\
      543	 & $6$	& $264$ &	$0.358$ &	\cellcolor{black!20}$0.155$ &	$0.203$ &	$0.351$ &	\cellcolor{red!33}$0.144$ &	$0.206$ &	$0.37$ &	\cellcolor{blue!33}$0.175$ &	$0.196$ &	\cellcolor{purple!33}$0.031$ \\
      \bottomrule
    \end{tabular}
    \label{app:subtable:fprW-minus-fprNW}
    \end{subtable}

    \vspace{0.5cm}
    \begin{subtable}[h]{\linewidth}
    \centering
        \subcaption{The top-$3$ most unfair models by subgroup-specific $\hatfnr$, when  $\hatfnr_\text{NW} > \hatfnr_\text{W}$ (i.e., unfair toward NW).}
    \begin{tabular}{lllcccccccccc}
      \toprule
      {Run \#} & $s$ & $b$ & $\haterr$ & $\hatfpr$ & $\hatfnr$ & $\haterr_\text{NW}$ & $\hatfpr_\text{NW}$  & $\hatfnr_\text{NW}$ & $\haterr_\text{W}$  & $\hatfpr_\text{W}$ & $\hatfnr_\text{W}$ & $\hatfnr_\text{NW} - \hatfnr_\text{W}$ \\
      \midrule
      246	& $3$ &	$141$ &	$0.379$ &	$0.166$ &	\cellcolor{black!20} $0.213$ &	$0.398$ &	$0.169$ &	\cellcolor{blue!33}$0.229$ &	$0.345$ &	$0.161$ &	\cellcolor{red!33}$0.184$ &	\cellcolor{purple!33}$0.045$ \\
      506 &	$6$	& $42$ & $0.367$ &	$0.17$ &	\cellcolor{black!20} $0.197$ &	$0.386$ &	$0.175$ &	\cellcolor{blue!33}$0.211$ &	$0.332$ &	$0.161$ &	\cellcolor{red!33}$0.171$ &	\cellcolor{purple!33}$0.04$\\
      204	& $3$	& $15$ &	$0.384$ &	$0.185$ &	\cellcolor{black!20} $0.199$ &	$0.394$ &	$0.181$ &	\cellcolor{blue!33}$0.213$ &	$0.365$ &	$0.192$ &	\cellcolor{red!33}$0.173$ &	\cellcolor{purple!33}$0.04$ \\
      \bottomrule
    \end{tabular}
    \label{app:subtable:fnrNW-minus-fnrW}
    \end{subtable}
    
    \vspace{0.5cm}
    \begin{subtable}[h]{\linewidth}
    \centering
    \subcaption{The top-$3$ most unfair models by subgroup-specific $\hatfnr$, when  $\hatfnr_\text{W} > \hatfnr_\text{NW}$ (i.e., unfair toward W).}
    \begin{tabular}{lllcccccccccc}
      \toprule
      {Run \#} & $s$ & $b$ & $\haterr$ & $\hatfpr$ & $\hatfnr$ & $\haterr_\text{NW}$ & $\hatfpr_\text{NW}$  & $\hatfnr_\text{NW}$ & $\haterr_\text{W}$  & $\hatfpr_\text{W}$ & $\hatfnr_\text{W}$ & $\hatfnr_\text{W} - \hatfnr_\text{NW}$ \\
      \midrule
      474	& $5$ &	$375$ & $0.373$ &	$0.175$ &	\cellcolor{black!20}$0.199$ &	$0.356$ &	$0.183$ &	\cellcolor{red!33}$0.174$ &	$0.406$ &	$0.159$ &	\cellcolor{blue!33}$0.247$ &	\cellcolor{purple!33}$0.073$ \\
      401 &	$5$	& $10$ &	$0.378$ &	$0.189$	& \cellcolor{black!20}$0.19$ &	$0.363$ &	$0.197$ &	\cellcolor{red!33}$0.167$ &	$0.406$ &	$0.173$ &	\cellcolor{blue!33}$0.233$ &	\cellcolor{purple!33}$0.066$\\
      52	& $1$	& $53$ &	$0.367$ &	$0.172$ &	\cellcolor{black!20}$0.196$ &	$0.351$ &	$0.178$ &	\cellcolor{red!33}$0.173$ &	$0.397$ &	$0.16$ &	\cellcolor{blue!33}$0.238$ &	\cellcolor{purple!33}$0.065$ \\
      \bottomrule
    \end{tabular}
    \label{app:subtable:fnrW-minus-fnrNW}
    \end{subtable}
  \label{app:table:compas-model-runs-diffs}
\end{table}

\newpage

This detailed view provides insight into how such a result is possible. Broadly speaking, individual runs have roughly similar error;\footnote{This should be taken relatively. In general, \texttt{COMPAS} demonstrates high error; the error is relatively tight given just how much error there is. The error fluctuates depending on the training data, but the average error rate across train/test splits is rather tight, despite the fluctuations in error within the $\boot$ runs of each split.} yet, the subgroup-specific error rates that compose the overall error can nevertheless vary widely depending on the underlying training data. This observation aligns with current interest in \emph{model multiplicity} in the algorithmic fairness community~\cite{black2022multiplicity, watson2023multiplicity}, which imports the idea from \citet{breiman2001multiplicity}. In this case, as suggested by Table~\ref{app:table:compas-model-runs-diffs}, there are models that demonstrate unfairness toward both subgroups with respect to each error rate metric $\haterr$, $\hatfpr$, and $\hatfnr$. When we move away from attempting to find a \emph{single} model that performs well (accurately or fairly) on \texttt{COMPAS}, and instead consider the information contained across different possible models, we yield the result that the average, expected behavior smooths over the variance in underlying models such that the result is close to fair: \emph{\textbf{The average of unfair models with high variance in subgroup error rates is essentially fair.}}

\vspace*{.25cm}
\custompar{Stability analysis} To verify the stability of this result, we re-execute our experiments for increasing numbers of train/test splits $S$ and replicates $\boot$. While our results for \texttt{COMPAS} are generally tight for small $S$ (e.g., Figures~\ref{subfig:compas-cdf-rfc} and~\ref{fig:compas-ens}), this was not the case for \texttt{German Credit}, for which it was difficult to estimate self-consistency consistently. As a result, for \texttt{COMPAS}, we did not expect markedly different results for increased $S$. Our results for $S=100, B=1001$ using logistic regression (Figure~\ref{app:fig:compas-cdf-rfc-100}, Table~\ref{app:table:compas-rfc-100})  confirm this intuition. 

\begin{minipage}{.55\linewidth}
\centering
        \includegraphics[width=.8\linewidth]{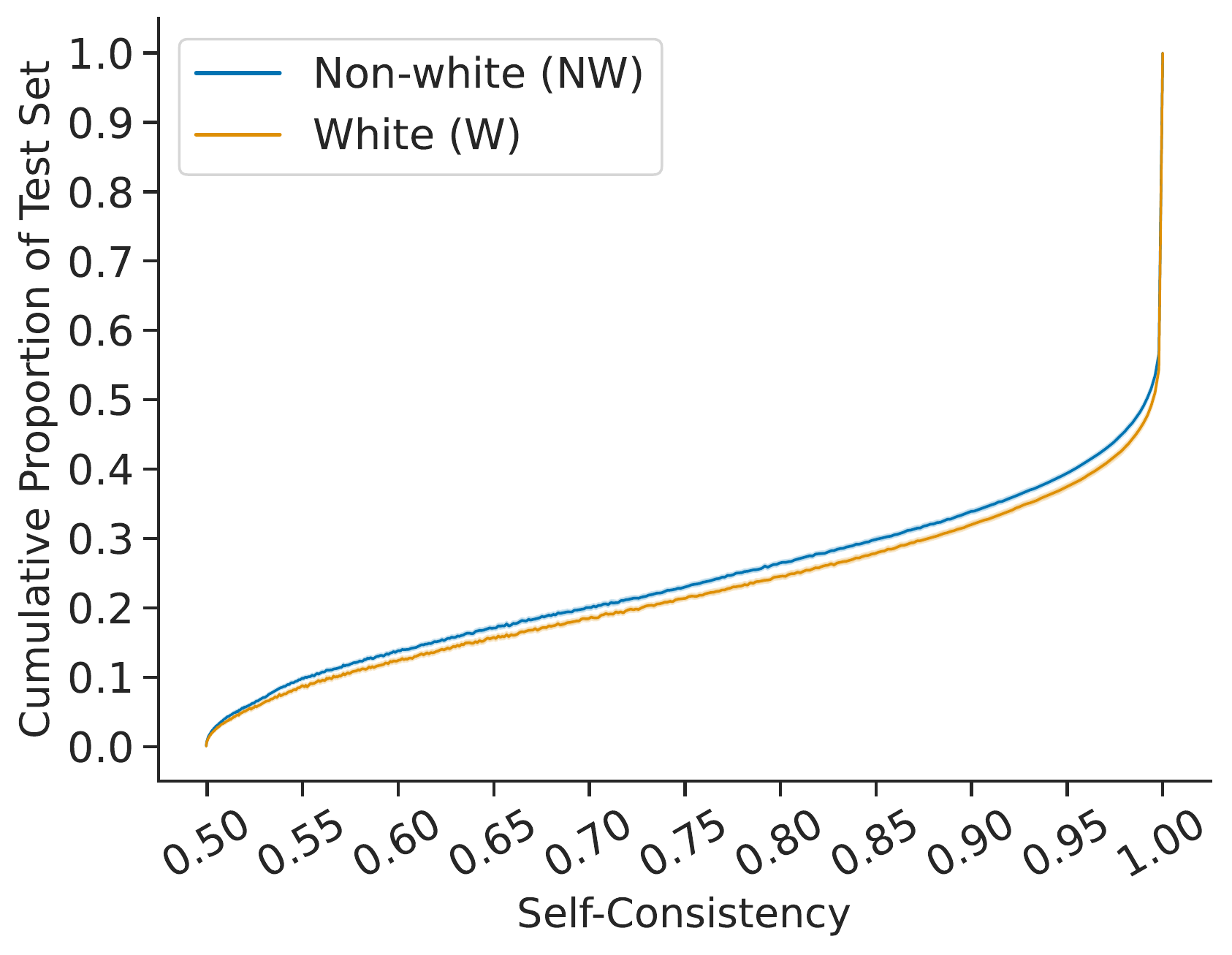}
        \vspace{-.3cm}
        \captionof{figure}{\texttt{COMPAS} split by $\group=\texttt{race}$, $B=1001, S=100$}
        \label{app:fig:compas-cdf-rfc-100}
\end{minipage}%
\hspace{-.5cm}
\begin{minipage}{.44\linewidth}
\centering
\vspace{-1.45cm}
      \begin{table}[H]
        \caption{Mean $\pm$ STD across $S=100$ train/test splits $\times\; B=1001$ runs.\looseness=-1}
        \label{app:table:compas-rfc-100}
        \scriptsize
        \begin{center}
        \begin{tabular}{lcccc}
		\toprule
		\multicolumn{5}{c}{\textbf{\texttt{COMPAS}}} \\ \cmidrule(lr){1-5}
		 & \textbf{$\haterr$}     & \textbf{$\hatfpr$}     & \textbf{$\hatfnr$}     & \textbf{$\hatsc$} \\ \midrule
		 \textbf{Total} & $.333\pm.008$ & $.14\pm.009$ & $.192\pm.01$ & $.883\pm.004$\\ \midrule
		 $g=\text{NW}$ & $.333\pm.01$ & $.148\pm.011$ & $.185\pm.012$ & $.88\pm.005$\\ \midrule
		 $g=\text{W}$ & $.332\pm.014$ & $.125\pm.013$ & $.207\pm.016$ & $.888\pm.006$\\ \bottomrule
	\end{tabular}
        \end{center}
        \end{table}
\end{minipage} 

\vspace*{.5cm}
We provide analogous results for \texttt{German Credit}, with $S=1000, B=1001$ using random forests (Figure~\ref{app:fig:german-cdf-rfc-1000}, Table~\ref{app:table:german-rfc-1000}). It takes an enormous number of runs to produce stable estimates of error and $\hatsc$ for \texttt{German Credit}, which indicate statistical equality across groups. Arguably, our results below for $1,001,000$ models still are very high variance (certainly with respect to error metrics). This task really has too few data points ($\approx600$) to generalize reliably. 
\vspace*{.5cm}

\begin{minipage}{.55\linewidth}
\centering
        \includegraphics[width=.8\linewidth]{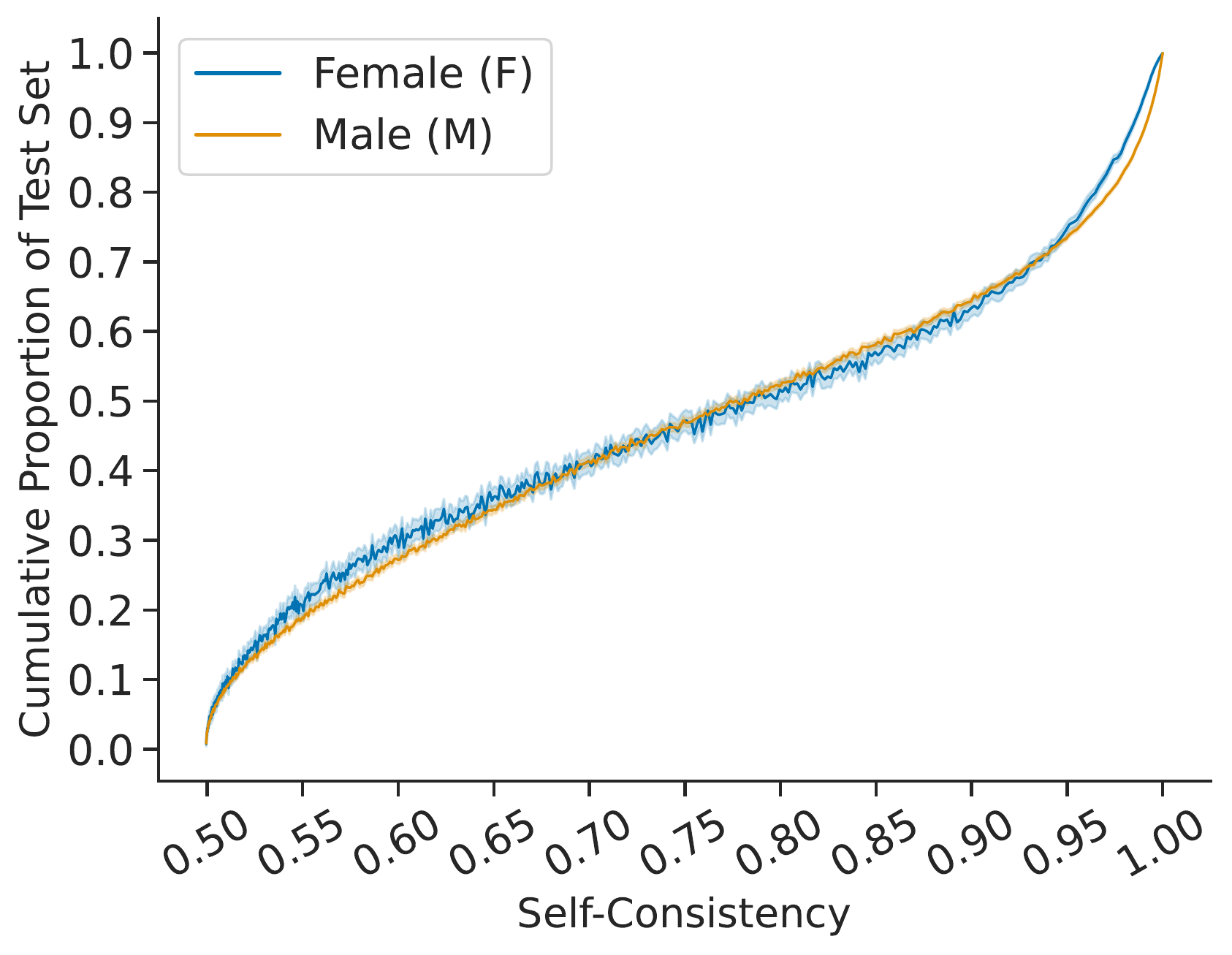}
        \vspace{-.3cm}
        \captionof{figure}{\texttt{German Credit} split by $\group=\texttt{sex}$, $S=1000, B = 100$\looseness=-1}
        \label{app:fig:german-cdf-rfc-1000}
\end{minipage}%
\hspace{-.5cm}
\begin{minipage}{.44\linewidth}
\centering
\vspace{-1.45cm}
      \begin{table}[H]
        \caption{Mean $\pm$ STD across $S=1000$ train/test splits $\times\; B=1001$ runs.\looseness=-1}
        \label{app:table:german-rfc-1000}
        \scriptsize
        \begin{center}
        \begin{tabular}{lcccc}
		\toprule
		\multicolumn{5}{c}{\textbf{\texttt{South German Credit}}} \\ \cmidrule(lr){1-5}
		 & \textbf{$\haterr$}     & \textbf{$\hatfpr$}     & \textbf{$\hatfnr$}     & \textbf{$\hatsc$} \\ \midrule
		 \textbf{Total} & $.28\pm.021$ & $.173\pm.028$ & $.107\pm.017$ & $.769\pm.015$\\ \midrule
		 $g=\text{F}$ & $.288\pm.064$ & $.183\pm.072$ & $.105\pm.037$ & $.766\pm.04$\\ \midrule
		 $g=\text{M}$ & $.279\pm.023$ & $.171\pm.029$ & $.108\pm.018$ & $.769\pm.016$\\ \bottomrule
	\end{tabular}
        \end{center}
        \end{table}
\end{minipage}

\section{\appfuture}\label{app:sec:future}

There are many interesting directions for future work that are out of scope for the present project. We address some topics below.

\custompar{Novel theory} We do not include extensive novel theory in this project. Nevertheless, our project raises interesting questions for theory in future work. Notably, we could compose our methodology with post-processing~\citep{hardt2016eo} for cases in which there is observed empirical unfairness. We could then investigate picking group-specific thresholds that take variance into account. We could reconfigure the formulations in \citep{hardt2016eo} and related work, with respect to the fairness-accuracy trade-off, as actually representing multiple such trade-off curves (that are a function of different models under consideration). There may be interesting directions for mathematical analysis in this direction. 

We could also extend traditional results on bagging and variance reduction for classifiers. While bagging has guarantees for variance reduction for regression, it does not have the same guarantees for classification~\citep{breiman1996bagging, breiman1998ac}. It generally is observed to work well in practice for variance reduction if the underlying classifiers are high variance --- which is indeed the regime we are in for this paper. However, there are interesting theory questions regarding abstention that we could investigate with theoretical tools, which could let us come up with other ways of reasoning about bagging and variance reduction.

Both of these directions are out of scope for the present paper. They are interesting, but do not have to do with our main experimental aims and contributions, and thus do not make it into a conference-length submission. We are not interested in novel theory in the present study. If anything, our work highlights how over-attention to theory can (directly or indirectly) bring about serious problems of mismeasurement in practice. That is a main takeaway for our work, which by nature does not involve novel theory.\looseness=-1

\custompar{Arbitrariness beyond algorithmic fairness} Our framework for reasoning about self-consistency and arbitrariness does not inherently have to do with algorithmic fairness. We could apply it to other domains. For example, it would be interesting to ask similar questions in deep learning and generative AI. We think that such work would be interesting, but is again out of scope for the present study. The first author of this project is in fact working on such questions as separate work. However, this project's research aims are inherently focused on fairness; the project was designed in response to observations in experimental practices in the fairness community, fairness definitions, and fairness theory.

\custompar{Experiments on synthetic data} Our results indicate that unfairness (as defined with respect to model error rates) is not frequently observed on common benchmark tasks in fair classification. Of course, there could be other datasets in fairness domains that are not currently used as benchmarks that more clearly demonstrate unfairness in practice. Hypothetically, there could be datasets for which we use Algorithm~\ref{algo:bagging-confidently} to reduce arbitrariness, and yet we still see significant systematic arbitrariness or differences in error rates (and thus unfairness) due to noise or bias. We just did not really see this for almost all of the tasks we investigate in this paper, which happen to be the ones that the fairness community uses for experiments. 

To study Algorithm~\ref{algo:bagging-confidently} in light of these other possibilities, we could develop synthetic datasets that retain unfairness after dealing with arbitrariness. We did not do this in the present study for two reasons. First, our focus was the practice of fairness research, as it currently stands, with a data-centric approach on the datasets people actually use for their research. We are not interested in synthetic data for this project. 

However, future theory results that extend our work could be vetted experimentally with synthetic data. The work we mention above regarding composition with post-processing, as well as revisting impossibility results from a distributional approach over possible models, may be very interesting to examine under data settings that we can control. 

\custompar{How to deal with abstention} Future work could also perform a deeper exploration of the trade-off between abstention rate and error. We could characterize a Pareto-optimal trade-off that is a function of the choice of self-consistency level $\kappa$, and also examine in experiments and analytically how absention leads to improvements in accuracy. Future work could also identify patterns in abstention sets beyond low self-consistency. In this, looking to metrics from model multiplicity may be helpful. Further, future work could combine human decision-making or other automated elements to see how we can root out arbitrariness. 

\custompar{Reproducibility} As mentioned in our Ethics Statement, we made attempts to reproduce prior work in fair classification, and often could not. We ultimately made reproducibility of specific papers out of scope for the present project, as we could make our contributions about arbitrariness and variance without such work. It would nevertheless be useful to focus future work on reproducing prior algorithmic fairness studies, and seeing if conclusions change in those works as a function of using Algorithm~\ref{algo:bagging-confidently} prior to introducing the proposed fairness intervention.

\custompar{Law and policy} As mentioned in our Ethics Statement, our work regarding arbitrariness raises concrete questions for the law around due process and automated decision-making. Such contributions are also out of the scope of the present work, but we are currently developing them for future submission to a law review journal.


\end{document}